\documentclass[10pt,twocolumn,letterpaper]{article}

\usepackage[final]{iccv}

\usepackage[T1]{fontenc}
\usepackage[dvipsnames]{xcolor}

\usepackage{multirow}
\usepackage{makecell}
\usepackage{listings}
\definecolor{iccvblue}{rgb}{0.21,0.49,0.74}
\usepackage[pagebackref,breaklinks,colorlinks,citecolor=iccvblue]{hyperref}

\usepackage{xspace}
\usepackage{framed}
\usepackage[accsupp]{axessibility}
\usepackage{graphicx}
\usepackage{makecell}
\usepackage{cuted}

\makeatletter
\DeclareRobustCommand\onedot{\futurelet\@let@token\@onedot}
\def\@onedot{\ifx\@let@token.\else.\null\fi\xspace}

\def\eg{\emph{e.g}\onedot} 
\def\ie{\emph{i.e}\onedot} 
 
\def\etc{\emph{etc}\onedot}

\makeatother

\newcommand{\Sec}{Sec.\xspace}

\newcommand{\myparagraph}[1]{\vspace{0.5em}\noindent\textbf{#1}}

\def\paperID{8342}
\def\confName{ICCV}
\def\confYear{2025}

\title{Kaputt: A Large-Scale Dataset for Visual Defect Detection}

\author{Sebastian H\"ofer$^{1}$ \and Dorian F. Henning$^{1}$ \and Artemij Amiranashvili$^{1}$ \and Douglas Morrison$^{1}$ \and Mariliza Tzes$^{1}$ \and Ingmar Posner$^{1,2}$ \and Marc Matvienko$^{1}$ \and Alessandro Rennola$^{1}$ \and Anton Milan$^{1}$\\
\\
$^{1}$Amazon, Fulfillment Technologies \& Robotics \quad $^{2}$University of Oxford, Applied AI Lab\\
{\tt\small \{hoefersh,doriahen,artemija,morridou,mtzes,ingmarp,mrcmtv,arenno,antmila\}@amazon.com}
}

\begin{document}
\maketitle
\begin{abstract}
We present a novel large-scale dataset for defect detection in a logistics setting. Recent work on industrial anomaly detection has primarily focused on manufacturing scenarios with highly controlled poses and a limited number of object categories. Existing benchmarks like MVTec-AD~\cite{bergmann_mvtec_2021} and VisA~\cite{zou2022spot} have reached saturation, with state-of-the-art methods achieving up to 99.9\% AUROC scores. In contrast to manufacturing, anomaly detection in retail logistics faces new challenges, particularly in the diversity and variability of object pose and appearance. Leading anomaly detection methods fall short when applied to this new setting.
To bridge this gap, we introduce a new benchmark that overcomes the current limitations of existing datasets. With over 230,000 images (and more than 29,000 defective instances), it is 40 times larger than MVTec and contains more than 48,000 distinct objects. To validate the difficulty of the problem, we conduct an extensive evaluation of multiple state-of-the-art anomaly detection methods, demonstrating that they do not surpass 56.96\% AUROC on our dataset. Further qualitative analysis confirms that existing methods struggle to leverage normal samples under heavy pose and appearance variation.
With our large-scale dataset, we set a new benchmark and encourage future research towards solving this challenging problem in retail logistics anomaly detection.
The dataset is available for download under \url{https://www.kaputt-dataset.com}.
\end{abstract}

\begin{figure*}[ht!]
    \centering
    \begin{minipage}[t]{0.615\textwidth}
        \centering
        \setlength{\tabcolsep}{1pt}
        \renewcommand{\arraystretch}{0.9}

        \begin{tabular}{@{}c@{}c@{}c@{}c@{}c@{}c@{}c@{}}
            \footnotesize Actuation & \footnotesize Deformation & \footnotesize Deconstruct. & \footnotesize Superficial & \footnotesize Penetration & \footnotesize Spillage & \footnotesize Missing Unit \\
            \hline
            \multicolumn{6}{c}{{\footnotesize  Minor Defects}} \\[2pt]
            \includegraphics[width=0.14\textwidth]{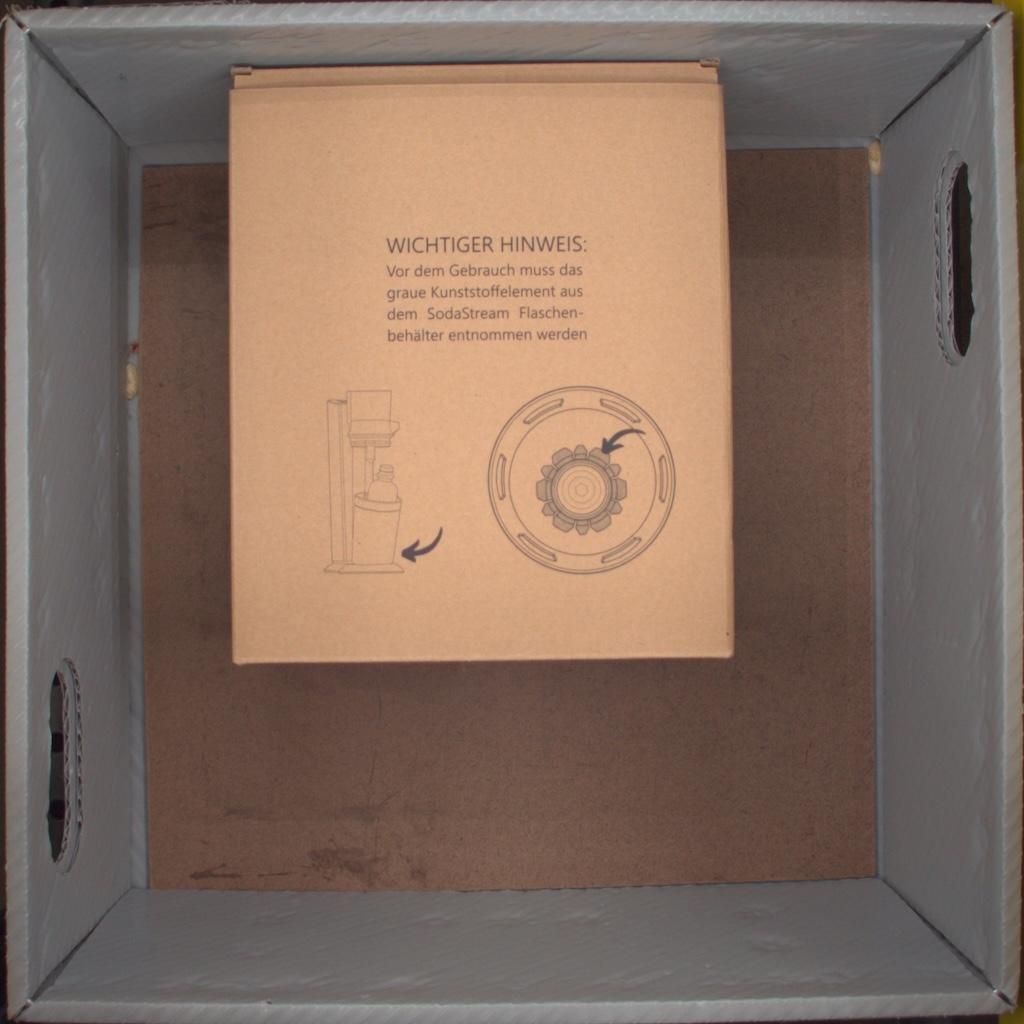} &
            \includegraphics[width=0.14\textwidth]{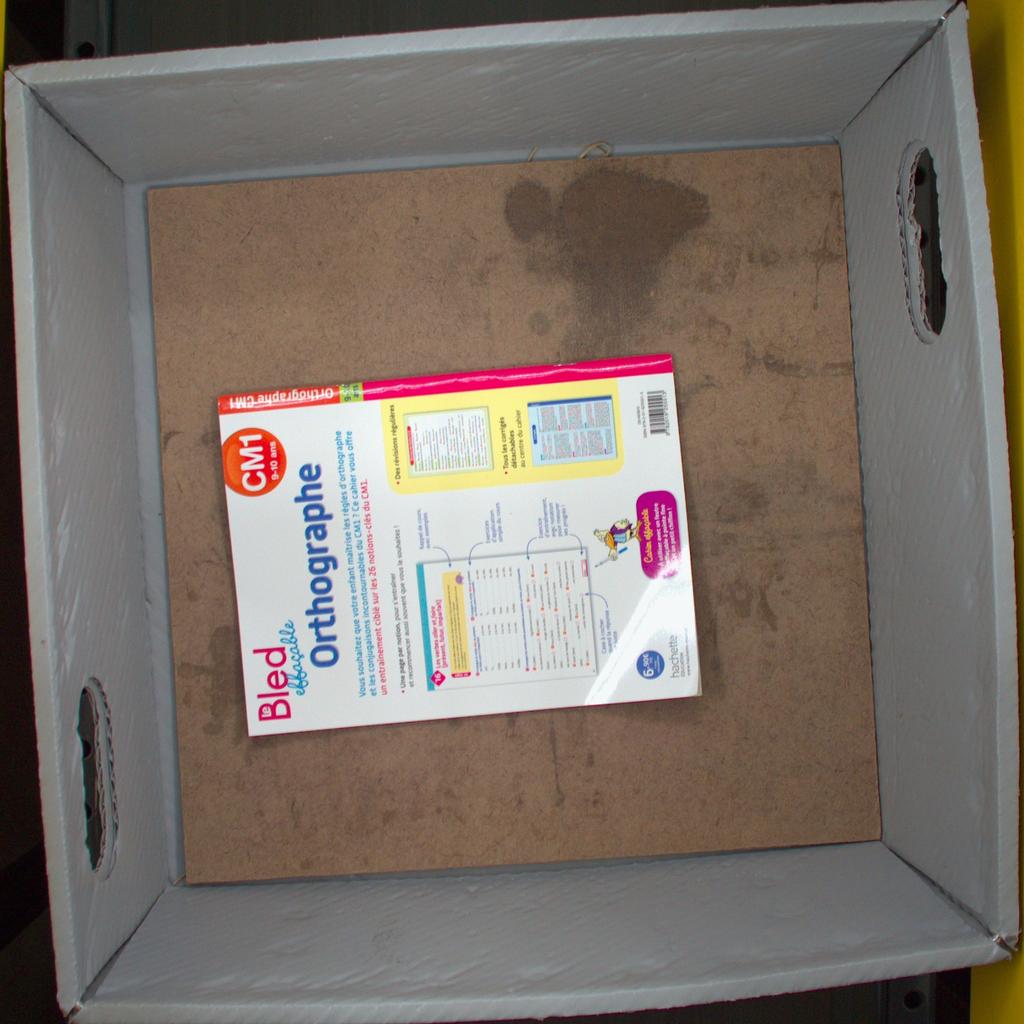} &
            \includegraphics[width=0.14\textwidth]{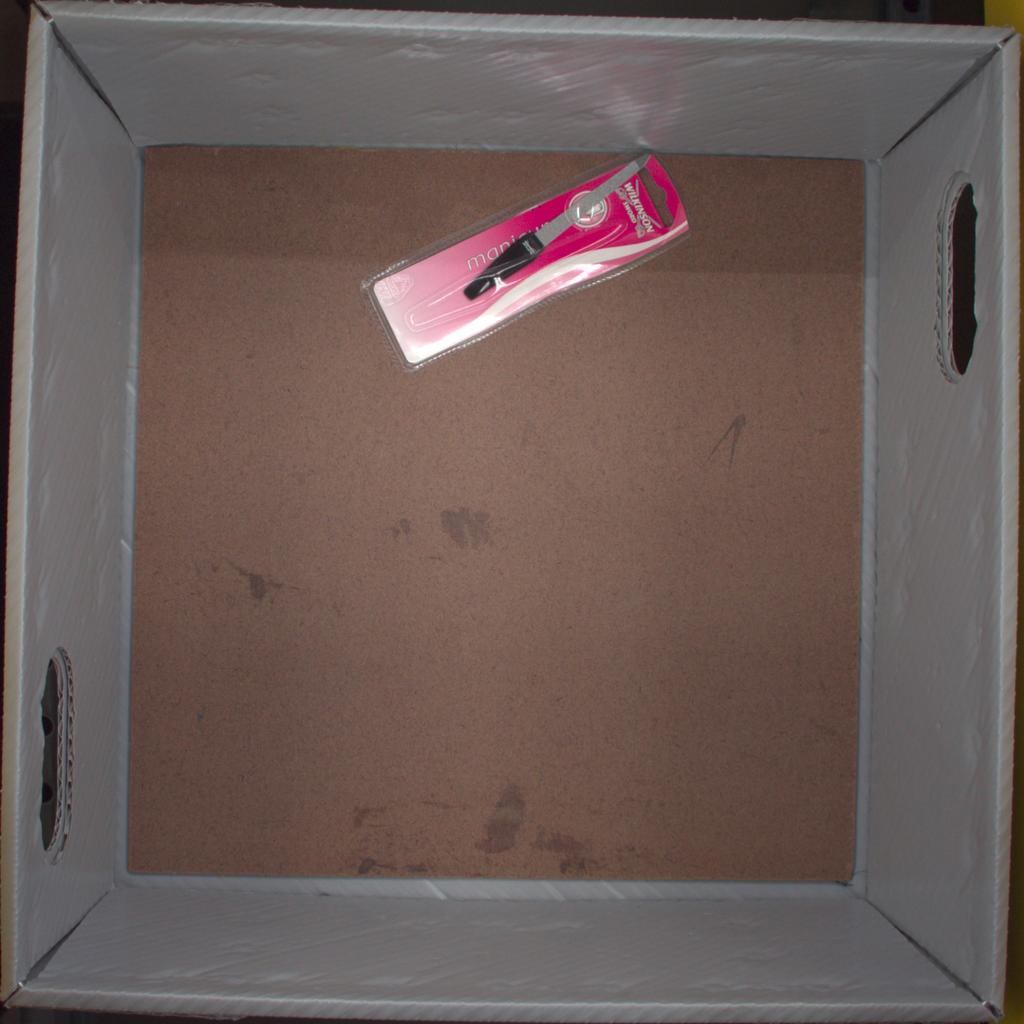} &
            \includegraphics[width=0.14\textwidth]{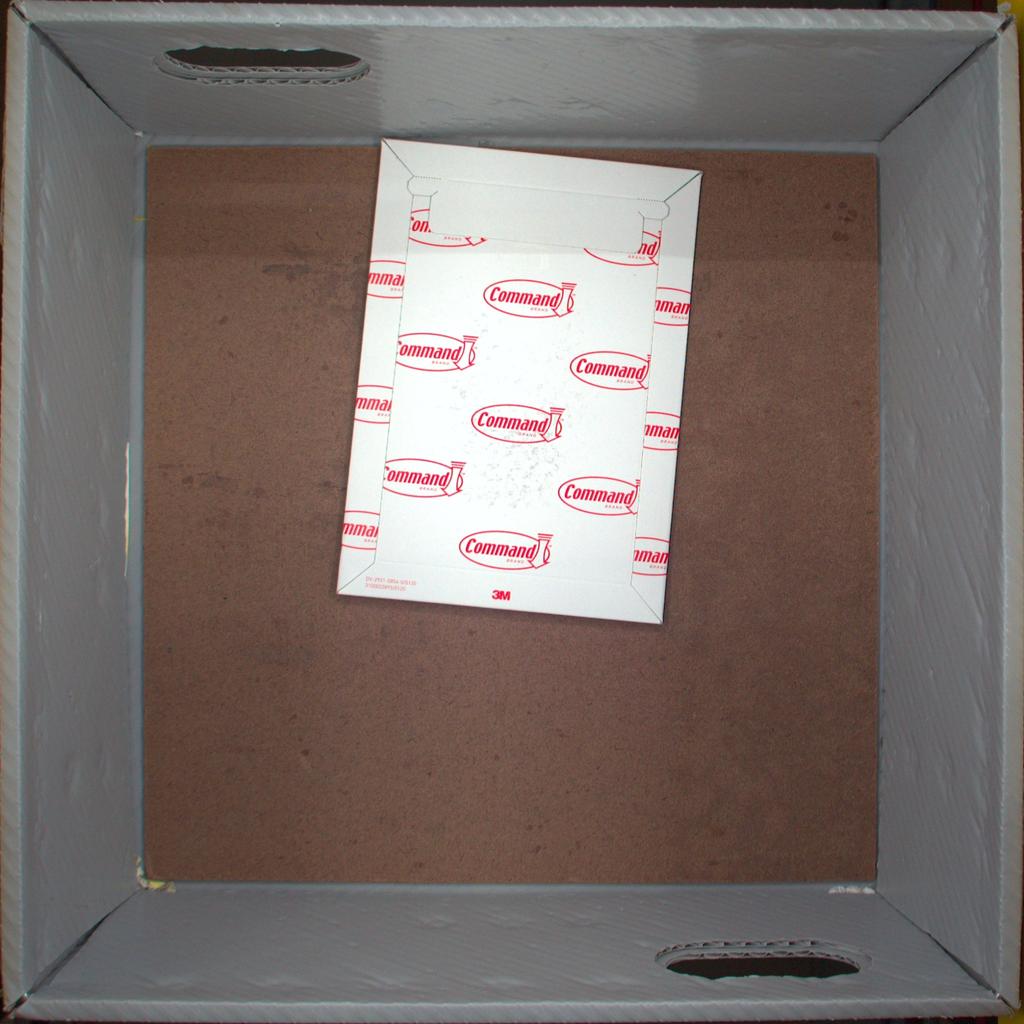} &
            \includegraphics[width=0.14\textwidth]{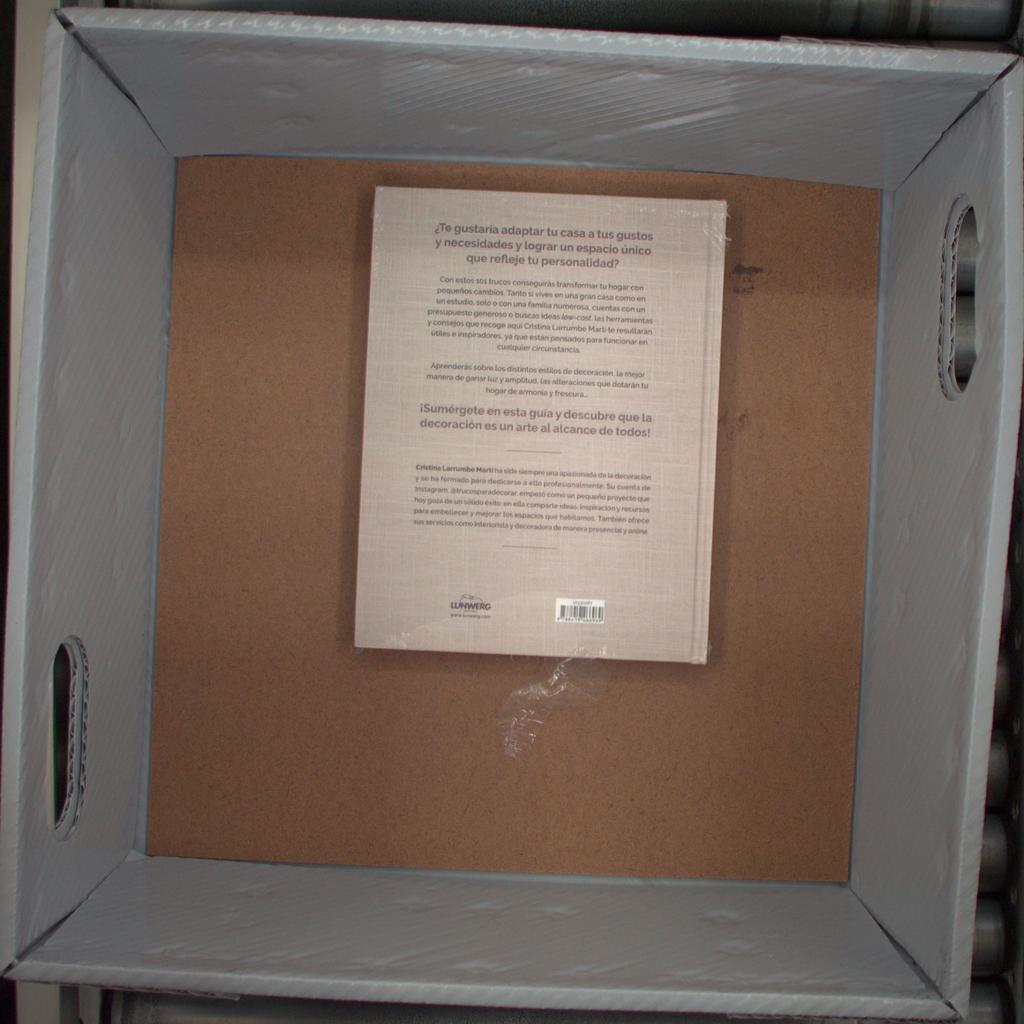} &
            \includegraphics[width=0.14\textwidth]{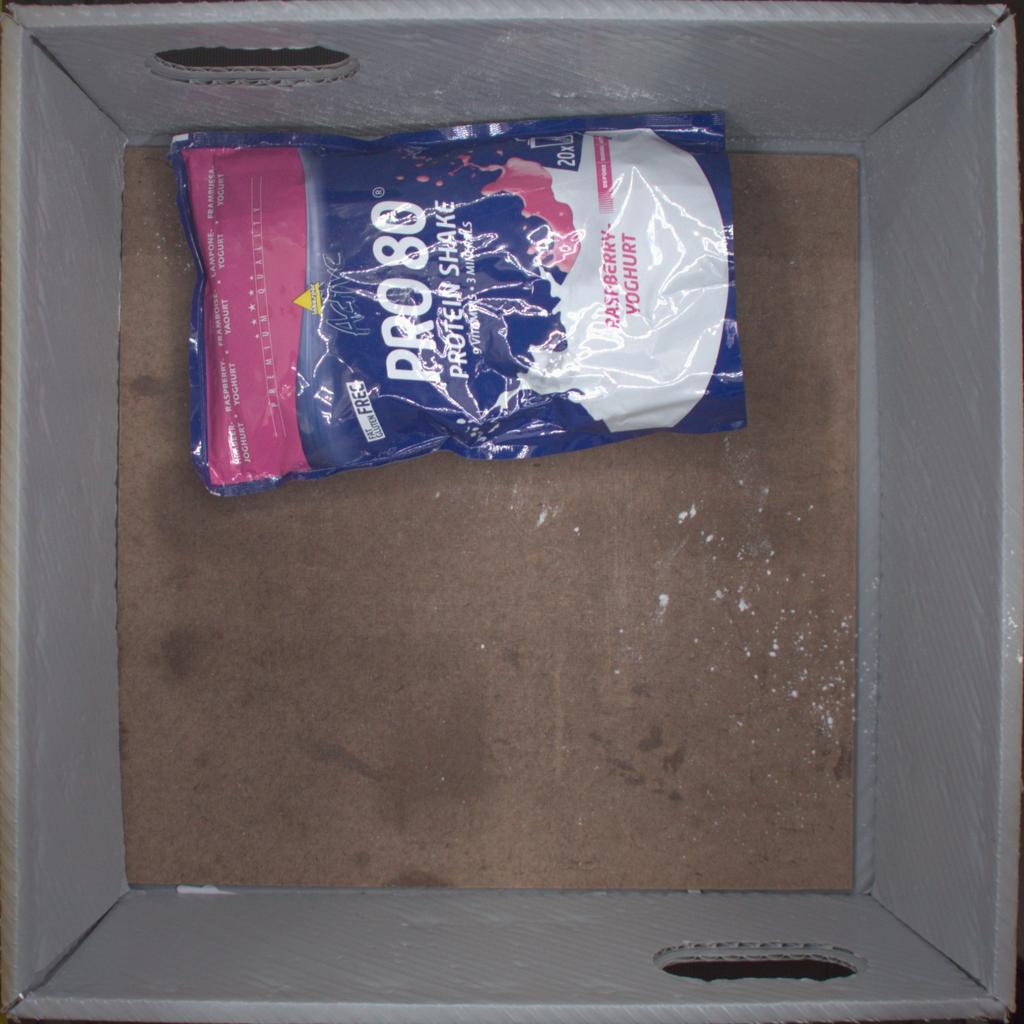}
            &

            \\[-3pt]
            \includegraphics[width=0.14\textwidth]{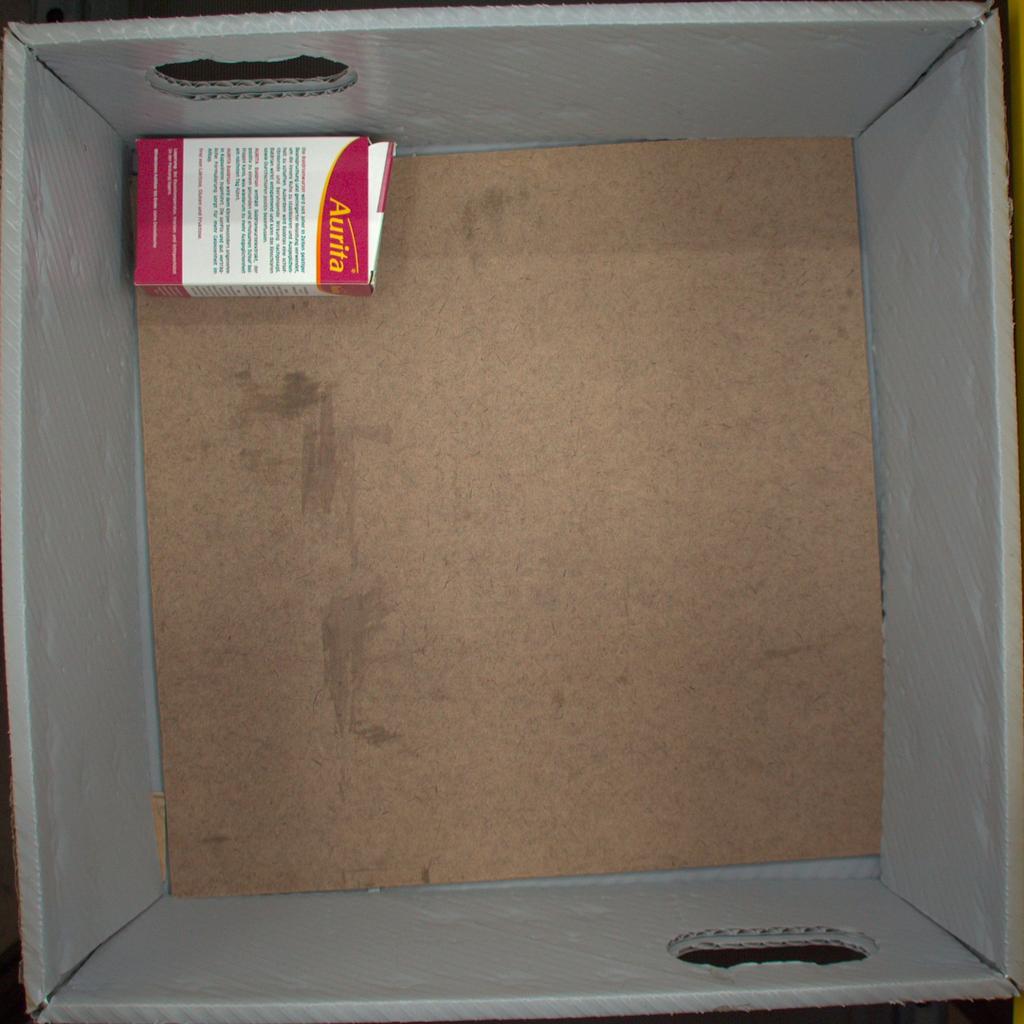} &
            \includegraphics[width=0.14\textwidth]{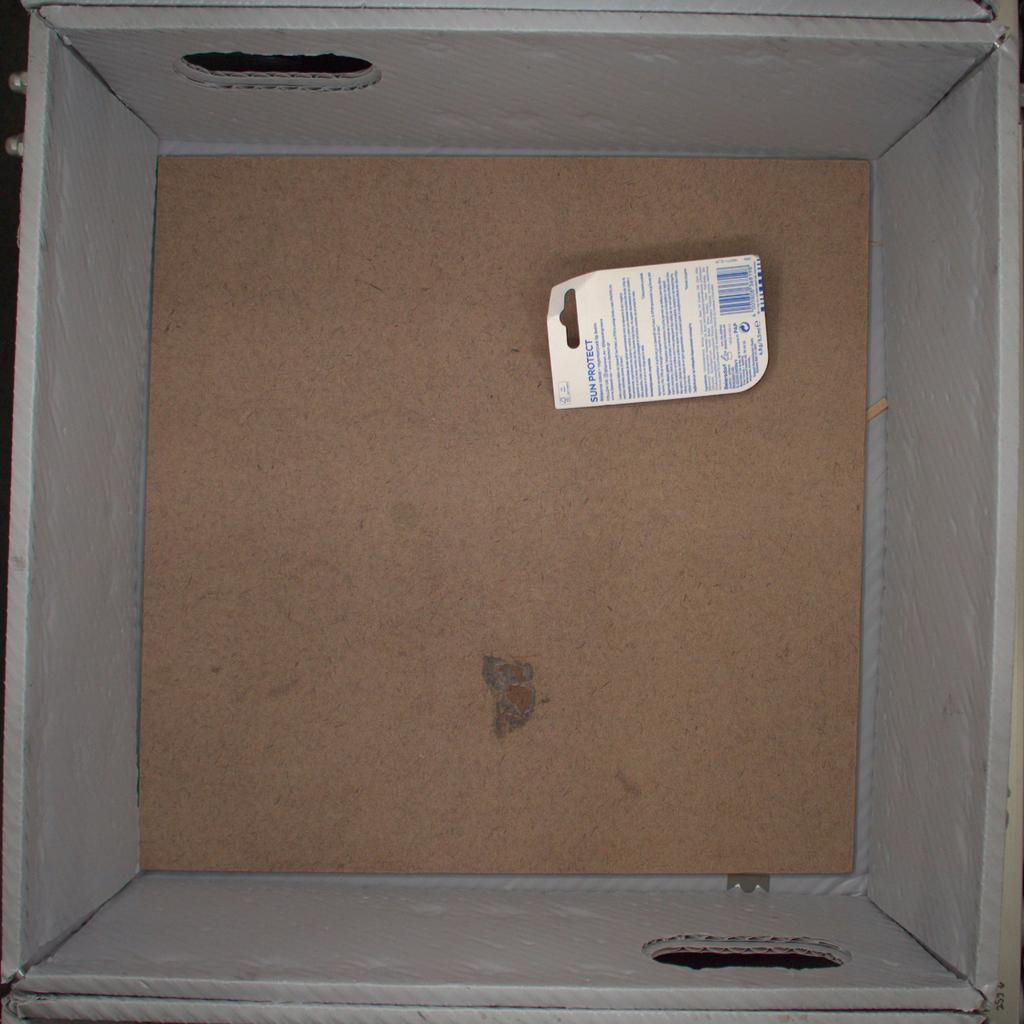} &
            \includegraphics[width=0.14\textwidth]{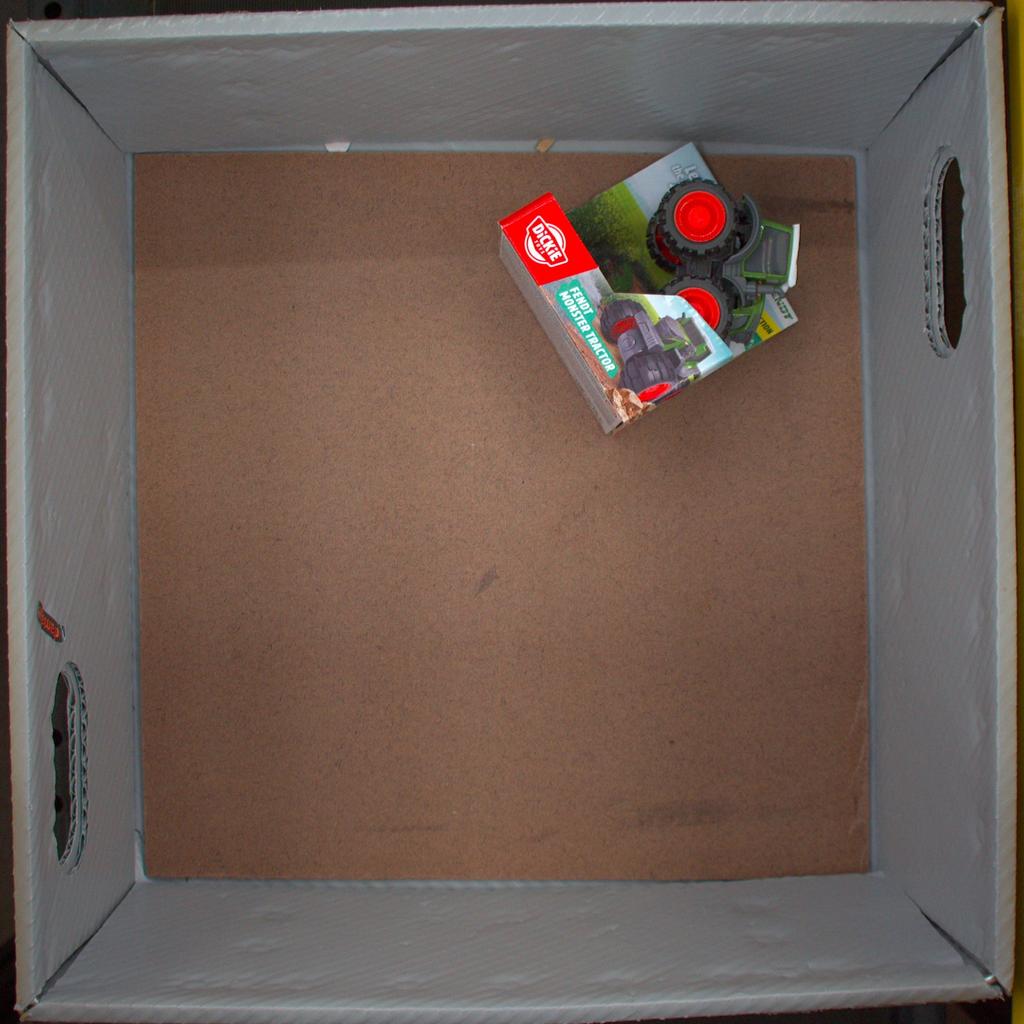} &
            \includegraphics[width=0.14\textwidth]{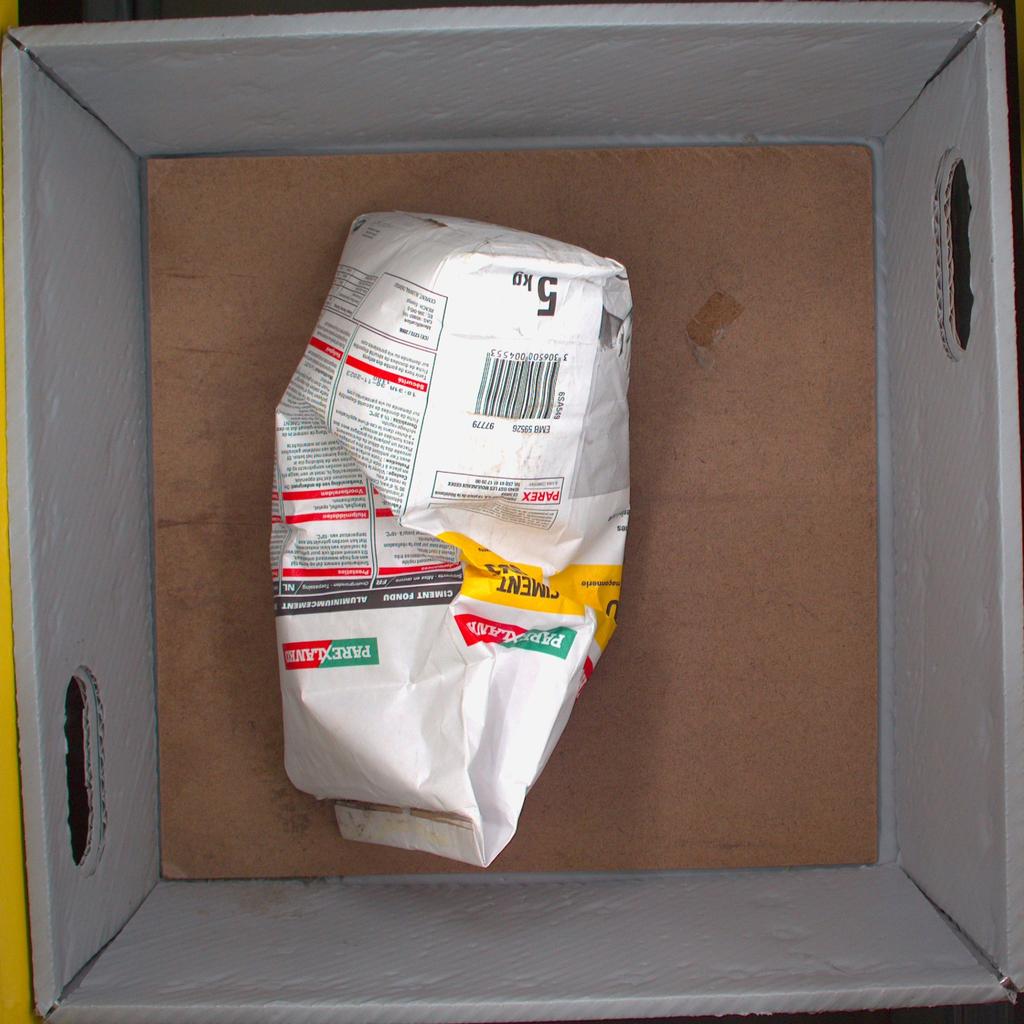} &
            \includegraphics[width=0.14\textwidth]{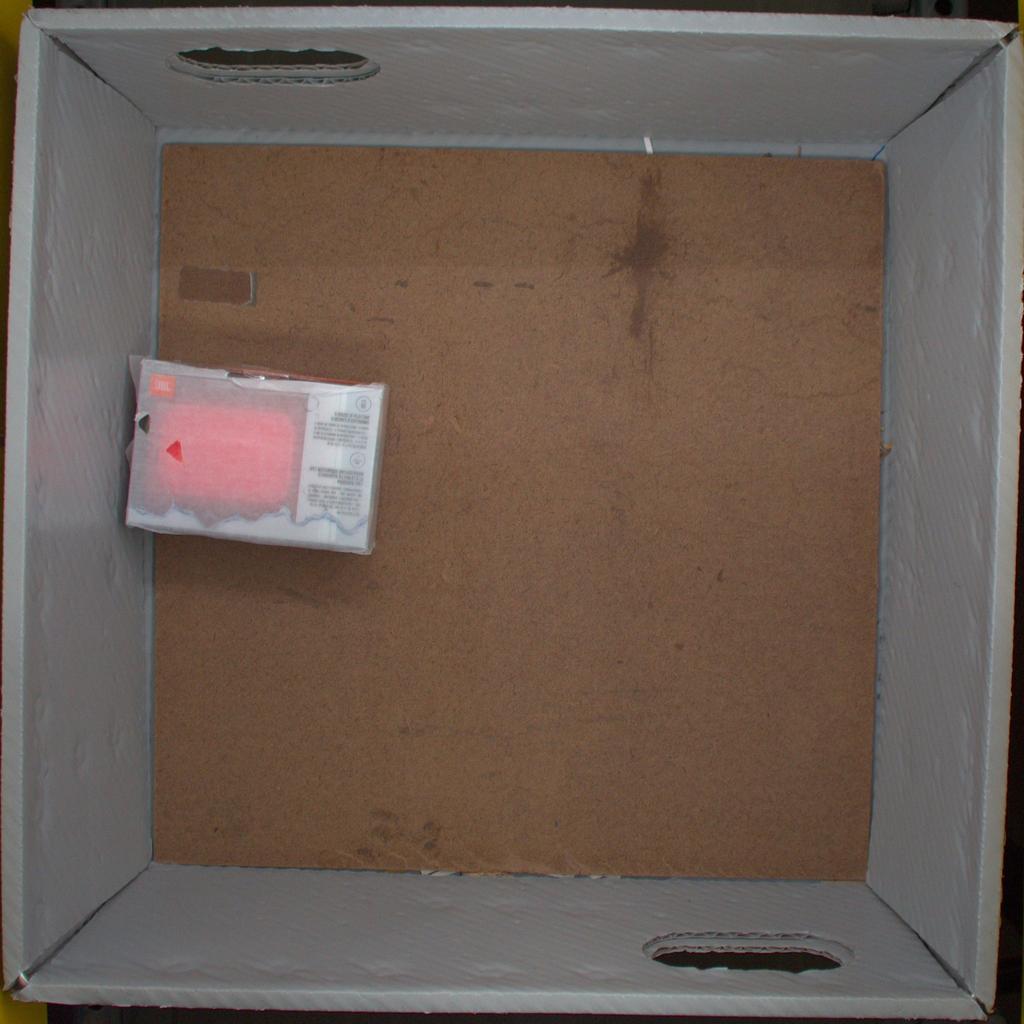} &
            \includegraphics[width=0.14\textwidth]{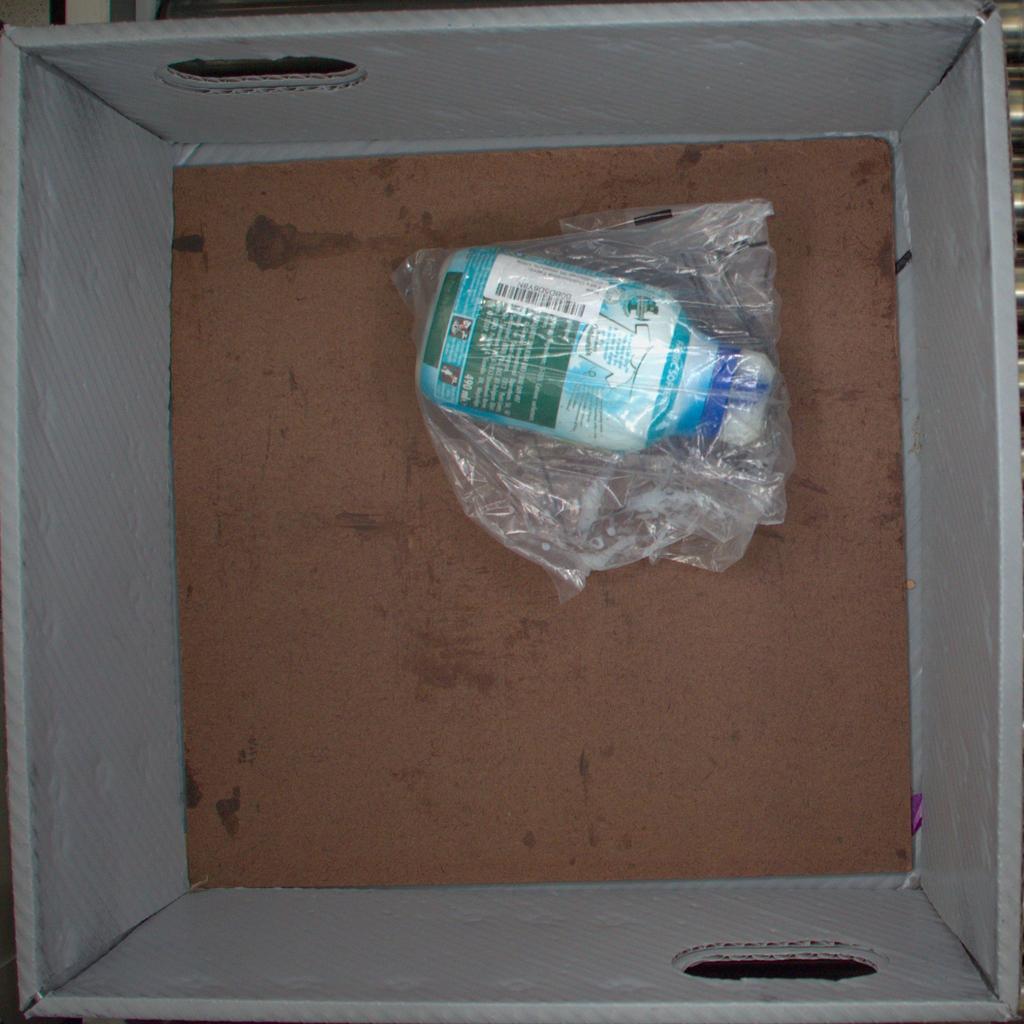}
            &

\\
            \hline
            \multicolumn{6}{c}{{\footnotesize Major Defects}} \\[2pt]
            \includegraphics[width=0.14\textwidth]{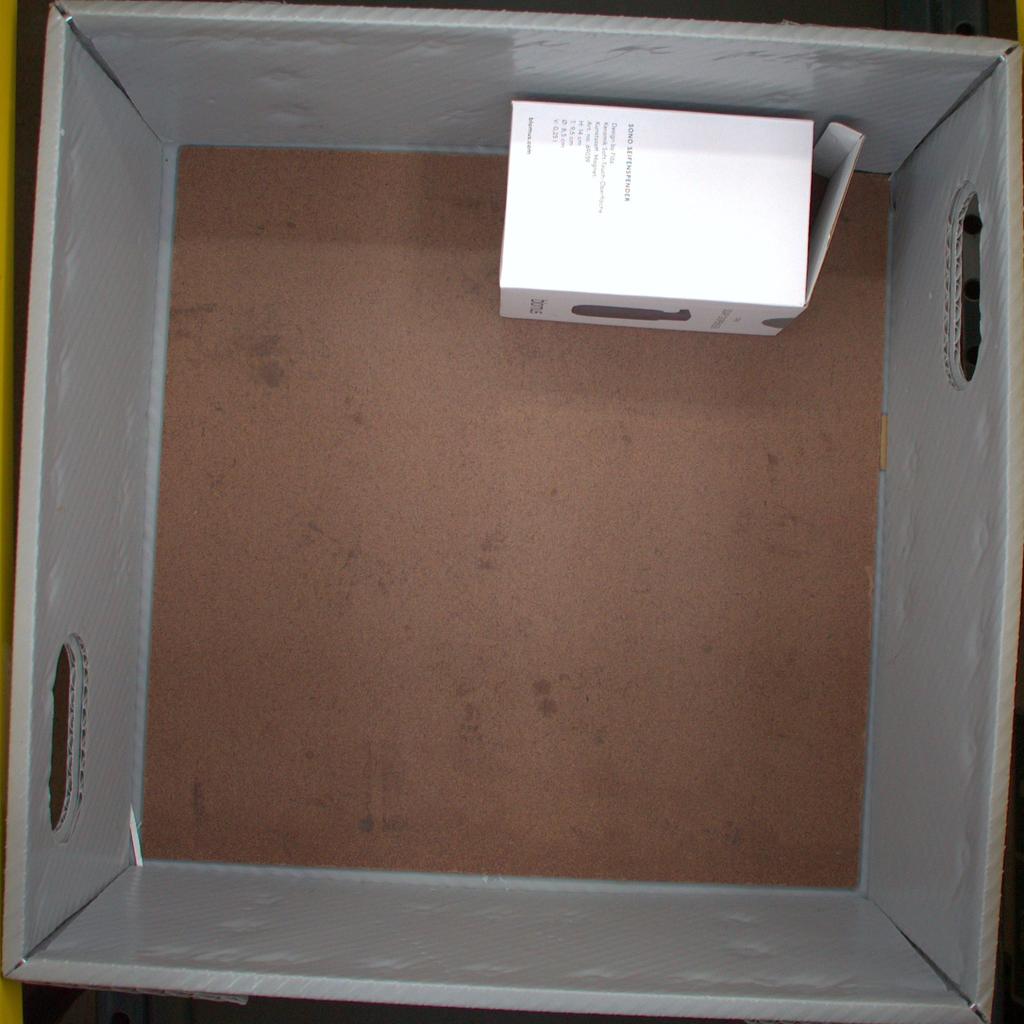} &
            \includegraphics[width=0.14\textwidth]{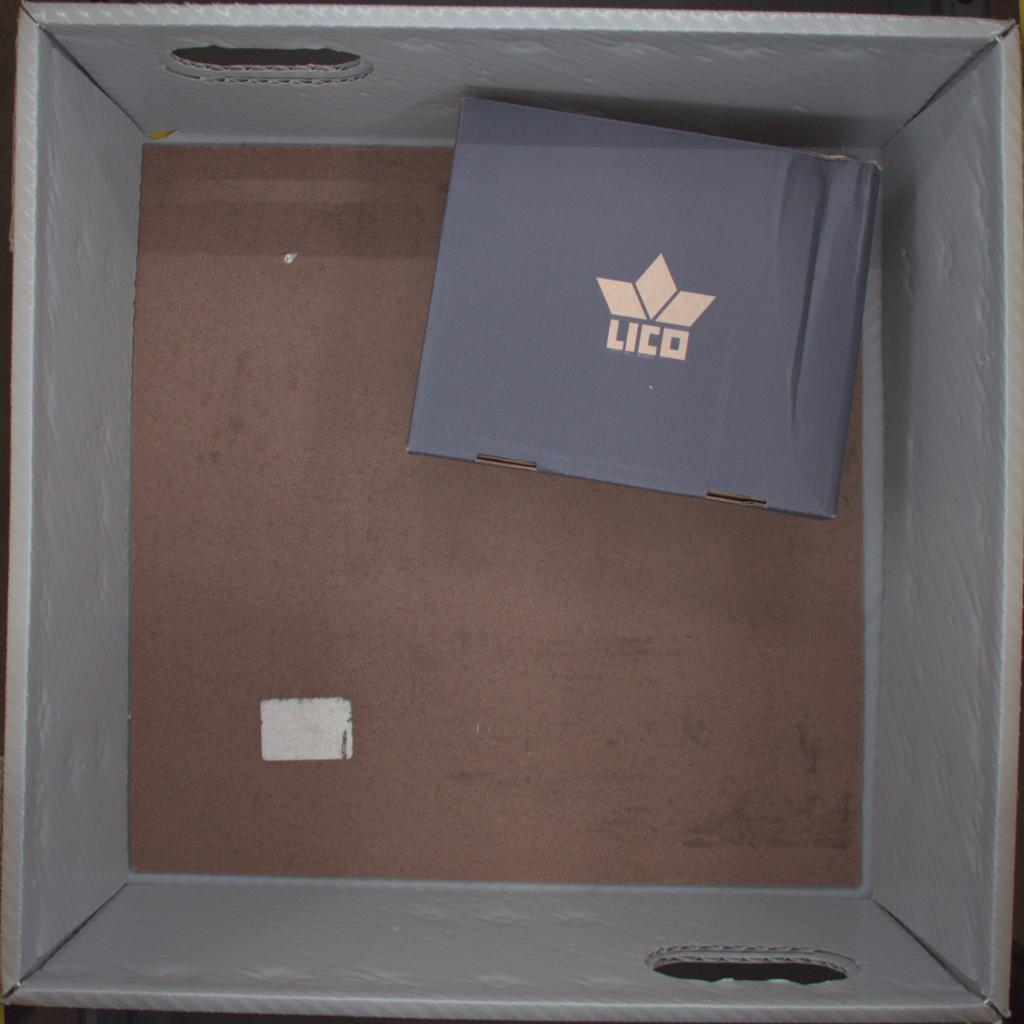} &
            \includegraphics[width=0.14\textwidth]{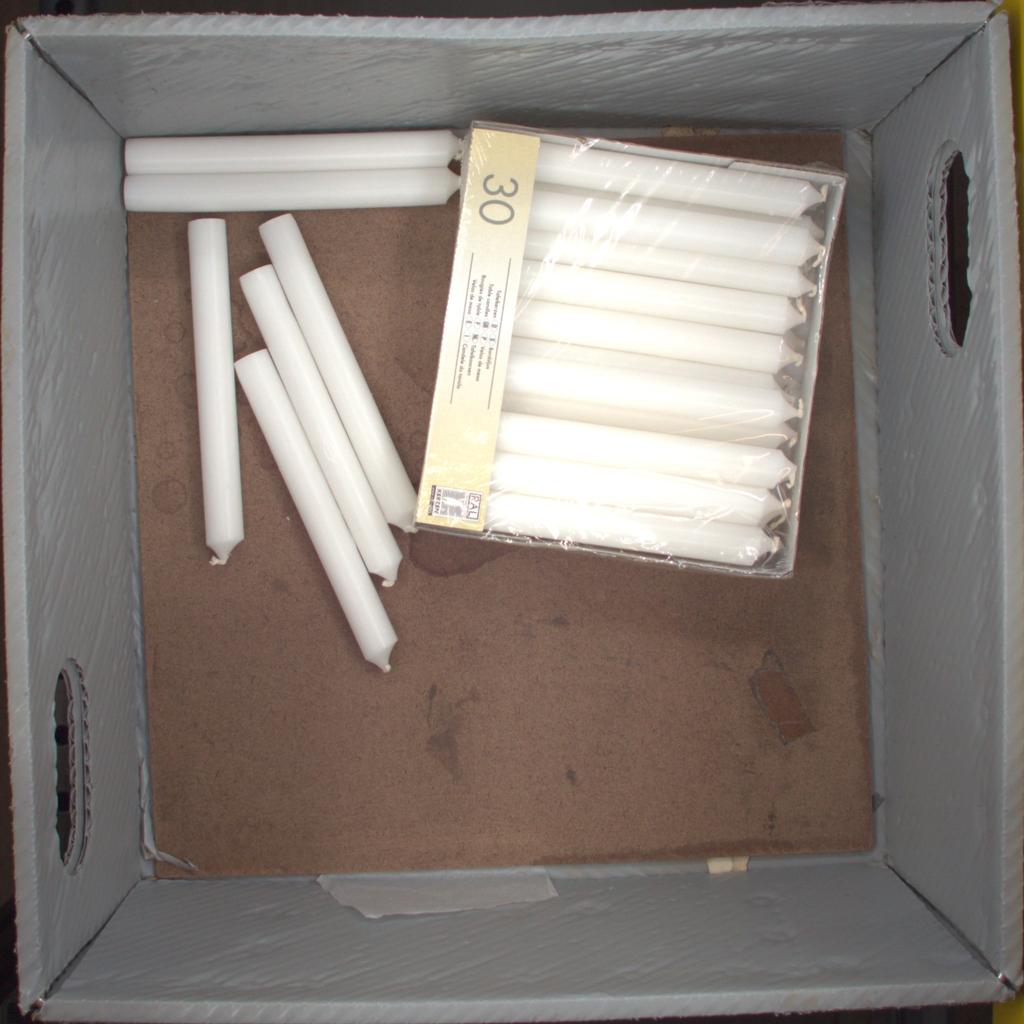} &
            \includegraphics[width=0.14\textwidth]{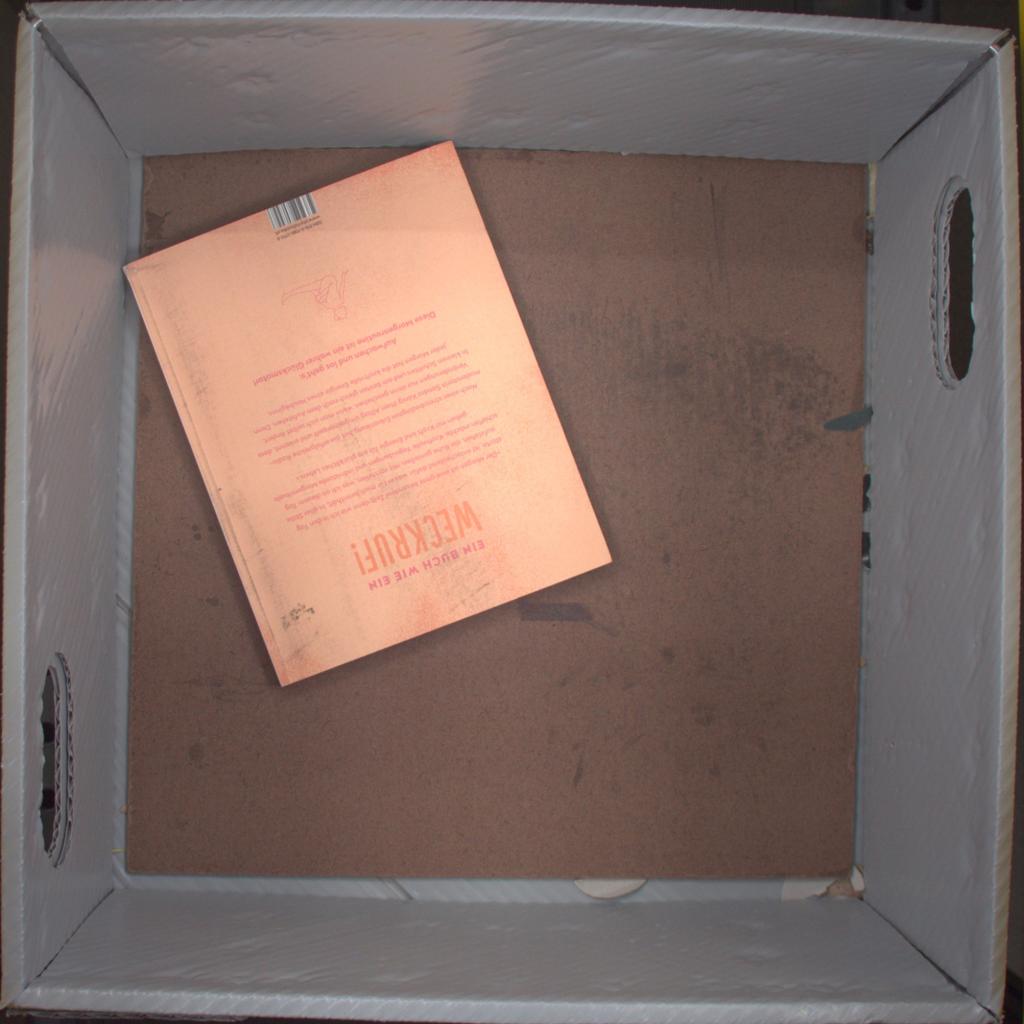} &
            \includegraphics[width=0.14\textwidth]{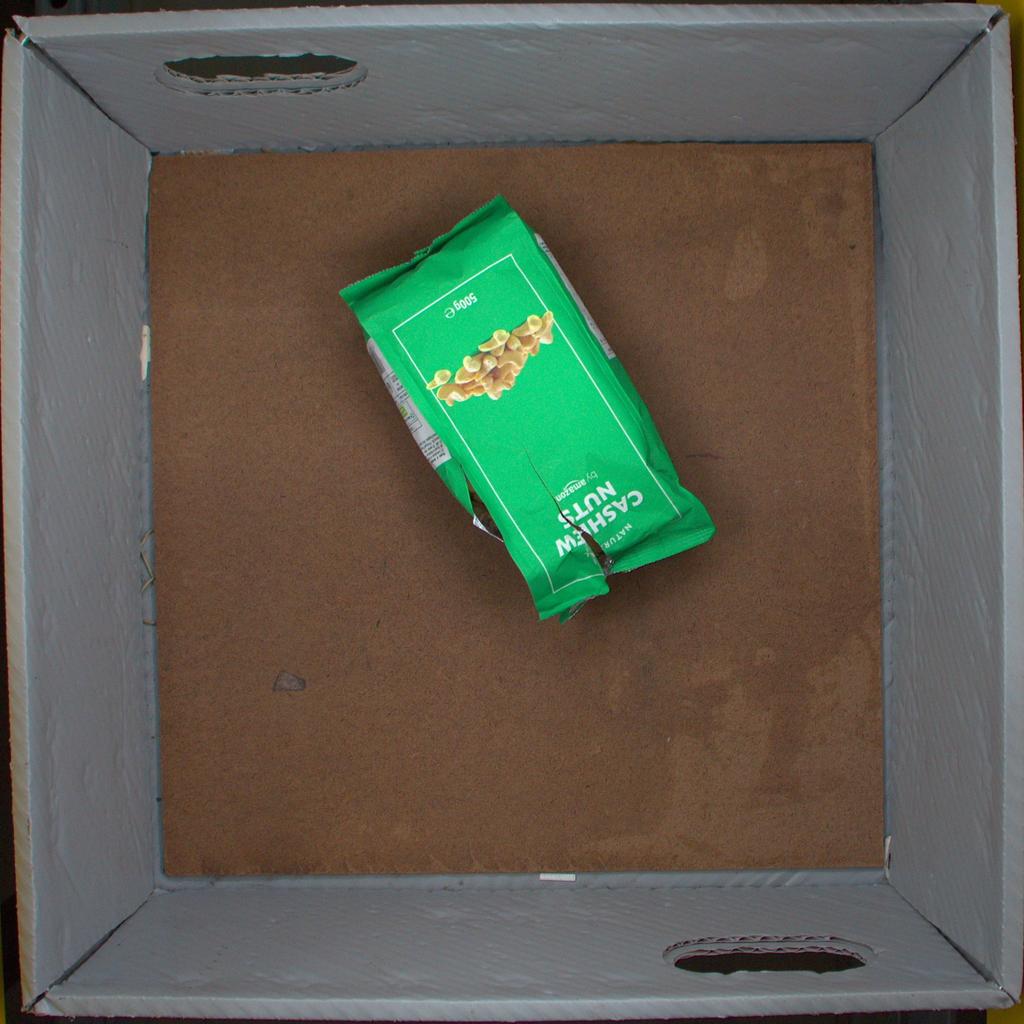} &
            \includegraphics[width=0.14\textwidth]{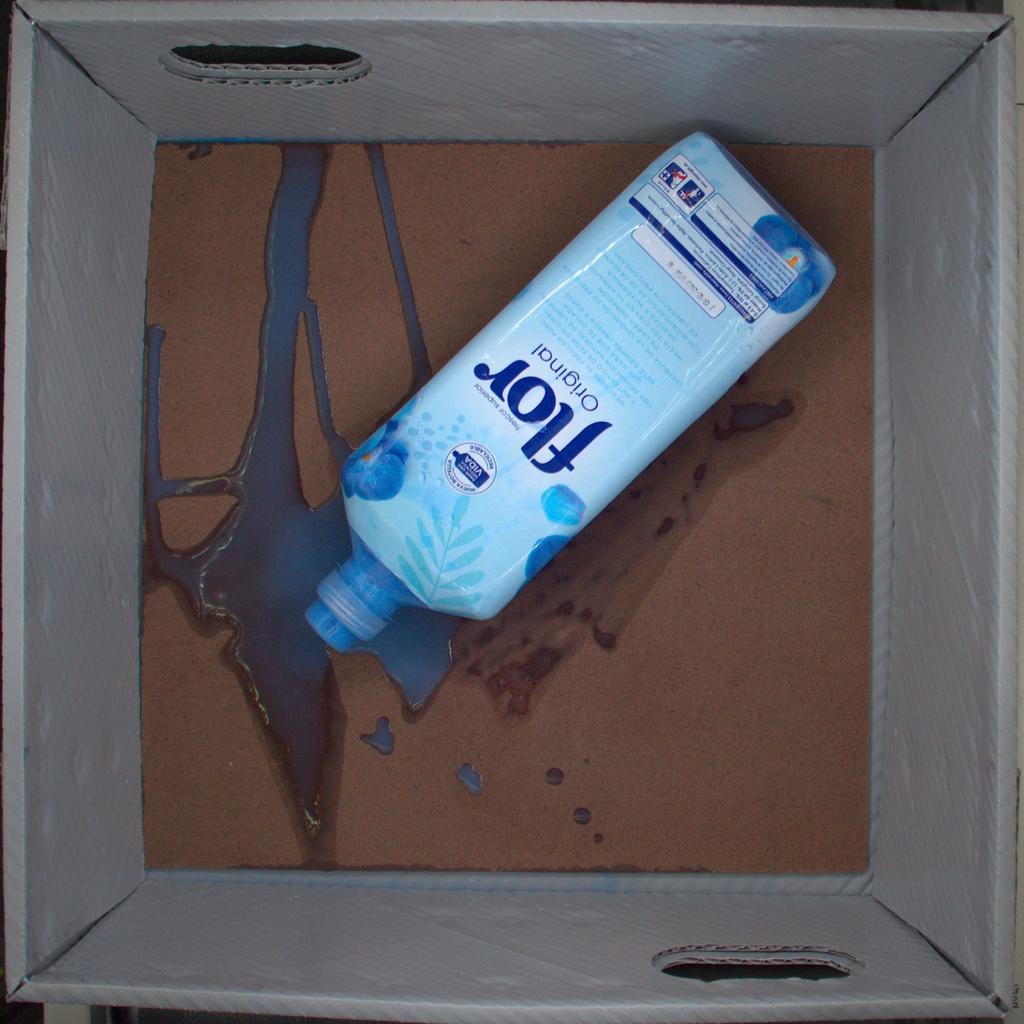} &
            \includegraphics[width=0.14\textwidth]{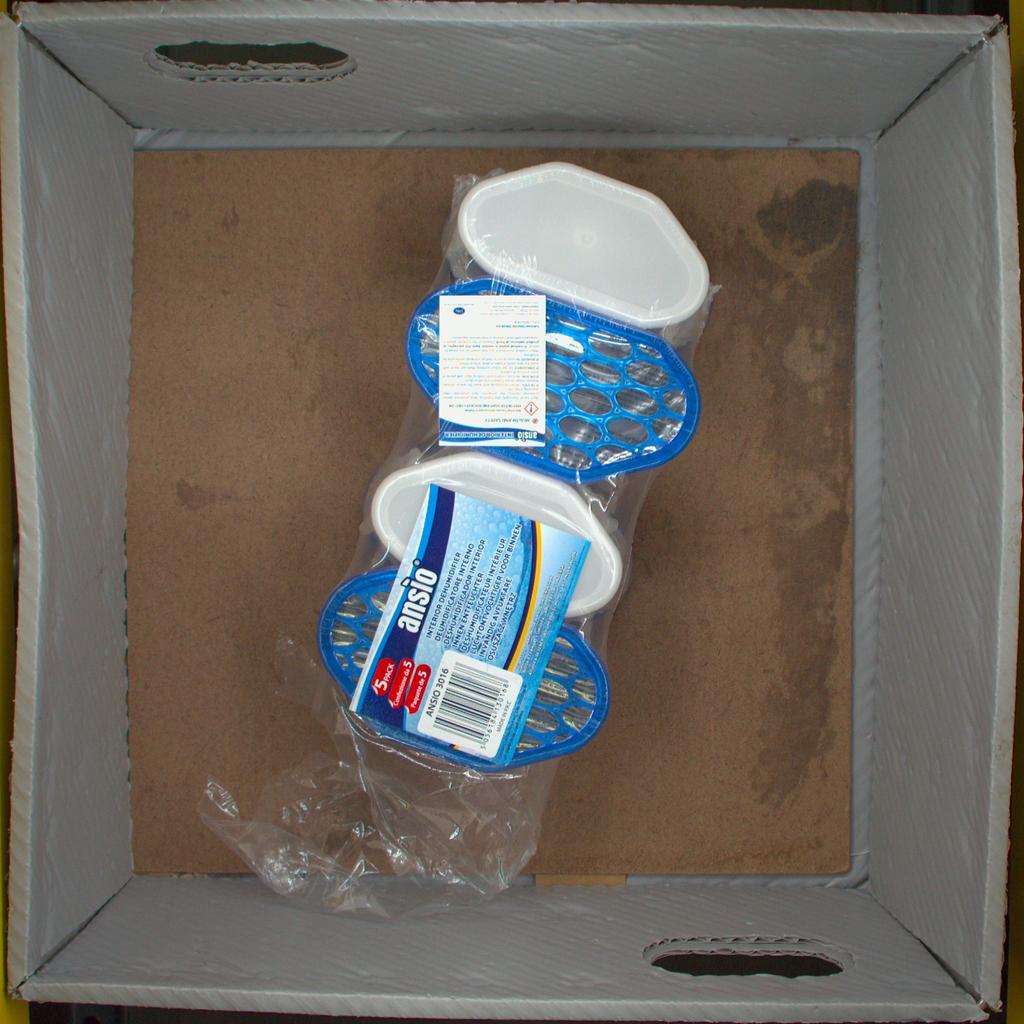} \\[-3pt]
            \includegraphics[width=0.14\textwidth]{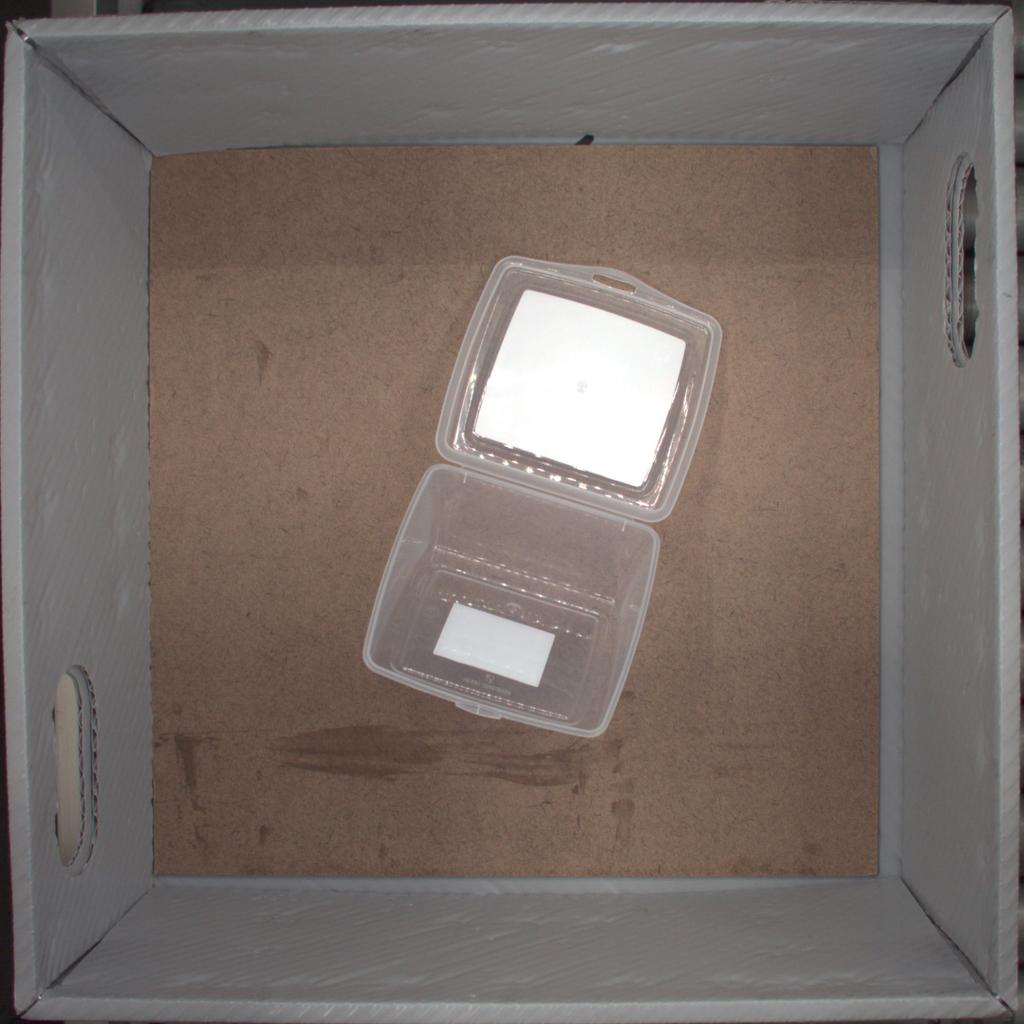} &
            \includegraphics[width=0.14\textwidth]{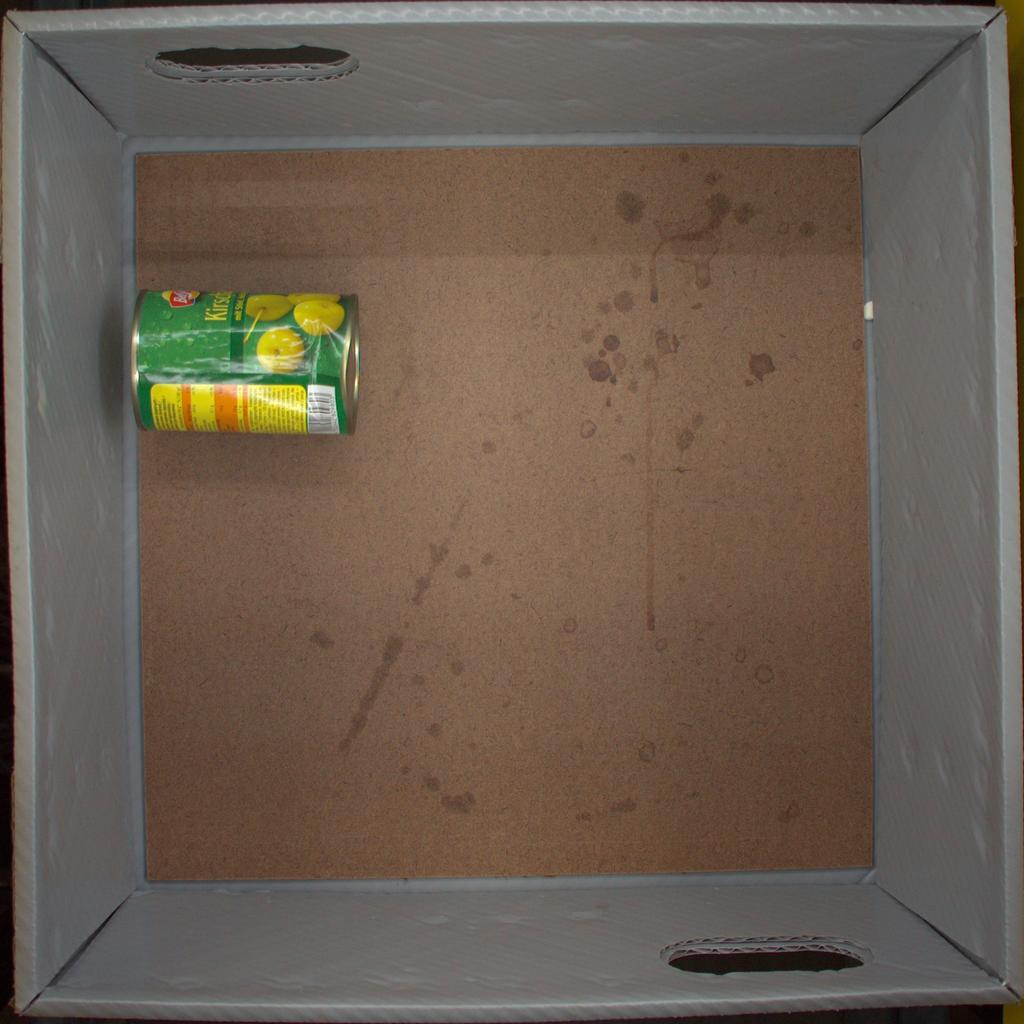} &
            \includegraphics[width=0.14\textwidth]{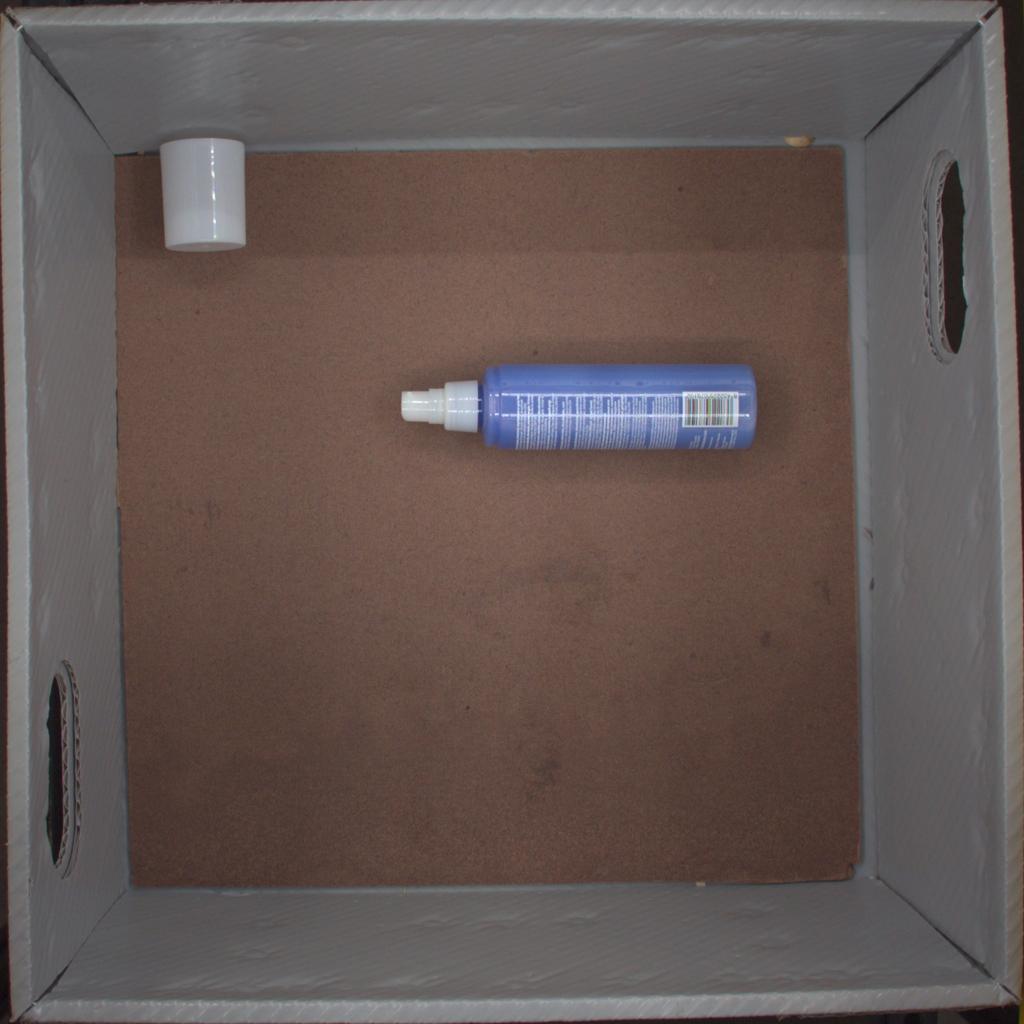} &
            \includegraphics[width=0.14\textwidth]{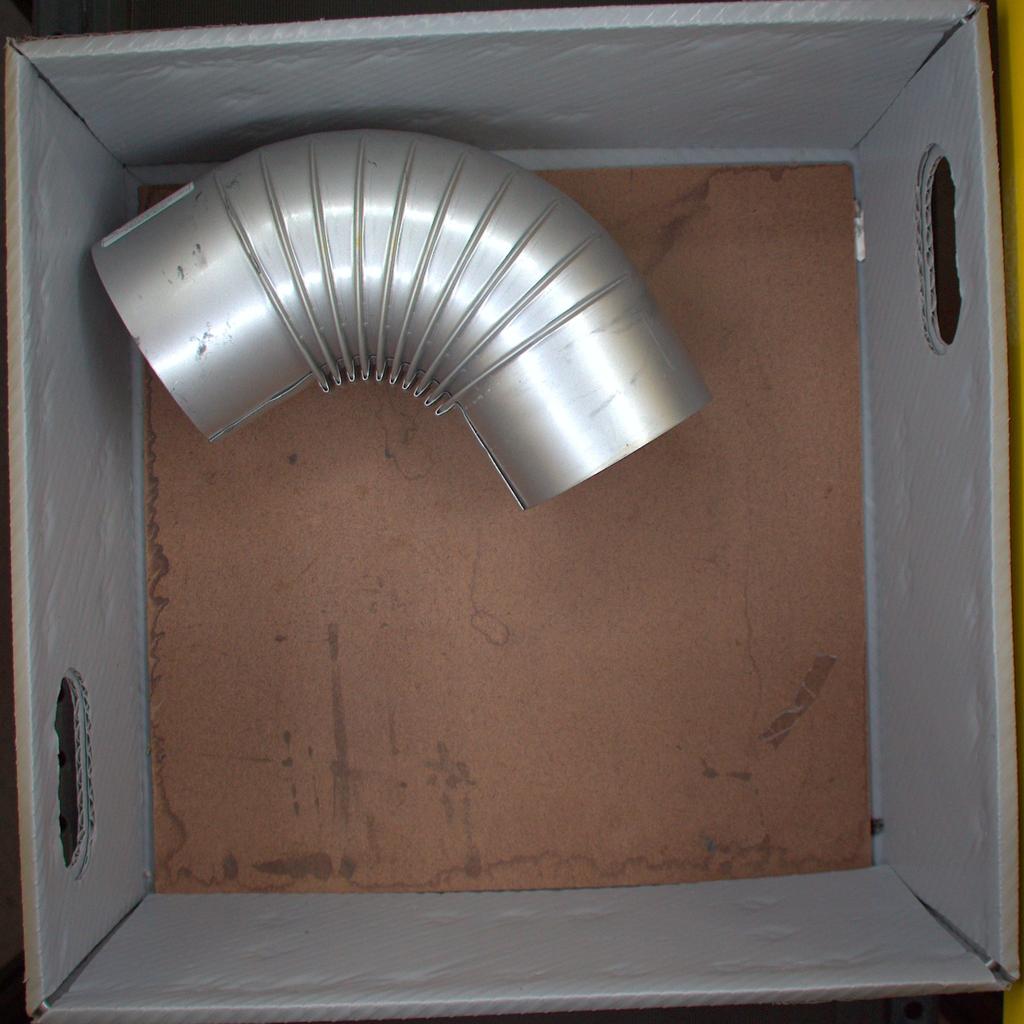} &
            \includegraphics[width=0.14\textwidth]{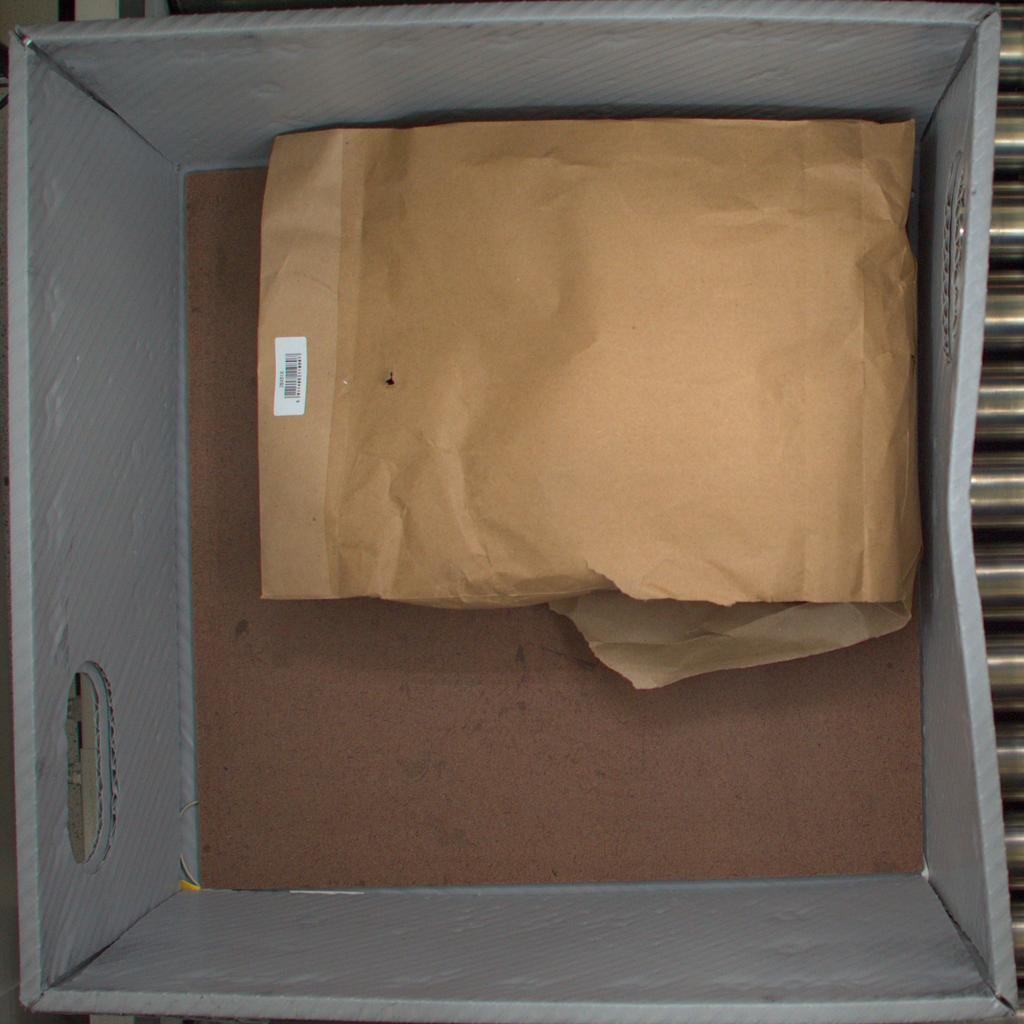} &
            \includegraphics[width=0.14\textwidth]{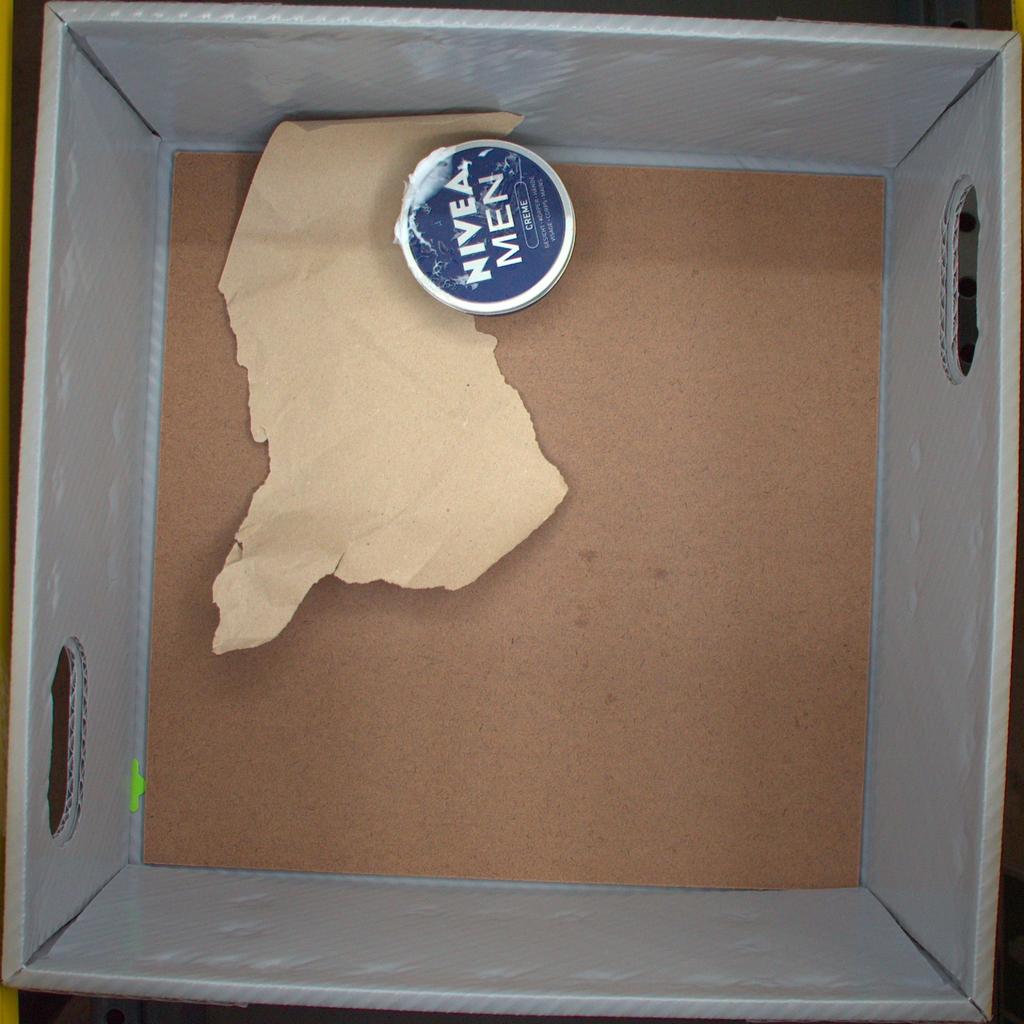}
            &
            \includegraphics[width=0.14\textwidth]{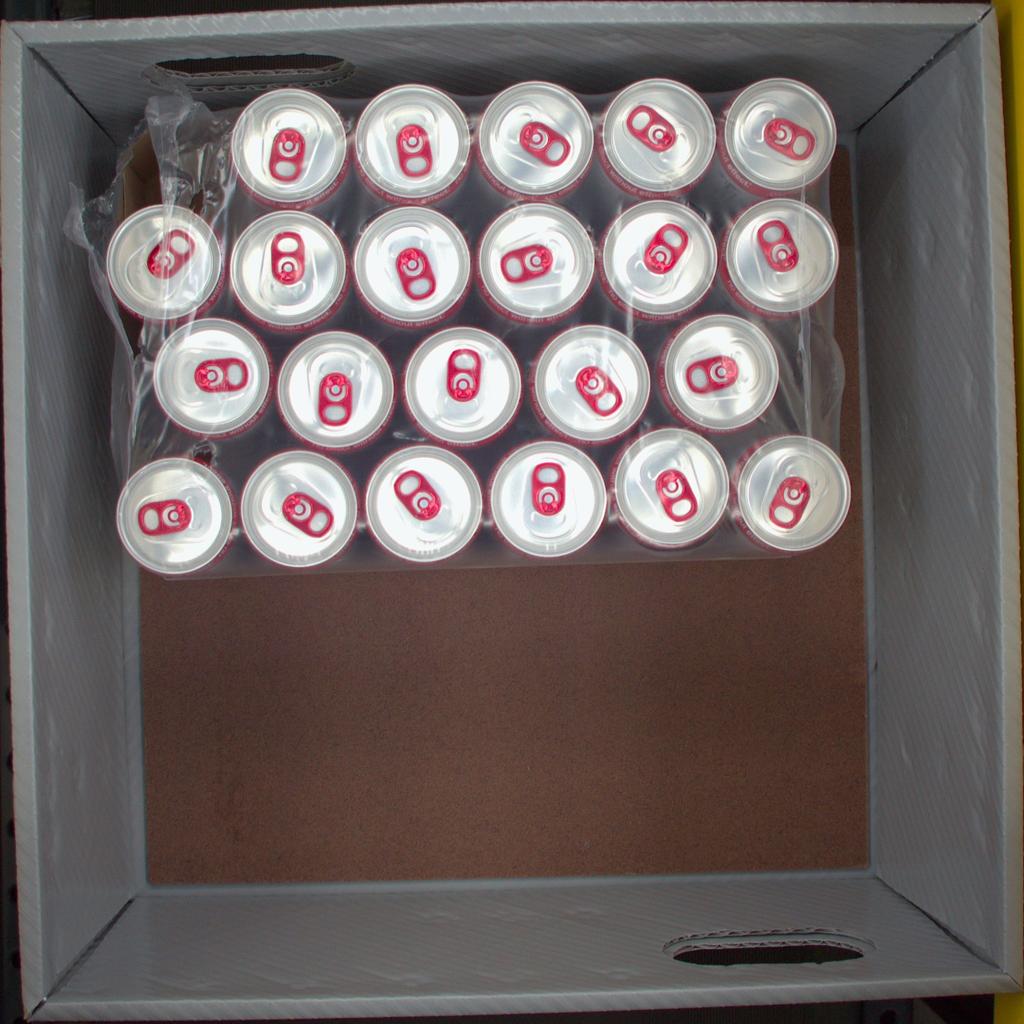} \\
            \hline
        \end{tabular}
        \caption{Overview of defect severities and defect types. Our dataset categorizes defective samples into two \emph{severity} classes: \emph{minor} (top two rows) and \emph{major} (bottom two rows). Additionally, each defective sample is assigned one or multiple defect \emph{types} (columns), which characterize the defect(s) an item exhibits in a more fine-grained manner. The figure shows two representative samples per defect type/severity combination.}
        \label{fig:damage_comparison}
    \end{minipage}
    \hfill
    \begin{minipage}[t]{0.356\textwidth}
    \centering
    \setlength{\tabcolsep}{1pt}
    \renewcommand{\arraystretch}{0.8}
    \begin{tabular}{@{}c|@{}c@{}c@{}c@{}}
        \footnotesize Query & \footnotesize Ref. 1 & \footnotesize Ref. 2 & \footnotesize Ref. 3 \\
        \hline
        \multicolumn{4}{c}{\footnotesize Sample 1} \\[-2pt]
        \includegraphics[width=0.24\textwidth]{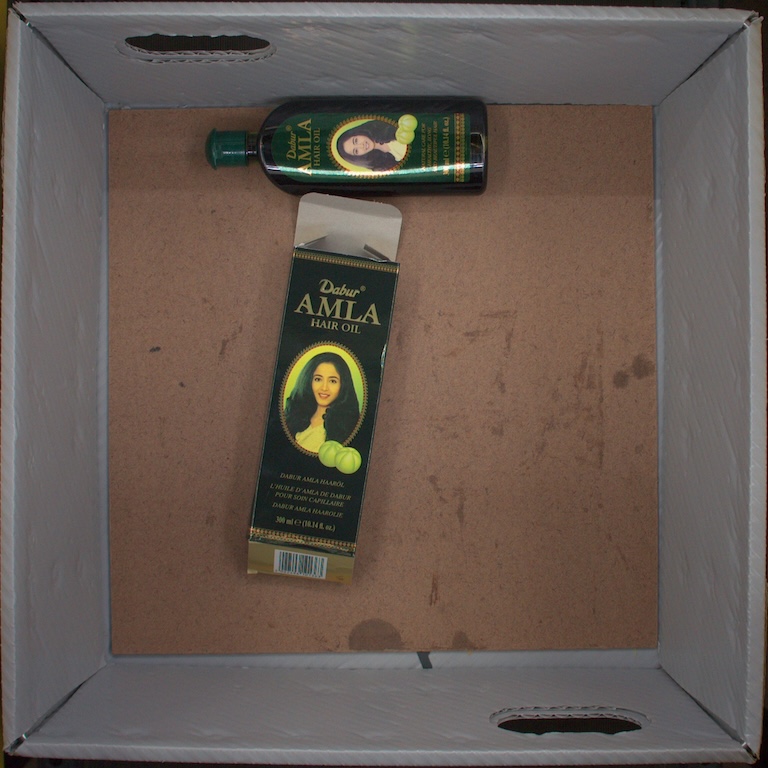} &
        \includegraphics[width=0.24\textwidth]{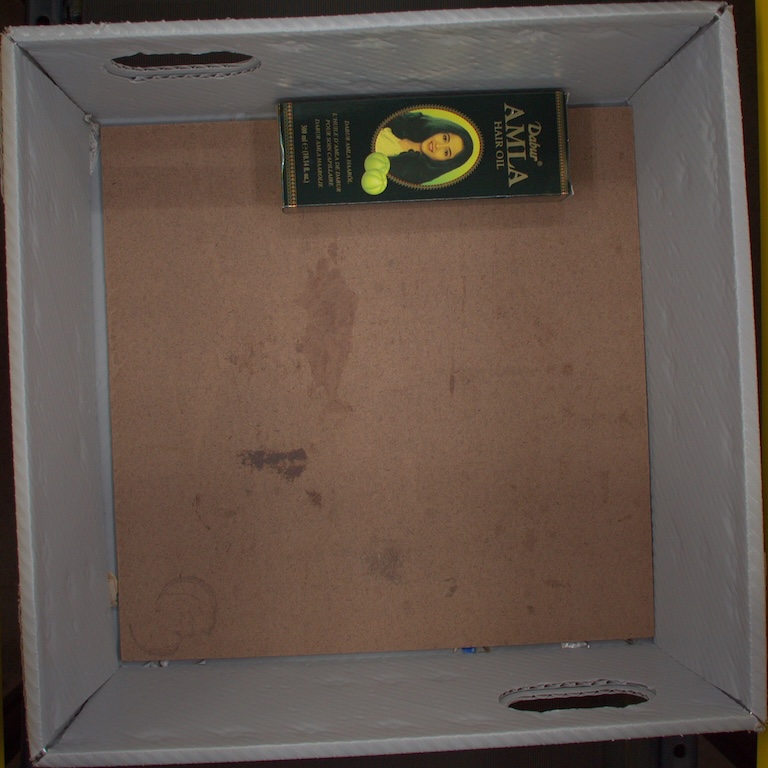} &
        \includegraphics[width=0.24\textwidth]{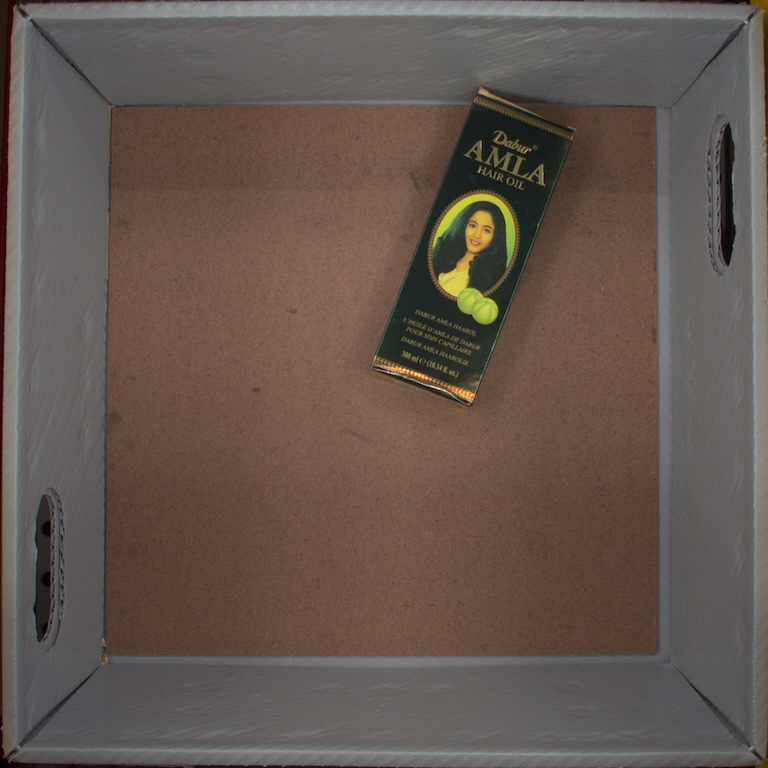} &
        \includegraphics[width=0.24\textwidth]{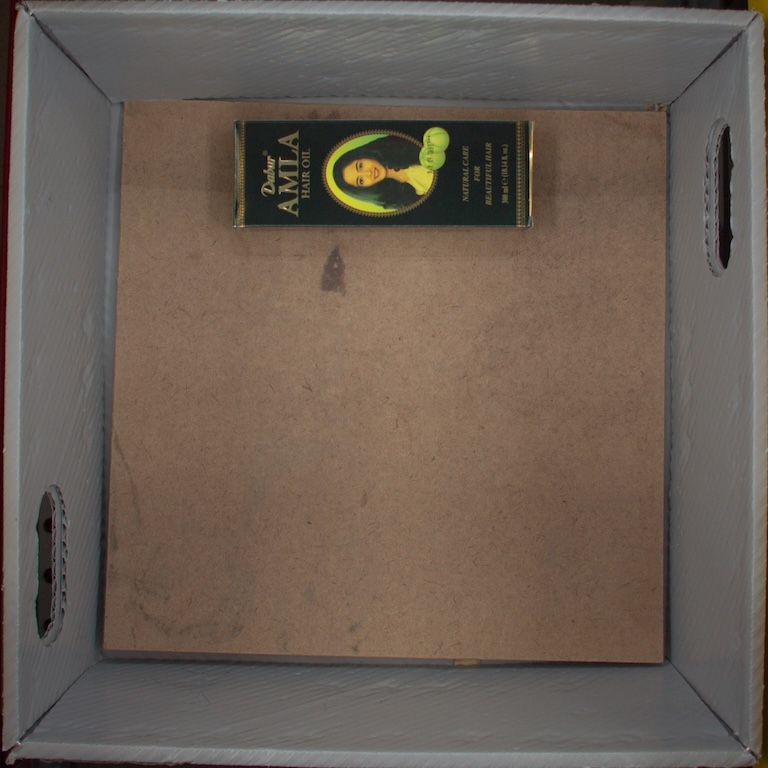}
        \\[-2pt]
        \hline
        \multicolumn{4}{c}{\footnotesize Sample 2} \\[-2pt]
        \includegraphics[width=0.24\textwidth]{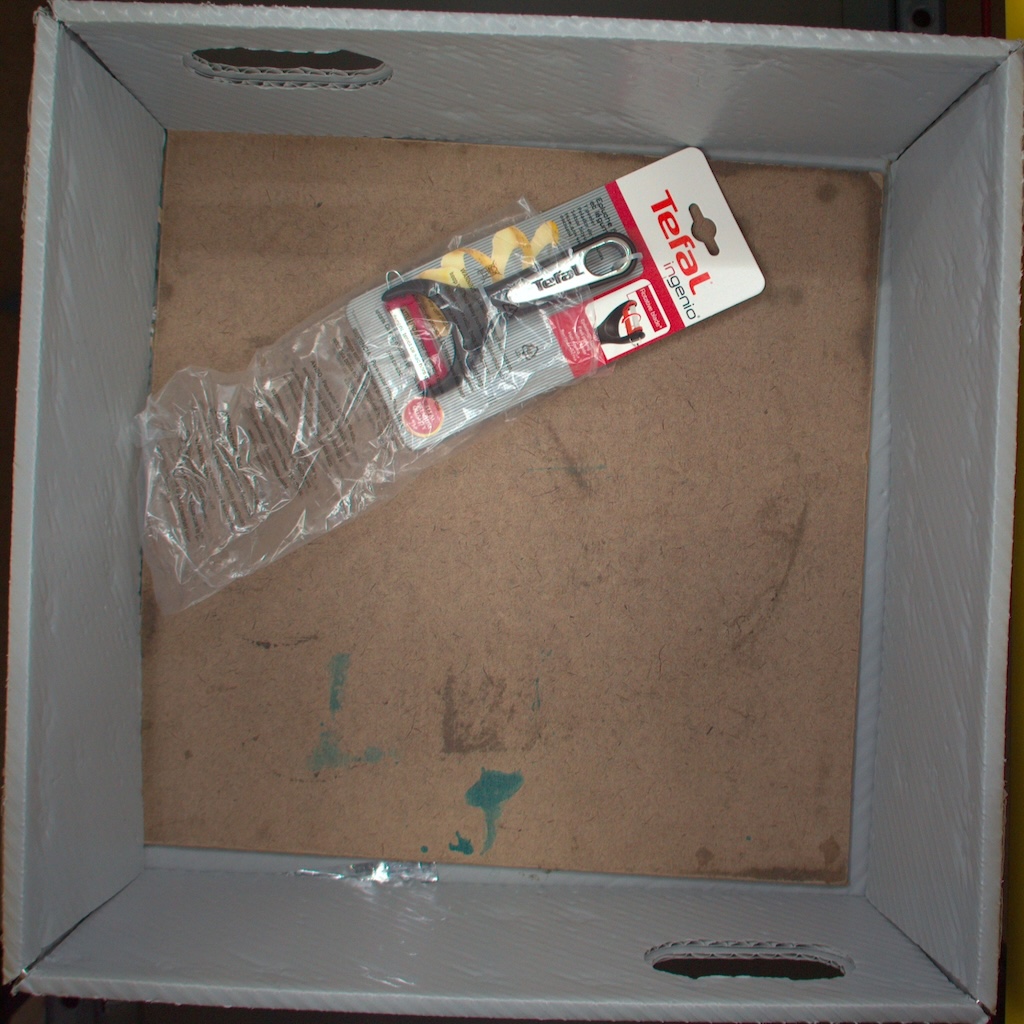} &
        \includegraphics[width=0.24\textwidth]{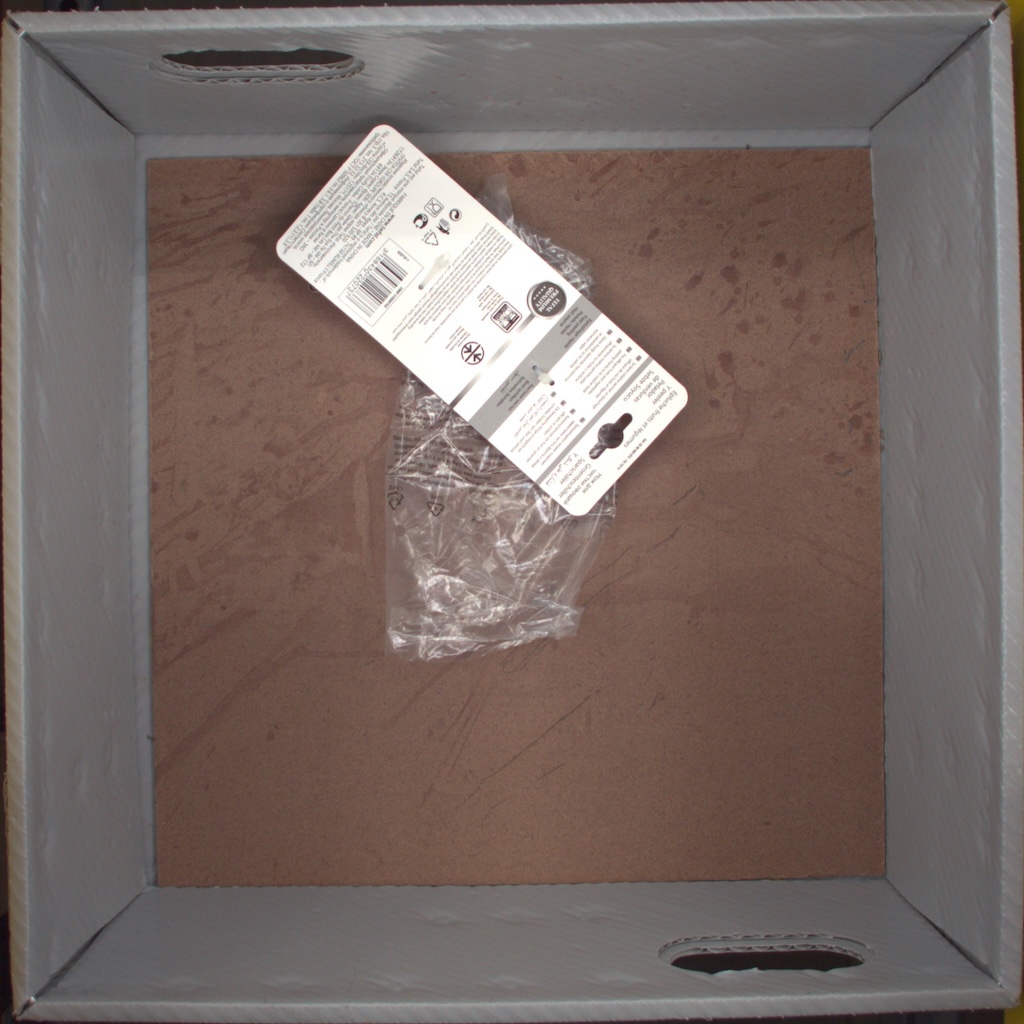} &
        \includegraphics[width=0.24\textwidth]{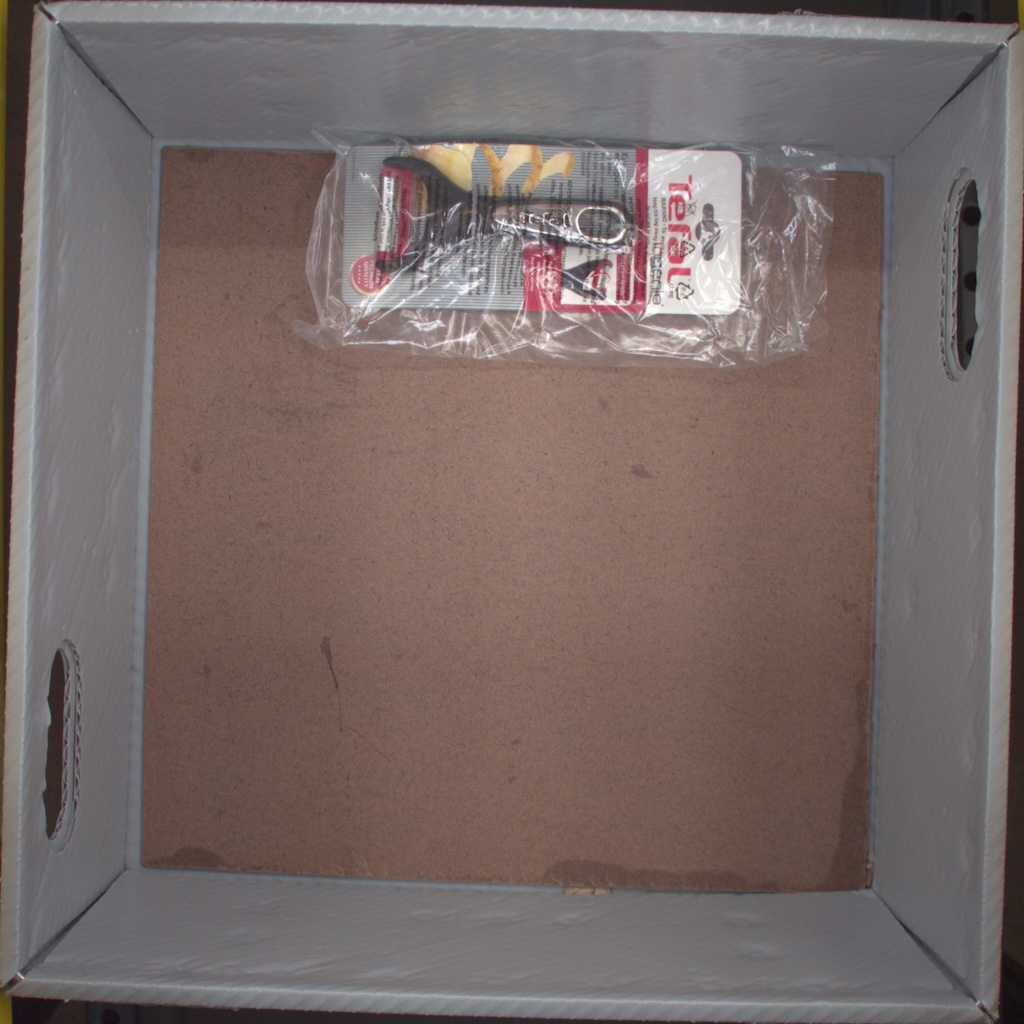} &
        \includegraphics[width=0.24\textwidth]{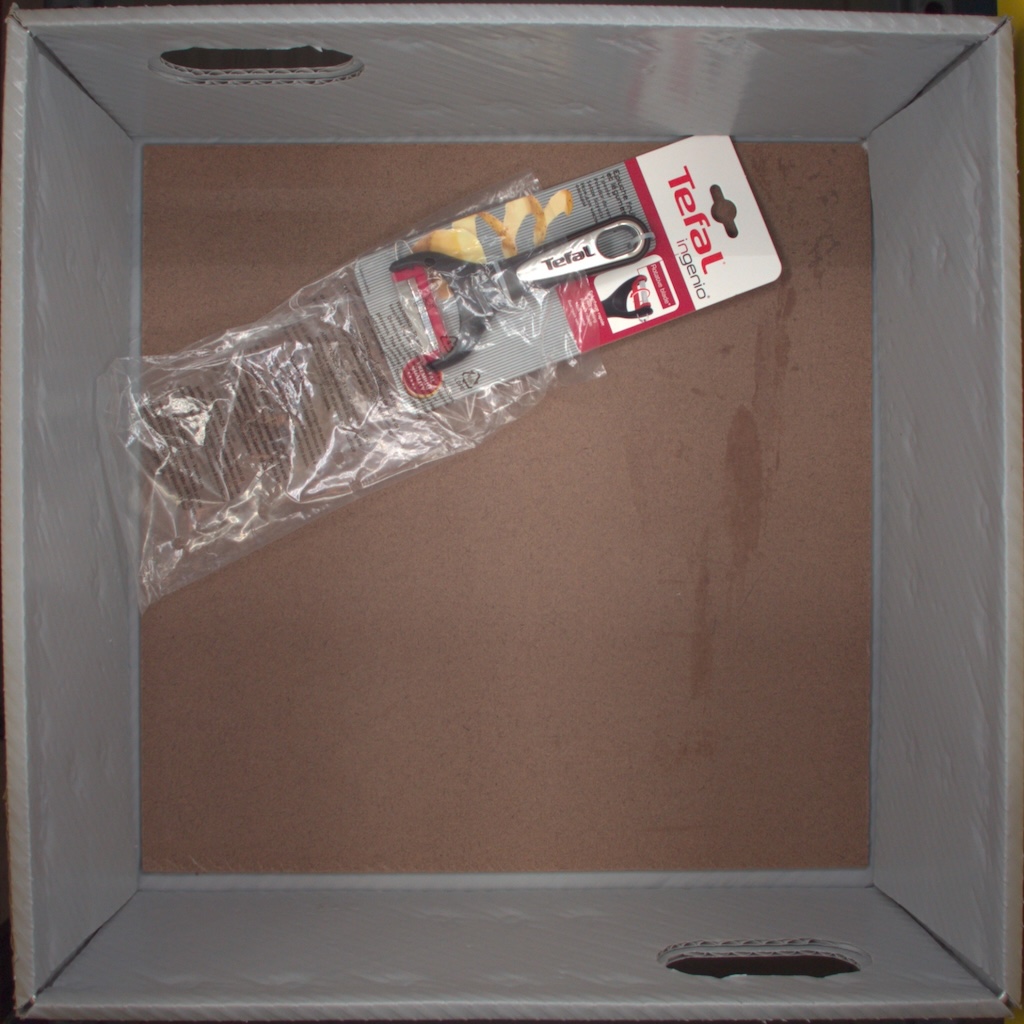}
        \\[-2pt]
        \hline

        \multicolumn{4}{c}{\footnotesize Sample 3} \\[-2pt]
        \includegraphics[width=0.24\textwidth]{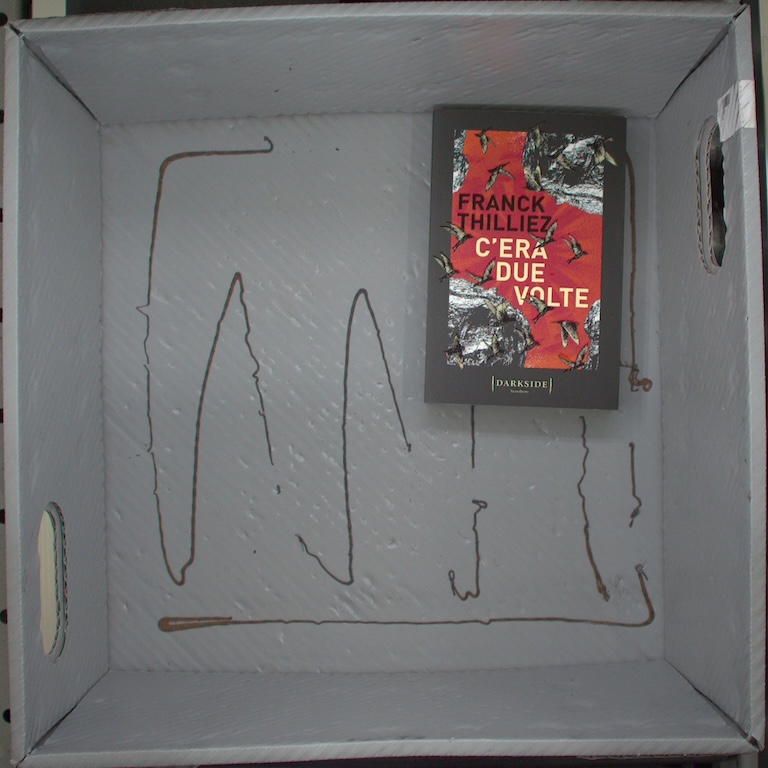} &
        \includegraphics[width=0.24\textwidth]{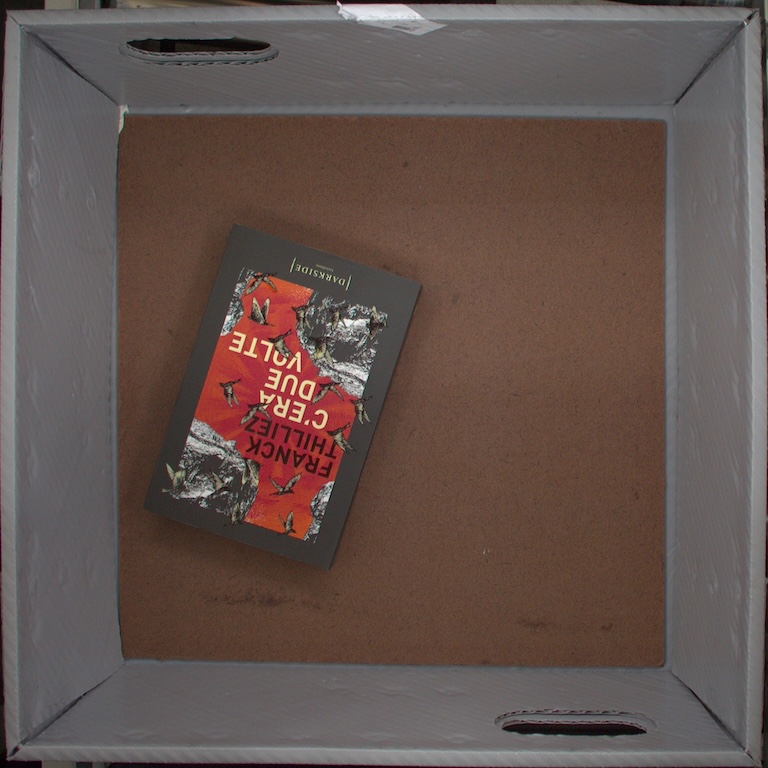} &
        \includegraphics[width=0.24\textwidth]{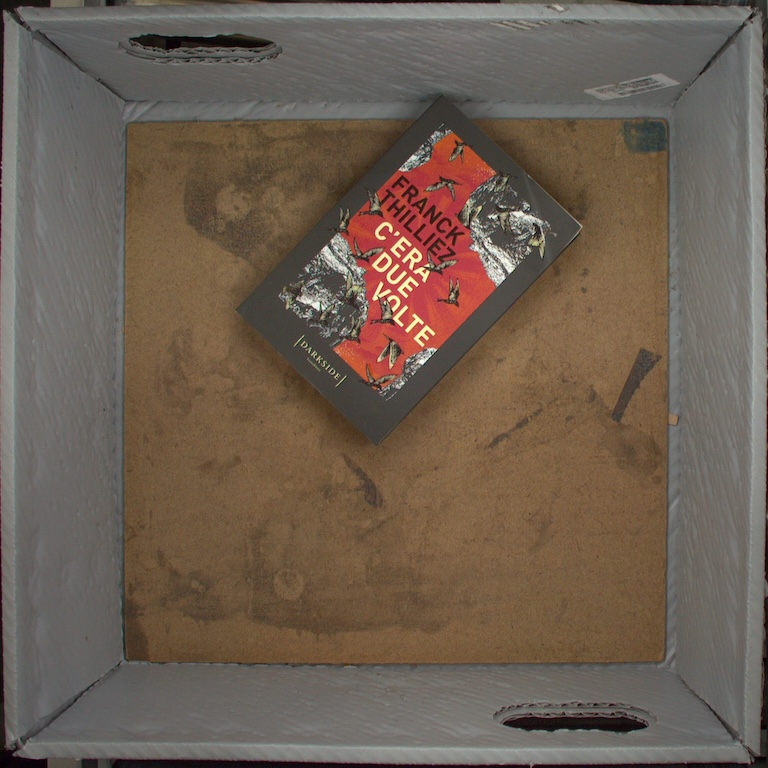} &

        \\[-2pt]
        \hline
        \multicolumn{4}{c}{\footnotesize Sample 4} \\[-2pt]
        \includegraphics[width=0.24\textwidth]{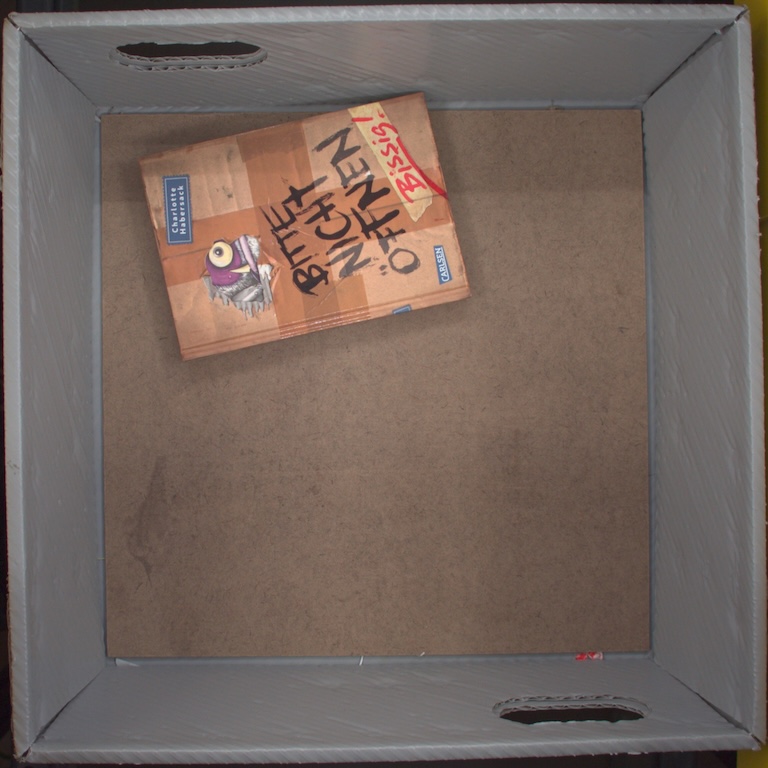} &
        \includegraphics[width=0.24\textwidth]{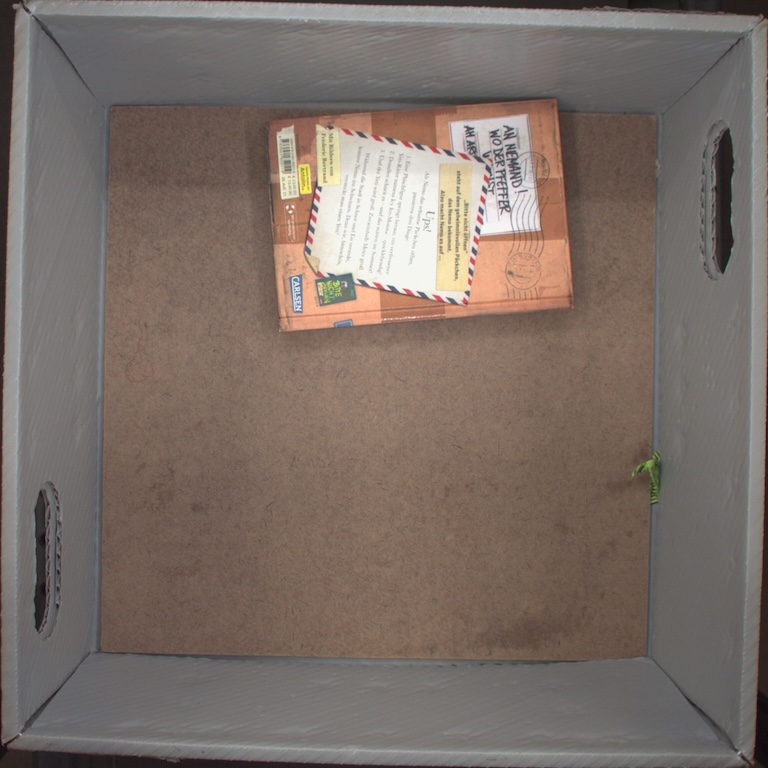} &
        \includegraphics[width=0.24\textwidth]{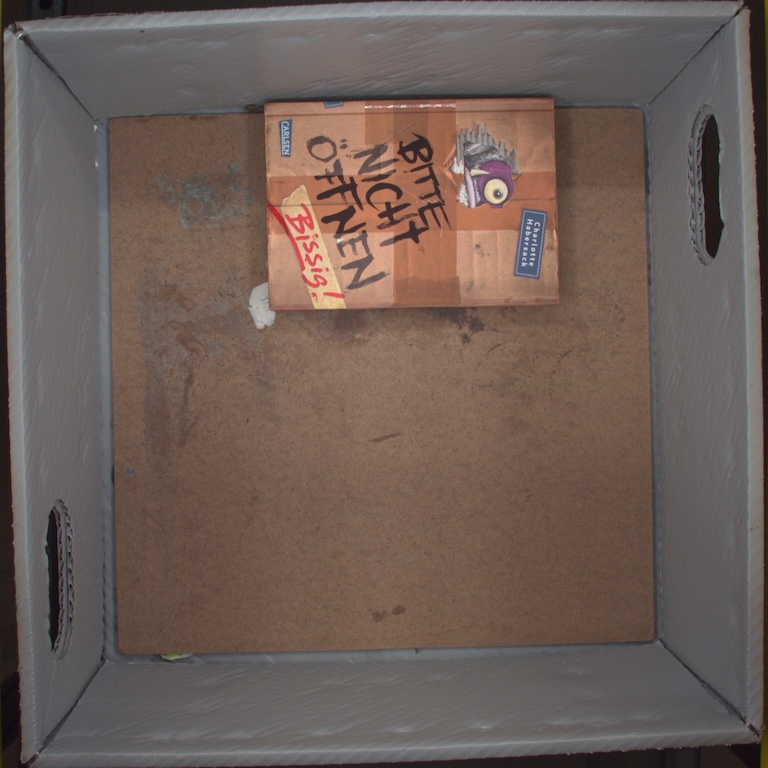} &
        \includegraphics[width=0.24\textwidth]{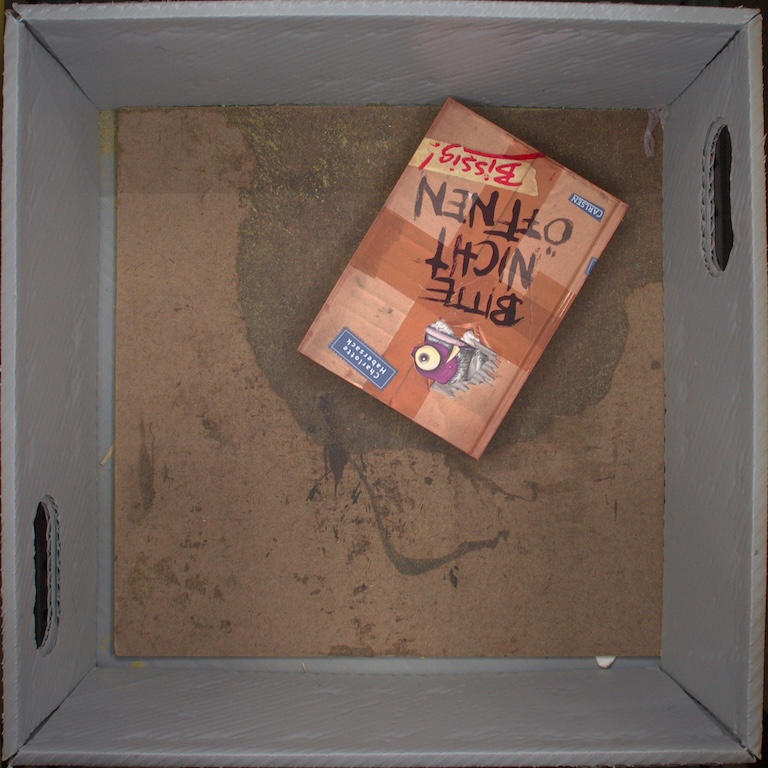} \\[-2pt]
        \hline

    \end{tabular}
    \caption{%
    Each \emph{query} image is associated with 1-3 \emph{reference} images which may exhibit significant variability: (1) Benign case. (2) Defective reference image ($<1$\% of all reference images). (3) Significant background variation, and $< 3$ reference images available. (4) Pose variability (front vs. back).}
    \label{fig:query_reference_samples}
    \end{minipage}

\end{figure*}

\section{Introduction}
\label{sec:intro}

Automated visual defect detection is critical for quality assurance in numerous industrial and logistics processes. Particularly at the scale of large retailers that handle millions of unique items, accurate detection of anomalies can significantly reduce costs, minimize waste, and enhance overall operational efficiency. However, developing robust visual defect detection systems in retail logistics applications presents significant challenges that have yet to be fully addressed by existing research.
The primary challenge stems from the diversity of items and the rarity of defects, which makes building supervised-learning datasets costly and time-consuming. This scarcity of training data leads to highly imbalanced datasets, necessitating unsupervised and anomaly-detection (AD) approaches.

State-of-the-art unsupervised and AD methods for visual defect detection achieve exceptional performance under controlled manufacturing conditions, reaching 99.9\% \cite{chen2024unified} and 99.5\% \cite{yao2024gladbetterreconstructionglobal} AUROC on MVTec-AD~\citep{bergmann_mvtec_2021} and VisA~\citep{zou2022spot} datasets, respectively. However, these methods struggle in complex logistics environments like Amazon's retail operations, where millions of diverse products flow through logistics centers. The challenges are multifaceted (Figure~\ref{fig:damage_comparison}): products range from consumables to electronics, each with distinct physical properties; defects vary from minor creases to major spillages, often with subtle manifestations that challenge even human inspectors; most items are observed only a few times, limiting both defective and non-defective sample availability; and significant pose variation occurs due to random product placement.

To enable researchers to overcome these challenges, we introduce a novel large-scale dataset for \emph{visual defect detection in retail logistics applications}. Our dataset significantly advances the field by addressing key limitations of existing benchmarks and poses the following key question: How can we build generalizable visual defect detection methods under challenging conditions such as limited instances per item, limited availability of both defective and non-defective samples per item, and significant intra-class variation?

Our key contribution is a challenging defect detection dataset with unparalleled scale and diversity of products, structured to enable the development of novel supervised, unsupervised, and hybrid approaches. The dataset comprises 238,421 images of 48,376 unique items. Items are presented in random poses and orientations, closely mirroring real-world retail logistics scenarios. The dataset is split into annotated and unannotated portions: the annotated \emph{query image} dataset contains 100,267 images, including 29,316 defective instances. For all query images, we provide qualitative defect severity and fine-grained defect type annotations, reflecting the subjectivity present in defect assessment. Additionally, each query image is associated with up to three unannotated \emph{reference images} that depict items in a ``normal`` condition (Figure~\ref{fig:query_reference_samples}). We feature diverse product categories, seven distinct defect types, and high-resolution images capturing obvious and subtle defects.

To substantiate the challenge posed by our dataset, we evaluate numerous state-of-the-art baselines. First, we demonstrate that supervised baselines~\citep{resnet2016,dosovitskiy_vit_2020} with access to a large training set of defective instances achieve up to 94.27\% AUROC. These methods achieve high performance by learning common defect pattern priors, while struggling in edge cases and ``adversarial'' items, such as items with damage-like designs (\eg a printed hole) or deformable items where creases may arise naturally without negatively impacting the product. We then demonstrate that these supervised methods fall short of yielding such performance under more realistic conditions, when only few defective samples are available for training. In such scenarios, unsupervised and anomaly detection baselines~\citep{roth2022towards,jeong2023winclip} were shown to excel~\citep{yao2024gladbetterreconstructionglobal} on related datasets like MVTec-AD~\citep{bergmann_mvtec_2021} or VisA~\citep{zou2022spot}. However, we demonstrate that these methods, as well as state-of-the-art vision-language models~\citep{claude-model-card, agrawal2024pixtral12b}, fail to surpass 56.96\% AUROC on our dataset. Qualitative analyses confirm that these methods struggle with item and pose variability as well as limited access to non-defective samples of the query item.

These results underscore the relevance of our dataset to the anomaly detection community in developing more robust and generalizable methods. By introducing this comprehensive dataset, we aim to stimulate progress in visual defect detection for retail logistics applications. We believe this unique resource will enable researchers and practitioners to develop more robust and generalizable models, capable of handling the complexities and nuances of real-world defect detection tasks.
The dataset is available for download under \url{https://www.kaputt-dataset.com}.

\section{Related Work}
\label{sec:related-work}

\begin{table*}[th!]
\begin{center}
\small
\caption{Overview of representative defect and anomaly detection datasets. Our dataset provides a unique new challenge to the defect detection field due to the amount of defective samples and intra-class variance within the dataset.}
\begin{tabular}{lp{2.5cm}p{2.5cm}p{2cm}p{3cm}p{2.3cm}}
\toprule
\textbf{Dataset} & \textbf{Labeled Samples}:\newline Total (Anomalous) & \textbf{Item}\newline \textbf{Categories} & \textbf{Unlabeled}\newline\textbf{Samples} & \textbf{Defect}\newline\textbf{Labels} & \textbf{Pose/Viewpoint}\newline\textbf{Variance} \\
\midrule
ARMBench \cite{mitash2023armbench} & 100,000+ (6,786) & N/A & - & Classes & \textbf{yes} \\
Kolektor \cite{Tabernik2019JIM} & 399 (52) & - & - & Classes & no \\
BTAD \cite{mishra21-vt-adl} & 2830 (1799) & 3 & - & Classes  & no \\

MVTec-AD \cite{bergmann_mvtec_2021} & 5,354 (1,258) & 15 & - & Classes & no \\
VisA \cite{zou2022spot} & 10,821 (1,200) & 12 & - & Classes, Segmentations  & no \\
\midrule
\textbf{Ours} & 100,267 \textbf{(29,316)} & \textbf{48,376} & 138,154 & Classes  & \textbf{yes} \\
\bottomrule
\end{tabular}
\label{tbl:datasets}
\end{center}
\end{table*}

\textbf{Defect detection applications}. Defect detection is an important and widely studied field due to its many commercial applications, including detecting defective parts in industrial manufacturing \cite{bergmann_mvtec_2021}, inspecting civil infrastructure such as bridges \cite{rubio_multi-class_2019,shi_improvement_2021}, vehicle damage \cite{zhang_vehicle-damage-detection_2020}, and medical applications \cite{kong_multi-task_2021}.  However, our use case differs from the standard industrial manufacturing applications, mainly in terms of item variation and defect variability.  While industrial applications typically focus on a single, known item or part, we are concerned with the much more open-ended problem of detecting defects for the millions of constantly changing items handled in retailers like Amazon, which may also exhibit significant intra-class variation (\eg packaging variations and random poses).  Thus, our work differs from much of the literature, in terms of data requirements and methods.

\myparagraph{Datasets}. The variety of defect detection applications has led to the development of a number of bespoke datasets in this domain \cite{han2022adbench, akcay2022anomalib}. Most relevant to our application is ARMBench~\citep{mitash2023armbench}. While targeting a similar domain and comparable in total size, ARMBench only contains one quarter of the defective samples our dataset offers, and features only two (open and deconstruction) compared to seven defect types. Strongly related are datasets targeting manufacturing defects, such as MVTec-AD\cite{bergmann_mvtec_2021} and VisA~\cite{zou2022spot}. These contain images of items with a wide variety of defects such as dents, contaminations, and structural changes. At this point however, the performance on these datasets is close to being saturated, with state-of-the-art methods achieving well over 99\% AUROC~\citep{yao2024gladbetterreconstructionglobal}. Our dataset offers one order of magnitude more data both in terms of annotations and anomalous instances, enabling researchers to leverage the dataset for developing and benchmarking various types of approaches. At the same time, significant pose variation render the dataset significantly more challenging than related ones. Table~\ref{tbl:datasets} presents a comprehensive comparison to existing defect detection datasets.

\myparagraph{Models}.
The defect detection problem has been approached in different ways, using \emph{supervised}, \emph{unsupervised}, and \emph{anomaly-detection} methods. The most straightforward approach is supervised learning, which aims to learn distinctive defect patterns given samples of both non-defective \emph{and} defective instances. For these approaches, the supervision signal takes the form of an image label \cite{rubio_multi-class_2019,shi_improvement_2021,zhang_vehicle-damage-detection_2020}, a segmentation mask, or both \cite{kong_multi-task_2021}, casting the machine learning problem as classification, segmentation or multi-task learning, respectively. A key limitation of these approaches is that they require access to a large dataset of \emph{defective} items for training, which are typically rare and difficult to collect.

This limitation motivates the use of alternative approaches that can leverage non-defective, ``normal'' samples more effectively by identifying defective instances as deviations from the expected normal appearance. While the exact distinction between the underlying paradigms is blurry, these approaches are usually categorized as \emph{unsupervised} learning, \emph{anomaly} detection, or \emph{outlier} detection. This includes methods that classify outliers directly given some representational space (\eg using one-class SVMs~\citep{ruff2018deep,yi2020patch}) or those that threshold a per-pixel or per-image patch reconstruction error~\citep{bergmann2018improving}. Similarly, deep generative approaches have been used to compute outlier statistics both on image reconstructions as well as in the learnt latent representation via Generative Adversarial Networks~\citep{akcay2019ganomaly,schlegl2019f}, diffusion models~\citep{yao2024gladbetterreconstructionglobal}, or pre-trained Vision Transformers~\citep{jeong2023winclip}. Exemplar-based methods compute outliers directly by constructing a more targeted reference dataset on the fly, via a nearest-neighbour approach~\citep{roth2022towards} and then computing image/patch-level feature distances relative to this set.

Such approaches are well suited to industrial applications, where examples of anomaly-free items in an identical, nominal pose are plentiful. However, they are prone to false positives predictions by flagging non-defect-related image variations as anomalous. As our experiments demonstrate, this makes current approaches impractical for real-world retail logistics applications where we are faced with significant intra-class variation, \eg due to differing poses or packaging.
More recently, \citet{jiang2025mmad} investigated whether this limitation could be addressed by leveraging the inherent visual understanding capabilities of Multimodal Large Language Models (MLLMs). Their findings, however, demonstrate that current MLLMs' performance falls short of industrial requirements: while excelling at \emph{object} analysis and description tasks, these models lack robust \emph{anomaly} detection capabilities. Our experiments using both commercial and open-source MLLMs~\citep{claude-model-card, agrawal2024pixtral12b} corroborate these findings.

\section{Dataset}
\label{sec:dataset}

Our dataset consists of top-down RGB images of retail items, each accompanied by categorical labels and segmentation masks. In the following sections, we detail the dataset's structure, collection, and annotation methodology.

\subsection{Dataset Structure}
\label{subsec:dataset-structure}

The dataset is organized into \emph{query} and \emph{reference} sets:

\begin{enumerate}
    \item \emph{Query dataset}: Contains image captures with associated:
    \begin{enumerate}
        \item \emph{Item identifier} (unique per item)
        \item \emph{Defect severity} (no defect, minor, major)
        \item \emph{Defect type(s)} for defective items (\eg, penetration, spillage)
        \item \emph{Item material} (\eg cardboard, plastic, books)
        \item \emph{Item segmentation mask}
    \end{enumerate}
    \item \emph{Reference dataset}: Contains 1-3 image captures per item identifier, primarily non-defective but not guaranteed.
    \begin{enumerate}
        \item \emph{Item identifier} (unique per item)
        \item \emph{Item segmentation mask}
    \end{enumerate}
\end{enumerate}

The dataset is further divided into training, validation, and test splits, each consisting of a unique query/reference set pair.  To test model generalization capabilities and prevent overfitting to specific items, we ensure that each item only appears exclusively in one of the splits, i.e. identifiers do not overlap between splits.

\subsection{Images}
\label{subsec:images}

For image capture, we use a data collection station equipped with a 12~MP RGB camera with an f/12mm lens. The camera is positioned top-down to capture the singulated item located inside a logistics container ("tray"). To provide uniform diffuse illumination while minimizing reflections commonly induced by plastic materials, we enclose the station with side walls and ensure constant lighting using LED panels. We provide a schematic drawing of the data collection setup in the Supplementary Material (Section~\ref{sec:dataset_details}).

We further post-process the acquired images by applying a square crop that includes only the tray, and resize the images to $2048 \times 2048$px. We also provide item segmentation masks/crops (Section~\ref{subsec:segmentation-masks}), but retain the full tray images as item boundaries are not always clearly defined due to dangling or protruding parts, and certain defects may only be visible on the tray surface (\eg, liquid spillages; see Figure~\ref{fig:damage_comparison}, examples 1 and 3, respectively).

\subsection{Data Collection}
\label{subsec:collection}

A major challenge in creating defect detection datasets is the rarity of defect events, making the acquisition of positive (defective) samples extremely time-consuming. We address this through a two-stage collection strategy: First, we collect items flagged as defective by human operators for annotation. Second, we implement an iterative mining process where a binary classifier, trained on previously annotated images (Section~\ref{subsec:annotation}), identifies potential defect candidates for further annotation.

The resulting initial query dataset of defective and non-defective images undergoes further curation based on the following criteria:
(1) \emph{Quality control} through manual filtering of low-quality images, particularly those with missing or off-center items. (2) \emph{Diverse item range} with maximum 15 samples per item to ensure variety. (3) \emph{Balanced defect rate} (28.6\%) that aligns with existing benchmarks like MVTec-AD~\cite{bergmann_mvtec_2021} and VisA~\cite{zou2022spot} since defective samples are more valuable for training and evaluation than non-defective samples. (4) \emph{Exclusion} of items lacking non-defective samples to prevent model overfitting.

The curated query dataset is split by item identifier into training (85\%), test (10\%), and validation (5\%) sets, each supplemented with up to three reference images from a separate unlabeled dataset. We remove missing or off-center reference images but exclude defect type and severity labels. This approach, including the limit of three reference images per sample, reflects real-world retail conditions where most items sell infrequently and creating a perfect reference database is impractical. Consequently, some limited amount of reference images may exhibit different packaging or contain defects.

\subsection{Annotation}
\label{subsec:annotation}

Next, we describe the labels and annotation process.

\subsubsection{Item Segmentation Masks}
\label{subsec:segmentation-masks}

The images depict the item inside the full tray, but some baseline methods may require item crops to perform optimally. We thus generate item segmentation masks using a U-Net~\cite{ronneberger_u-net_2015} model trained on 17,000 manually annotated masks, and create \emph{square} item-crops with 10\% padding. The generated masks and item-crop images are released as part of the dataset. Moreover, we evaluate the baselines on both full and item-cropped images.

\begin{figure}
    \centering
        \includegraphics[width=0.15\textwidth]{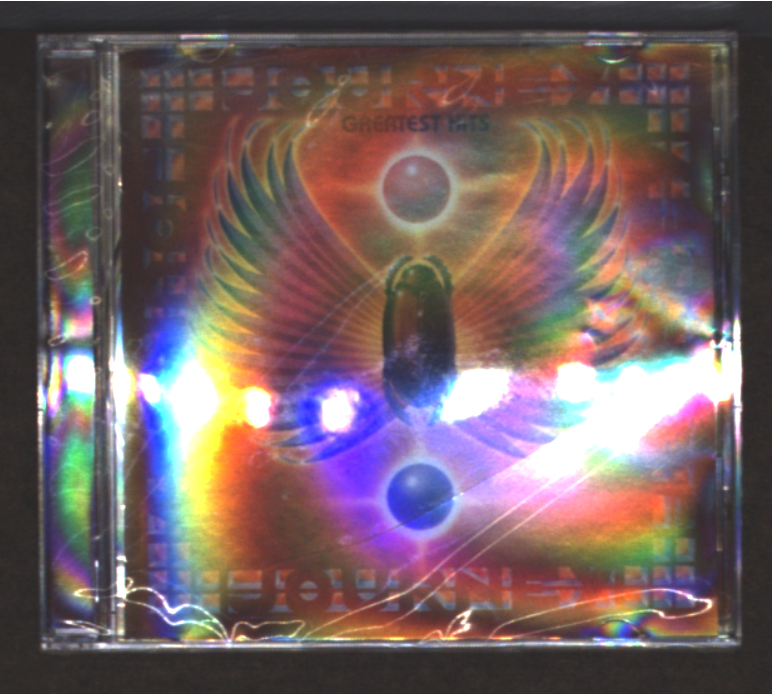}
        \hfill
        \includegraphics[width=0.134\textwidth]{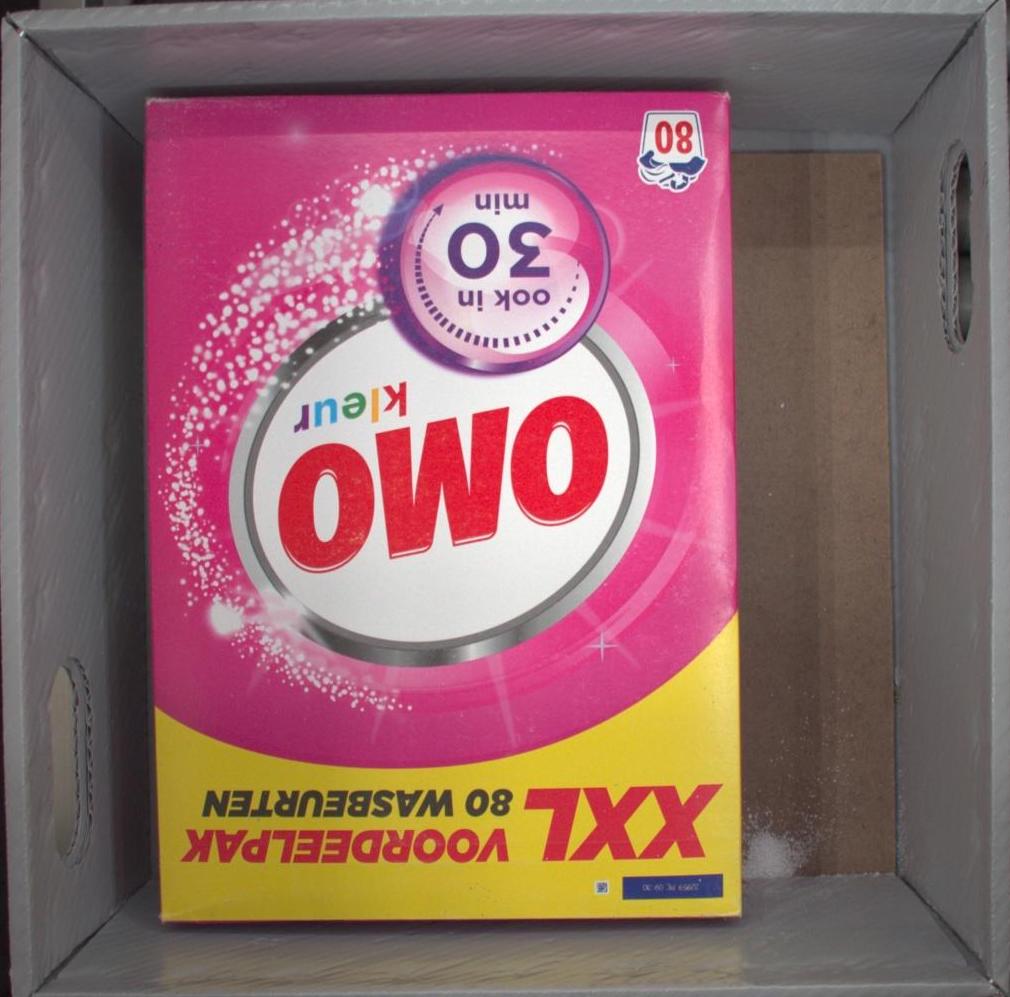}
        \hfill
        \includegraphics[width=0.12\textwidth]{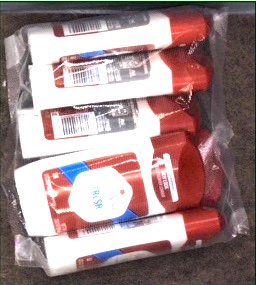}
    \caption{
 Examples for challenging defective cases (from left to right). (1) Unobservable cases. A small stripe in the bottom half of the CD could be either a reflection or a crack in the cover. (2) Complex cases. The detergent pack looks intact, but at a second look the  powder on the tray next to it item indicates a spillage defect. (3) Ambiguous cases. The multi-pack is complete but its units are unordered, which is acceptable but has different visual appearance than the corresponding reference image.
\vspace*{-0.5cm}
}
    \label{fig:label-noise-samples}
\end{figure}

\begin{figure*}[th!]
    \centering
       \includegraphics[trim = 1mm 1mm 1mm 1mm, clip, width=0.495\textwidth]{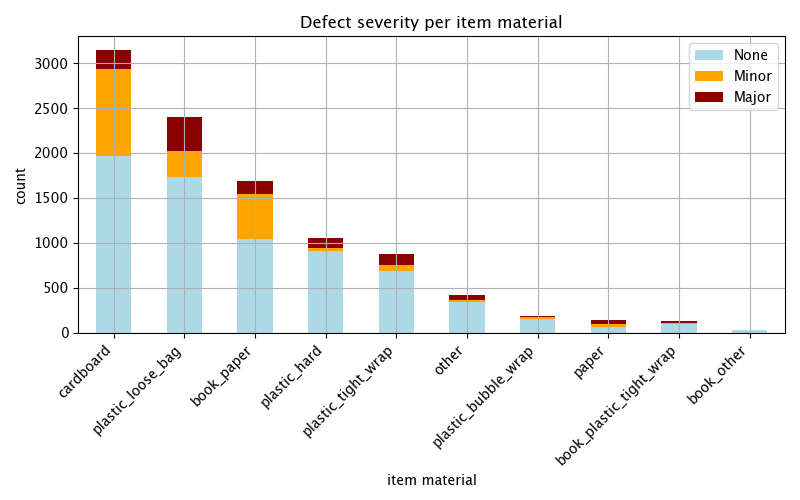}
        \hfill
       \includegraphics[trim = 1mm 1mm 1mm 1mm, clip, width=0.495\textwidth]{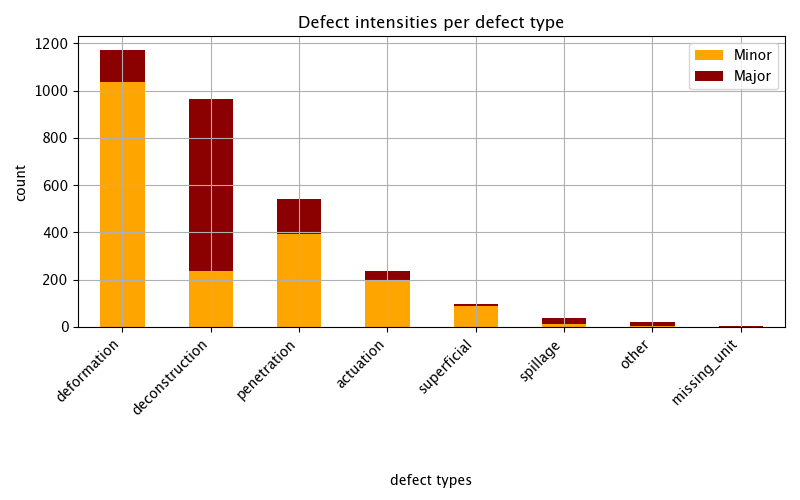}
    \caption{\textbf{Left}: Distribution of item material types and defect severities. We observe that items with cardboard material dominate the dataset, followed by plastic bags/cases and books.
    \textbf{Right}: Distribution of defect types per defect severity. We find that \emph{deformation} is the most common defect type, however, it mostly results in minor defect severity, similar to \emph{penetration}, \emph{actuation} and \emph{superficial}. In contrast, \emph{deconstruction} and \emph{spillage} commonly result in major defect severity.
    \label{fig:item-type-damage-type-distributions}
    }
\end{figure*}

\subsubsection{Categorical Labels}
\label{subsec:damage-taxonomy}

Both the query and reference datasets are curated to avoid low-quality images, as described above. Additionally, each sample in the query datasets is manually annotated with the following categorical labels: \emph{defect severity}, \emph{defect type}, and \emph{item material}.

\myparagraph{Defect severity.} Each sample is annotated with a defect severity label: \emph{no defect}, \emph{minor}, or \emph{major}. Major defect compromises the item's integrity (\eg significant crush or puncture) or risks doing so (\eg fully opened box lid). Minor defect renders the item not pristine but potentially acceptable (\eg small dents on cardboard packaging). Acknowledging the subtle boundaries between these categories, our benchmark (\Sec~\ref{sec:benchmark}) focuses on detecting \emph{any} defect (minor and major), with ablation studies model performance on major defect detection.

\myparagraph{Defect type}. For each defective sample exhibiting at least minor defect severity, we annotate one or more \emph{defect types}: \emph{penetration} (\eg holes, tears, cuts), \emph{deformation} (\eg dents, crushes), \emph{actuation} (\eg open box/bag/book), \emph{deconstruction}, \emph{spillage} (liquid, powder, \etc), \emph{superficial} (\eg dirt, scratches), \emph{missing unit}. Assigning multiple defect types per item is explicitly permitted, as items may incur multiple defects at the same or different spatial locations.

\myparagraph{Item material}. Each item is categorized according to its primary outer material: \emph{cardboard}, \emph{plastic (loose bag)}, \emph{plastic (hard)}, \emph{plastic (bubble wrap)}, \emph{plastic (tight wrap)}, \emph{paper}, \emph{book (paper)}, \emph{book (plastic)}, \emph{other}. The distribution per item material is shown in Figure~\ref{fig:item-type-damage-type-distributions} (left), indicating higher volumes of cardboard and plastic packaging, followed by books.

Each sample is independently labeled by three annotators. We then aggregate annotations through majority voting. During our baseline evaluation experiments, we found some wrongly labeled samples, which we manually corrected. We observe that annotation errors primarily arise from the following issues, exemplified in Figure~\ref{fig:label-noise-samples}: (1) \emph{unobservable} cases where defects cannot be detected due to sensing limitations or lack of non-defective reference images; (2) \emph{complex} cases where defects are present but so subtle that they get overlooked by annotators; (3) \emph{ambiguous} cases where an anomaly is visible but it is not clear whether it qualifies as a defect. We acknowledge that, despite best efforts, the dataset may still contain mislabeled samples, but demonstrate that these do not negatively affect the evaluated baselines (see Supplementary Material).

We visualize the resulting distribution of defect severities and types in Figure~\ref{fig:item-type-damage-type-distributions} (right)\footnote{The figure presents a slightly simplified version of the defect type distribution, as we only assume one single defect type per sample, which holds true for 72\% of all defective samples in the dataset. For samples with multiple defect types, we select one based on a predefined priority list, where more severe defects (such as spillage) take precedence over less severe ones (such as actuation).}.

\section{Benchmark}
\label{sec:benchmark}

To demonstrate the challenge posed by our dataset and establish baseline performance, we evaluate various state-of-the-art models. We define four distinct evaluation scenarios, based on whether an approach uses the training data, the reference images, none or both. We choose methods in such a way to cover a wide variety of relevant approaches, favoring established and widely adopted methods over their latest variants.

\subsection{Evaluation Metrics}
\label{subsec:metrics}
To compare the different baselines, we formulate the task as a binary classification problem between \emph{no defect} and \emph{any defect} (\ie defect severity minor or major). Formally, we evaluate a classifier
$f(
    \textbf{x}_\textrm{q}^\textrm{ID},
    \{
        \textbf{x}_\textrm{ref}^{\textrm{ID}(1)},
        \ldots,
        \textbf{x}_\textrm{ref}^{\textrm{ID}(G_\textrm{ID})}
    \}
)
= \hat{y}$
given a (labeled) test query image $(\textbf{x}_\textrm{q}^\textrm{ID}, y) \in \mathcal{D}_\textrm{q}^\textrm{test}$ with item identifier ID and binary defect label $y \in \{0, 1\}$ (0 = non-defective, 1 = defective), and a set of $G_\textrm{ID}$ corresponding (unlabeled) test reference images with the same item identifier from $\mathcal{D}_\textrm{ref}^\textrm{test}$. Note that both the number of query and reference images per item identifier varies, and also that models (\eg the methods listed in Section~\ref{subsec:with-training-no-reference}) may ignore reference images altogether.

Apart from the query and reference test sets, our dataset also comprises equivalent subset pairs for model training $\mathcal{D}_\textrm{q}^\textrm{train}$, $\mathcal{D}_\textrm{ref}^\textrm{train}$ and validation $\mathcal{D}_\textrm{q}^\textrm{valid}$, $\mathcal{D}_\textrm{ref}^\textrm{valid}$. The training datasets are used by the supervised and combined methods, while the validation datasets are used for hyperparameter tuning and decision threshold selection.

To compare different models $f$, we use Average Precision (AP) on \textbf{any} (minor or major) defect (\textbf{AP\textsubscript{any}}), as our key performance metric, and additional auxiliary metrics:
\begin{itemize}
    \item Average Precision (AP) on major defect (\textbf{AP\textsubscript{major}}), computed only on the subset of either non-defective or items with major defects,
    \item Area under Receiver Operator Characteristic (\textbf{AUROC}),
    \item Recall at 50\% Precision (\textbf{R@50\%P}), and
    \item Recall at 1\% False Positive Rate (\textbf{R@1\%FPR}).
\end{itemize}
Some methods perform better on full images and others on item-cropped images. For the sake of brevity, we report numbers only for the best-performing variant of each method, indicating what type of image is used in Table~\ref{tab:num_results}.

\subsection{Scenarios}
\label{subsec:scenarios}

\begin{table*}[!ht]
\centering
\begin{tabular}{rcccccc}
    \textbf{Baseline} & \textbf{Tray/Item} & \textbf{AP\textsubscript{any}} [\%] & \textbf{AP\textsubscript{major}} [\%] & \textbf{AUROC} & \textbf{R@50\%P} [\%] & \textbf{R@1\%FPR} [\%] \\ \hline \hline 
\texttt{\textit{Random}} & - & 31.84 & 14.00 & 50.00 & 0.00 & 1.08 \\
\hline
\multicolumn{7}{c}{\textbf{No training, no references} (zero-shot, few-shot)}\\ 
\hline
\texttt{CLIP} & item & 36.20 & 17.15 & 56.05 & \textbf{0.56} & \textbf{1.53} \\
\texttt{POMP} & item  & 32.98 & 18.17 & 50.44 & 0.00 & 1.28 \\
\texttt{WinCLIP-zero} & item & 33.87 & 19.11 & 52.30 & 0.03 & 1.37 \\ 
\texttt{Claude-icl} & tray & \textbf{36.57} & \textbf{24.76} & \textbf{56.96} & 0.00 & 0.31 \\
\texttt{Pixtral-zero} & tray & 32.75 & 16.42 & 50.93 & 0.00 & 0.81 \\
\texttt{Pixtral-icl} & tray & 32.18 & 15.83 & 50.86 & 0.00 & 0.69 \\
\hline
\multicolumn{7}{c}{\textbf{No training, \emph{with} references} (few-shot, non-parametric, in-context learning)} \\ 
\hline
\texttt{PatchCore50} & item & \textbf{35.86} & 17.80 & \textbf{54.69} & \textbf{2.46} & \textbf{2.18} \\ 
\texttt{WinCLIP-few} & item & 34.05 & \textbf{19.29} & 52.41 & 0.66 & 1.56 \\
\hline
\multicolumn{7}{c}{\textbf{\emph{With} training, no references} (supervised/instruction fine-tuning)}\\ 
\hline
\texttt{ResNet50} & tray & 81.06 & 74.93 & 88.36 & 91.98 & 30.01 \\ 
\texttt{ViT-S}  & tray  & \textbf{90.67} & \textbf{91.45} & \textbf{94.27} & \textbf{97.69} & \textbf{59.36} \\
\texttt{Pixtral-ft} & tray & 33.43 & 17.19 & 51.44 & 3.62 & 3.62 \\ 
\texttt{AutoGluonMM}    & item & 87.77 & 86.10 & 92.47 & 96.76 & 46.26 \\ 
\hline
\multicolumn{7}{c}{\textbf{\emph{With} training, \emph{with} references} (supervised with references, non-parametric with fine-tuning)}\\ 
\hline
\texttt{PatchCore50-ft} & item & 40.18 & 20.98 & 60.14 & 6.52 & 2.37 \\
\texttt{AutoGluonMM-ref}    & item & \textbf{71.21} & \textbf{61.45} & \textbf{84.29} & \textbf{89.83} & \textbf{13.32} \\ 
\hline \hline
\end{tabular}
\caption{Results of the evaluated baseline methods on the test set split with 10067 total samples (minor defect: 2089, major defect: 1117).
}
\label{tab:num_results}
\end{table*}

We present four distinct evaluation scenarios that each explore unique aspects of our dataset.

(1) \emph{No training and no reference images}. Such approaches are commonly referred to as zero shot, leveraging strong general-purpose vision-language models, here
CLIP~\citep{pmlr-v139-radford21a}, Claude~\citep{claude-model-card}, and Pixtral~\citep{agrawal2024pixtral12b}.

(2) \emph{No training and with reference images}. Here, models have no access to the labeled training set but they leverage (unlabeled) reference images at test time.
In the anomaly detection (AD) context, this approach is commonly referred to as few-shot AD~\citep{yi2020patch} or few-normal-shot AD~\citep{jeong2023winclip}.

(3) \emph{With training and without reference images}. These are purely supervised methods~\citep{resnet2016,dosovitskiy_vit_2020} that leverage the annotated training data to train a binary classifier, ignoring the reference datasets.

(4) \emph{With training and with reference images}. These methods combine approaches (2) and (3) by both training a model (backbone) and leveraging reference images.

Next, we detail the methods we evaluate as part of each of the four scenarios. All model and training details required to replicate the results are provided in the Supplementary Material~\ref{sec:model_training_details}.
Note that some methods can be applied in multiple scenarios, as we point out in the following.

\subsubsection{No Training and No Reference Images}
\label{subsec:no-training-no-reference}

In this scenario, we test whether and to which extent strong general-purpose image understanding capabilities translate to zero-shot defect detection performance.

\myparagraph{CLIP.} We test the vanilla CLIP model~\cite{pmlr-v139-radford21a} (\texttt{CLIP}), and CLIP with fine-tuned prompts~\citep{ren2023prompt} (\texttt{POMP}). For vanilla CLIP, we perform manual prompt optimization on the validation set, and ended up with the following prompts for classification: \texttt{Image of an item without problems} and \texttt{Image of an item with problems}, for non-defective and defective samples, respectively. For POMP, we use the labels \texttt{undamaged} and \texttt{damaged} for the respective classes.

\myparagraph{WinCLIP.} WinCLIP \cite{jeong2023winclip} extends the original CLIP model for anomaly detection, by (1) providing a diverse set of text prompts representing defective and healthy samples, and (2) using multi-scale image feature extraction and comparison. In the zero-shot setting, WinCLIP only uses query image and text prompts (\texttt{WinCLIP-zero}).

\myparagraph{Claude.} We evaluate Anthropic's Claude 3.5 Sonnet~\citep{claude-model-card}, a public Vision-Language Model (VLM), in the zero-shot setting in two ways.
\texttt{Claude-zero} evaluates the model when provided with a text prompt and the query image, and ask the model to inspect the image for defects using Chain-of-Thought and finally grade the defect severity on a scale from 0 to 10. We tested different prompts on a small held-out set, and apply the best-performing prompt to the entire dataset (see Supplementary Material \ref{sec:vlm_prompts}). In the second setting (\texttt{Claude-icl}), we apply the few-shot in-context learning (ICL) scenario, where we additionally provide five samples as positive \emph{defective} classes with respective example answers. Due to computational constraints, prompt images are rescaled to $512 \times 512$.

\myparagraph{Pixtral.} In addition to Claude, we evaluate the recent open-source Pixtral-12B model \cite{agrawal2024pixtral12b} as another VLM on our dataset. Similarly, we evaluate both a zero-shot (\texttt{Pixtral-zero}), and an in-context learning (\texttt{Pixtral-icl}) setting.
The best-performing prompts can be found in the Supplementary Material \ref{sec:vlm_prompts}.

\subsubsection{No Training and With Reference Images}
\label{subsec:no-training-with-reference}

We evaluate two AD approaches that leverage reference images of known item categories at test time.

\myparagraph{PatchCore}~\cite{roth2022towards} is a state-of-the-art anomaly detection method that leverages patch-level features from an image, comparing each patch's feature to a memory bank of normal patches and identifying anomalous samples through patch-level feature distance. The image-level anomalous score is computed as the maximum of the patch-level anomaly scores across all patches in the image. We use the reference test set to create the individual memory banks of features for different items and to compute the anomaly score. As a backbone for feature extraction, we test ResNet50 pretrained on ImageNet (\texttt{PatchCore50}).

\myparagraph{WinCLIP.} We test WinCLIP in the few-shot setting, by enabling it to perform visual feature comparison with reference images (\texttt{WinCLIP-few}).

\subsubsection{With Training and No Reference Images}
\label{subsec:with-training-no-reference}

Here, we focus on common supervised methods that leverage the annotated training data to learn how to recognize the appearance of visual defects. To do so, we fine-tune two common types image backbones for defect classification using a binary cross entropy (BCE) loss, feeding only the query images as input.

\myparagraph{Convolutional Networks}. We fine-tune a ResNet50 model~\citep{resnet2016} backbone, pretrained on ImageNet \cite{imagenet2009}, on our training data  (\texttt{ResNet50}).
Preliminary experiments demonstrated that a high resolution is necessary to prevent subtle or small defects from being obscured or lost due to downsampling, and we thus use a resolution of $1024 \times 1024$ pixels. Training is conducted for 20 epochs with an initial learning rate of $5 \times 10^{-5}$ and batch size 48.

\myparagraph{Vision Transformers}. We fine-tune a ViT-small pretrained on DINOv2 \cite{oquab2024dinov} with patch size $14 \times 14$ px \cite{dosovitskiy2021imageworth16x16words} at $1024 \times 1024$ px for 30 epochs, with an initial learning rate of $5 \times 10^{-6}$ and batch size 8 (\texttt{ViT-S}). Additionally, we test an AutoML approach using the AutoGluon MultiModal framework \cite{tang2024autogluon} (\texttt{AutoGluonMM}).

\myparagraph{Pixtral fine-tuned}. In addition to the zero-shot and few-shot variants of Pixtral, we \emph{instruct-finetuned} the model on question-answer pairs from our dataset in order to adapt the model to our domain (\texttt{Pixtral-ft}). We run LoRA fine-tuning for one epoch on 10,000 samples from the training set with a fixed learning rate of $3 \times 10^{-5}$.

\subsubsection{With Training and With Reference Images}
\label{subsec:with-training-with-reference}

Finally, we study whether access to both training data and reference images improves performance.

\myparagraph{PatchCore with a fine-tuned backbone}. We test PatchCore as explained in Section~\ref{subsec:no-training-with-reference}, but replace the ResNet50 backbone fine-tuned on ImageNet with a ResNet50 backbone fine-tuned on our training dataset from Section~\ref{subsec:with-training-no-reference} (\texttt{PatchCore50-ft}) similar to~\cite{koshil2024apc}.

\myparagraph{AutoGluonMM}. To handle both query and reference images of the same item at train and test time, we use AutoGluon MultiModal~\cite{tang2024autogluon} by passing  all images (query \emph{and} references) for each sample through the same image backbone and averaging their respective embeddings to obtain the final representation  (\texttt{AutoGluon-ref}).

\subsection{Results}

We summarize the results of all baseline methods in
Tables~\ref{tab:num_results} and~\ref{tab:reduce_results}, with a detailed error analysis provided in the Supplementary Material~\ref{sec:error_analysis}. Our experiments aim to answer the following questions. (1) How well do methods with access to (all) defective instances at training time perform? (2) How does performance deteriorate when fewer defective instances are available for training? (3) How well do unsupervised and anomaly detection methods without access to defective instances for training perform?

\myparagraph{(1) Upper bound with access to a large number of defective instances at training time}.
The supervised baselines perform well, with \texttt{ViT-S} reaching 90.67\% \textbf{AP\textsubscript{any}}. While models effectively detect major defects like deconstructions, penetrations, and deformations, they struggle with subtle anomalies, rare defect types (spillage), and reference-dependent defects (missing unit). False positives primarily occur with oddly-shaped items and ``adversarial'' items featuring damage-like designs. Notably, methods using both training data and references (\texttt{PatchCore50-ft} and \texttt{AutoGluonMM-ref}) underperform compared to reference-free approaches. This suggests that naive reference usage actually hinders model performance, likely due to feature averaging across input images complicating the learning task. This hypothesis is supported by inspecting their training set performance (96\% \textbf{AP\textsubscript{any}} without references versus 87\% with).

\myparagraph{(2) Reduced access to defective instances at training time}.
Table~\ref{tab:reduce_results} shows how supervised baselines perform in a more realistic scenario with limited number of defective training samples, with only 1\% defective rate in the training set. Unsurprisingly, performance drops significantly from 90.67\% \textbf{AP\textsubscript{any}} to 57.7\% \textbf{AP\textsubscript{any}} for fully supervised methods (\texttt{Query only}). As before, the model is not able to leverage the non-defective samples (\texttt{Query + ref}).

\myparagraph{(3) No defective instances at training time}. No method without training surpasses 36.57\% \textbf{AP\textsubscript{any}} (\texttt{Claude-icl}), with both zero-shot/in-context learning models (\texttt{CLIP}/\texttt{POMP}, \texttt{Pixtral-*}) and anomaly detection models (\texttt{PatchCore50}, \texttt{WinCLIP-*}) performing only slightly above chance.
Out of the CLIP-based approaches the original CLIP model performs best. We find that the model seems to occasionally read the text on the items and wrongly associates it with defect predictions.
VLMs provide a reasonable overall description and can catch egregious defects like gross deconstruction, but fail to capture the intricacy and variety of minor defects concerning deformable items, stickers/dirt on trays, and subtle anomalies, corroborating previous findings by \cite{jiang2025mmad}. Anomaly detection methods latch on to non-defect related visual differences, such as novel poses/viewpoints, background noise, and packaging variations.

Interestingly, PatchCore with a fine-tuned ResNet50 backbone (\texttt{PatchCore50-ft}) shows 4.32 ppts improvement compared to the ImageNet-based model \texttt{PatchCore50}, indicating the usefulness of leveraging defective instances for representation learning in anomaly detection~\cite{koshil2024apc}. However, it still struggles in detecting minor actuation and deconstruction, particularly when items are slightly displaced of their packaging. Moreover, some false negatives stem from faulty reference images incorrectly assumed to be non-defective, highlighting the extra pre-caution required in using unlabeled reference data in anomaly detection. False positives mostly arise from visual disparities between test and reference images, including variations in pose and product appearance.
These results highlight the need for improved anomaly detection methods with a more thorough understanding of defects and more sophisticated ways for using references for visual comparison.

\begin{table}[tb]
\centering

\begin{tabular}{rccc}
    \textbf{Input} & \textbf{AP\textsubscript{any}} [\%] & \textbf{AP\textsubscript{major}} [\%] & \textbf{AUROC}  \\ \hline

    \texttt{Query only} & 57.7 & 40.5 & 74.4 \\
    \texttt{Query + ref} & 40.4 & 14.9 & 63.2 \\
    \hline
\end{tabular}

\caption{Classification performance on a reduced training set, with a defect rate of only 1\%. We compare a ViT using only query images and a late-fusion ViT using both query \emph{and} reference images.}
\vspace{-.75em}
\label{tab:reduce_results}
\end{table}

In summary, supervised methods perform best when given access to large amounts of defective instances during training, but still struggle with edge cases such as deformable and adversarial items. Adding reference images naively degrades rather than improves performance. Unsupervised and anomaly detection methods fall short by a significant margin, but improve with access to training data.
\section{Outlook and Conclusion}
\label{sec:conclusion}

We presented a large-scale dataset for visual defect and anomaly detection in retail logistics. Comprising 238,421 images including 29,316 defective samples, it captures challenges of retail logistics processes and represents one of the largest and most diverse datasets of its kind.  The dataset overcomes critical limitations in existing benchmarks and enables the research community to address the remaining challenges in visual defect and anomaly detection.
It allows for benchmarking methods in various scenarios, with and without training and reference images. We demonstrate the complexity of the proposed task by evaluating a number of state-of-the-art approaches and highlight the need for more robust solutions, particularly in anomaly-detection settings.

This dataset marks a significant step towards developing defect detection systems capable of handling real-world scenarios, setting a new standard for research in retail logistics applications of visual inspection. We encourage future research to explore the dataset by developing novel approaches. Key questions for future work include but are not limited to:
(1) How can anomaly detection methods be generalized to deal with significant item and pose variability?
(2) How can methods effectively leverage both training data and reference images?
(3) How can we create methods that not only detect defects but also explain their reasoning?

\subsection*{Acknowledgements}

We thank our collaborators in Amazon's operations, hardware and software engineering, as well as our annotation teams. Their invaluable contributions to hardware development, software implementation, data collection, and labeling efforts were essential to the success of this work.

{
    \small
    \bibliographystyle{ieeenat_fullname}
    \bibliography{main}
}

\clearpage
\setcounter{page}{1}
\maketitlesupplementary

\renewcommand{\thesection}{\Roman{section}}
\setcounter{section}{0}

\label{sec:appendix}

\section{VLM Prompts}
\label{sec:vlm_prompts}
We followed prompt engineering best practices for the zero-shot models and selected the one that performed best on a small validation set.
The following prompt was used for the VLMs \texttt{Pixtral} and \texttt{Claude} to evaluate performance our dataset:
\begin{framed}
You are a highly skilled subject matter expert for inventory quality assurance and control. The presented image shows an item inside a tray. You have to determine whether the item is in pristine condition and can be sold as new and shipped to the customer as is, or whether it is damaged in any way and needs further attention before it can be shipped. Consider the following damage categories: crushed, tear, hole, deformed, ripped, deconstructed. Typical defects also include open boxes, or damaged and ripped packaging. Sometimes if the packaging is damaged, the item itself may become deconstructed and parts of the content may fall out. The container itself may be dirty which should not count as damage. However, if there is spillage that originated from a liquid item, then it must be called out as damage. Pay close attention to books and especially to corners of front or back pages. Moreover, items that a deconstructed, i.e. where the original packaging is damaged or fell off, should be flagged as damaged. In addition to the final decision specified by DAMAGED or UNDAMAGED, please also provide the severity score on a scale from 0 (pristine condition) to 10 (completely destroyed). Think step-by-step and provide the final response as json with keys "condition" and "severity".
\end{framed}

\section{CLIP Prompts}
\label{sec:clip_prompts}

For the \texttt{CLIP} model 0-shot baseline, we tested the following prompts on the validation set:

\begin{itemize}
    \item \texttt{Image of an item with some damage.} and \texttt{Image of an item with no damage.}
    \item \texttt{Item without damage inside of a tray.} and \texttt{Item with damage inside of a tray.}
    \item \texttt{Image of an undamaged item.} and \texttt{Image of a damaged item.}
    \item \texttt{Image of an item without problems.} and \texttt{Image of an item with problems.}
\end{itemize}

The last prompt performed best on the validation set and was used for the test set evaluation.

\section{Dataset Details}
\label{sec:dataset_details}

\begin{figure}[bh!]
    \centering
         \includegraphics[width=0.45\textwidth]{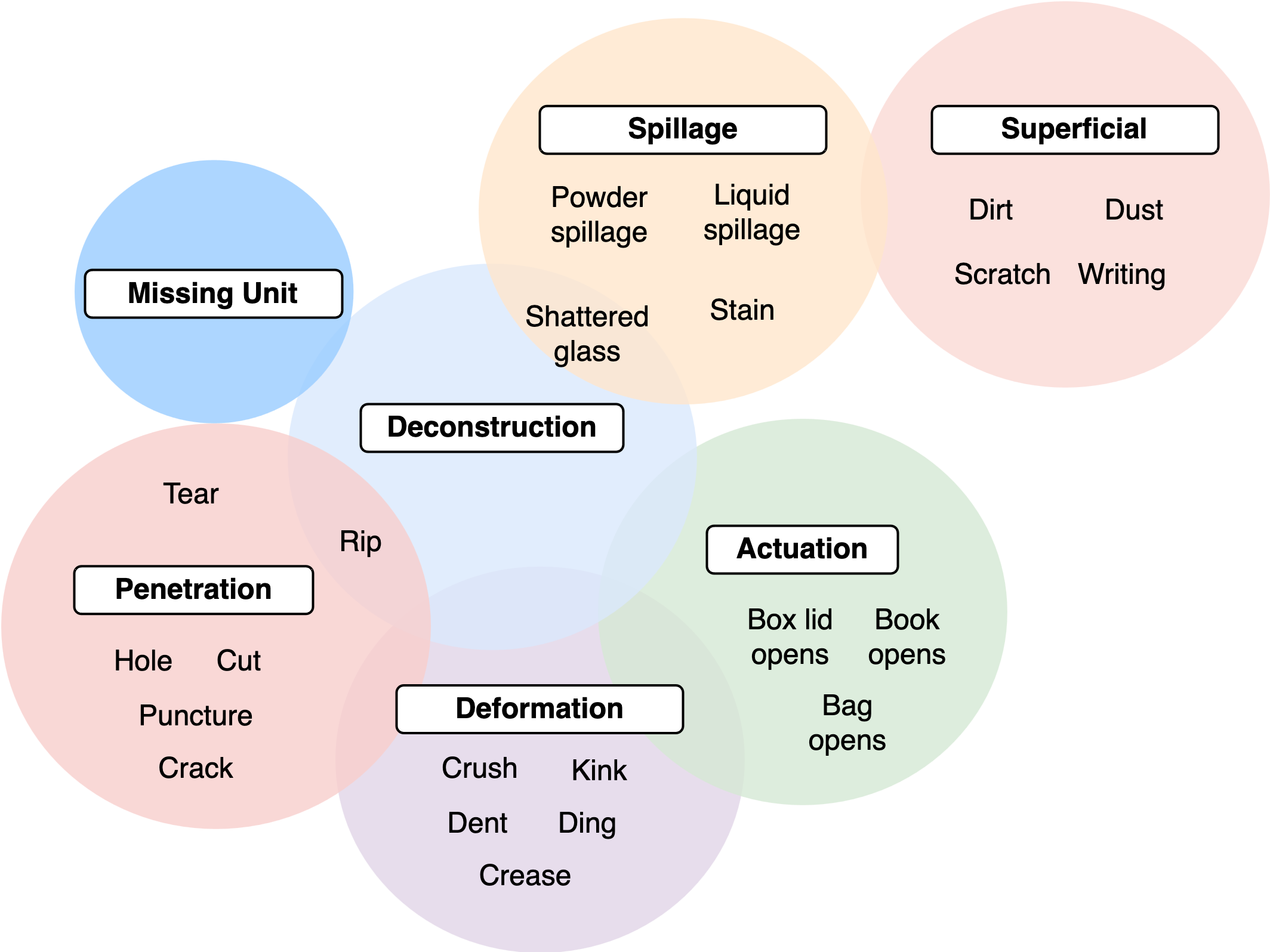}
    \caption{
    Overview of the defect types used to annotate defective samples (bold font) and related and colloquial characterization of the defect types (in bubbles). The proximity of the bubbles and their overlap indicates which defect types are similar/related\label{fig:damage-types}.
    }
\end{figure}

\begin{figure}[bh!]
    \centering
         \includegraphics[width=0.45\textwidth]{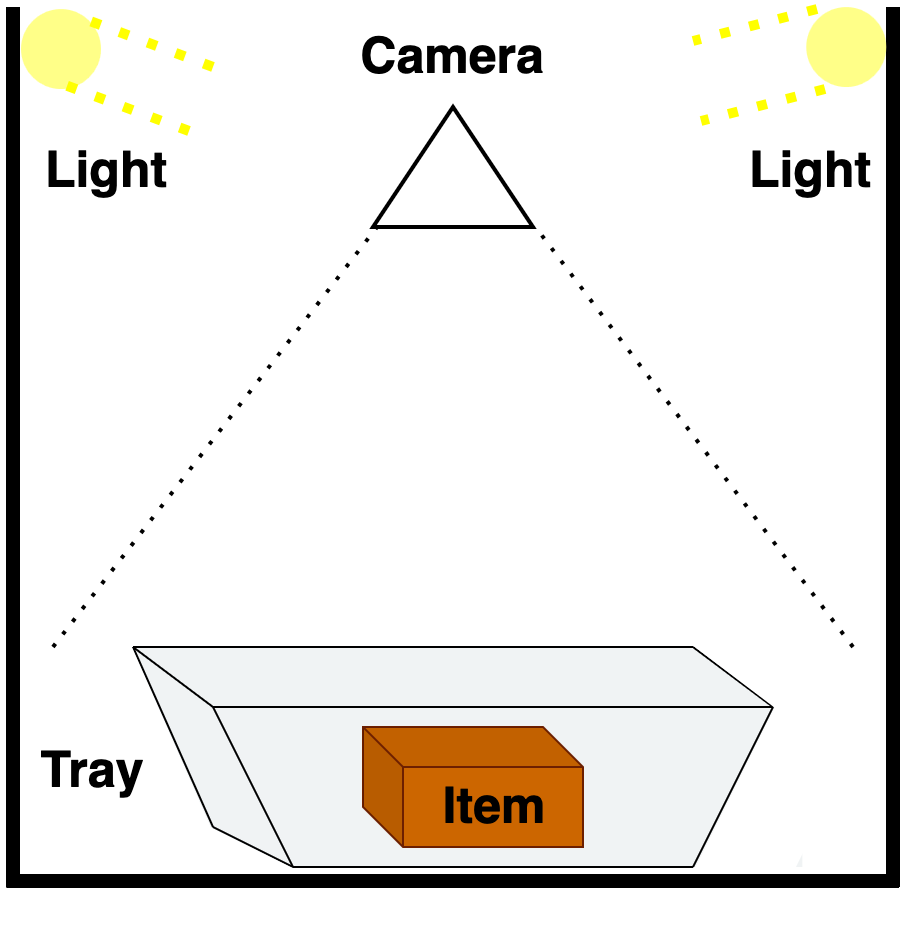}
    \caption{
    Schematic illustration of the data collection station. An overhead camera captures images from the top. Items reside in ``trays``, common logistics containers.\label{fig:rigspecs}}
\end{figure}

We provide additional details about our defect taxonomy underlying our dataset (Figure~\ref{fig:damage-types}) and experimental setup of our data collection station (Figure~\ref{fig:rigspecs}).

\clearpage

\begin{table*}[!ht]

\section{Error Analysis}
\label{sec:error_analysis}

The following tables systematically evaluate failure modes of all baselines. Table~\ref{ref:failuremodestats} provides quantitative statistics, while Tables 5-7 showcase exemplary correct and wrong predictions. Tables 8-35 further analyze model performance per item material and per defect type. Finally, Figures 7 and 8 show Precision-Recall and ROC curves.

\vspace{1cm}

\centering
\renewcommand{\arraystretch}{1.2}
\begin{tabular}{r|cccccc|c|ccccc|c}
 & \multicolumn{7}{c|}{\textbf{False Positives}} & \multicolumn{6}{c}{\textbf{False Negatives}} \\

\textbf{Baseline} &
   \rotatebox{90}{Adversarial} &
   \rotatebox{90}{Pack. Type} &
   \rotatebox{90}{Query Distr.} &
   \rotatebox{90}{Reference Var.} &
   \rotatebox{90}{Minuscule} &
   \rotatebox{90}{Unclear} &
   \rotatebox{90}{Total} &
   \rotatebox{90}{Subtle} &
   \rotatebox{90}{Conf. Wrong} &
   \rotatebox{90}{Reference Var.} &
   \rotatebox{90}{Missing Unit} &
   \rotatebox{90}{Unclear} &
   \rotatebox{90}{Total} \\ [2ex]
\hline \hline
\multicolumn{14}{c}{\textbf{No training, no references} (zero-shot, few-shot)}\\
\hline
\texttt{CLIP} & 1 & 0 & 0 & - & 5 & 44 & 50 & 4 & 41 & - & 0 & 5 & 50 \\
\texttt{POMP} & 3 & 0 & 0 & - & 4 & 43 & 50 & 17 & 29 & - & 0 & 4 & 50 \\
\texttt{WinCLIP-zero} & 0 & 32 & 8 & - & 0 & 10 & 50 & 37 & 10 & - & 2 & 1 & 50 \\
\texttt{Claude-icl} & 8 & 18 & 13 & - & 0 & 11 & 50 & 42 & 8 & - & 0 & 0 & 50 \\
\texttt{Pixtral-zero} & 4 & 20 & 11 & - & 0 & 15 & 50 & 39 & 11 & - & 0 & 0 & 50 \\
\texttt{Pixtral-icl} & 7 & 11 & 5 & - & 0 & 27 & 50 & 38 & 12 & - & 0 & 0 & 50 \\
\hline
\multicolumn{14}{c}{\textbf{No training, \emph{with} references} (few-shot, non-parametric, in-context learning)} \\
\hline
\texttt{PatchCore50} & 1 & 0 & 0 & 42 & 1 & 6 & 50 & 40 & 1 & 0 & 0 & 9 & 50 \\
\texttt{WinCLIP-few} & 2 & 20 & 9 & 9 & 0 & 10 & 50 & 44 & 5 & 0 & 0 & 1 & 50 \\
\hline
\multicolumn{14}{c}{\textbf{\emph{With} training, no references} (supervised/instruction fine-tuning)}\\
\hline
\texttt{ResNet50} & 8 & 16 & 11 & - & 13 & 2 & 50 & 33 & 17 & - & 0 & 0 & 50 \\
\texttt{ViT-S} & 5 & 6 & 2 & - & 15 & 22 & 50 & 27 & 19 & - & 2 & 2 & 50 \\
\texttt{Pixtral-ft} & 1 & 11 & 7 & - & 3 & 28 & 50 & 40 & 10 & - & 0 & 0 & 50 \\
\texttt{AutoGluonMM} & 10 & 7 & 6 & - & 20 & 7 & 50 & 31 & 19 & - & 0 & 0 & 50 \\
\hline
\multicolumn{14}{c}{\textbf{\emph{With} training, \emph{with} references} (supervised with references, non-parametric with fine-tuning)}\\
\hline
\texttt{PatchCore50-ft} & 1 & 3 & 2 & 36 & 1 & 7 & 50 & 38 & 5 & 0 & 7 & 0 & 50 \\
\texttt{AutoGluonMM-ref} & 3 & 9 & 3 & 16 & 4 & 15 & 50 & 21 & 20 & 3 & 0 & 6 & 50 \\
\hline \hline
\end{tabular}
\caption{Error Analysis of Different Baseline Models. For each evaluated baseline model, we analyzed the top 50 (by classification score) false positives and false negatives. Upon visual examination, we classified them into 6 groups for the false positives and 5 groups for the false negatives. False positives were either \emph{adversarial examples} (\eg books with printed creases, oddly shaped), difficult \emph{packaging types} (\eg paper or plastic that wrinkles), \emph{query distractions} (\eg trash or tray scribbles), \emph{reference variations} (\eg confused or defective references), \emph{minuscule defects}, or \emph{unclear} reasons. False negatives were categorized into \emph{confidently wrong} decisions by the baseline model, \emph{missing units} (\eg when multi-packs are captured with missing units that can only be identified by a model using references), or \emph{minuscule}, \emph{reference variations}, and \emph{unclear} reasons, analogue to false positive cases. \label{ref:failuremodestats}}
\end{table*}

\newcommand{\largeimage}[1]{
  \includegraphics[width=3cm,height=3cm]{#1}
}

\newcommand{\smallimage}[1]{
  \includegraphics[width=1.4cm,height=1.4cm]{#1}
}

\newcommand{\imagecell}[3]{
  \largeimage{#1}\par
  \smallimage{#2}\hfill\smallimage{#3}

}

\newcommand{\imagecellfour}[4]{
  \smallimage{#1}\hfill\smallimage{#2}\par
  \smallimage{#3}\hfill\smallimage{#4}
}

\begin{table*}
\centering
\caption{Failure and success modes of \texttt{AutoGluonMM} - With reference images. The images are arranged as squares in groups of four. The top-left example image shows the query image and the other three are the corresponding reference gallery images.}
\resizebox{0.95\textwidth}{!}{
\begin{tabular}{l|>{\centering\arraybackslash}p{3cm}|>{\centering\arraybackslash}p{3cm}|>{\centering\arraybackslash}p{3cm}|>{\centering\arraybackslash}p{3cm}}
\toprule
Model & TP & FP & TN & FN \\
\midrule

\multirow{1}{*}{\rotatebox{90}{\parbox{5.6cm}{\centering \texttt{AutoGluonMM}}}} &

\imagecellfour{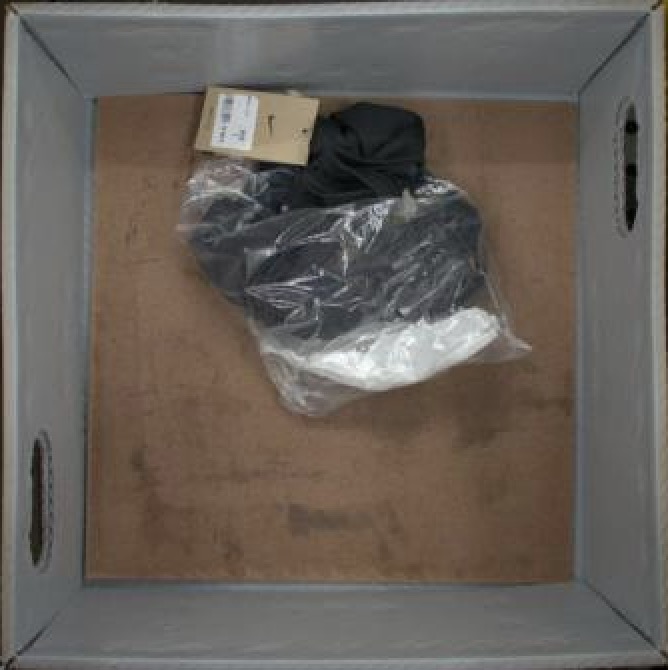}
{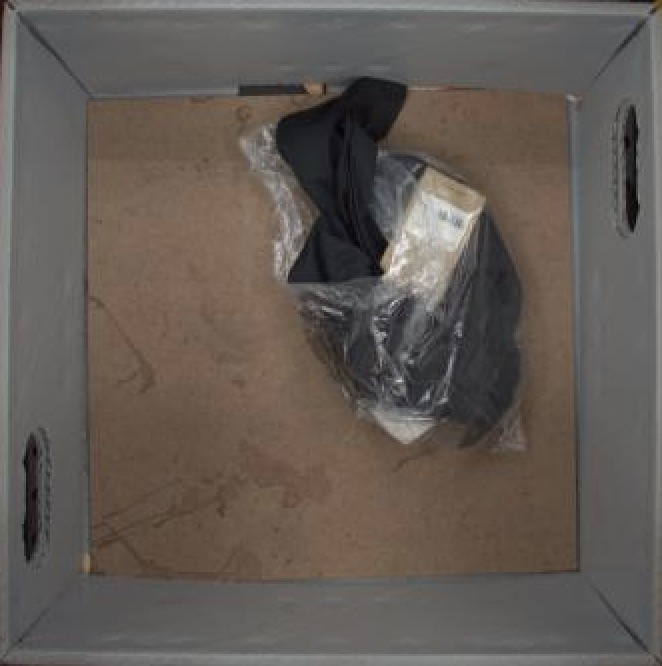}
{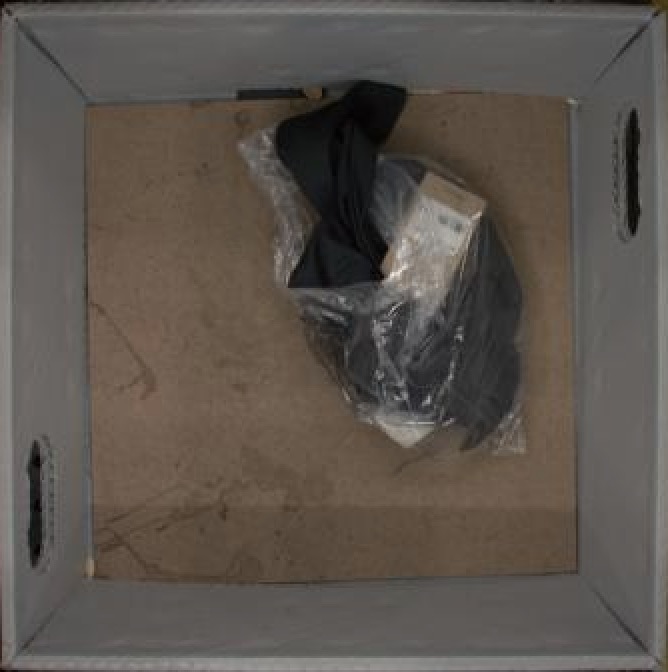}
{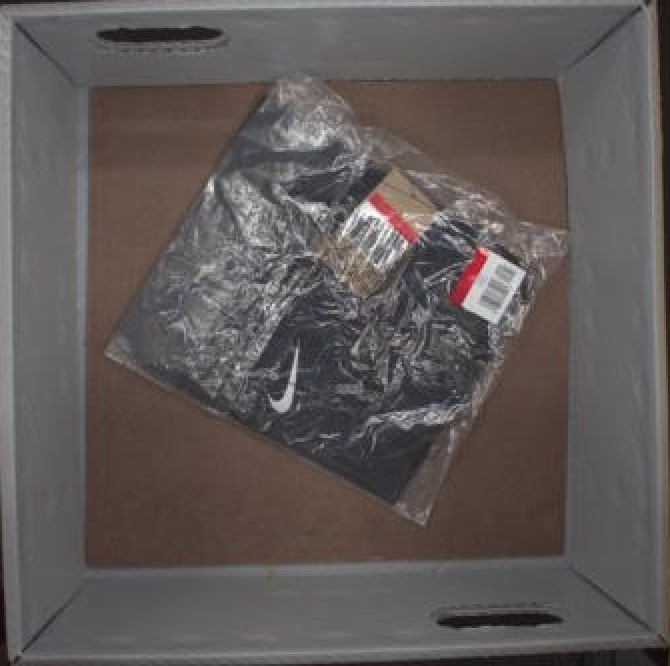}
\imagecellfour{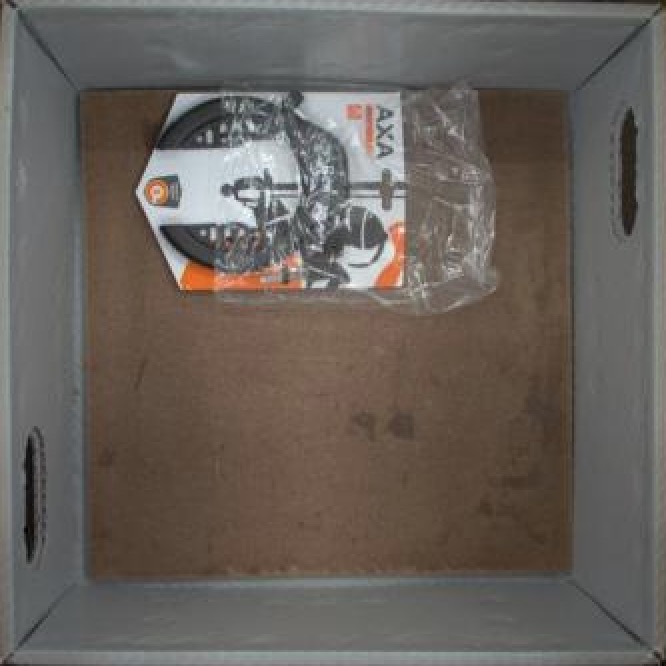}
{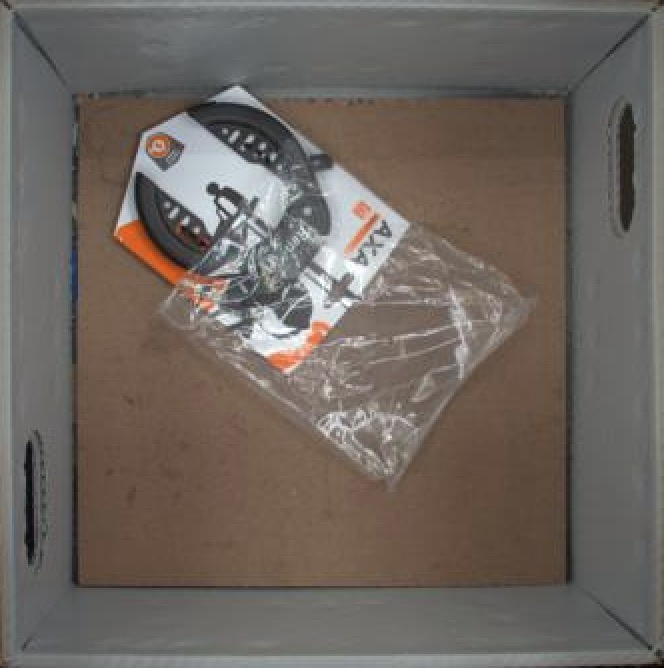}
{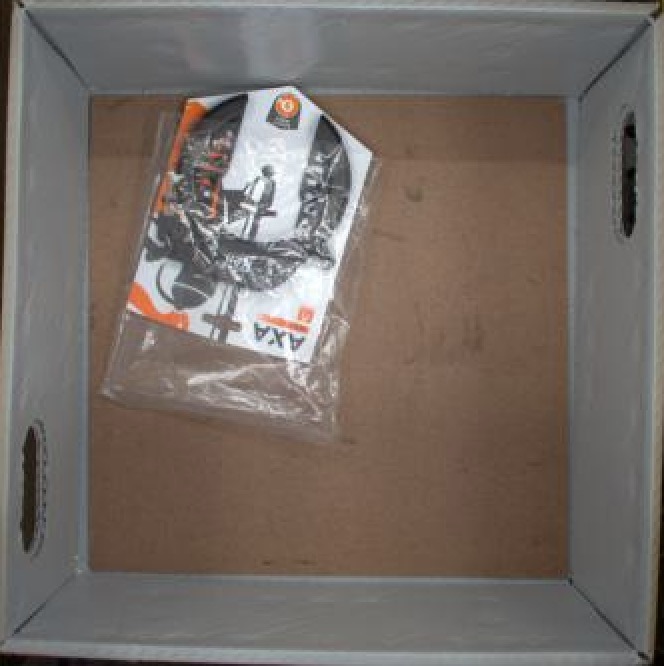}
{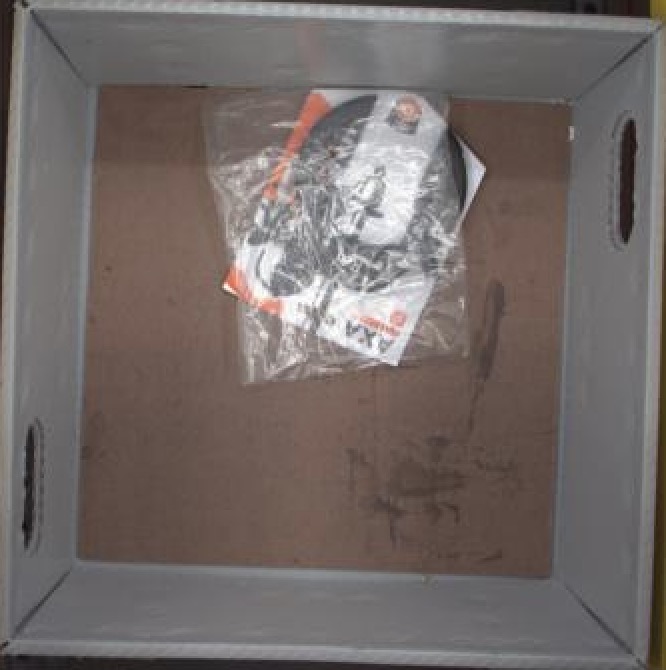}
&
\imagecellfour{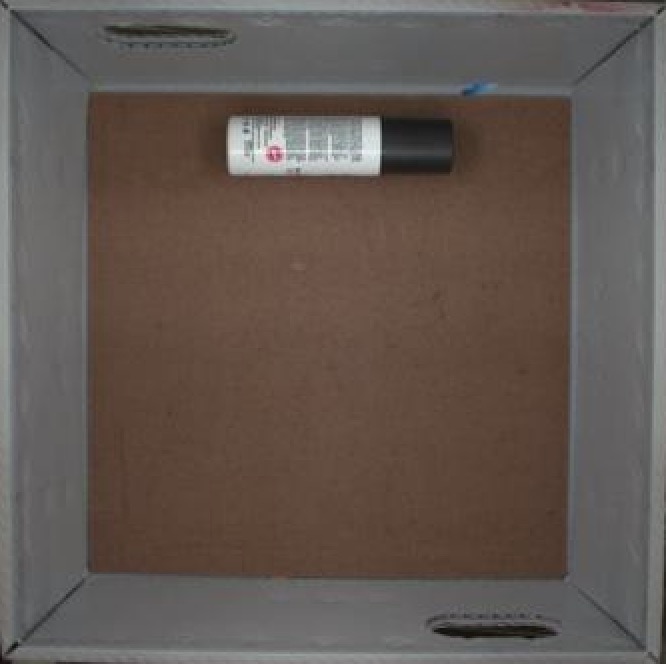}
{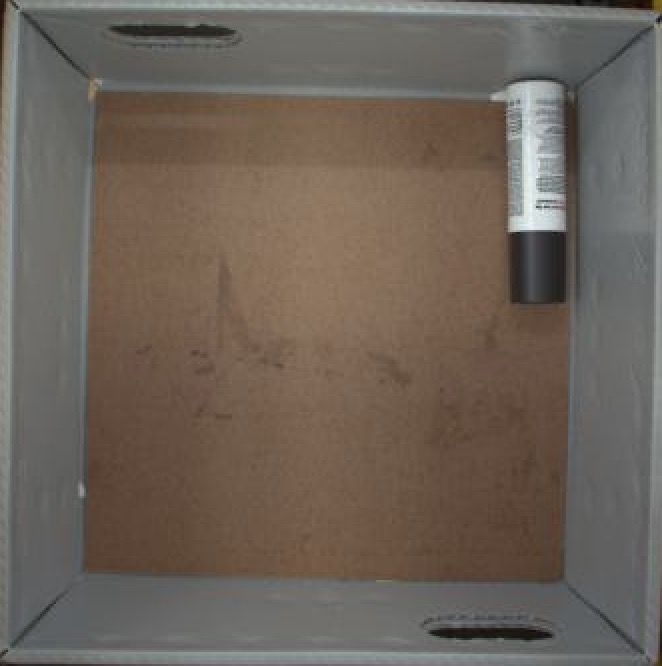}
{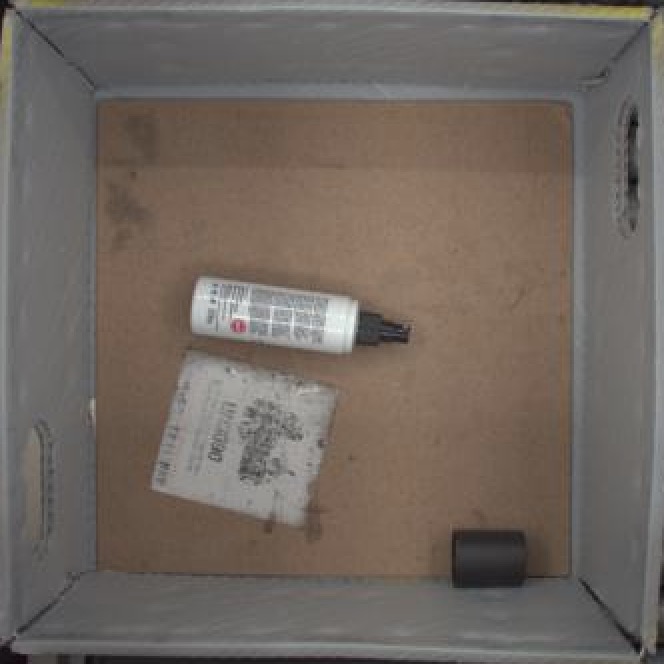}
{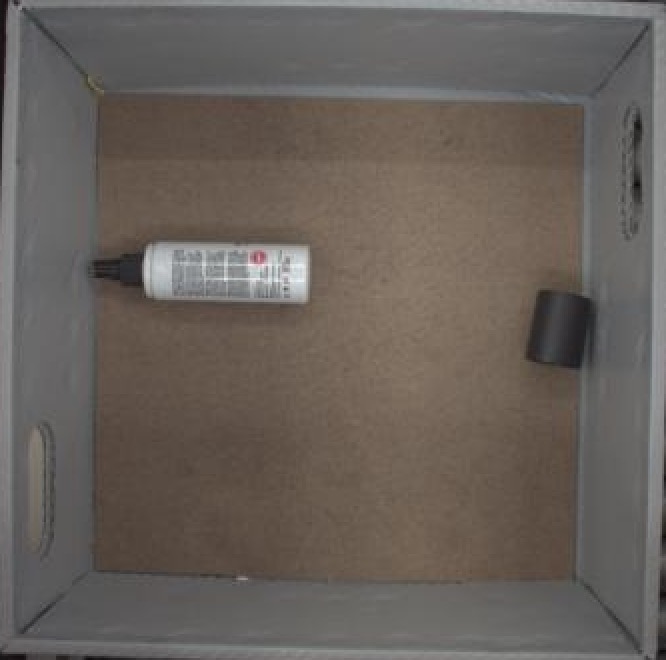}
\imagecellfour{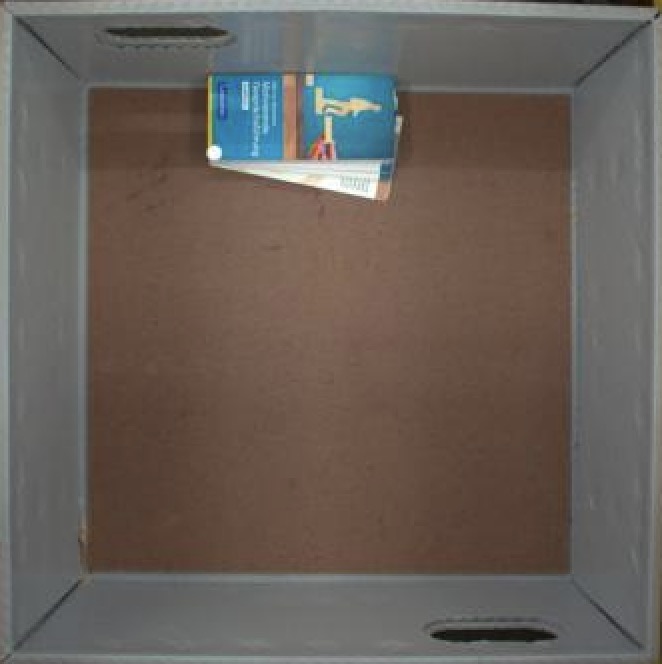}
{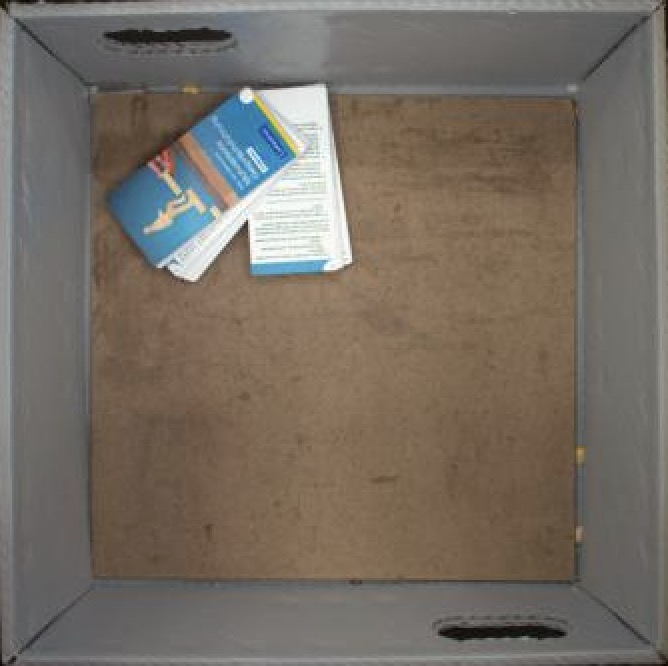}
{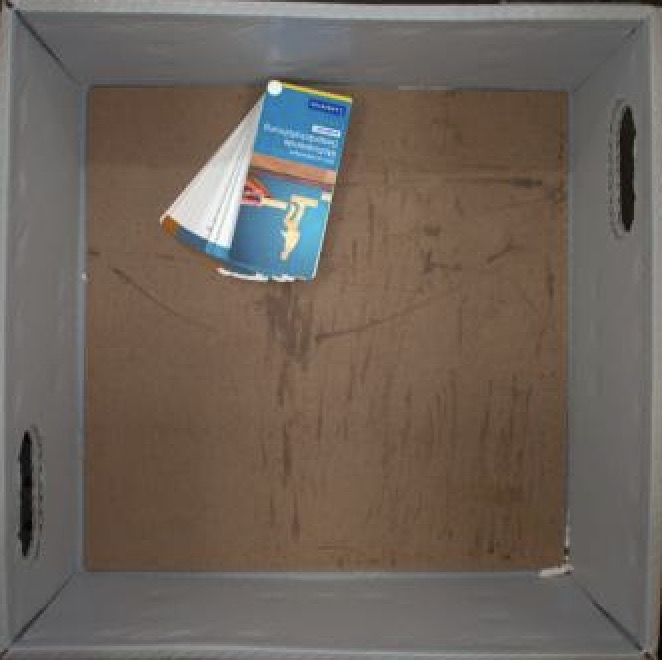}
{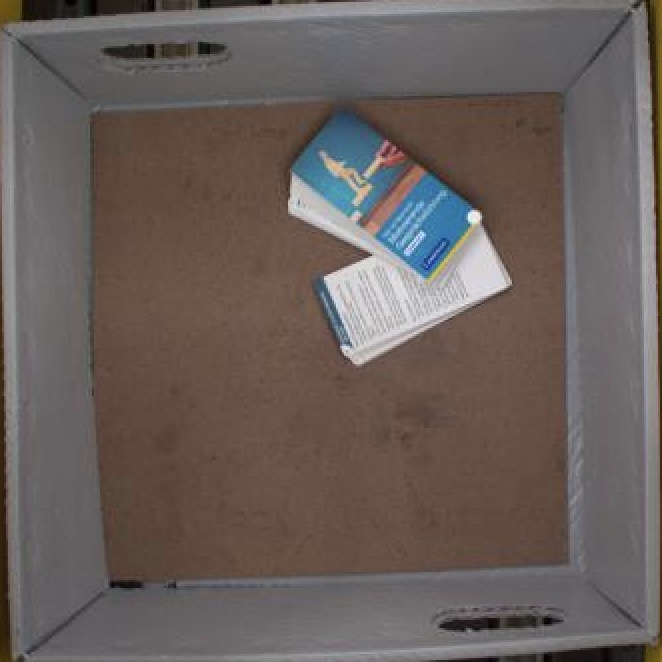}
&
\imagecellfour{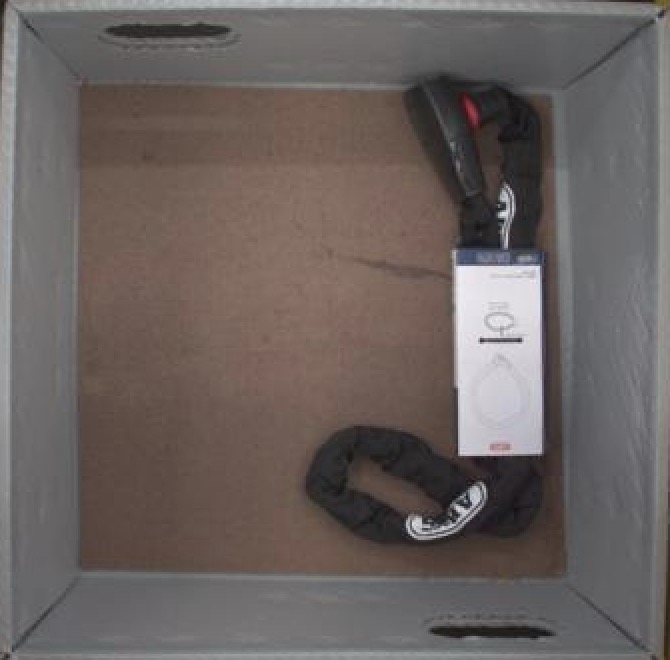}
{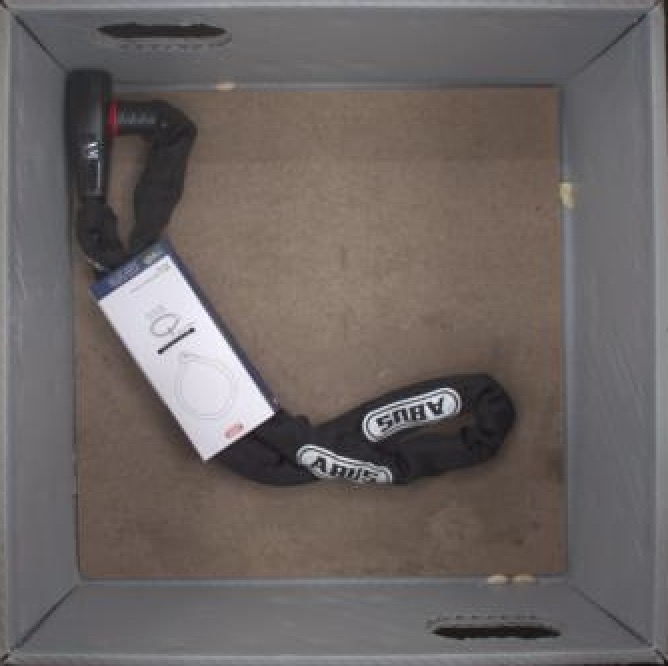}
{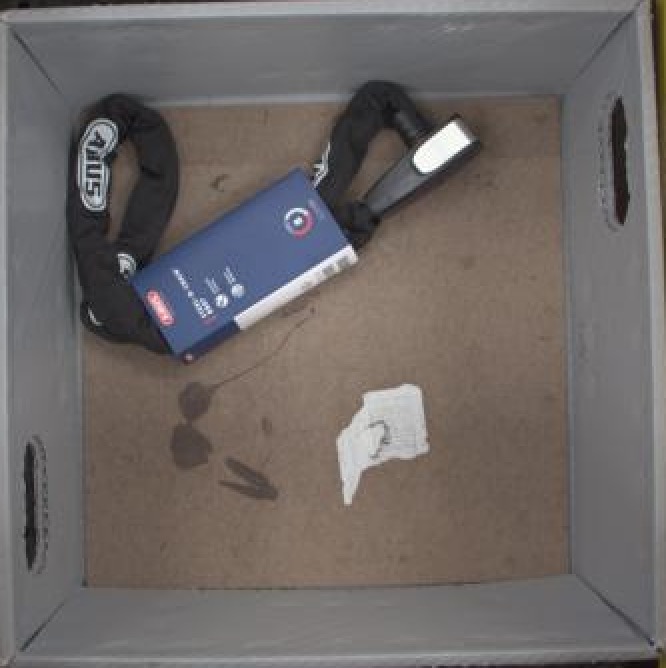}
{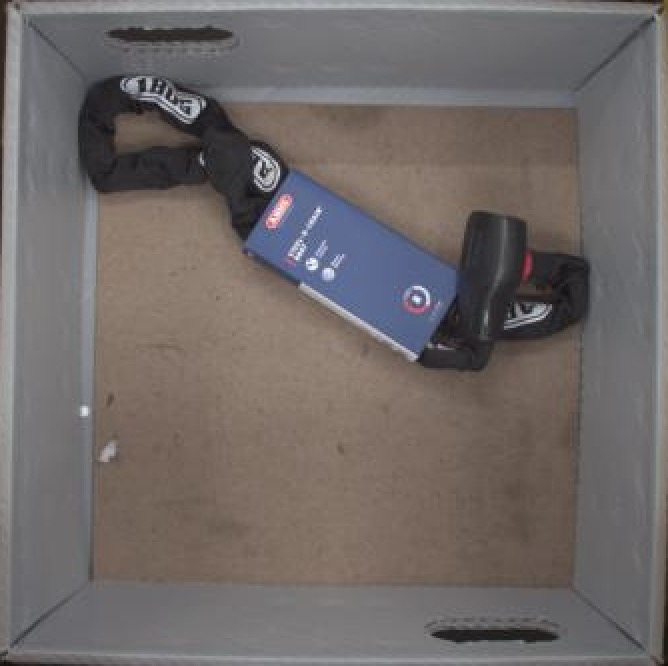}
\imagecellfour{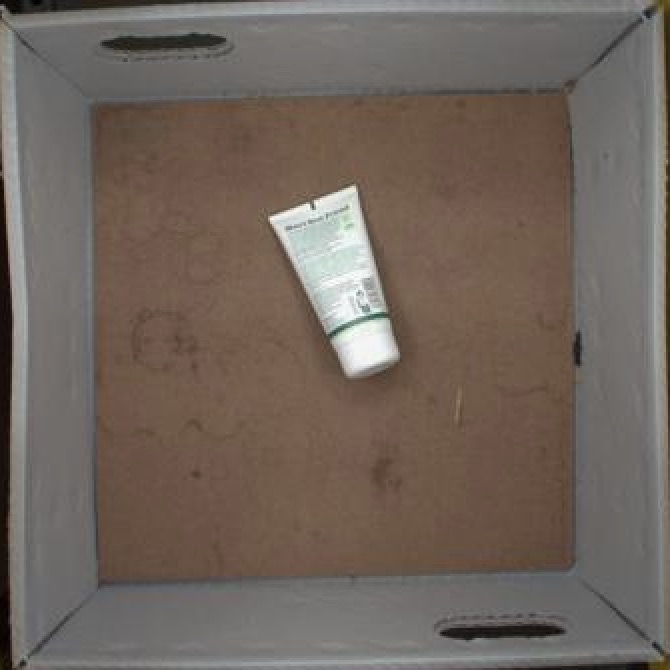}
{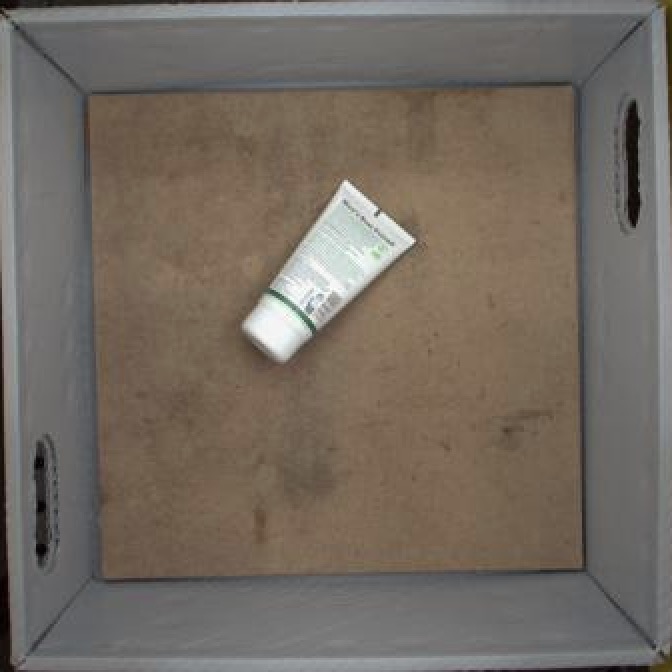}
{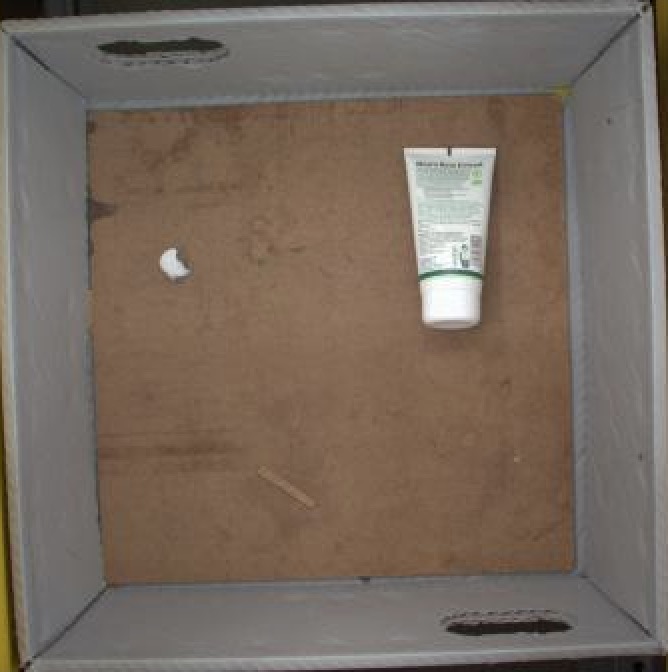}
{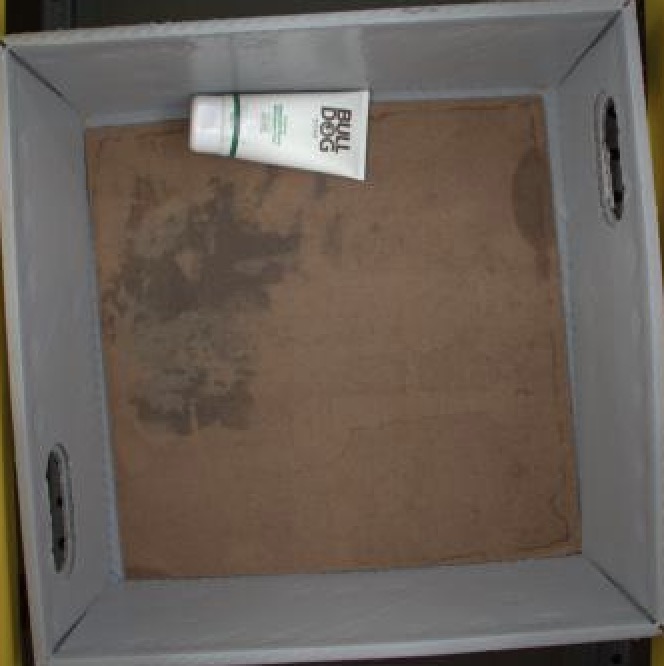}
&
\imagecellfour{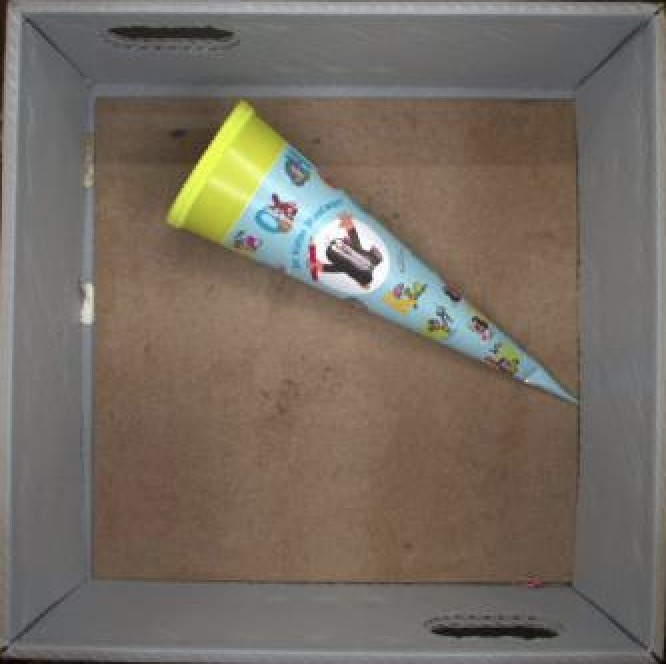}
{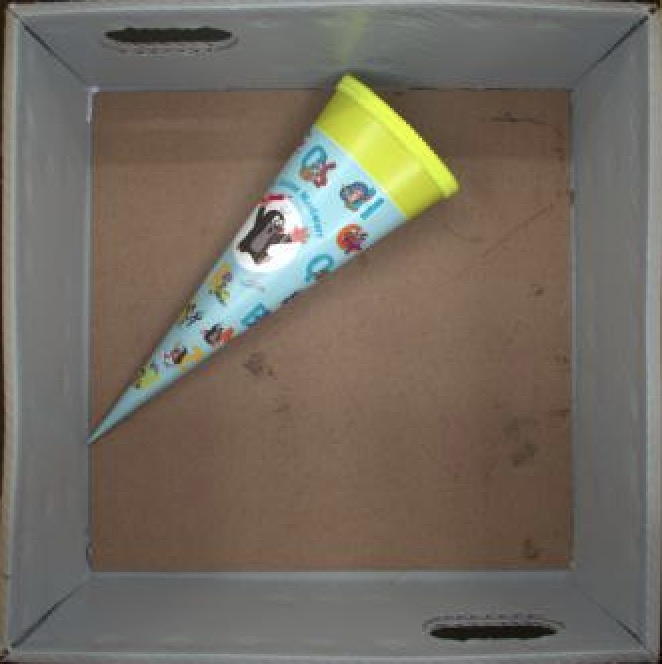}
{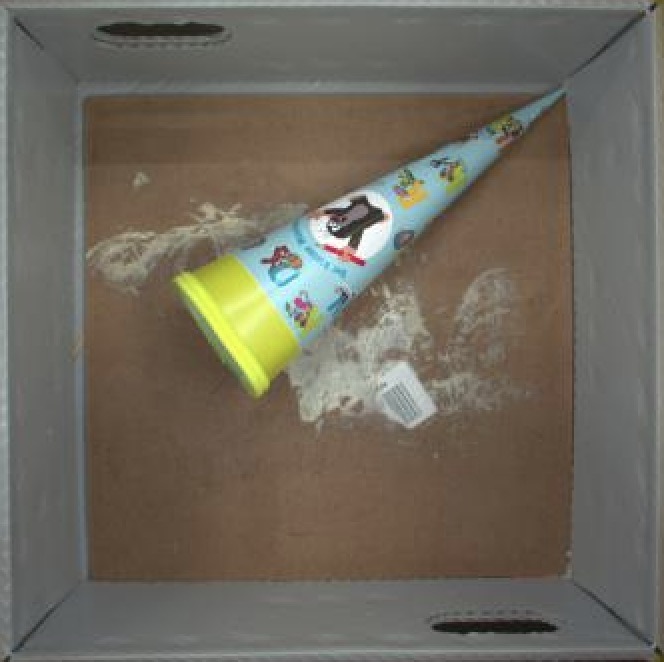}
{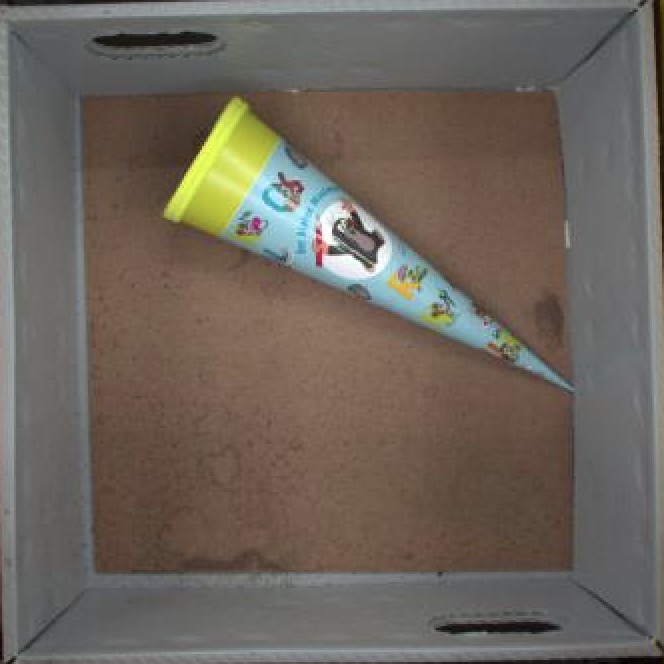}
\imagecellfour{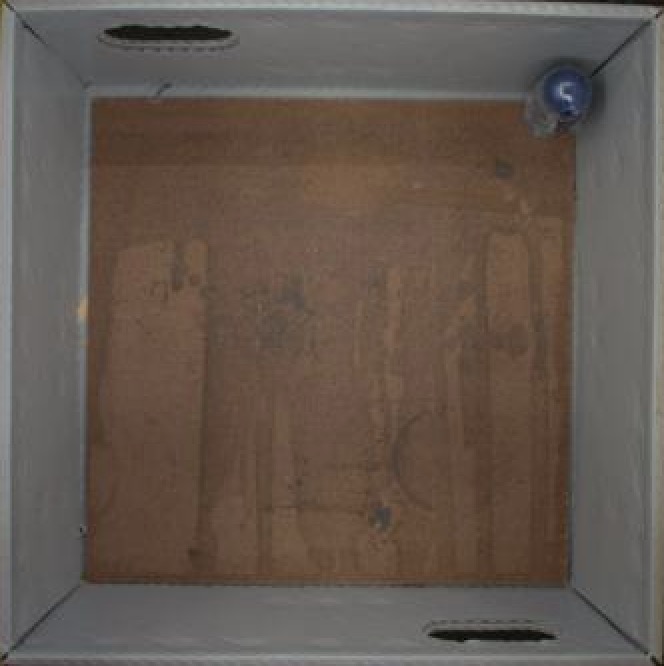}
{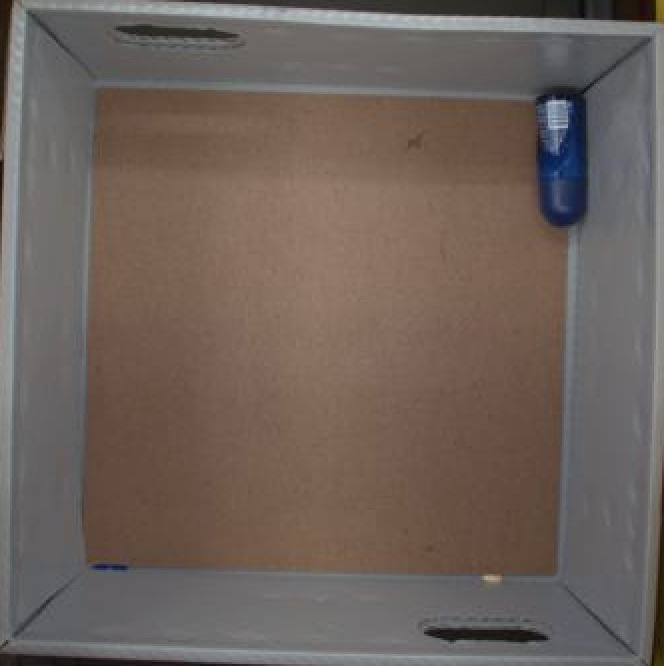}
{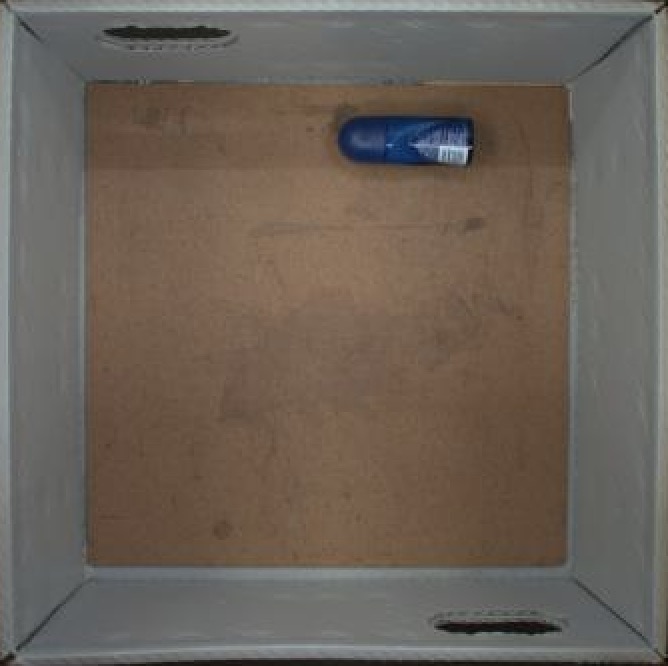}
{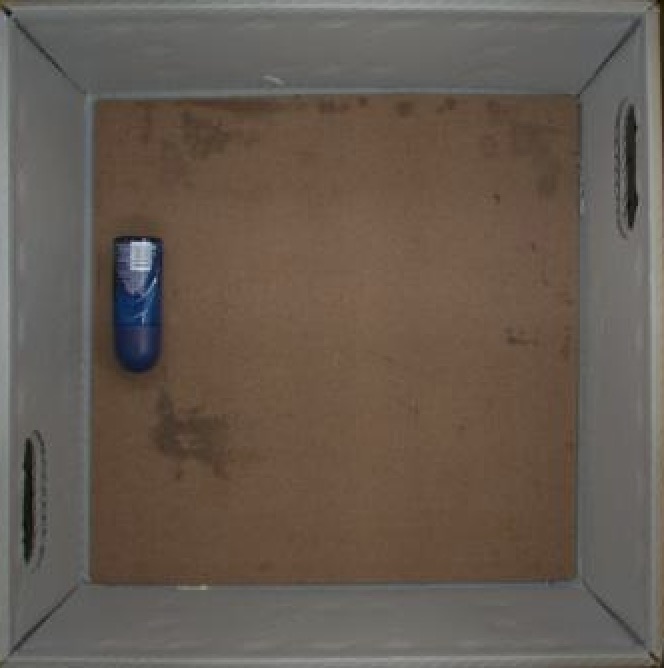}
\\
\bottomrule
\end{tabular}
}
\label{tab:failure_modes_autogluon}
\end{table*}

\begin{table*}
\centering
\caption{Failure and success modes of \texttt{PatchCore50-ft} and \texttt{WinCLIP} - The images are arranged as squares in groups of four. The top-left images shows the query image, the top-right image the query with overlaid activation map, and the two bottom images are two of the reference images.}
\resizebox{0.95\textwidth}{!}{
\begin{tabular}{l|>{\centering\arraybackslash}p{3cm}|>{\centering\arraybackslash}p{3cm}|>{\centering\arraybackslash}p{3cm}|>{\centering\arraybackslash}p{3cm}}
\toprule
Model & TP & FP & TN & FN \\
\midrule
\multirow{1}{*}{\rotatebox{90}{\parbox{5.6cm}{\centering \texttt{PatchCore50-ft}}}} &

\imagecellfour{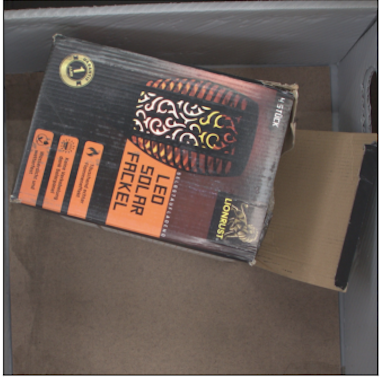}
{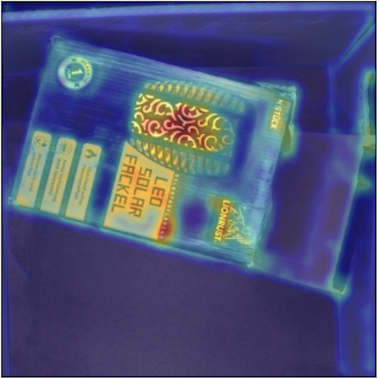}
{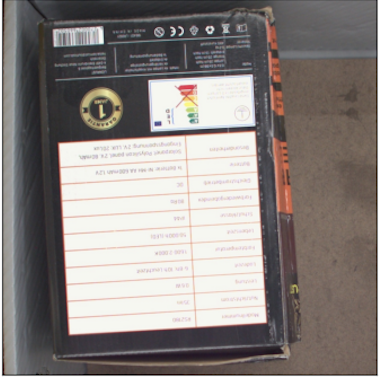}
{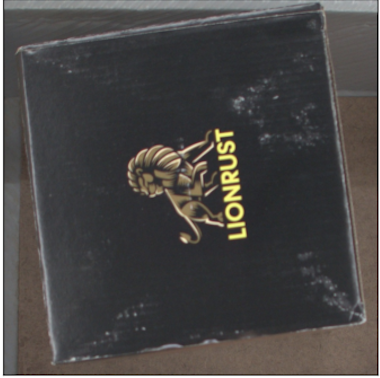}
\imagecellfour{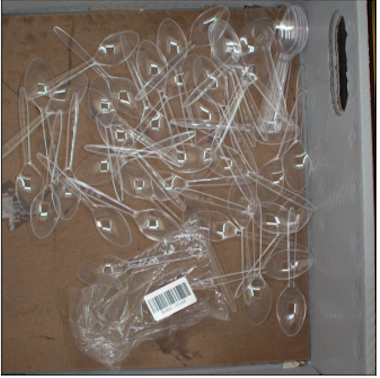}
{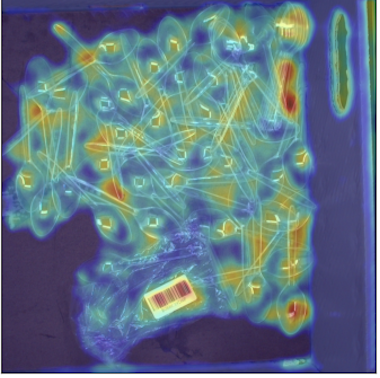}
{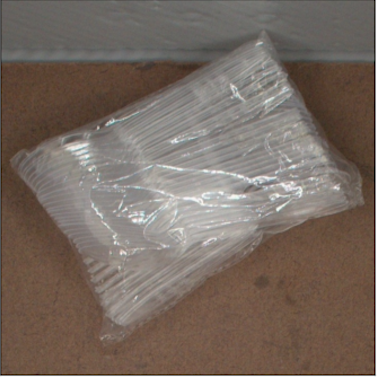}
{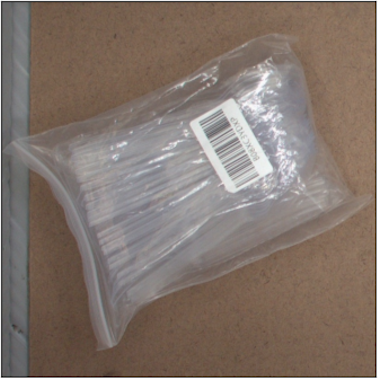}
&
\imagecellfour{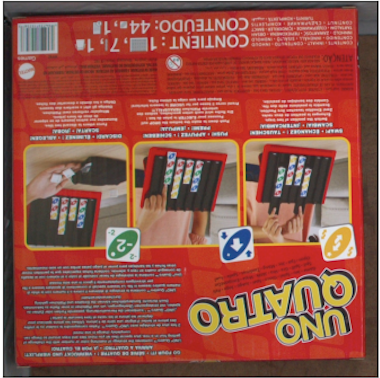}
{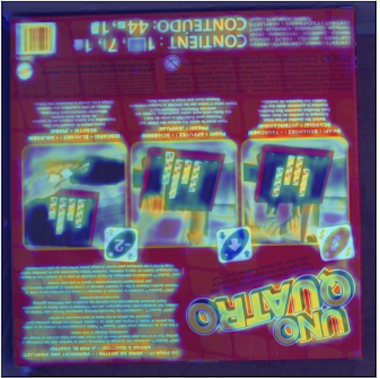}
{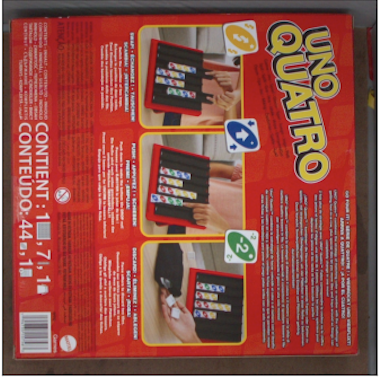}
{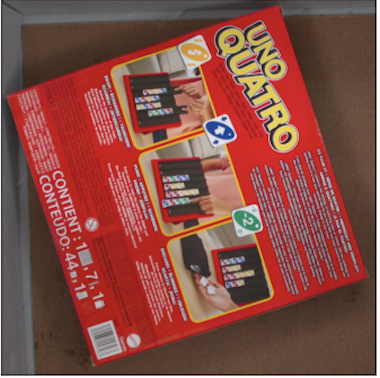}
\imagecellfour{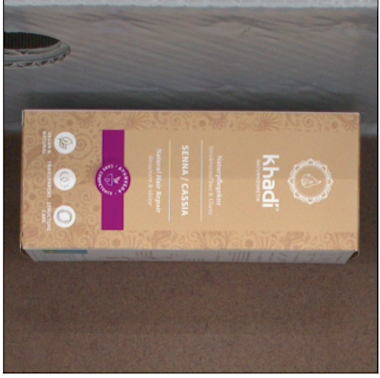}
{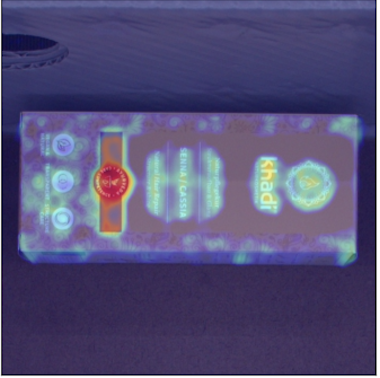}
{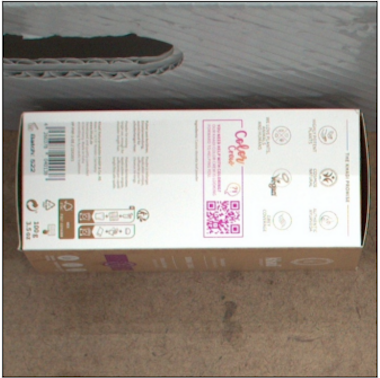}
{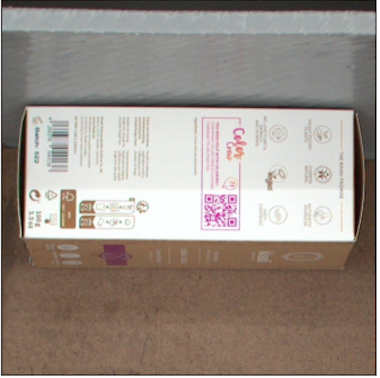}
&
\imagecellfour{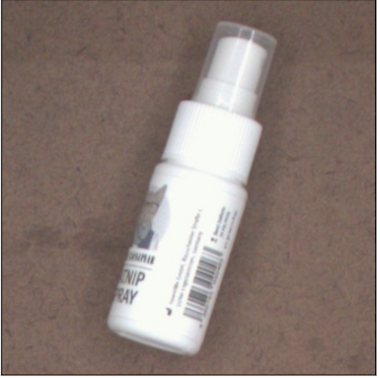}
{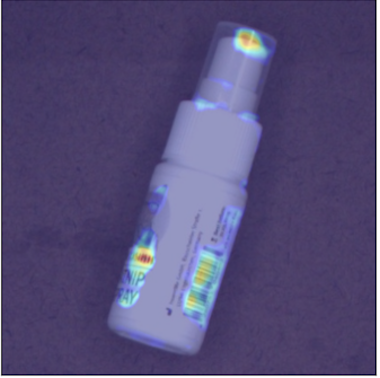}
{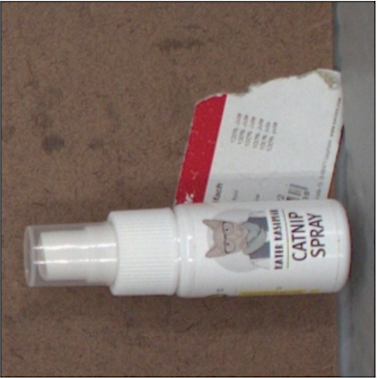}
{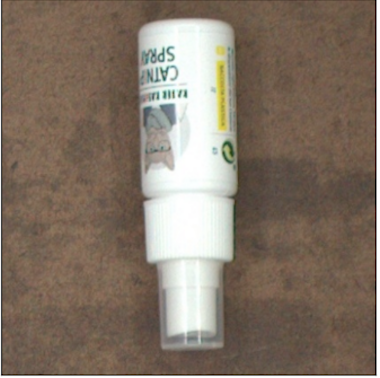}
\imagecellfour{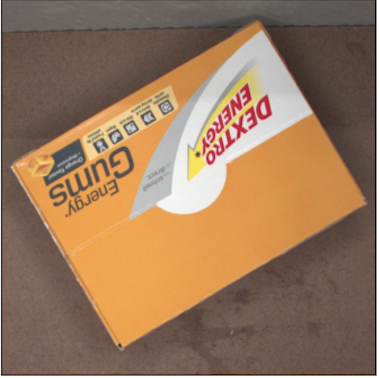}
{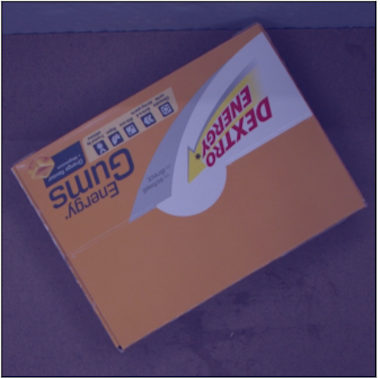}
{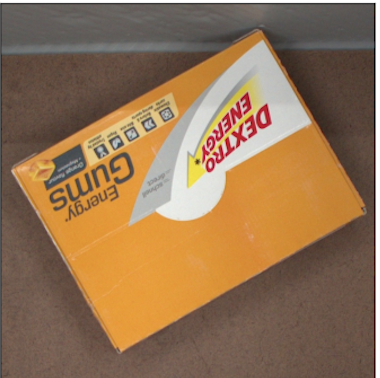}
{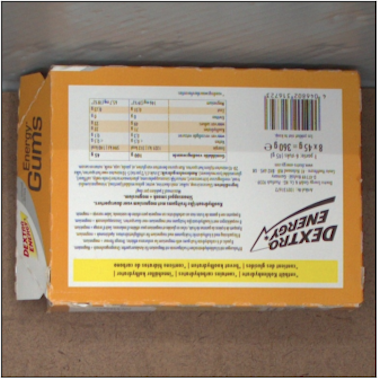}
&
\imagecellfour{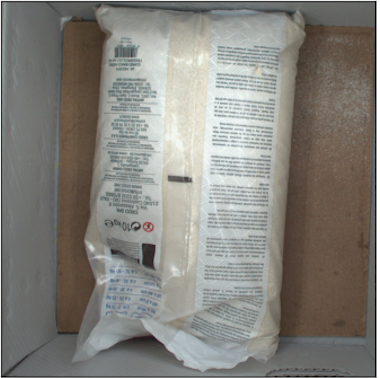}
{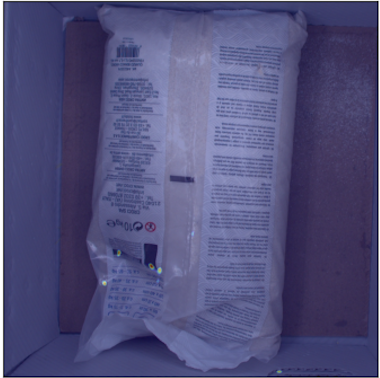}
{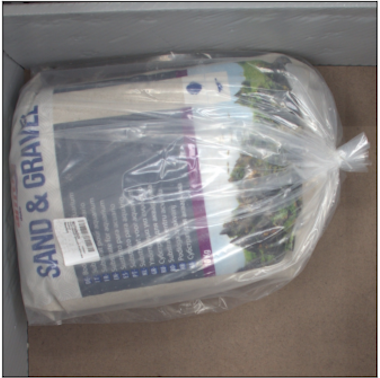}
{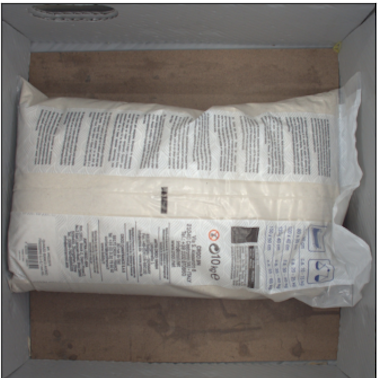}
\imagecellfour{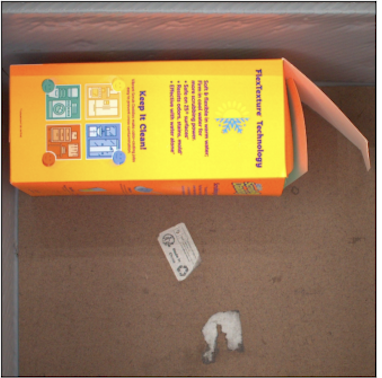}
{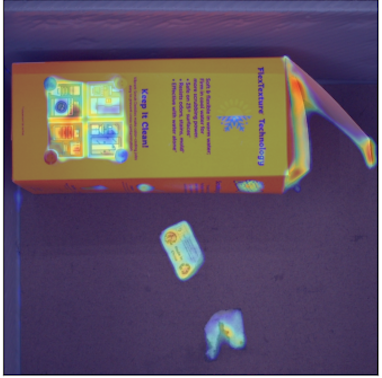}
{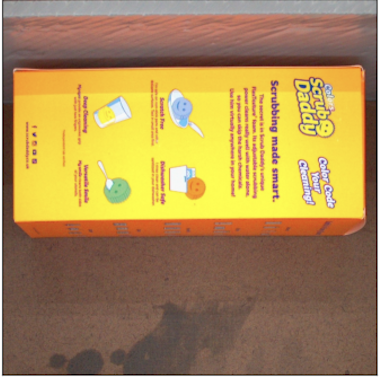}
{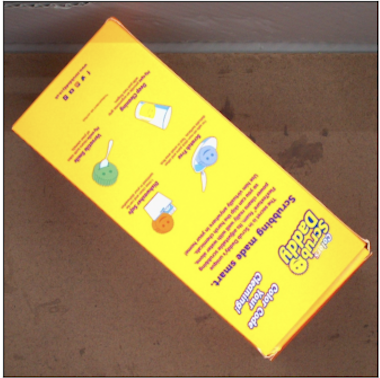}
\\
\midrule
\multirow{1}{*}{\rotatebox{90}{\parbox{5.6cm}{\centering \texttt{WinCLIP}}}} &

\imagecellfour{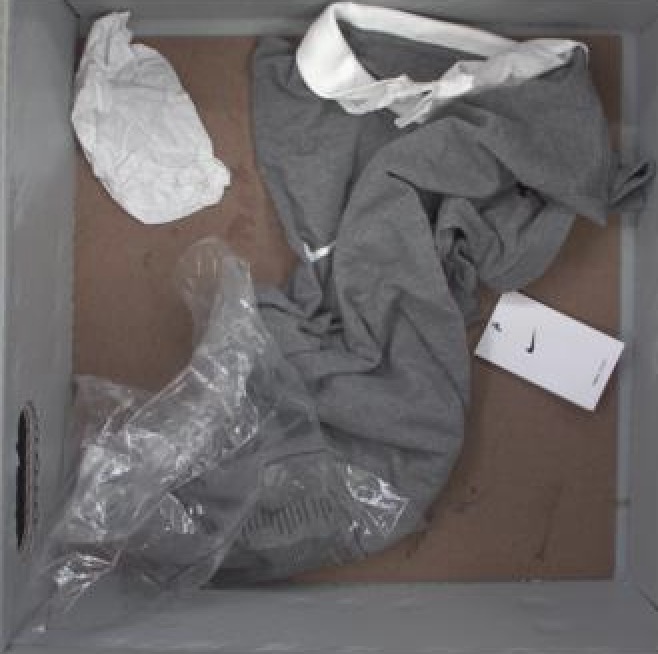}
{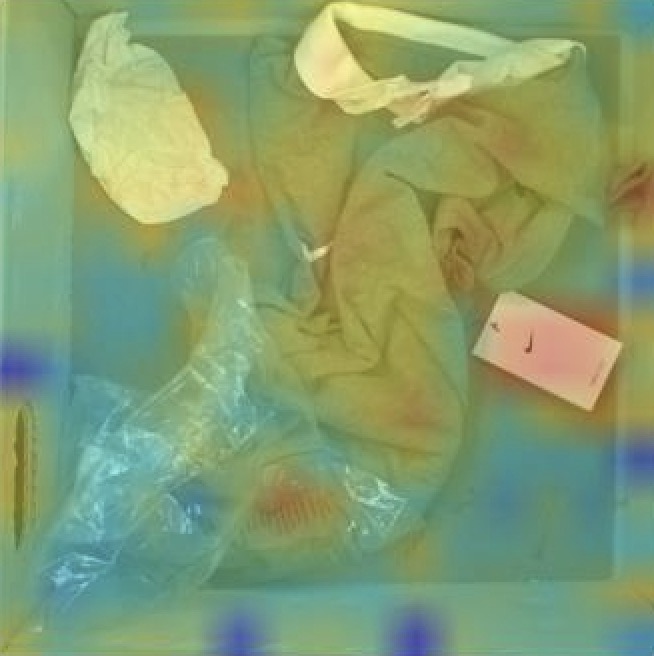}
{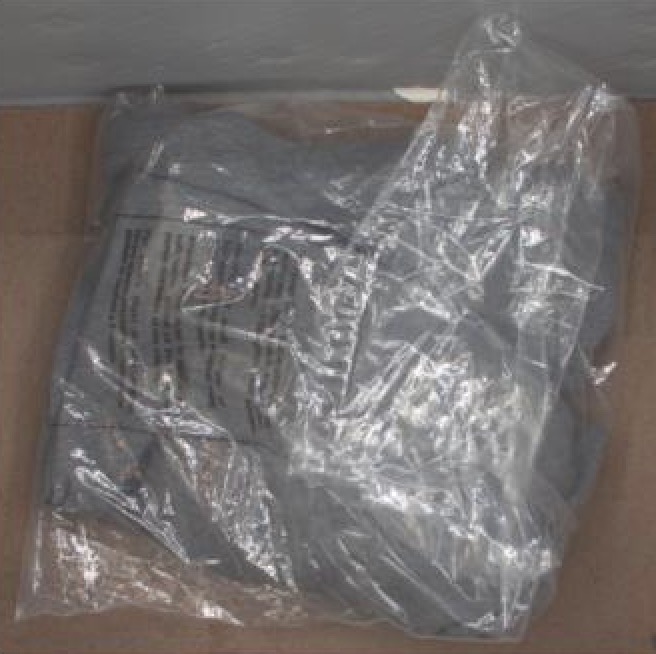}
{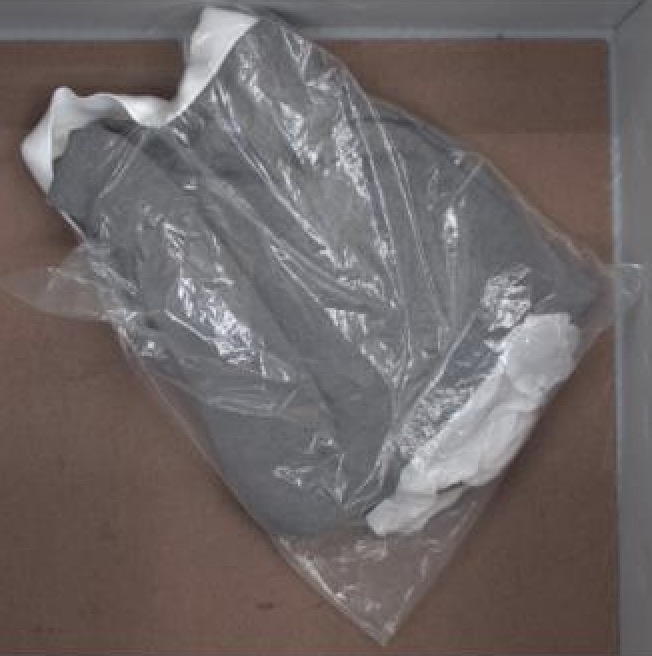}
\imagecellfour{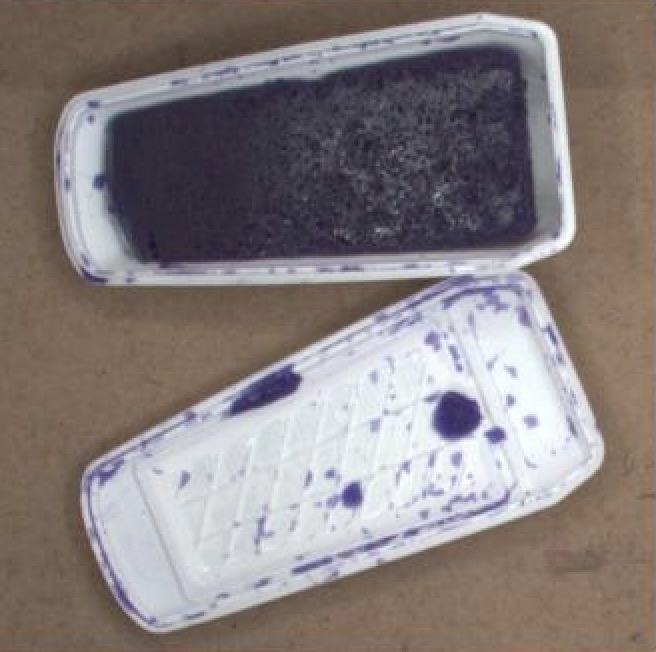}
{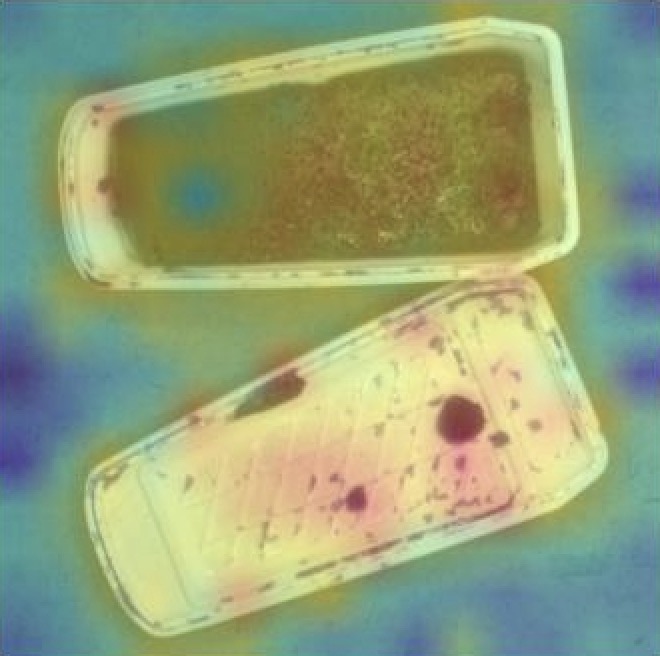}
{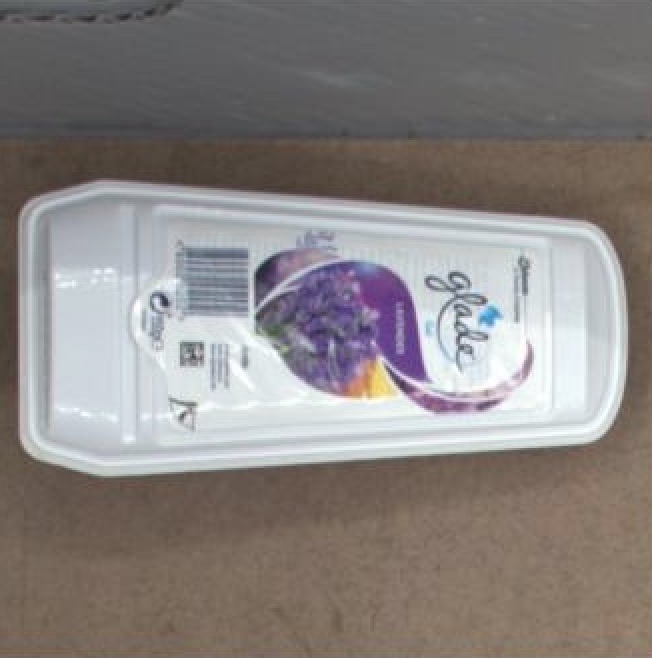}
{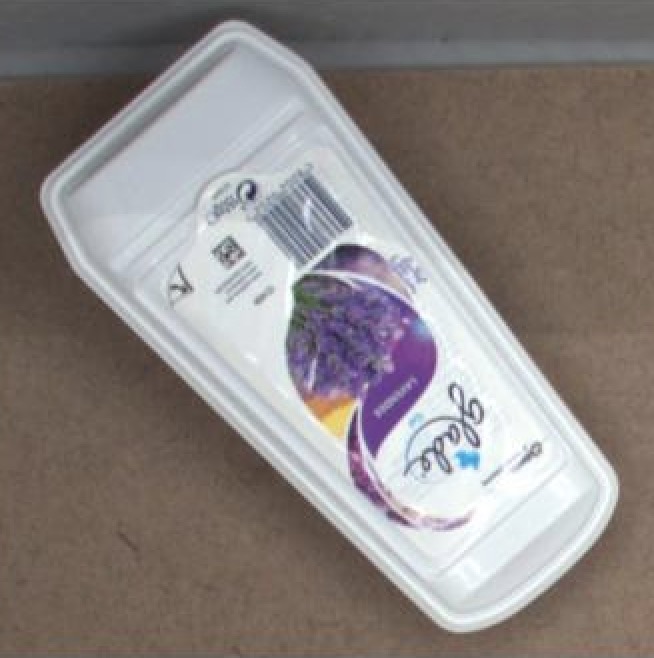}
&
\imagecellfour{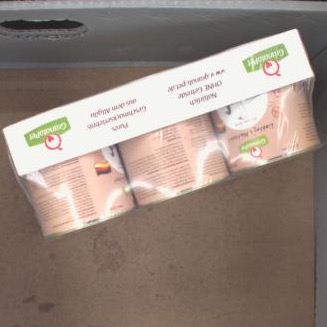}
{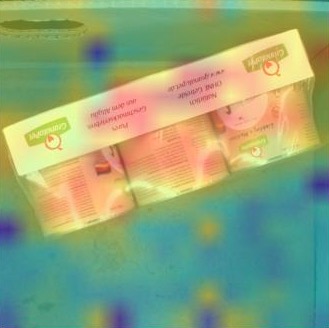}
{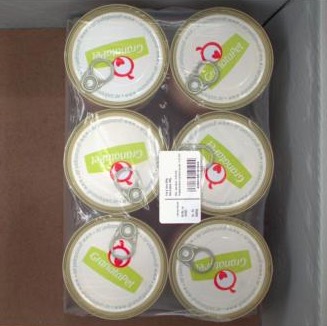}
{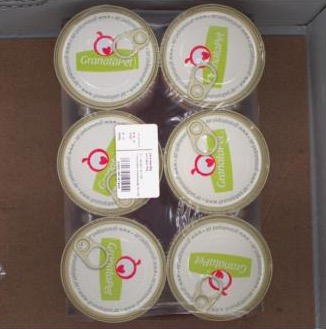}
\imagecellfour{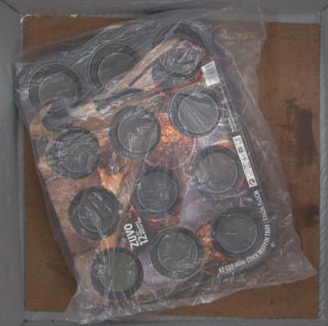}
{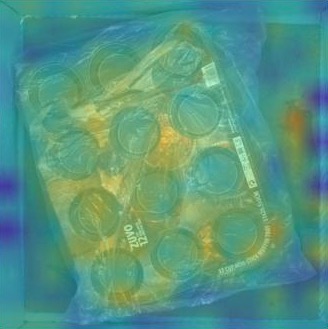}
{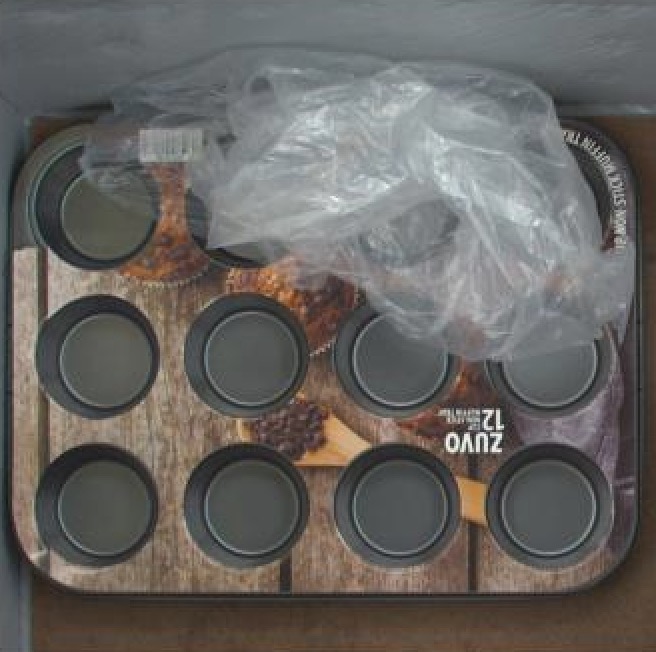}
{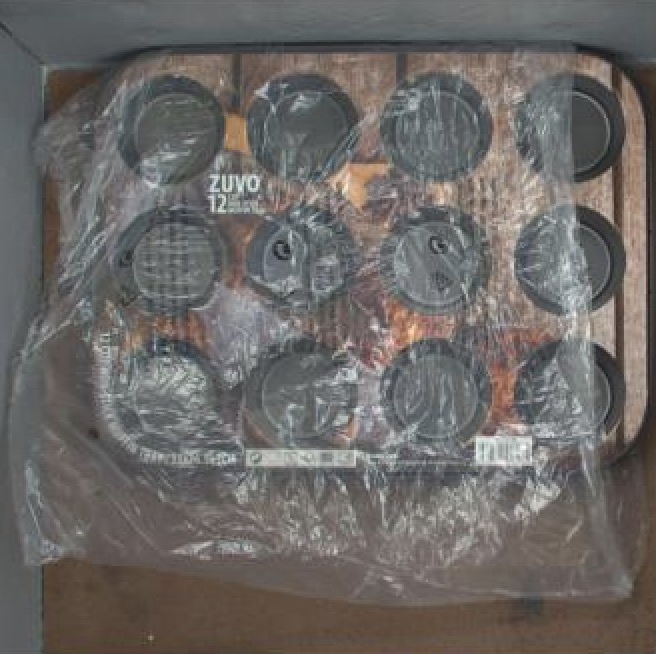}
&
\imagecellfour{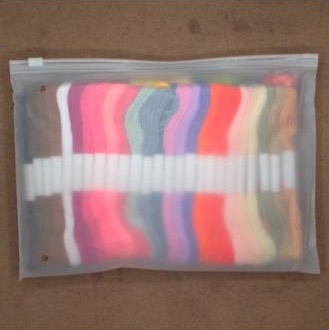}
{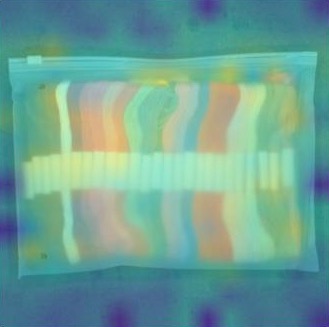}
{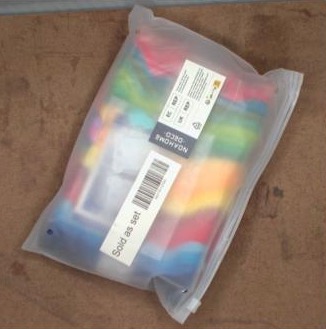}
{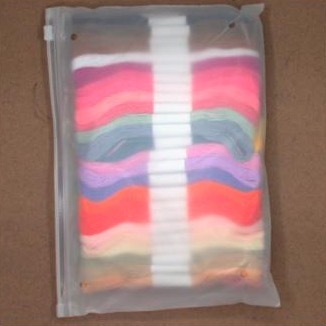}
\imagecellfour{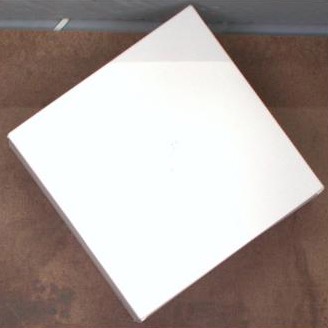}
{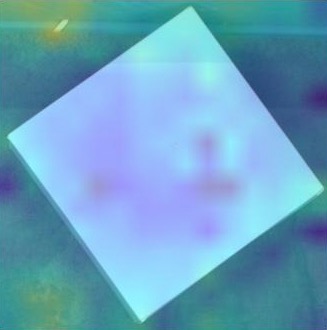}
{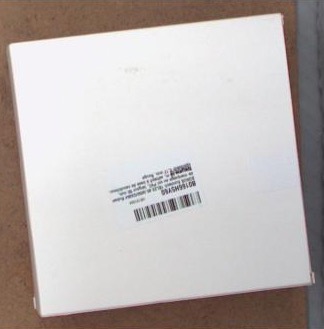}
{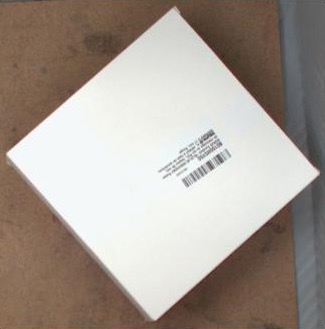}
&
\imagecellfour{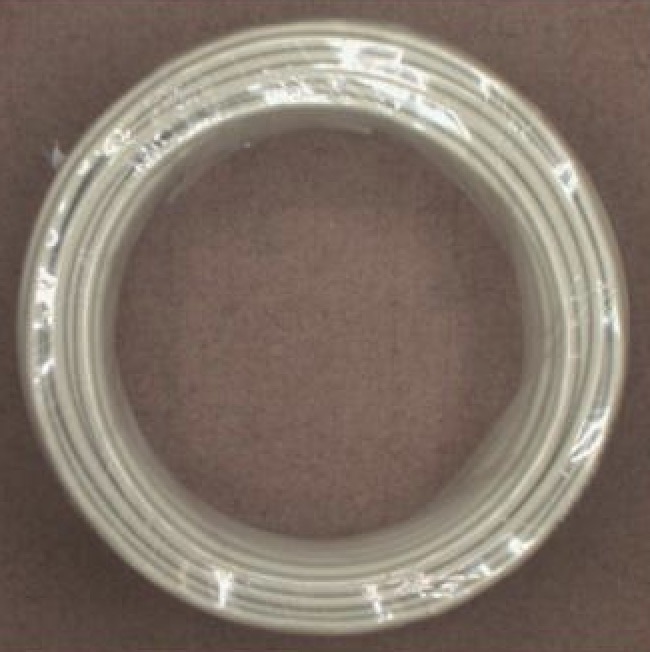}
{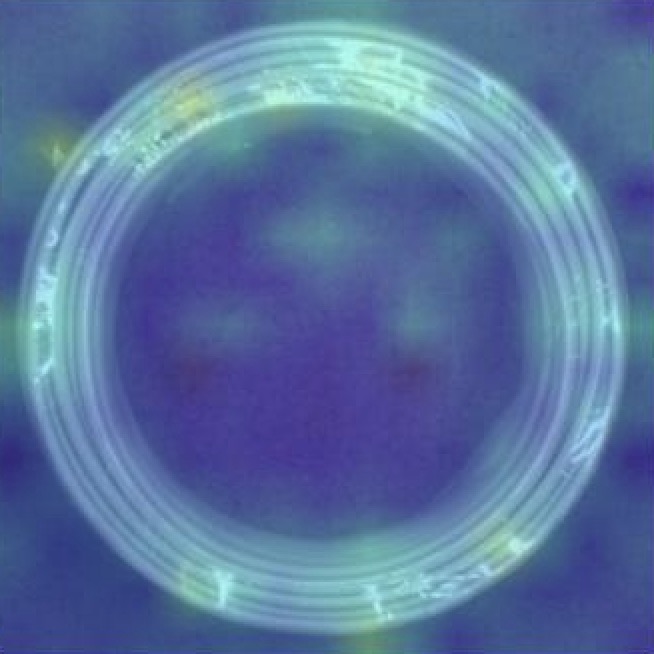}
{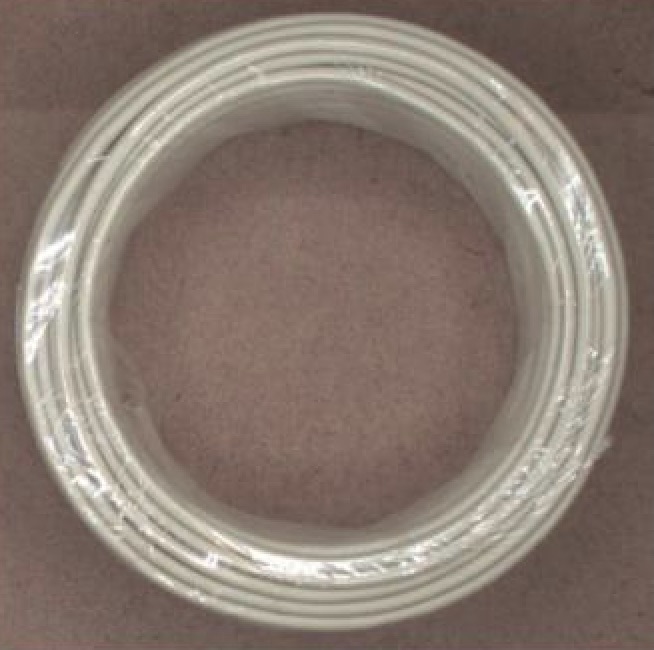}
{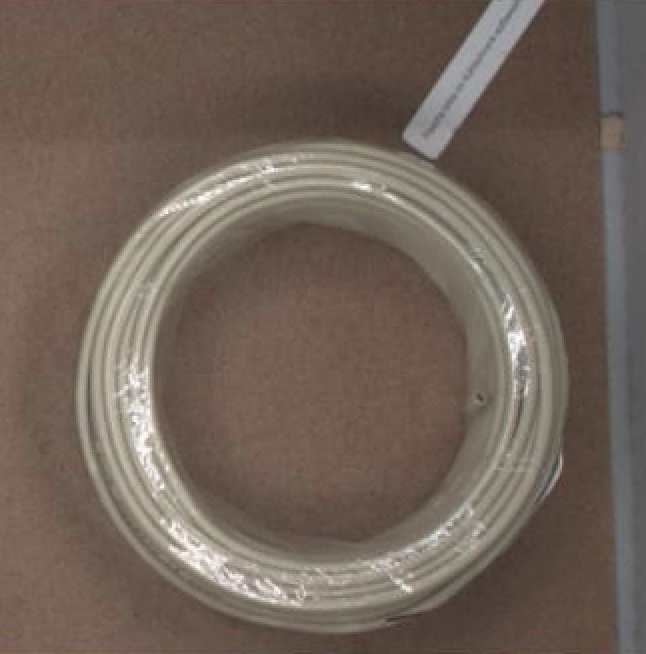}
\imagecellfour{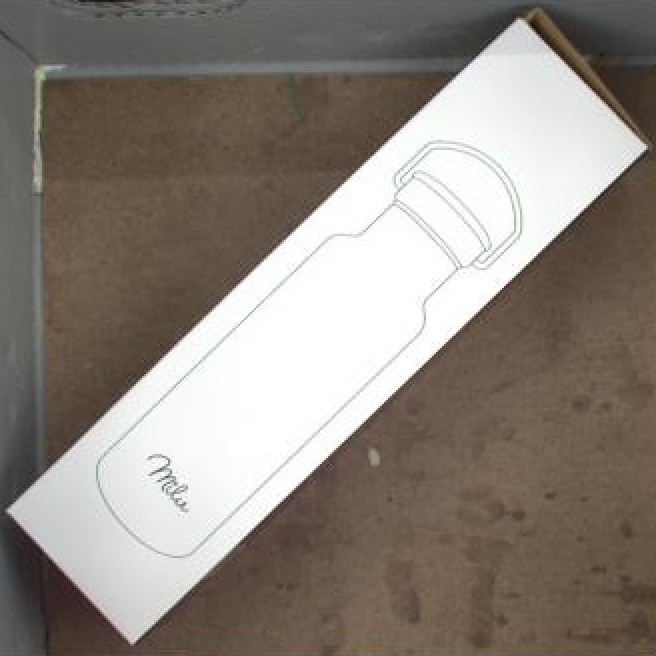}
{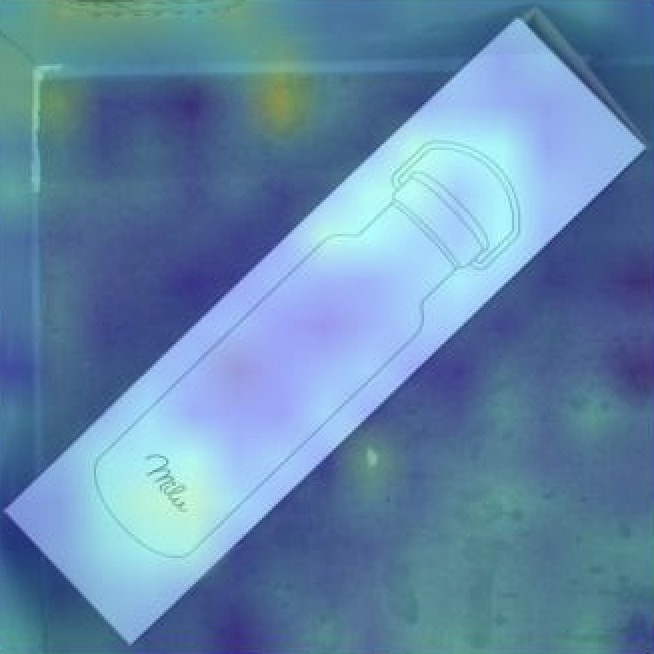}
{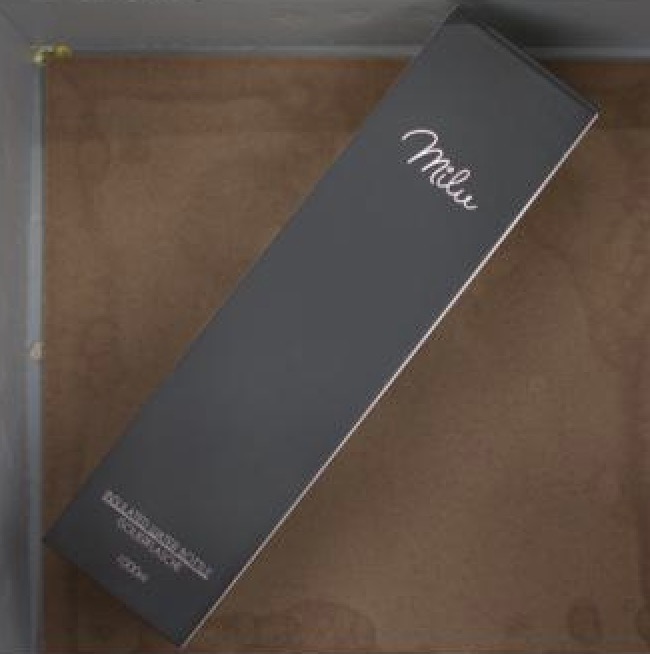}
{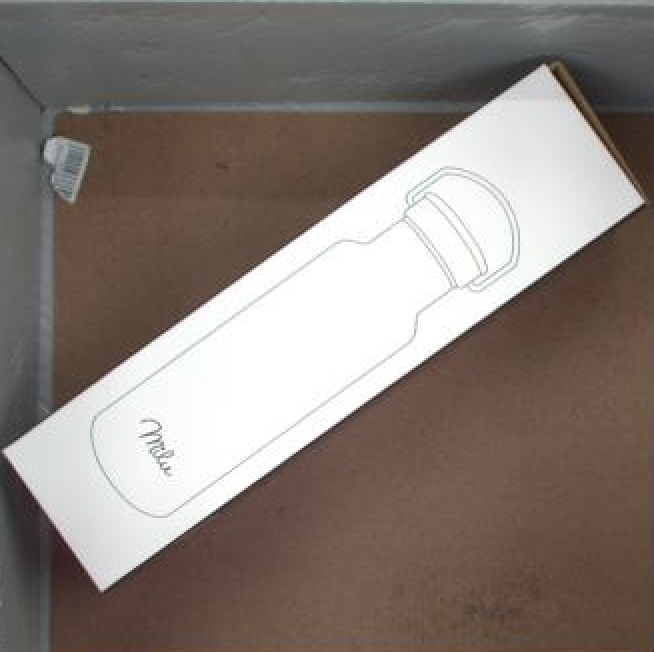}
\\
\bottomrule
\end{tabular}
}
\label{tab:failure_modes_with_training_with_references}
\end{table*}

\begin{table*}
\centering
\caption{Failure and success modes of \texttt{ResNet50}, \texttt{CLIP}, and \texttt{ViT-S} - Those models were trained and evaluated without reference images. Each image shows a different example.}
\resizebox{0.95\textwidth}{!}{
\begin{tabular}{c|>{\centering\arraybackslash}p{3cm}|>{\centering\arraybackslash}p{3cm}|>{\centering\arraybackslash}p{3cm}|>{\centering\arraybackslash}p{3cm}}
\toprule
 & TP & FP & TN & FN \\
\midrule
\multirow{1}{*}{\rotatebox{90}{\parbox{1cm}{\centering \texttt{ResNet50}}}} &

\imagecell{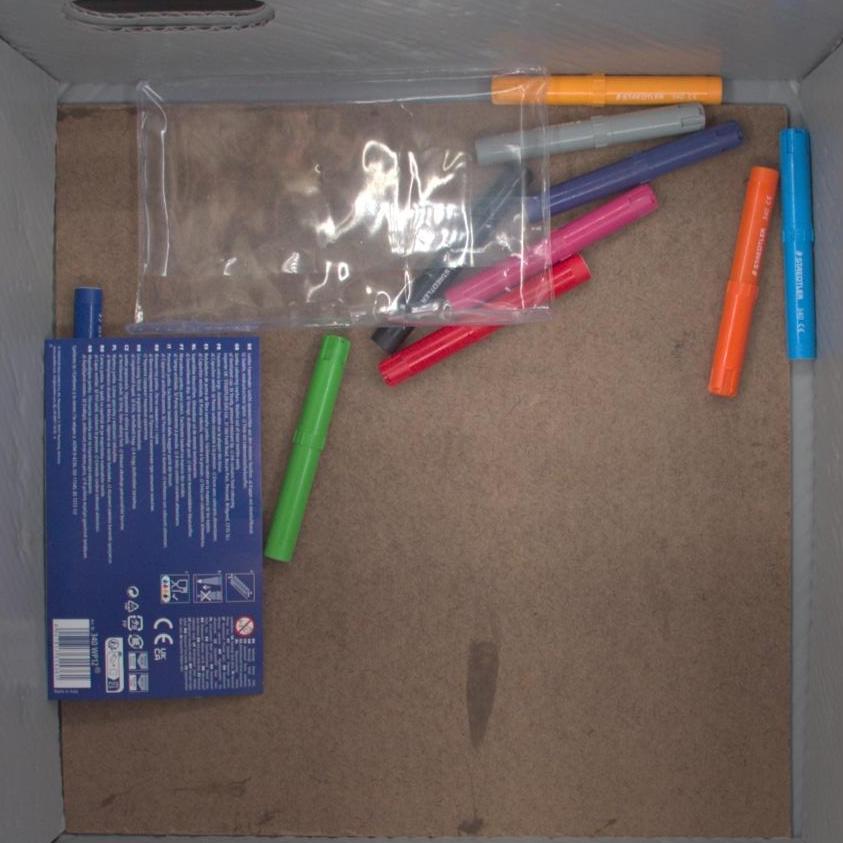}
{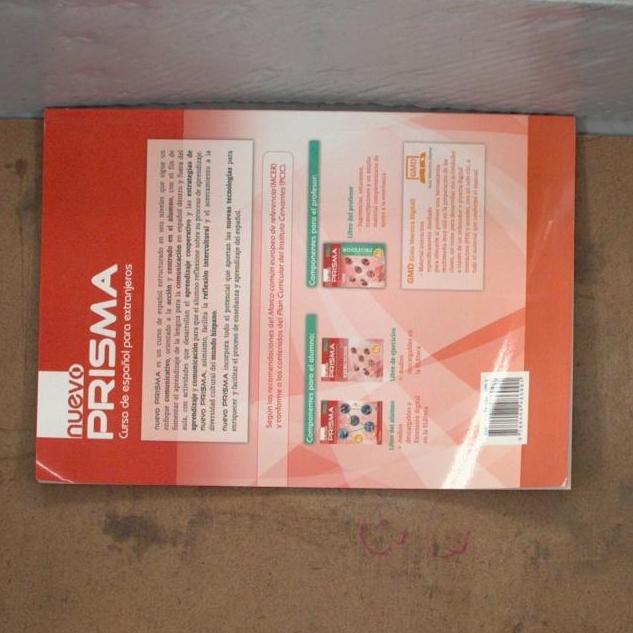}
{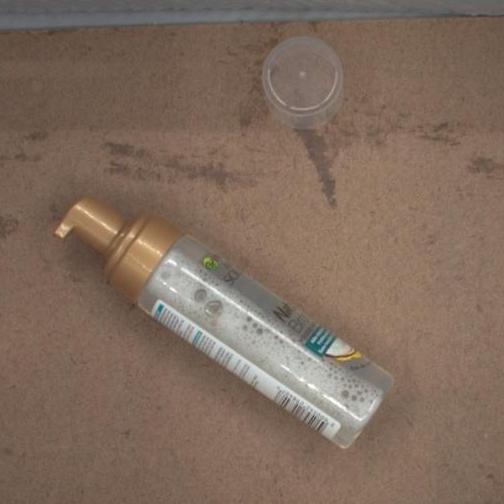}

&
\imagecell{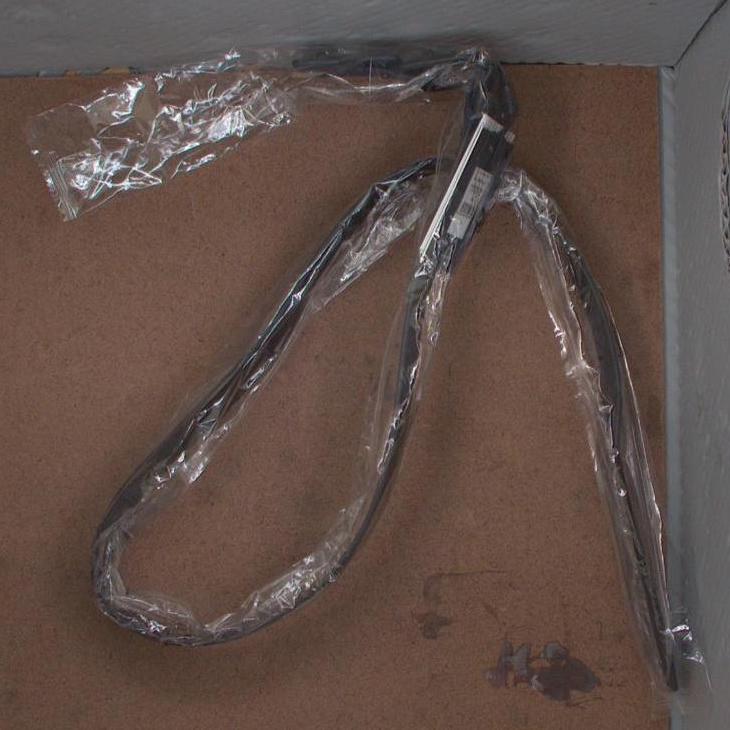}
{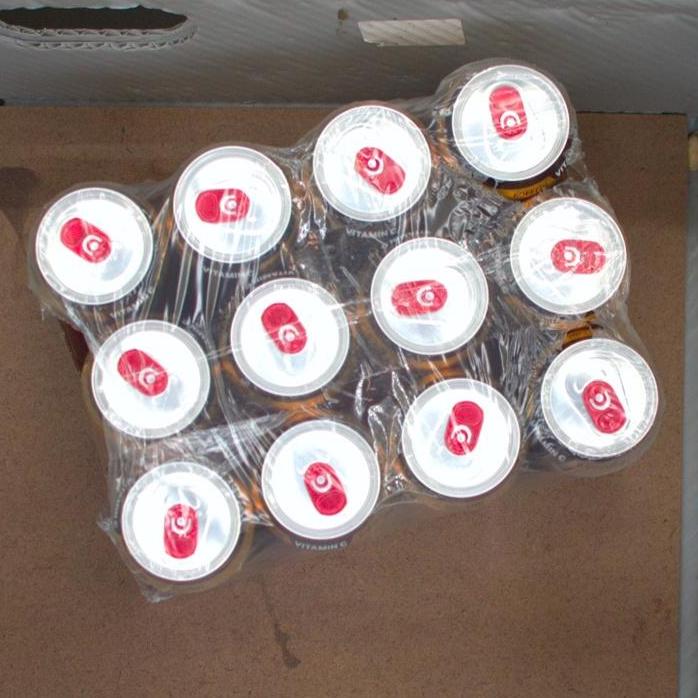}
{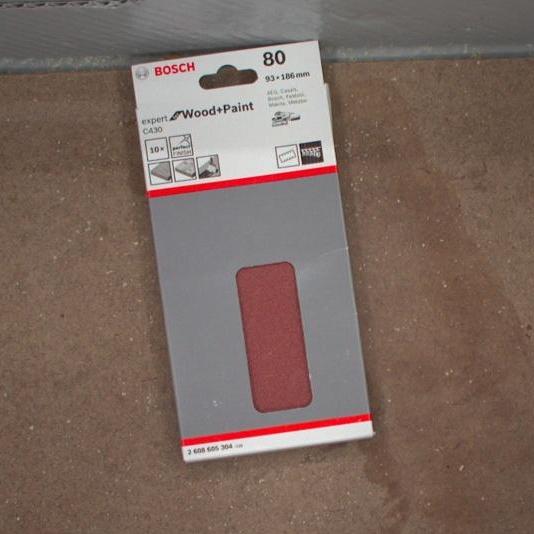}

&
\imagecell{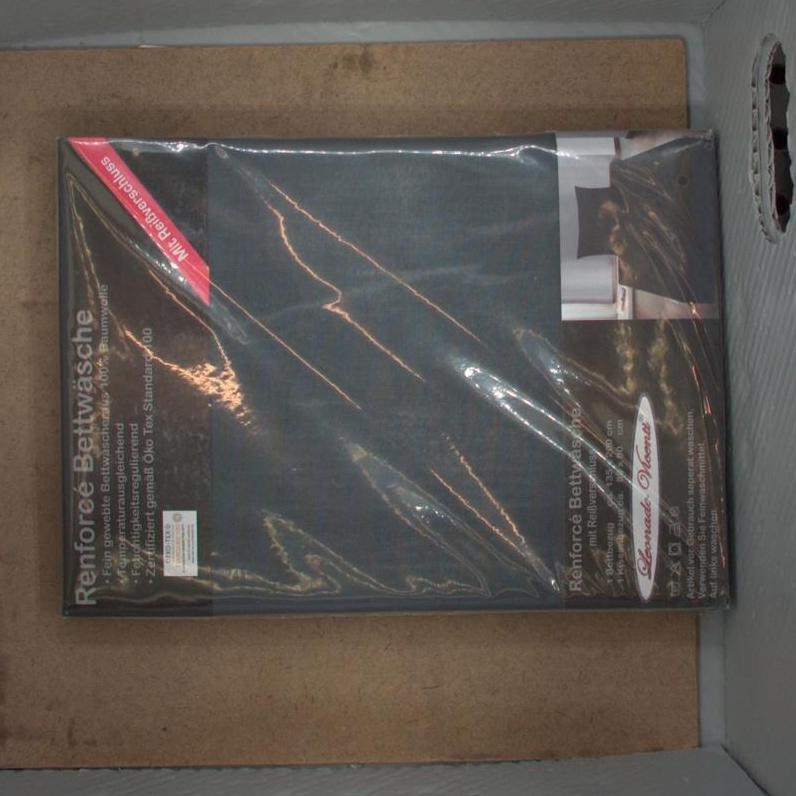}
{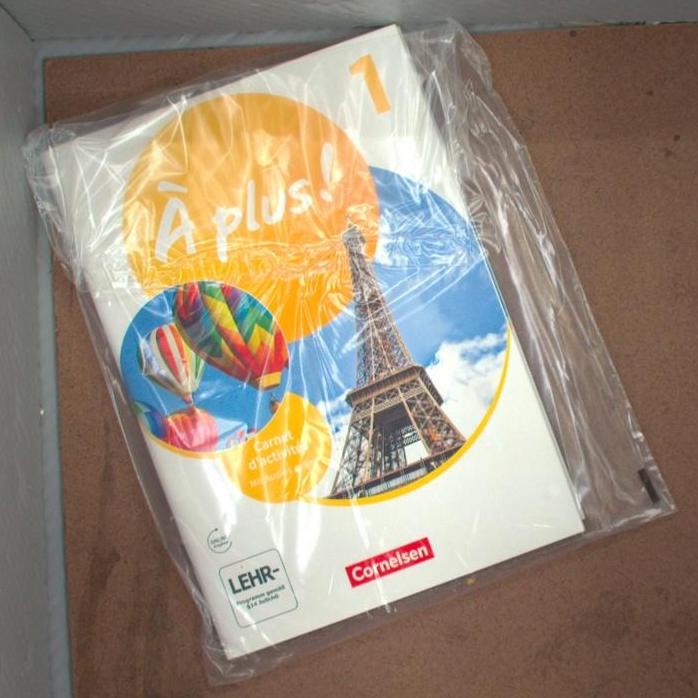}
{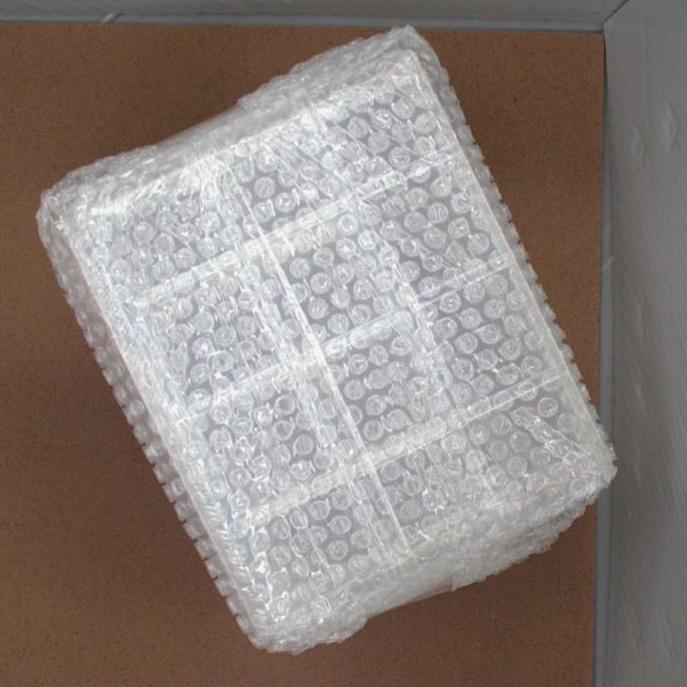}

&
\imagecell{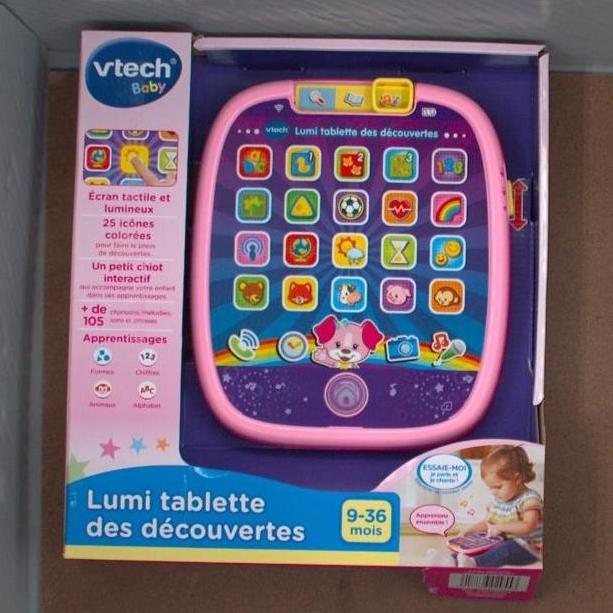}
{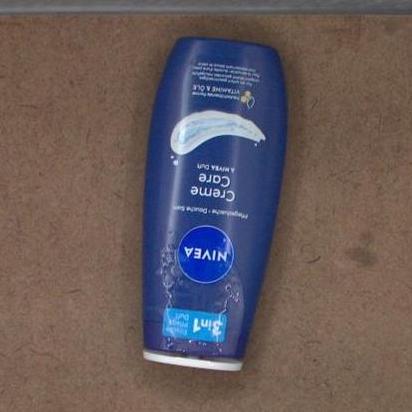}
{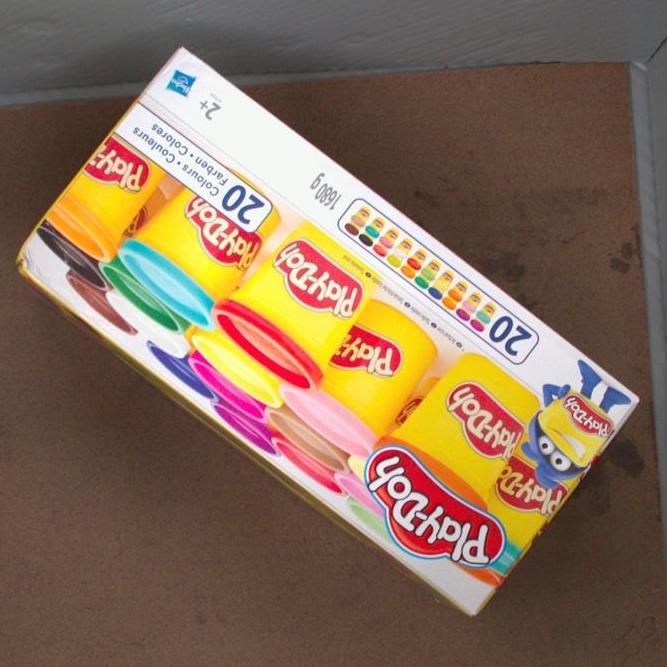}

\\
\midrule

\multirow{1}{*}{\rotatebox{90}{\parbox{1cm}{\centering \texttt{CLIP}}}} &

\imagecellfour{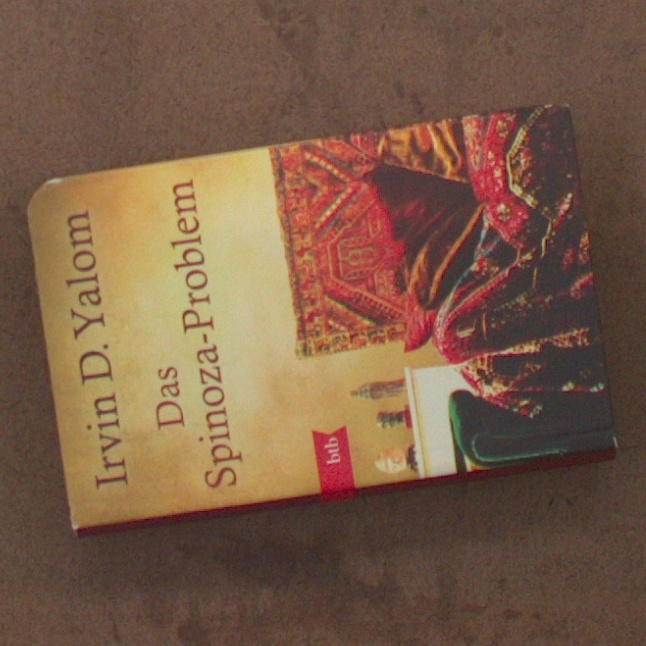}
{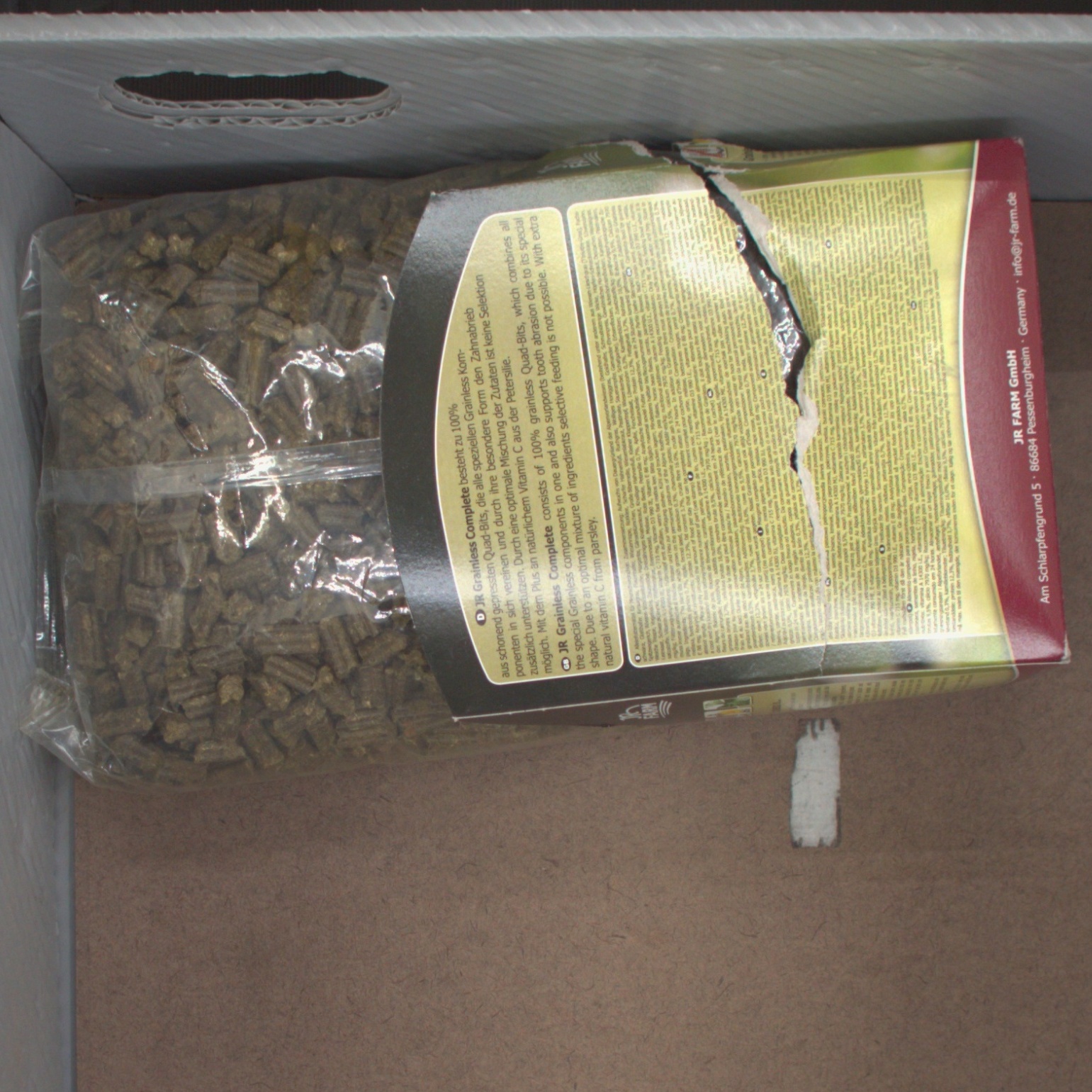}
{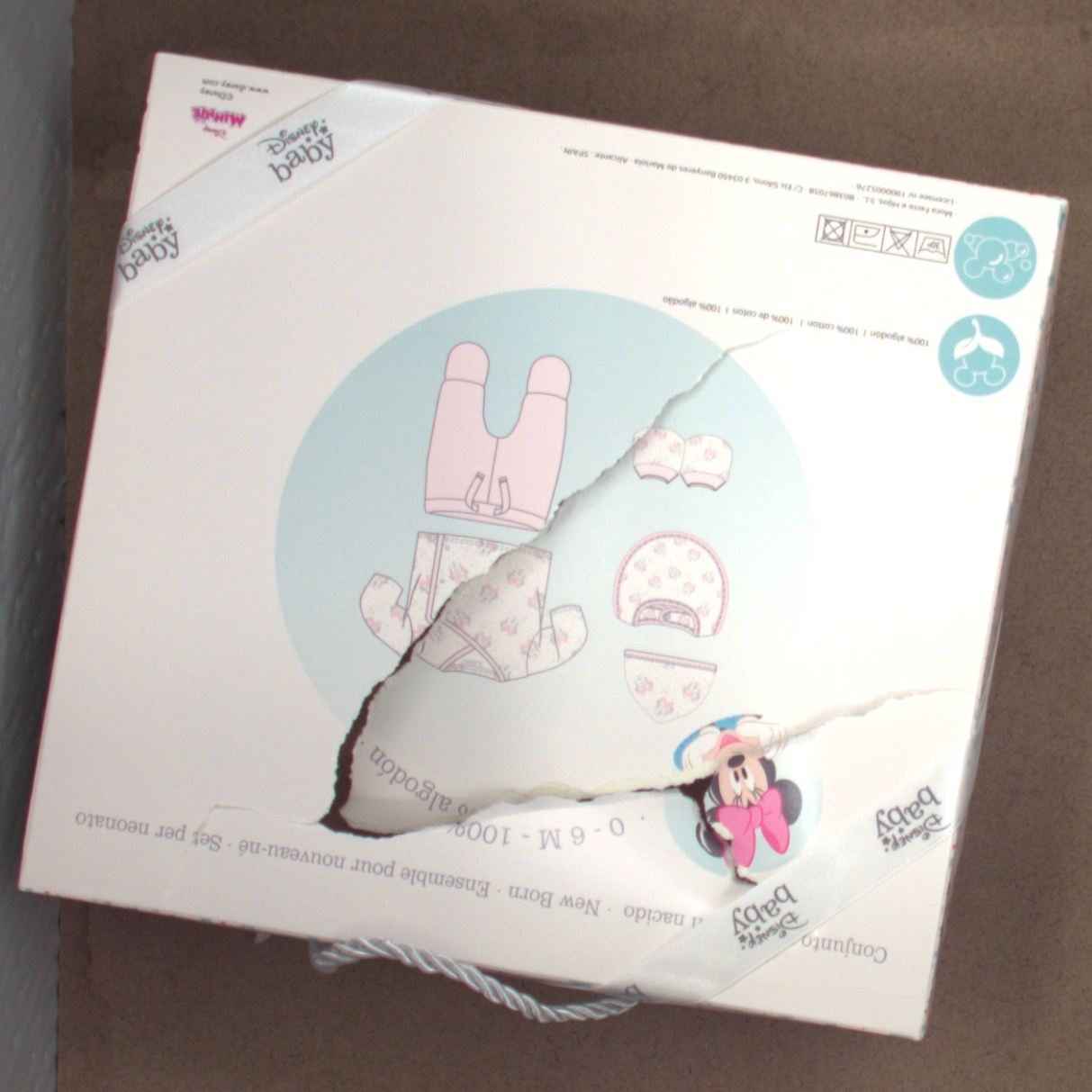}
{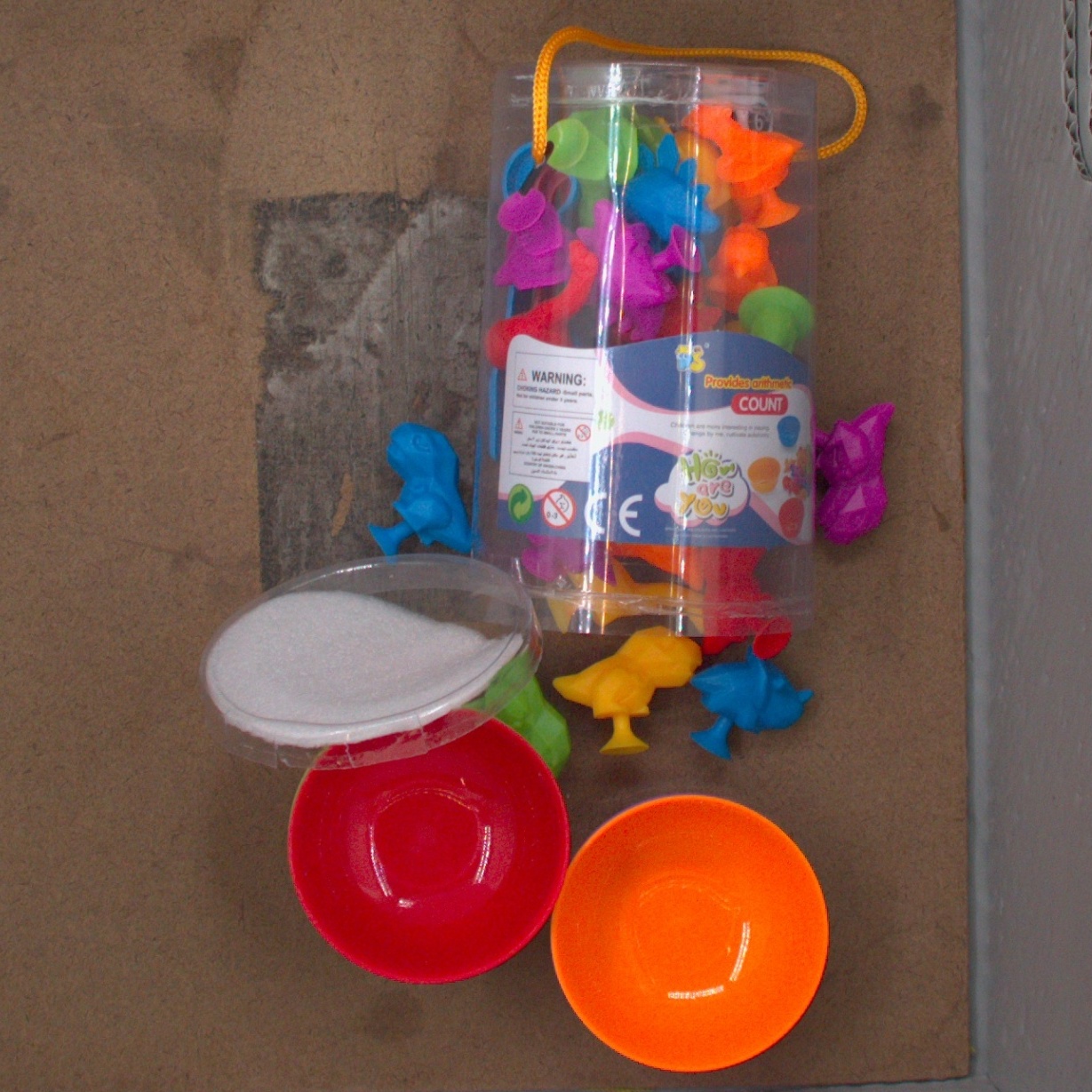}
&
\imagecellfour{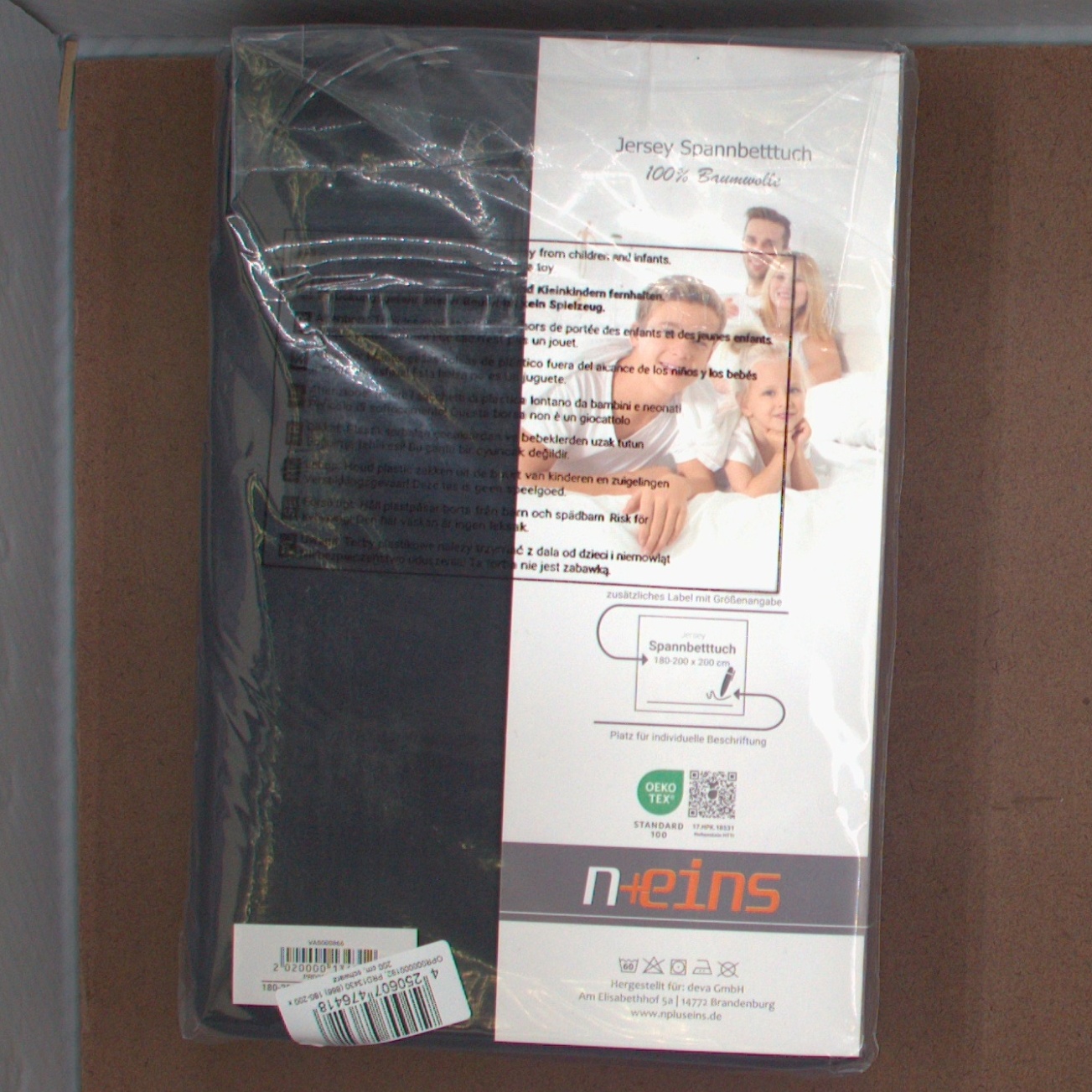}
{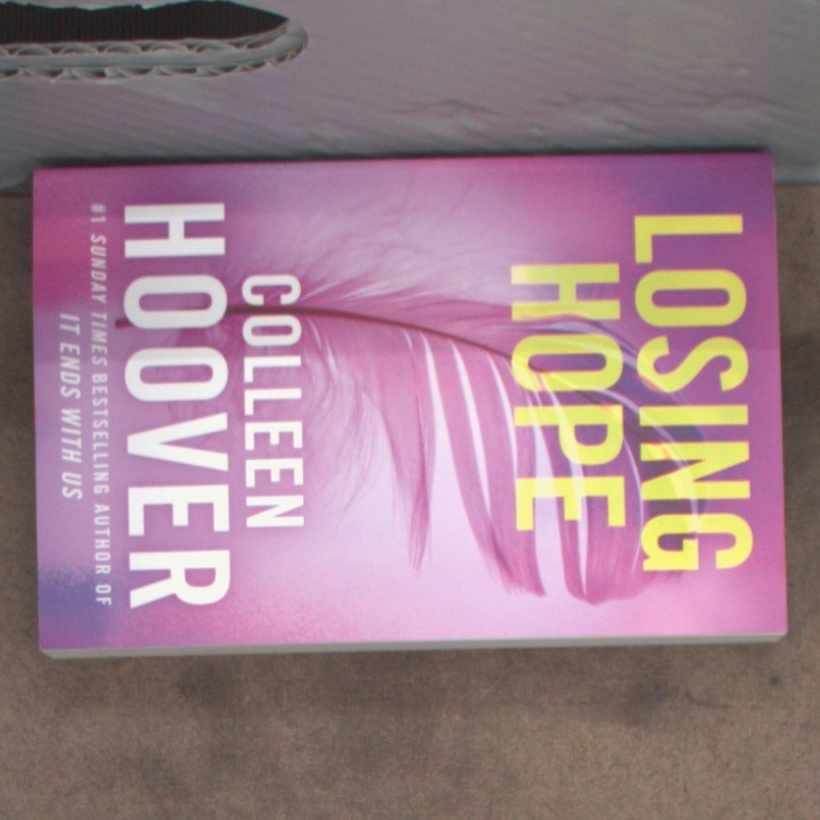}
{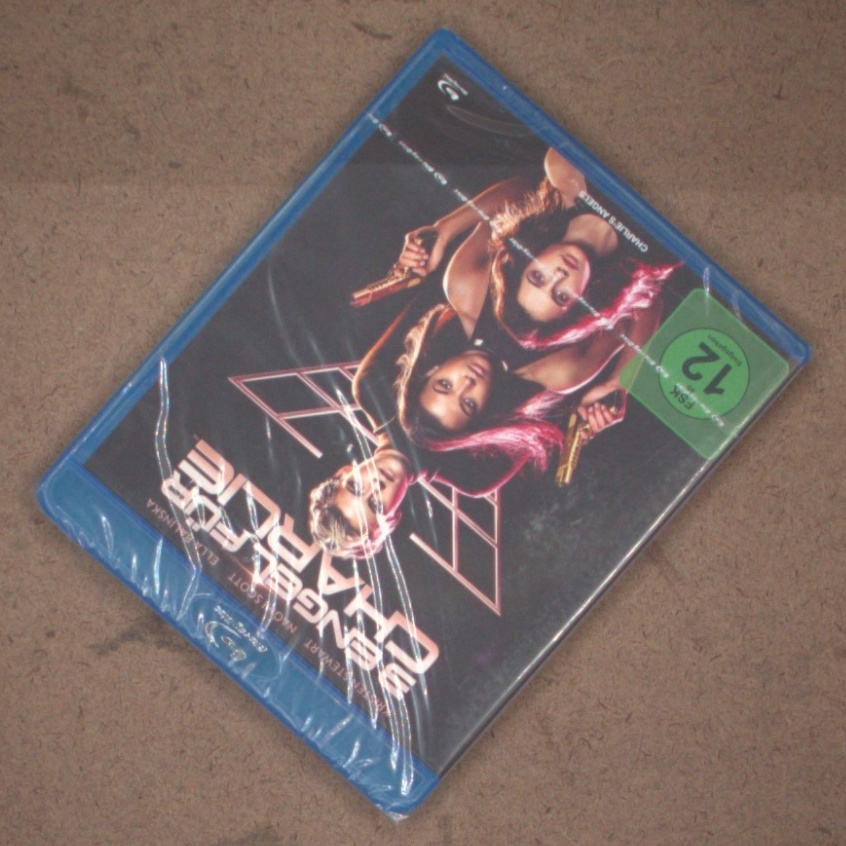}
{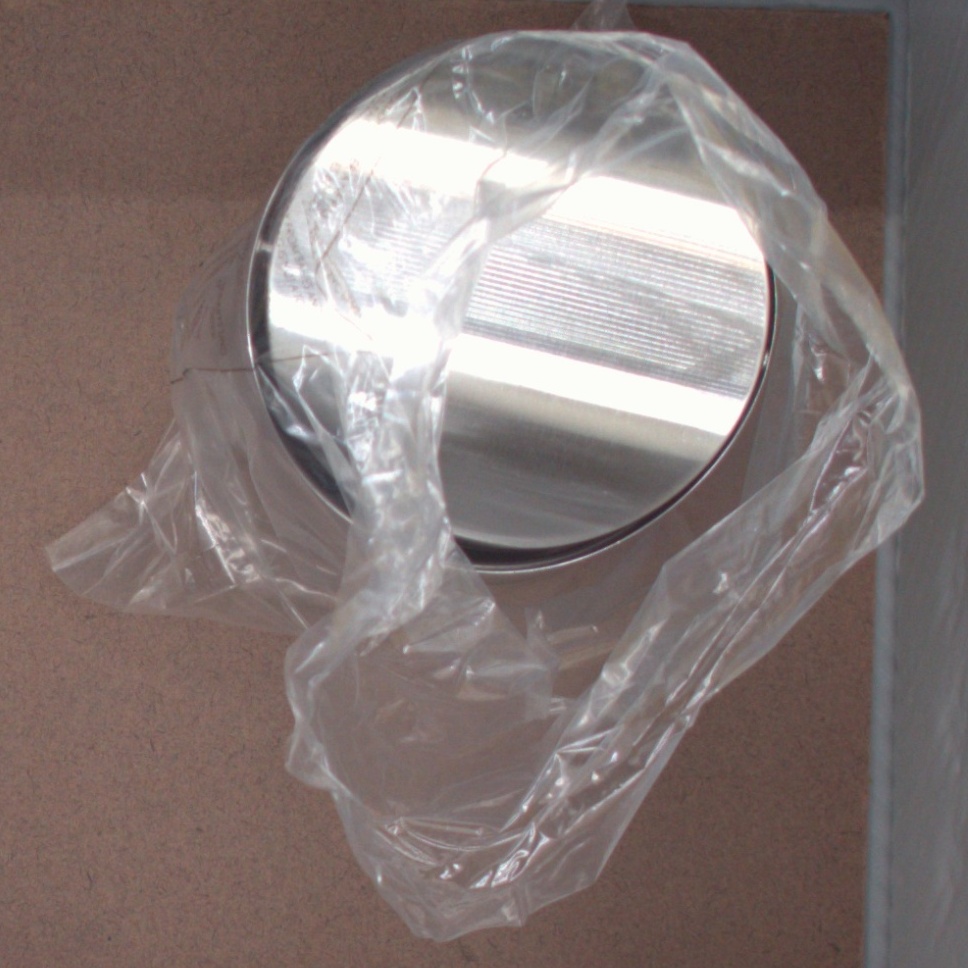}
&
\imagecellfour{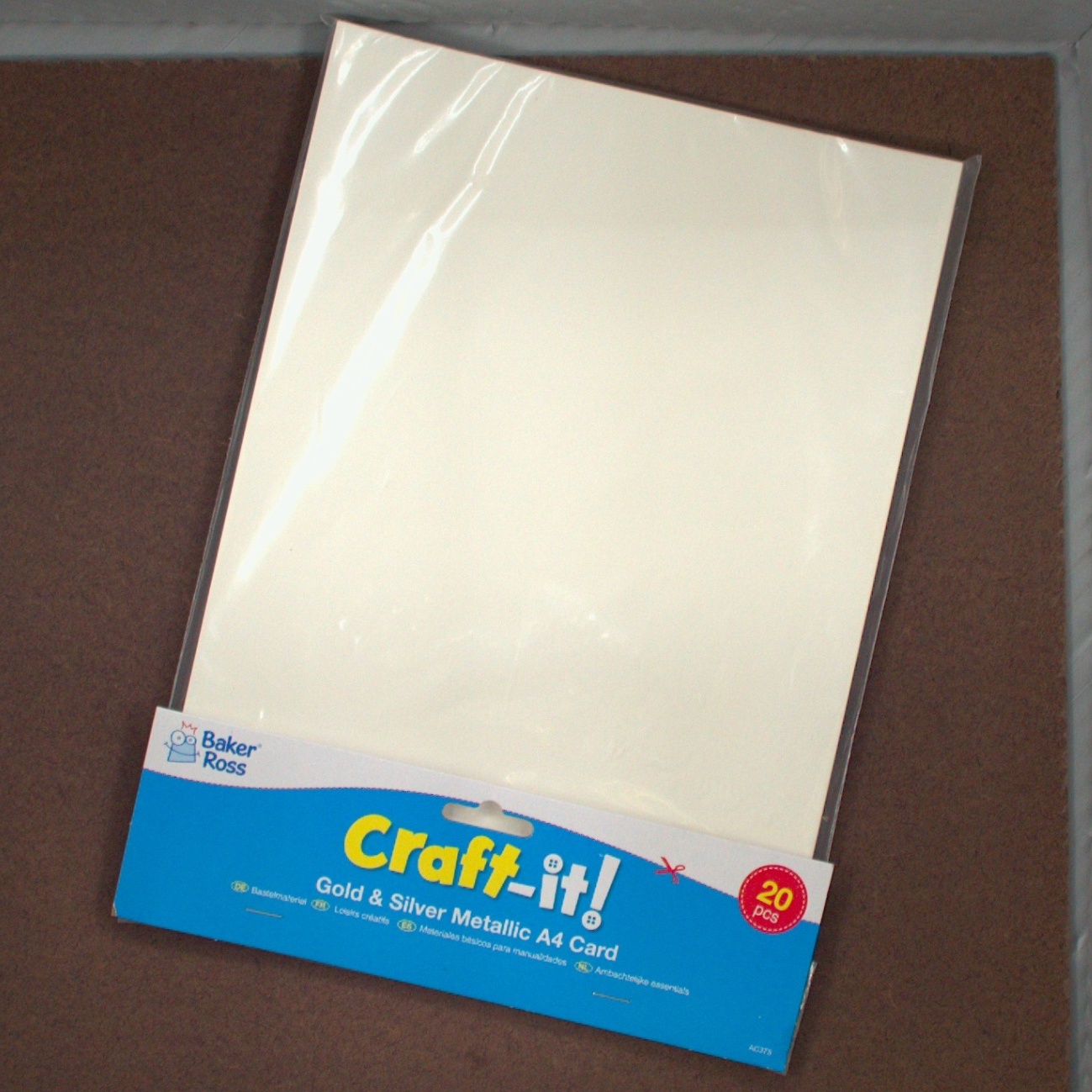}
{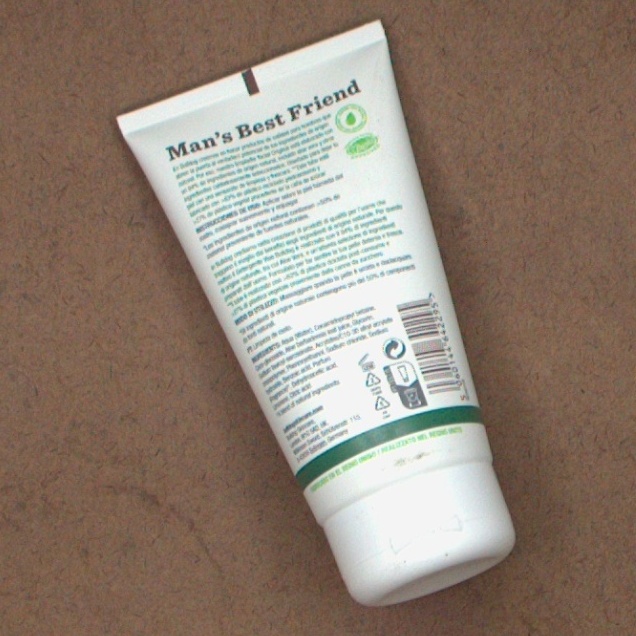}
{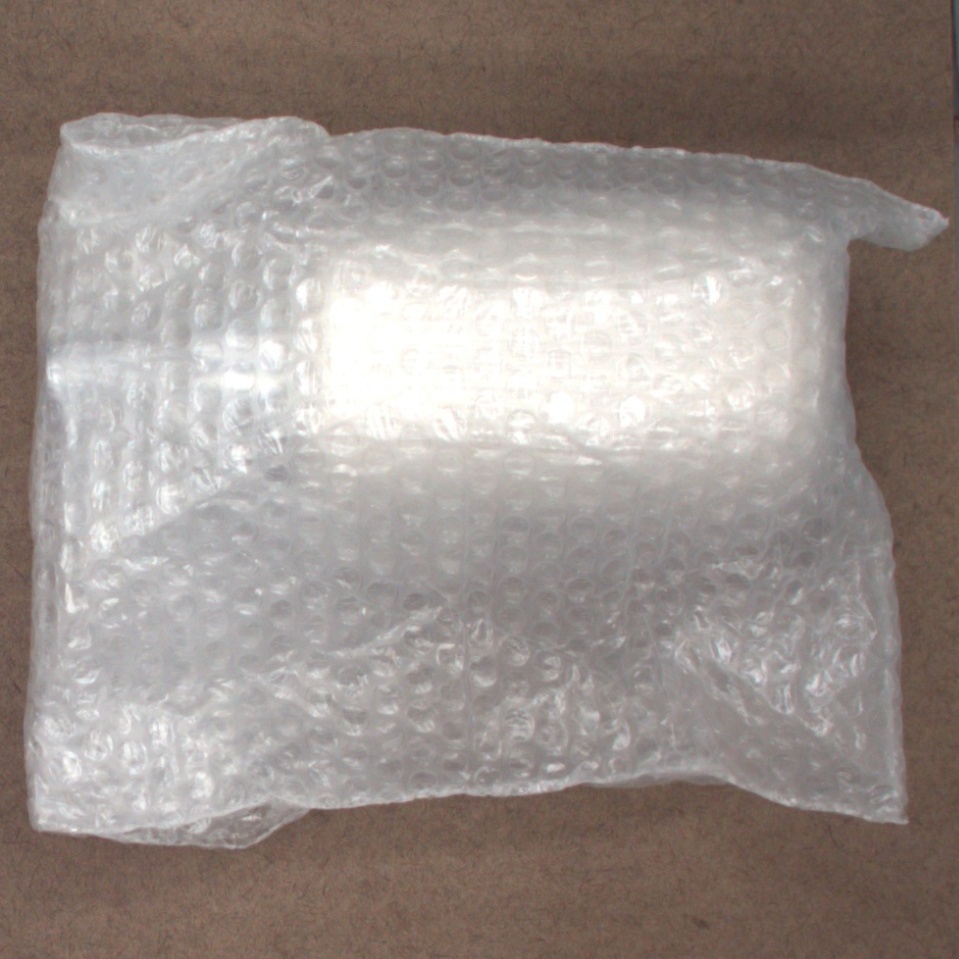}
{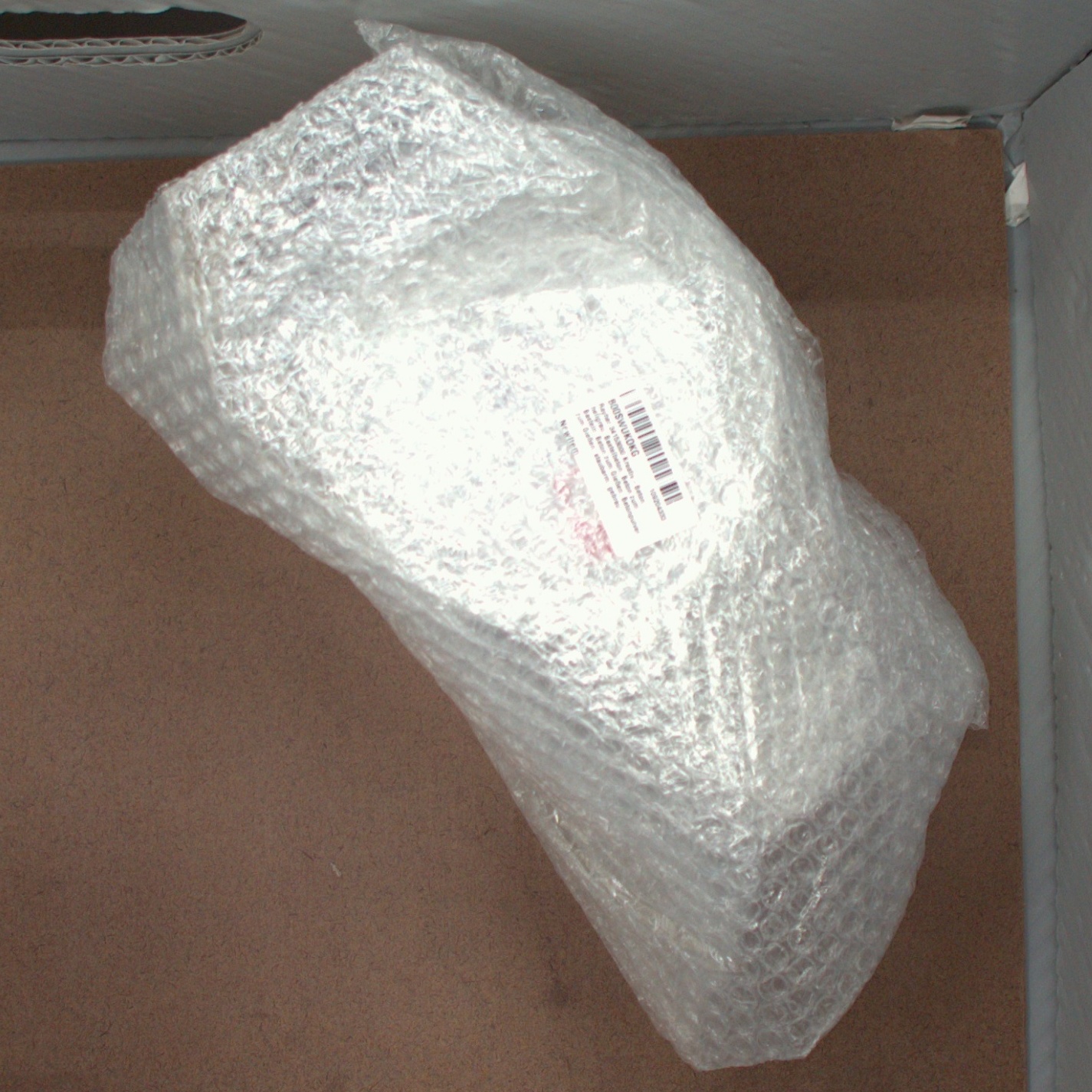}
&
\imagecellfour{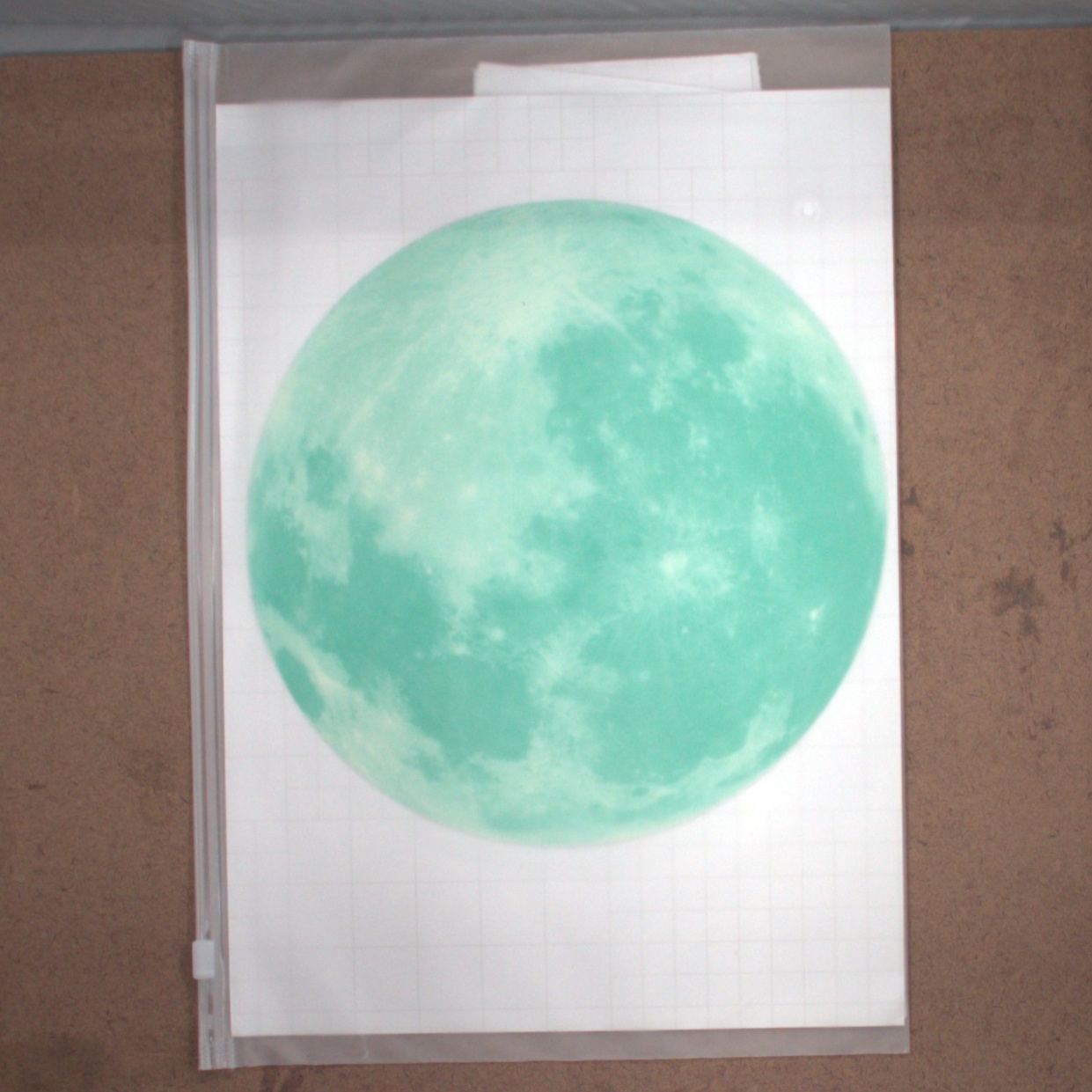}
{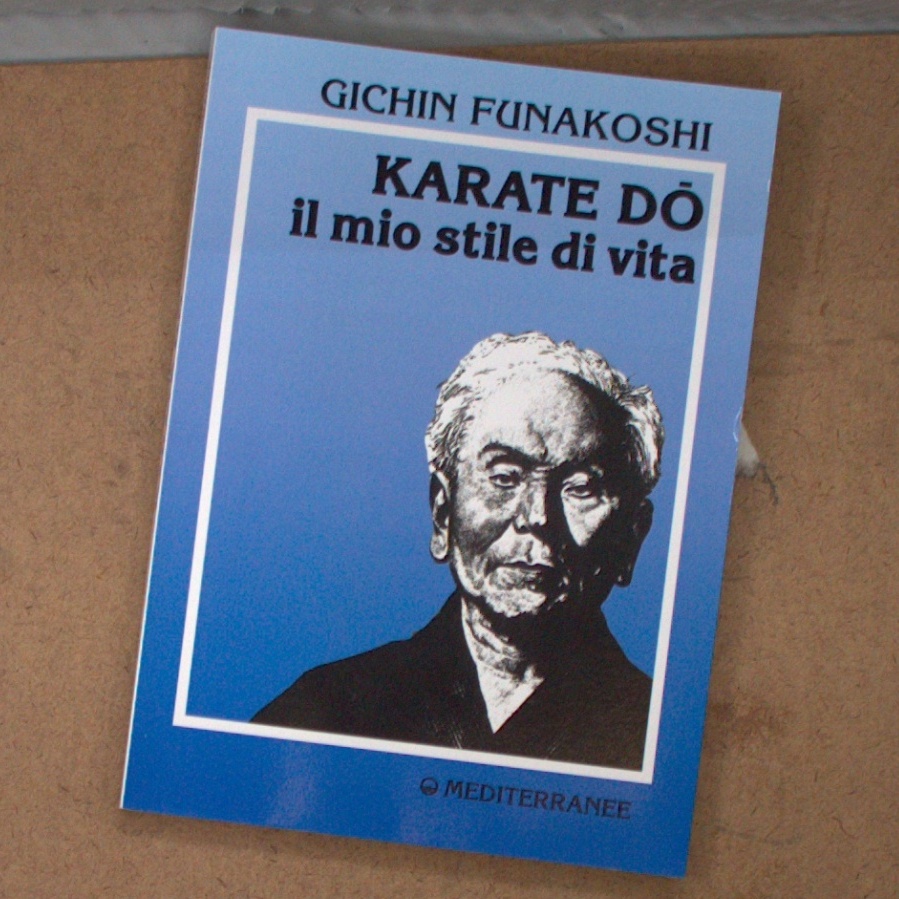}
{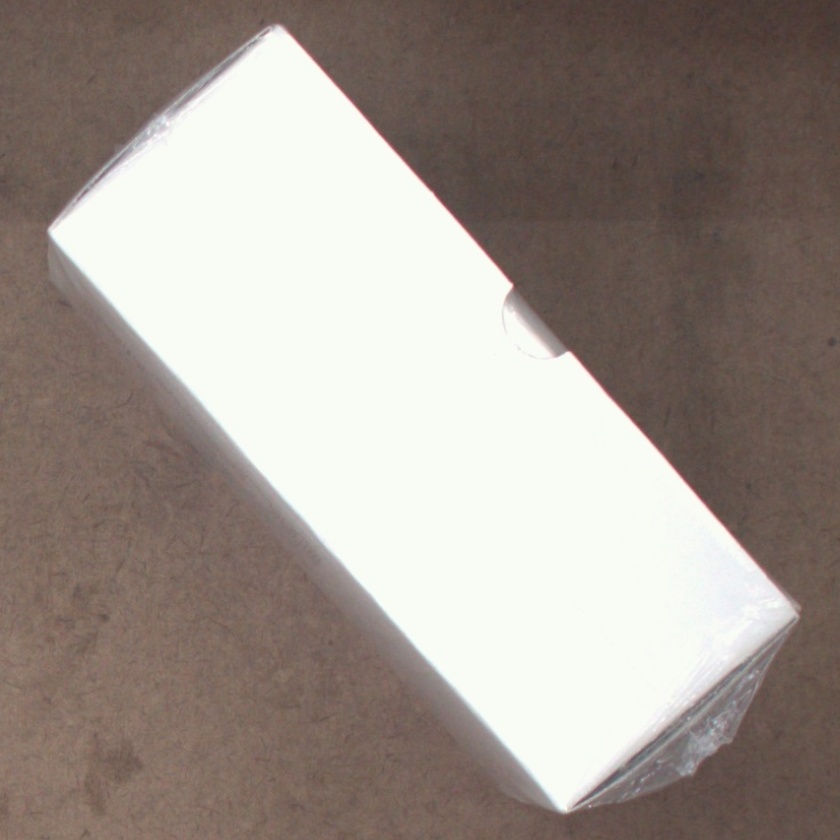}
{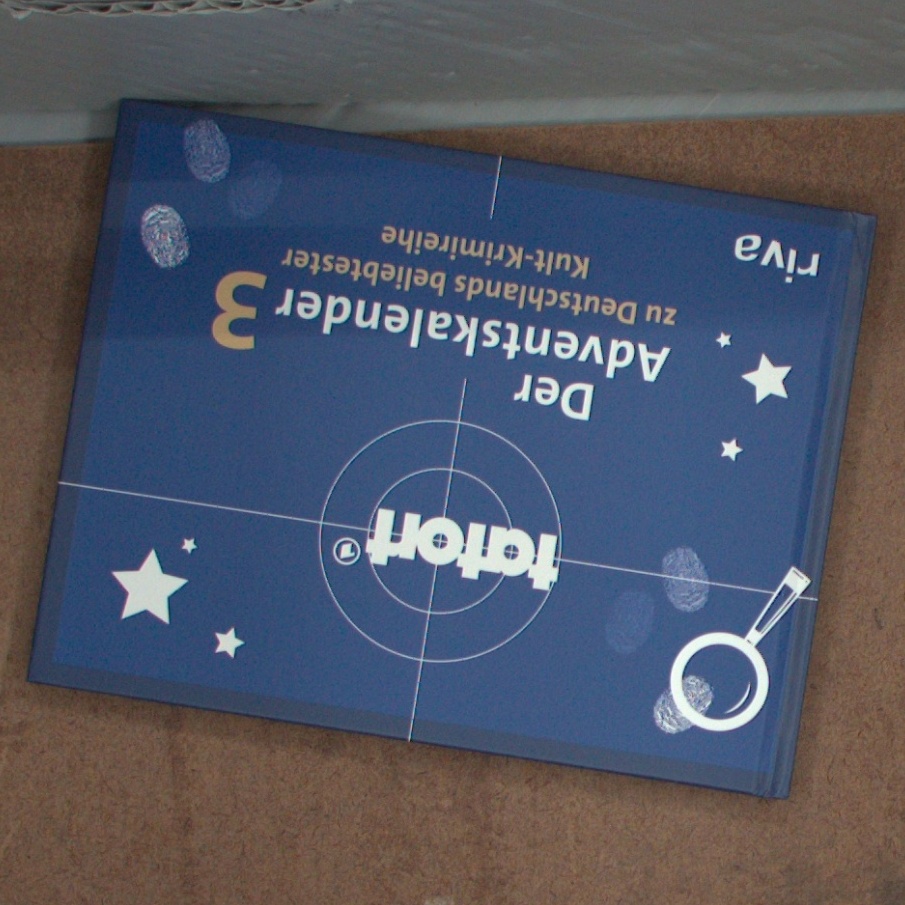}
\\
\midrule
\multirow{1}{*}{\rotatebox{90}{\parbox{1cm}{\centering \texttt{ViT-S}}}} &

\imagecellfour{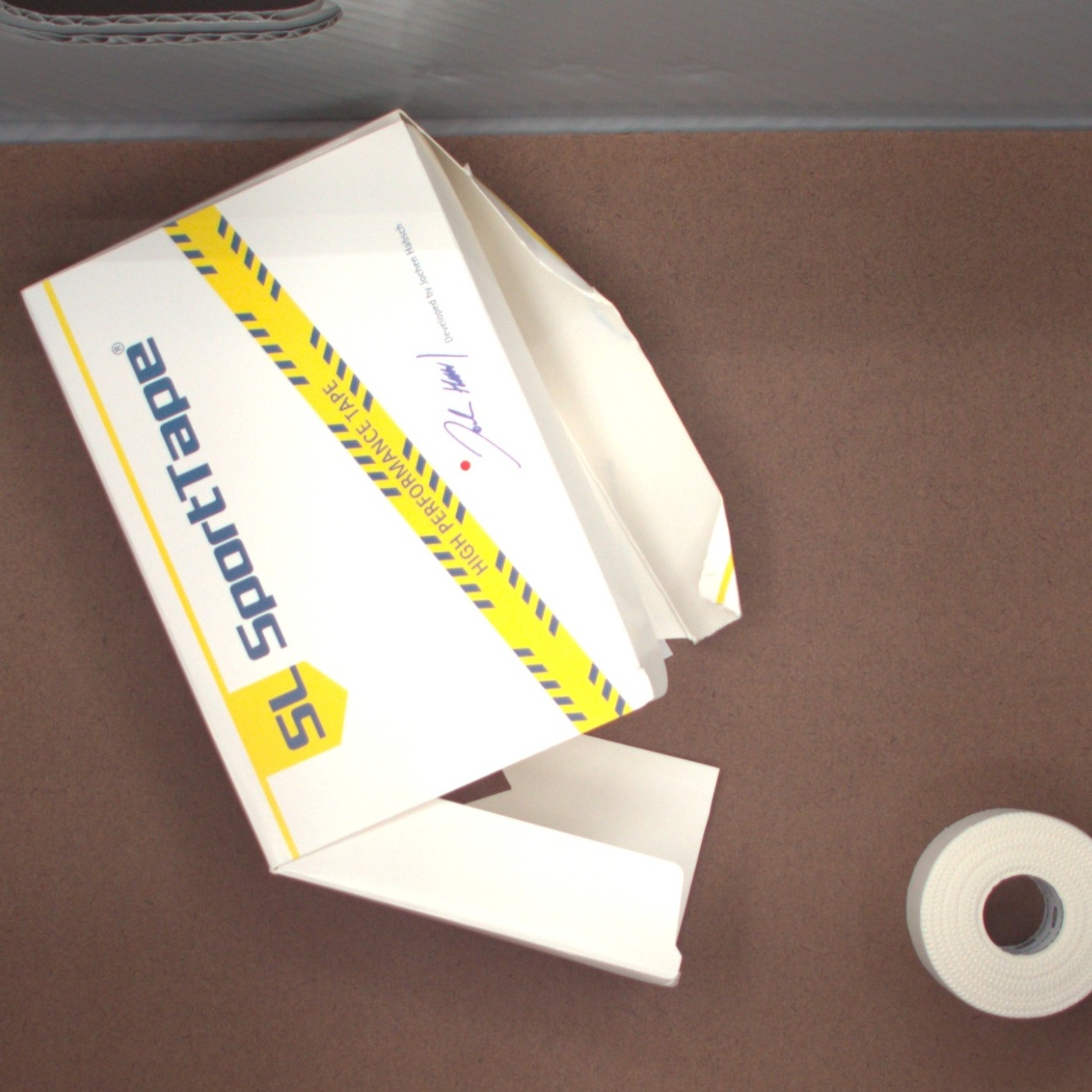}
{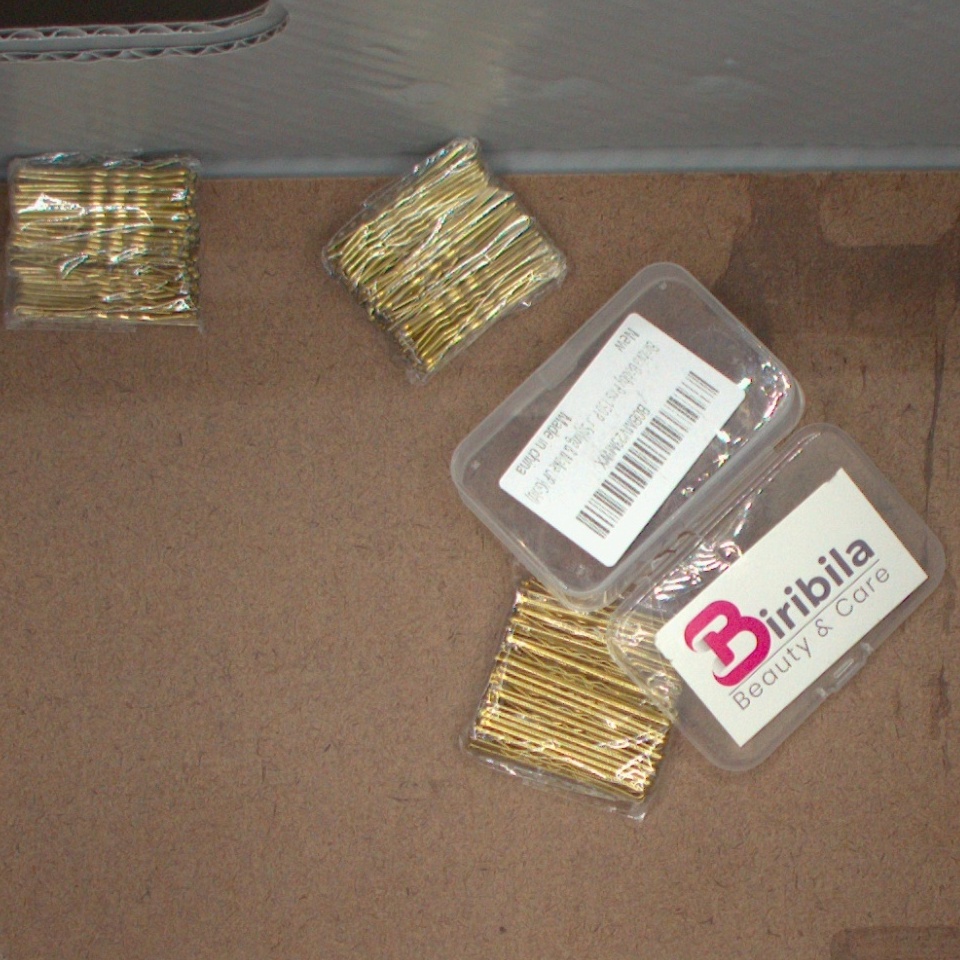}
{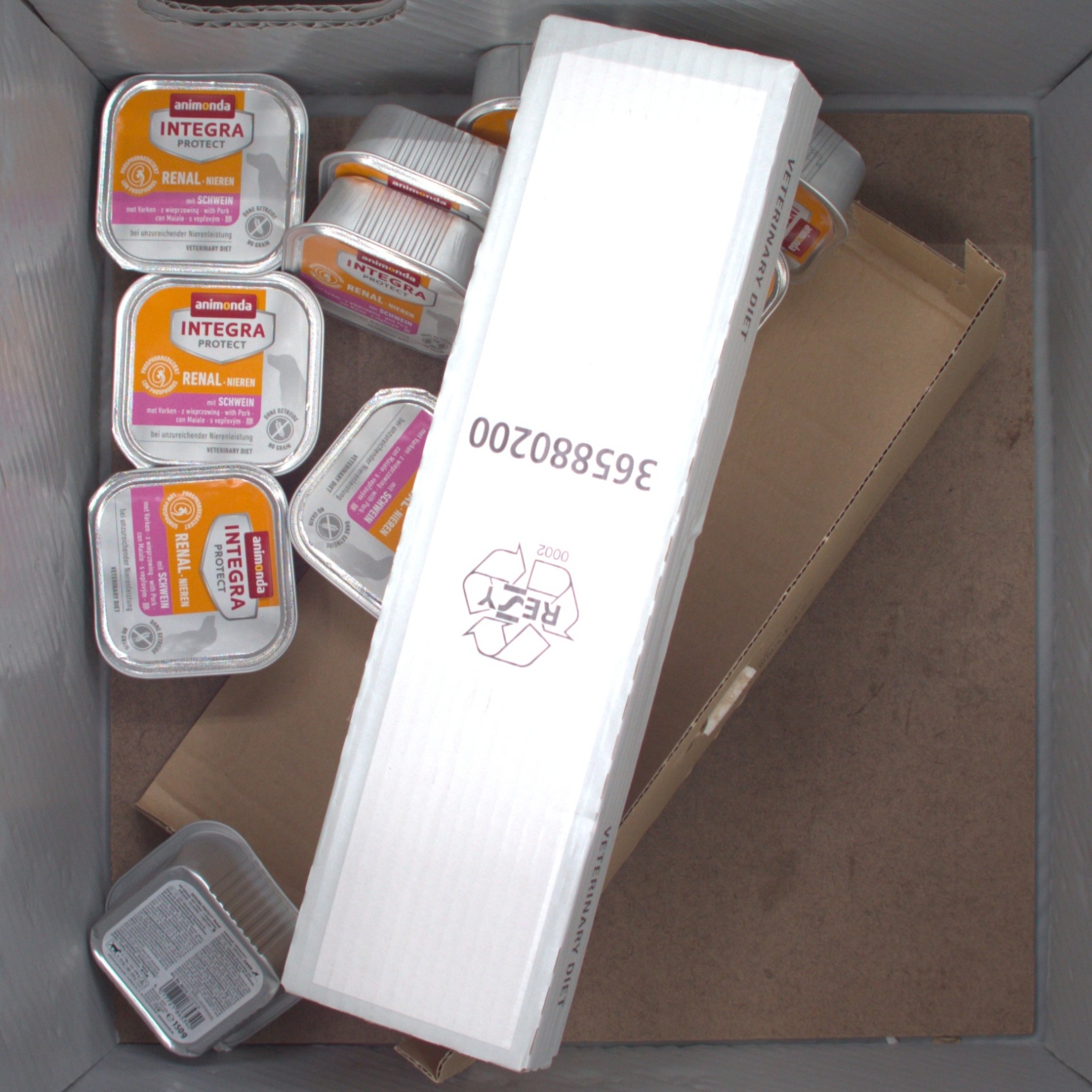}
{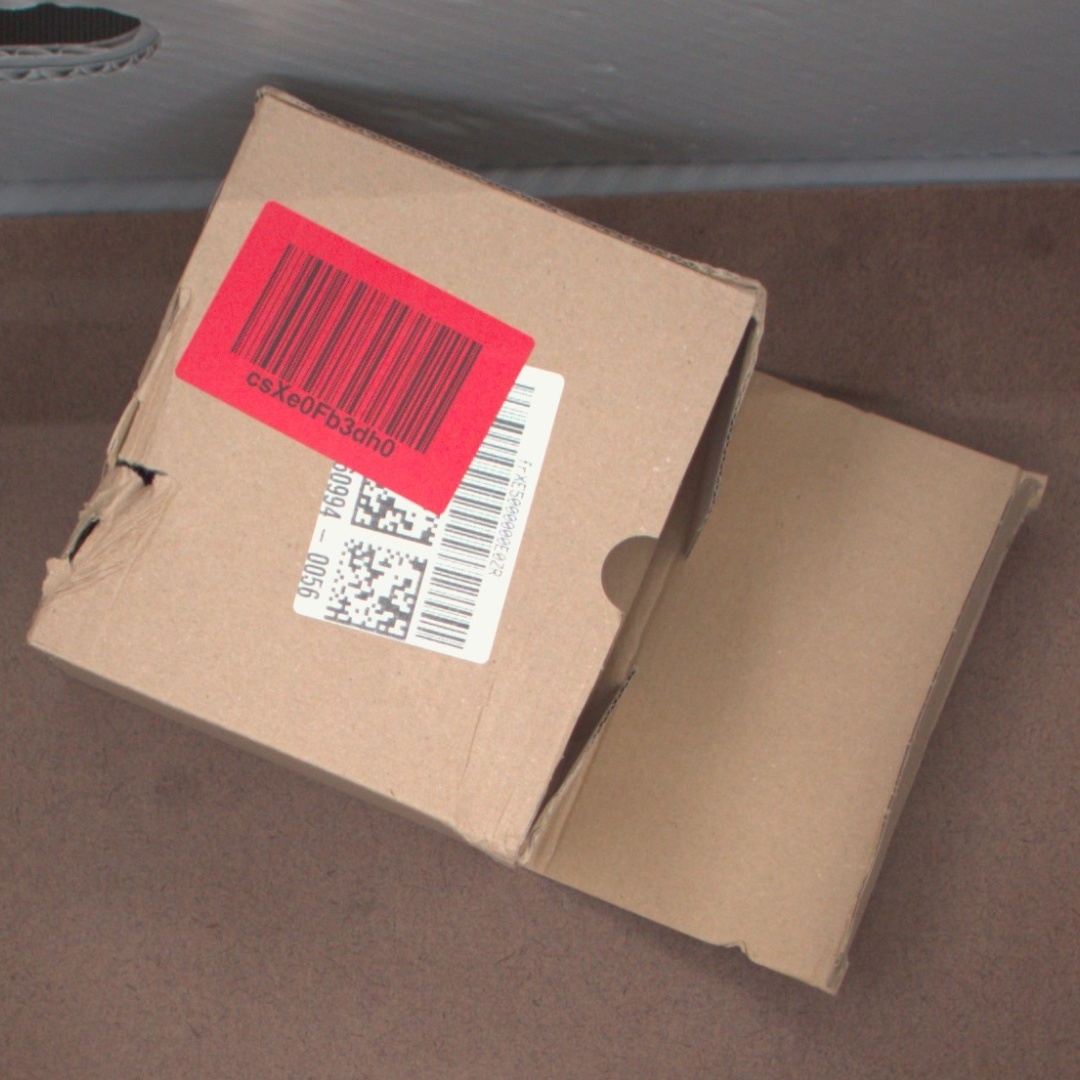}
&
\imagecellfour{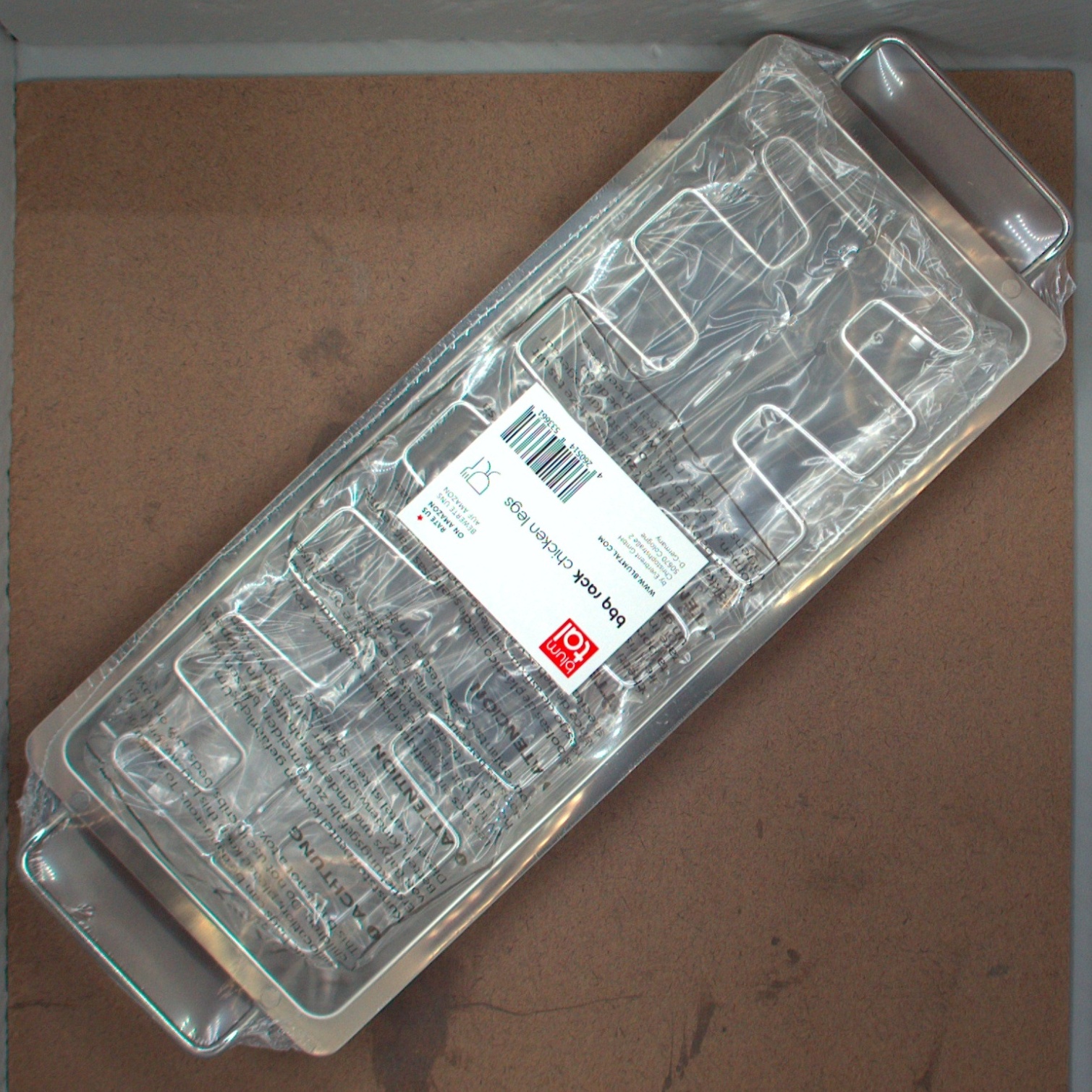}
{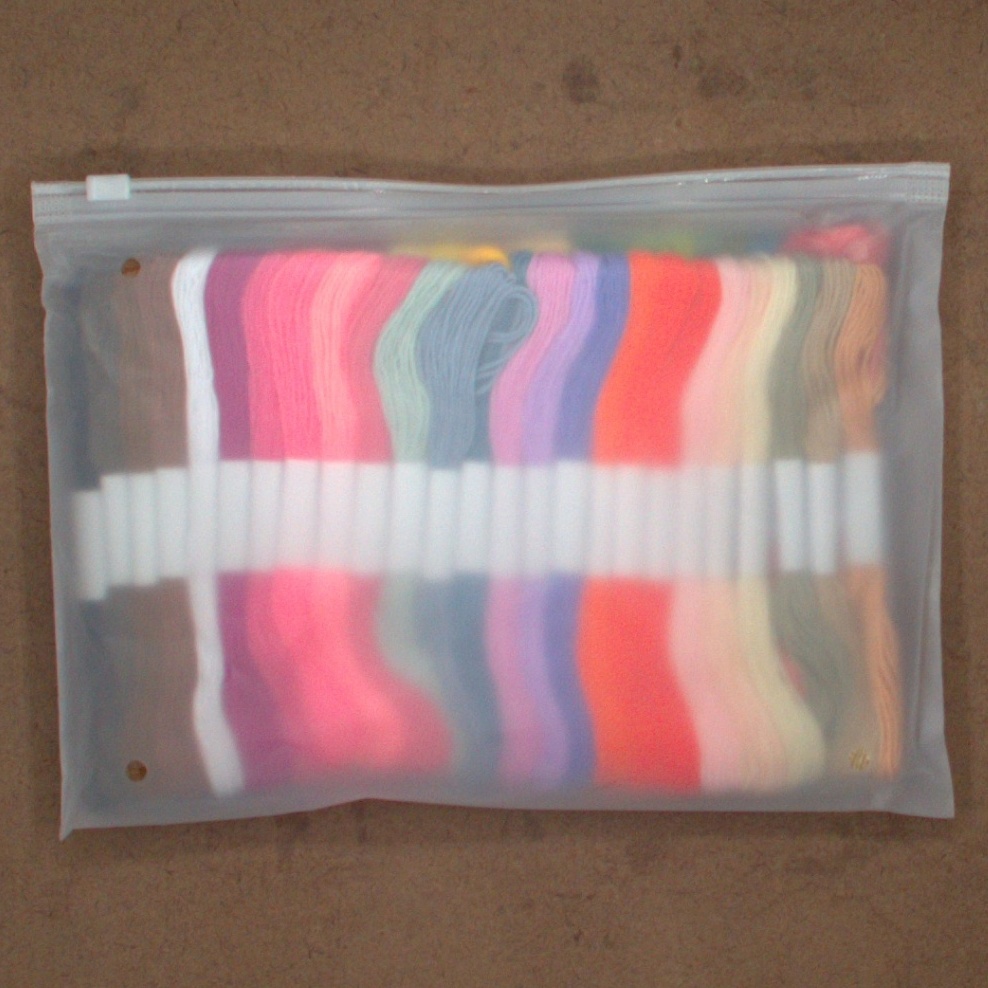}
{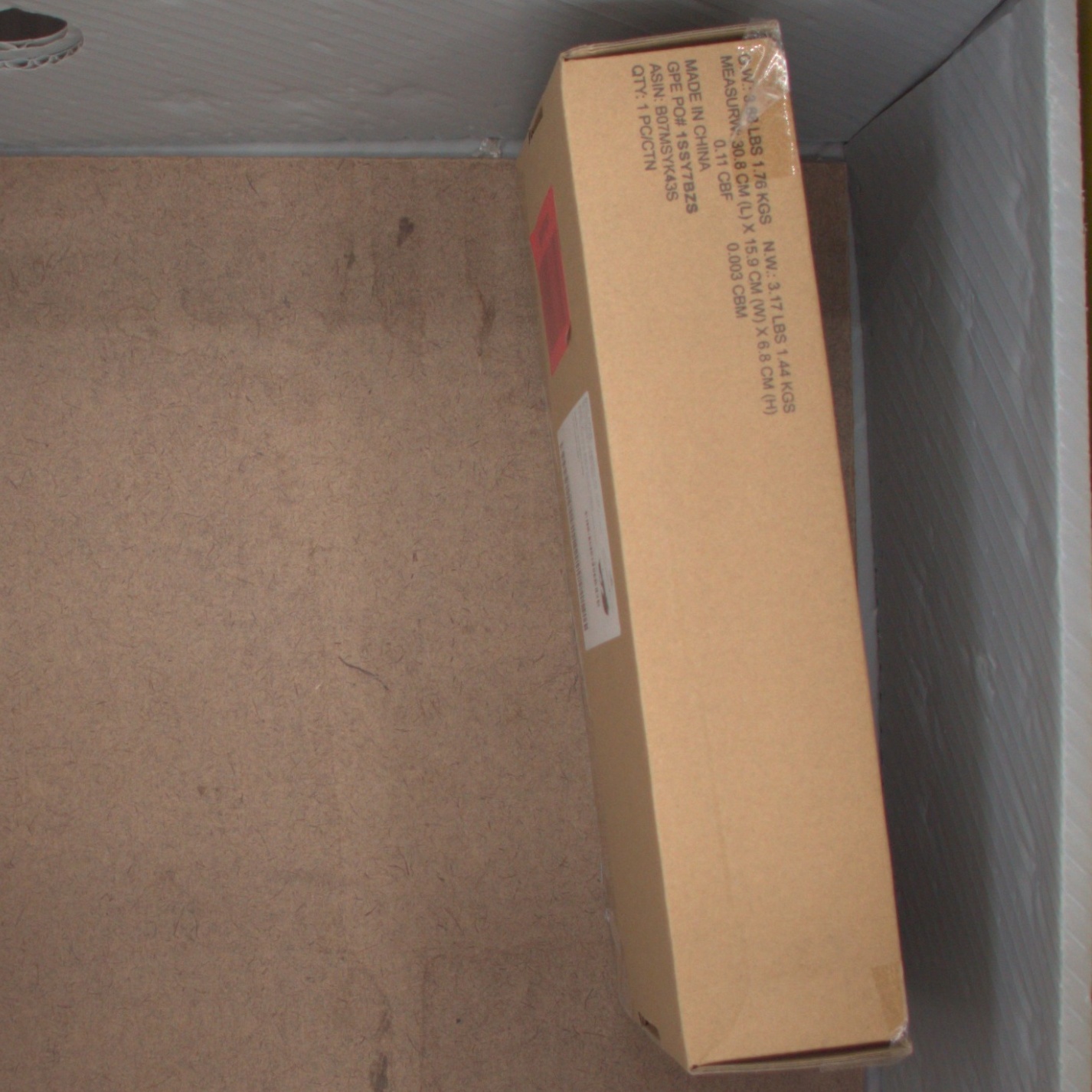}
{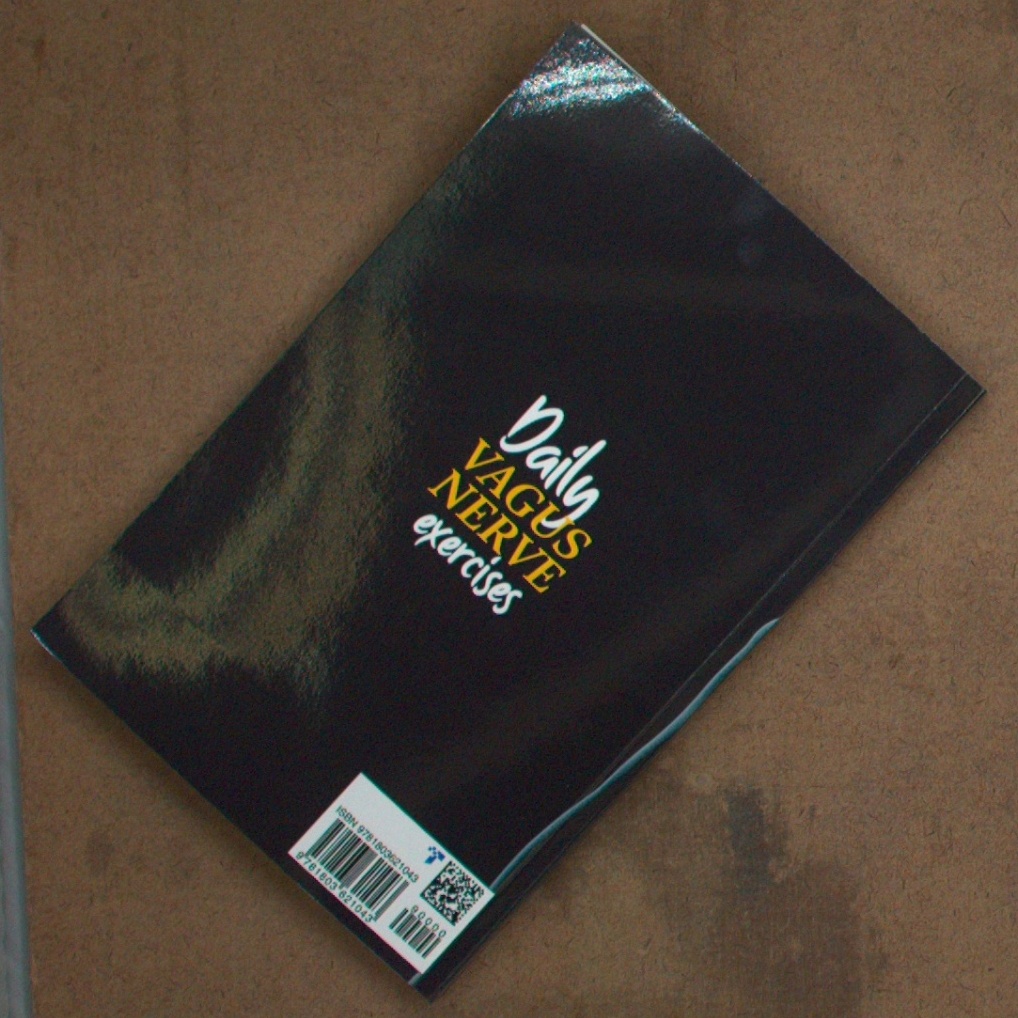}
&
\imagecellfour{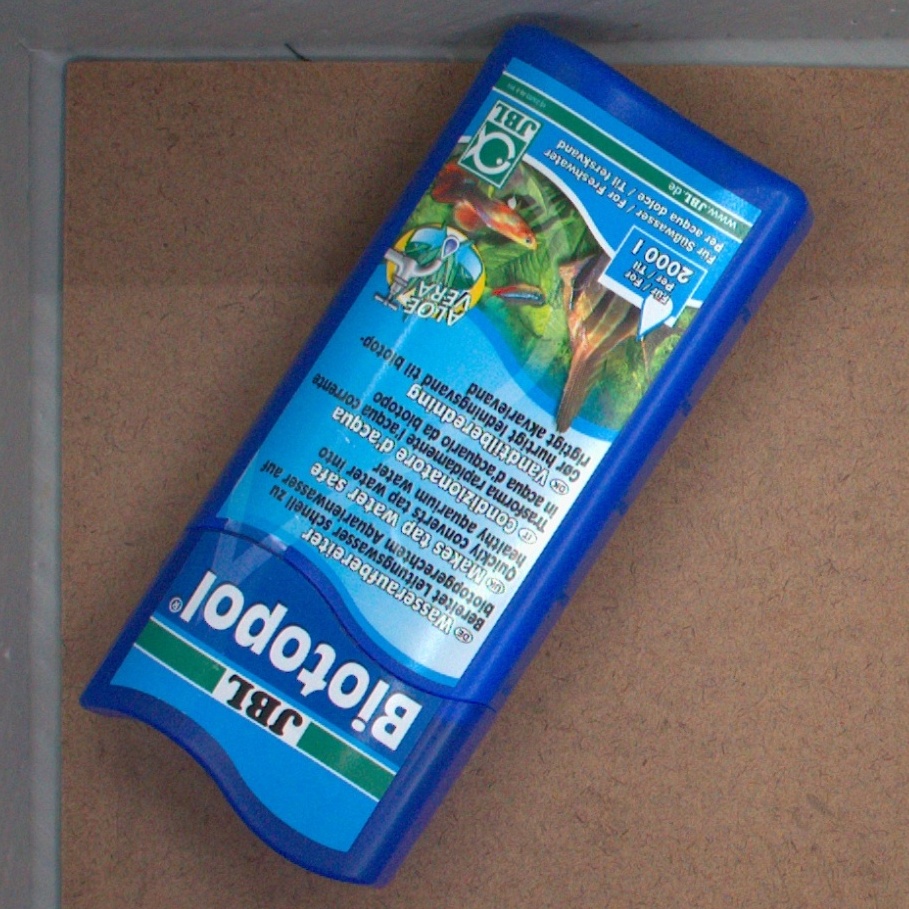}
{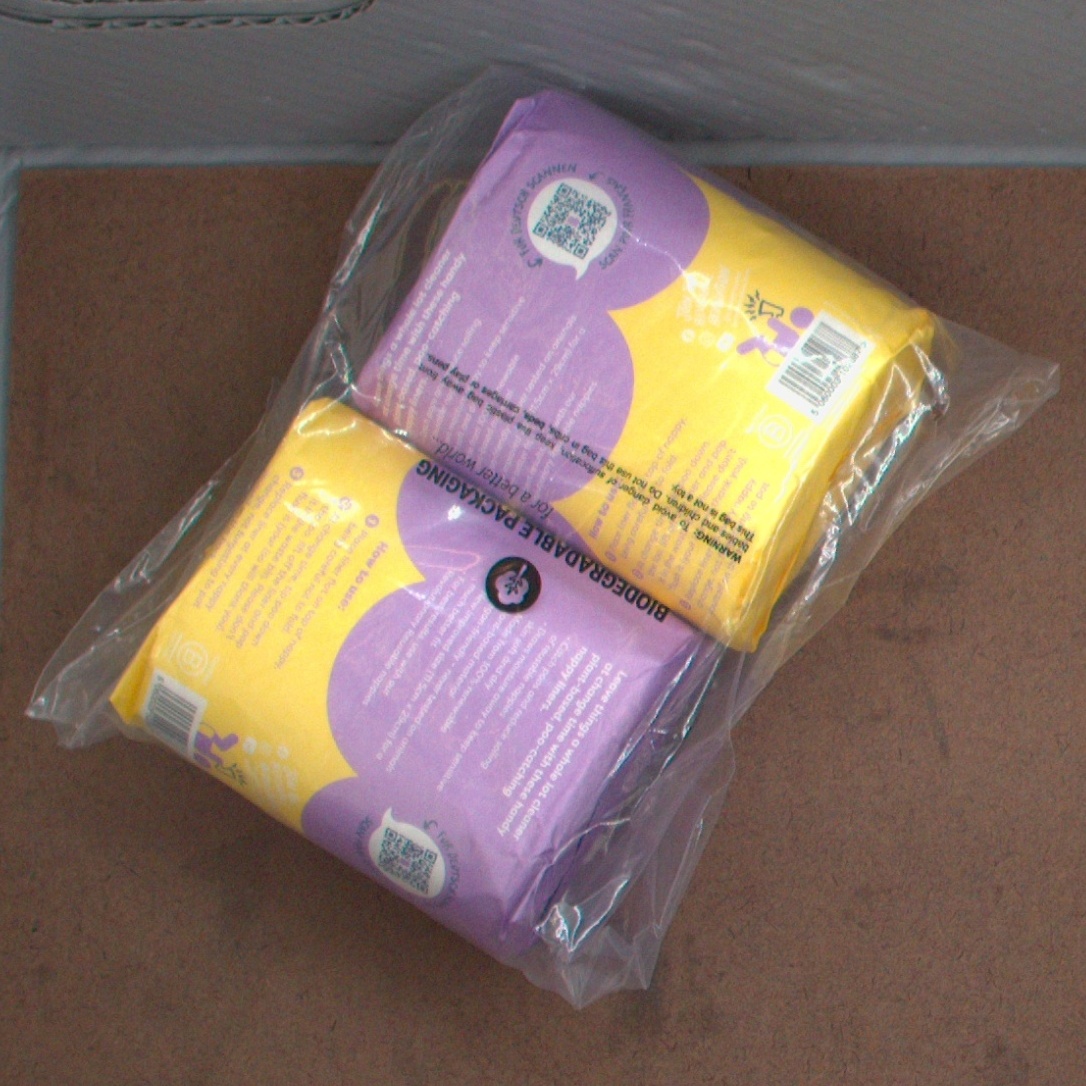}
{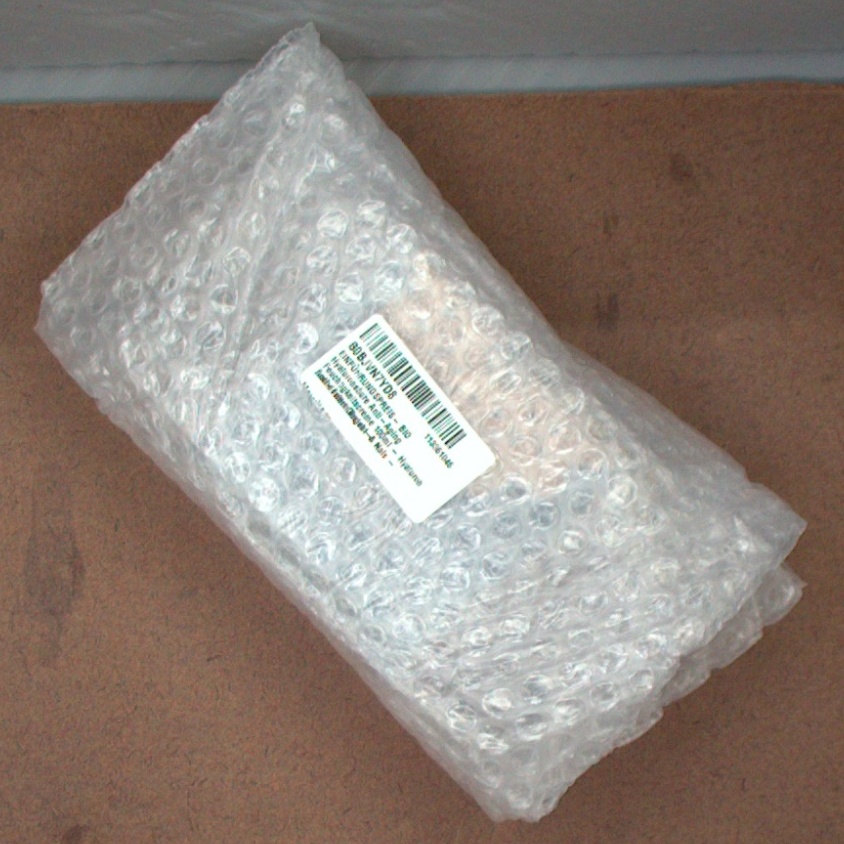}
{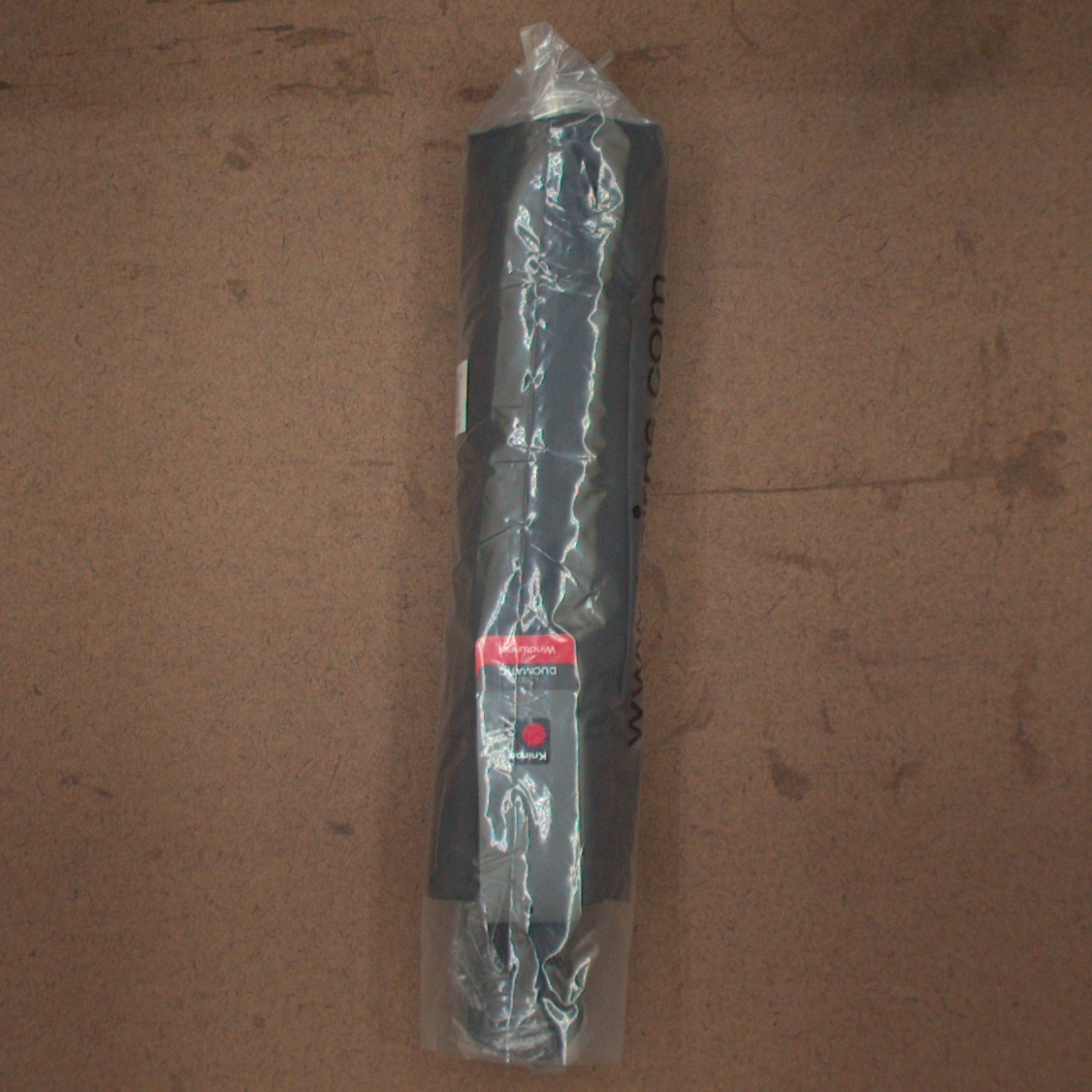}
&
\imagecellfour{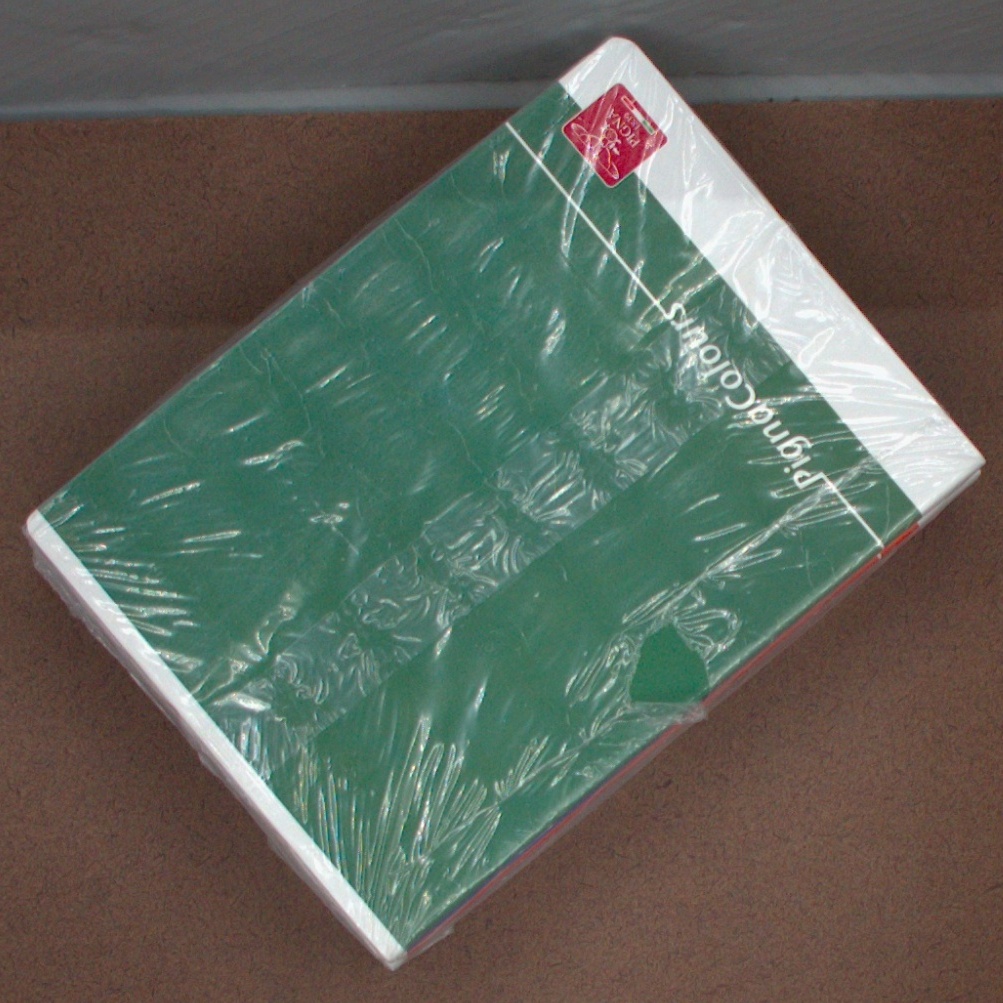}
{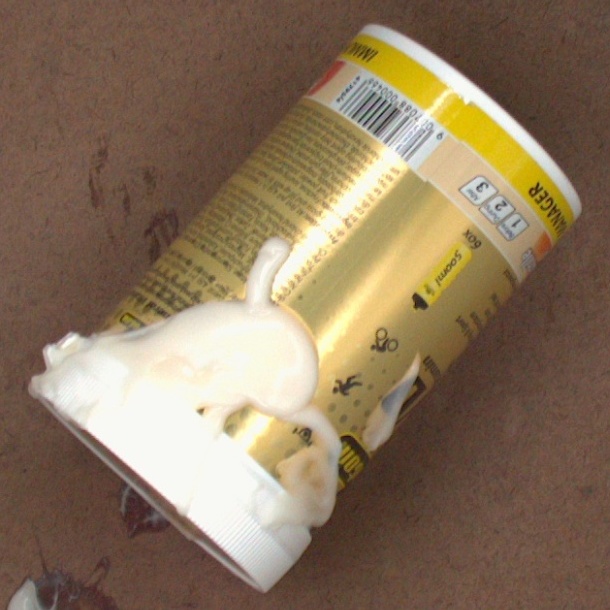}
{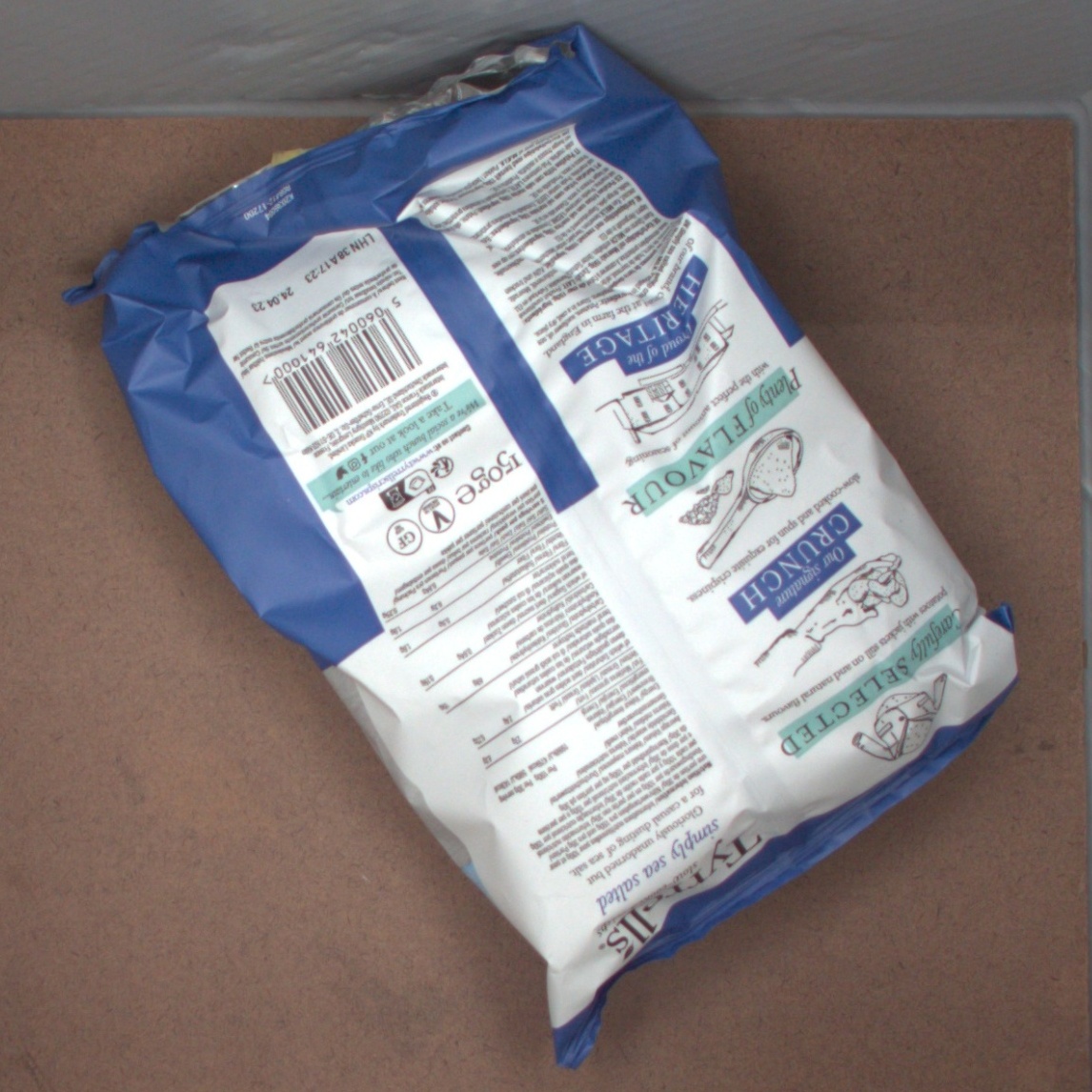}
{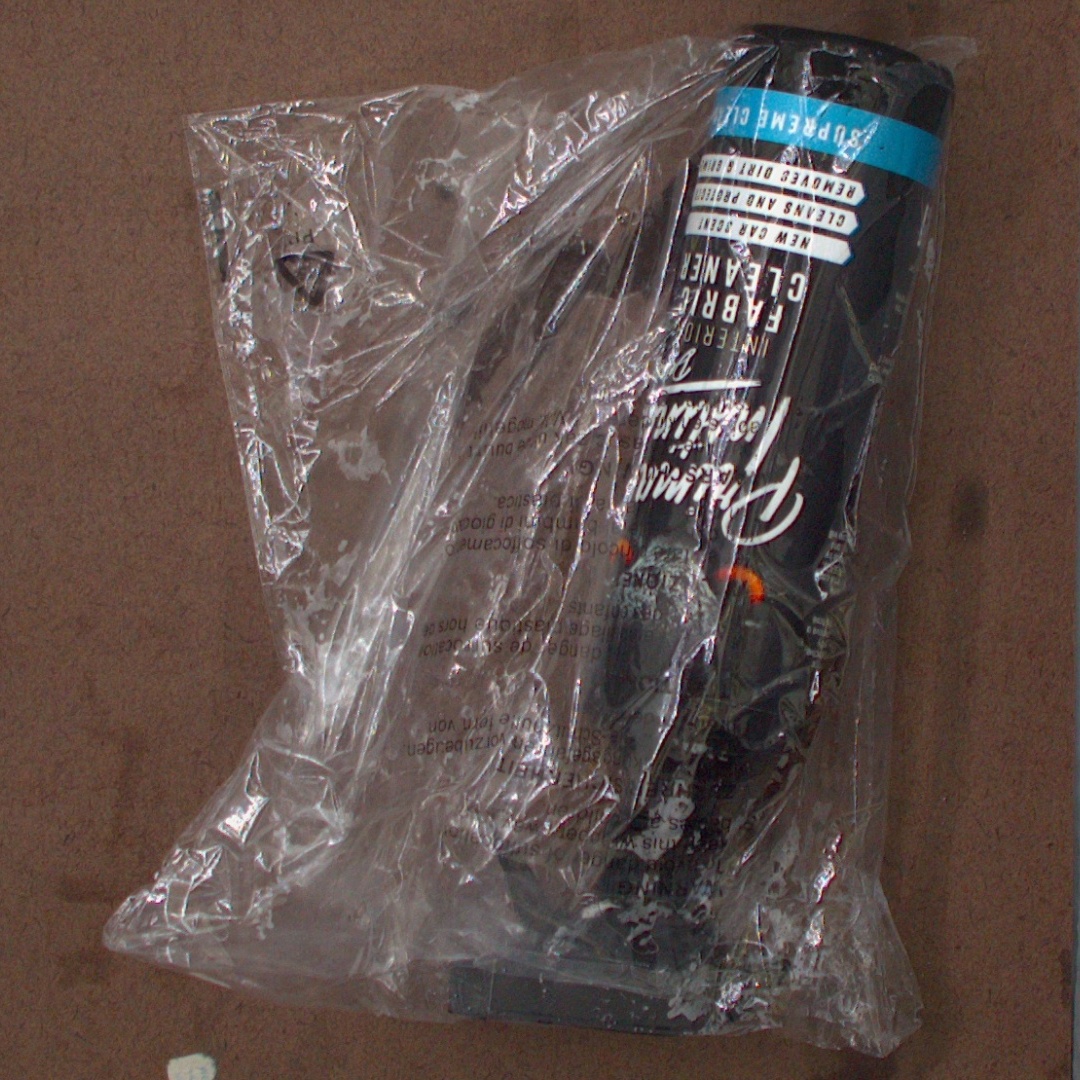}
\\
\bottomrule
\end{tabular}
}
\label{tab:failure_modes_no_references}
\end{table*}

\clearpage

\begin{table*}[!ht]
\centering
\caption{Item Material KPIs for \texttt{CLIP}}
\begin{tabular}{rccccccc}
\textbf{Material} & \textbf{Total} & \textbf{Defects} & \textbf{AP\textsubscript{any}} [\%] & \textbf{AP\textsubscript{major}} [\%] & \textbf{AUROC} & \textbf{R@50\%P} [\%] & \textbf{R@1\%FPR} [\%] \\ \hline \hline
\texttt{book other} & 36 & 10 & 38.36 & 50.00 & 63.27 & 0.00 & - \\
\texttt{book paper} & 1664 & 671 & 46.85 & 18.81 & 57.02 & 11.92 & 1.79 \\
\texttt{book plastic tight wrap} & 127 & 38 & 30.80 & 23.08 & 52.69 & 0.00 & - \\
\texttt{cardboard} & 3212 & 1261 & 41.84 & 12.77 & 53.43 & 0.79 & 1.11 \\
\texttt{other} & 409 & 90 & 28.71 & 16.62 & 55.10 & 11.11 & 3.33 \\
\texttt{paper} & 131 & 73 & \textbf{62.01} & \textbf{48.82} & 60.37 & \textbf{100.00} & 2.74 \\
\texttt{plastic bubble wrap} & 187 & 36 & 35.59 & 20.39 & \textbf{70.37} & 36.11 & - \\
\texttt{plastic hard} & 1032 & 148 & 17.27 & 14.35 & 53.98 & 0.00 & 1.35 \\
\texttt{plastic loose bag} & 2419 & 684 & 29.94 & 19.52 & 52.66 & 0.15 & 1.32 \\
\texttt{plastic tight wrap} & 850 & 195 & 29.05 & 21.55 & 59.73 & 0.51 & 1.03 \\
    \hline \hline
\end{tabular}
\end{table*}

\begin{table*}[!ht]
\centering
\caption{Damage Type KPIs for \texttt{CLIP}}
\begin{tabular}{rccccccc}
\textbf{Damage Type} & \textbf{Total} & \textbf{Defects} & \textbf{AP\textsubscript{any}} [\%] & \textbf{AP\textsubscript{major}} [\%] & \textbf{AUROC} & \textbf{R@50\%P} [\%] & \textbf{R@1\%FPR} [\%] \\ \hline \hline
\texttt{penetration} & 7827 & 966 & 15.22 & 9.28 & 57.96 & 0.21 & 1.55 \\
\texttt{deformation} & 8646 & 1785 & \textbf{24.52} & 7.30 & \textbf{56.55} & \textbf{0.22} & 1.74 \\
\texttt{actuation} & 8403 & 1542 & 21.43 & \textbf{11.83} & 55.72 & 0.00 & 1.75 \\
\texttt{superficial} & 7585 & 724 & 11.25 & 2.29 & 56.08 & 0.00 & 1.38 \\
\texttt{spillage} & 6943 & 82 & 1.33 & 0.92 & 52.15 & 0.00 & 2.44 \\
\texttt{deconstruction} & 7888 & 1027 & 16.15 & 11.69 & 56.97 & 0.00 & 1.85 \\
\\texttt{missing unit} & 6886 & 25 & 0.35 & 0.35 & 49.43 & 0.00 & 0.00 \\
    \hline \hline
\end{tabular}
\end{table*}

\begin{table*}[!ht]
\centering
\caption{Item Material KPIs for \texttt{POMP}}
\begin{tabular}{rccccccc}
\textbf{Material} & \textbf{Total} & \textbf{Defects} & \textbf{AP\textsubscript{any}} [\%] & \textbf{AP\textsubscript{major}} [\%] & \textbf{AUROC} & \textbf{R@50\%P} [\%] & \textbf{R@1\%FPR} [\%] \\ \hline \hline
\texttt{book other} & 36 & 10 & 29.22 & 16.67 & 52.31 & 0.00 & - \\
\texttt{book paper} & 1664 & 671 & 42.50 & 15.11 & 54.02 & 0.00 & 0.30 \\
\texttt{book plastic tight wrap} & 127 & 38 & 40.90 & 25.23 & 64.87 & 2.63 & 2.63 \\
\texttt{cardboard} & 3212 & 1261 & 38.55 & 16.30 & 48.40 & 0.00 & 0.87 \\
\texttt{other} & 409 & 90 & 24.89 & 17.41 & 52.19 & 0.00 & 2.22 \\
\texttt{paper} & 131 & 73 & \textbf{56.09} & \textbf{46.61} & 49.71 & \textbf{100.00} & - \\
\texttt{plastic bubble wrap} & 187 & 36 & 30.86 & 18.04 & \textbf{65.34} & 5.56 & \textbf{5.56} \\
\texttt{plastic hard} & 1032 & 148 & 22.96 & 21.72 & 59.09 & 2.03 & 3.38 \\
\texttt{plastic loose bag} & 2419 & 684 & 30.97 & 22.56 & 52.76 & 0.00 & 1.32 \\
\texttt{plastic tight wrap} & 850 & 195 & 22.91 & 16.45 & 51.02 & 0.00 & 0.51 \\
    \hline \hline
\end{tabular}
\end{table*}

\begin{table*}[!ht]
\centering
\caption{Damage Type KPIs for \texttt{POMP}}
\begin{tabular}{rccccccc}
\textbf{Damage Type} & \textbf{Total} & \textbf{Defects} & \textbf{AP\textsubscript{any}} [\%] & \textbf{AP\textsubscript{major}} [\%] & \textbf{AUROC} & \textbf{R@50\%P} [\%] & \textbf{R@1\%FPR} [\%] \\ \hline \hline
\texttt{penetration} & 7827 & 966 & 13.10 & 8.72 & 51.11 & 0.00 & 1.14 \\
\texttt{deformation} & 8646 & 1785 & \textbf{20.50} & 6.84 & 47.39 & 0.00 & 1.29 \\
\texttt{actuation} & 8403 & 1542 & 19.89 & 11.79 & 52.37 & 0.00 & 1.10 \\
\texttt{superficial} & 7585 & 724 & 8.26 & 1.47 & 42.67 & 0.00 & 1.11 \\
\texttt{spillage} & 6943 & 82 & 3.04 & 2.85 & \textbf{65.33} & 0.00 & \textbf{3.66} \\
\texttt{deconstruction} & 7888 & 1027 & 15.82 & \textbf{12.32} & 57.18 & 0.00 & 1.46 \\
\texttt{missing unit} & 6886 & 25 & 0.54 & 0.54 & 57.76 & 0.00 & 0.00 \\
    \hline \hline
\end{tabular}
\end{table*}

\begin{table*}[!ht]
\centering
\caption{Item Material KPIs for \texttt{WinCLIP-zero}}
\begin{tabular}{rccccccc}
\textbf{Material} & \textbf{Total} & \textbf{Defects} & \textbf{AP\textsubscript{any}} [\%] & \textbf{AP\textsubscript{major}} [\%] & \textbf{AUROC} & \textbf{R@50\%P} [\%] & \textbf{R@1\%FPR} [\%] \\ \hline \hline
\texttt{book other} & 36 & 10 & 51.57 & 33.33 & 65.77 & 60.00 & - \\
\texttt{book paper} & 1664 & 671 & 42.26 & 14.91 & 53.46 & 0.00 & 0.89 \\
\texttt{book plastic tight wrap} & 127 & 38 & 36.84 & 23.23 & 57.61 & 5.26 & 2.63 \\
\texttt{cardboard} & 3212 & 1261 & 42.02 & 15.74 & 51.22 & 5.79 & 2.22 \\
\texttt{other} & 409 & 90 & 28.96 & 22.61 & 58.06 & 0.00 & 1.11 \\
\texttt{paper} & 131 & 73 & \textbf{75.00} & \textbf{72.04} & 66.36 & \textbf{100.00} & \textbf{13.70} \\
\texttt{plastic bubble wrap} & 187 & 36 & 39.00 & 25.27 & \textbf{72.46} & 16.67 & 8.33 \\
\texttt{plastic hard} & 1032 & 148 & 23.12 & 19.58 & 60.02 & 0.00 & 2.70 \\
\texttt{plastic loose bag} & 2419 & 684 & 31.13 & 20.46 & 54.61 & 1.32 & 1.61 \\
\texttt{plastic tight wrap} & 850 & 195 & 23.87 & 17.21 & 47.32 & 2.05 & 2.05 \\
    \hline \hline
\end{tabular}
\end{table*}

\begin{table*}[!ht]
\centering
\caption{Damage Type KPIsfor \texttt{WinCLIP-zero}}
\begin{tabular}{rccccccc}
\textbf{Damage Type} & \textbf{Total} & \textbf{Defects} & \textbf{AP\textsubscript{any}} [\%] & \textbf{AP\textsubscript{major}} [\%] & \textbf{AUROC} & \textbf{R@50\%P} [\%] & \textbf{R@1\%FPR} [\%] \\ \hline \hline
\texttt{penetration} & 7827 & 966 & 15.87 & 11.44 & 55.90 & 0.10 & 2.17 \\
\texttt{deformation} & 8646 & 1785 & 18.29 & 5.30 & 45.77 & 0.00 & 0.39 \\
\texttt{actuation} & 8403 & 1542 & \textbf{22.36} & 13.46 & 56.72 & \textbf{0.07} & 1.88 \\
\texttt{superficial} & 7585 & 724 & 8.49 & 1.21 & 45.85 & 0.00 & 0.14 \\
\texttt{spillage} & 6943 & 82 & 2.12 & 1.76 & \textbf{64.97} & 0.00 & 1.22 \\
\texttt{deconstruction} & 7888 & 1027 & 19.40 & \textbf{14.90} & 62.12 & 0.00 & \textbf{2.82} \\
\texttt{missing unit} & 6886 & 25 & 0.85 & 0.85 & 49.95 & 0.00 & 4.00 \\
    \hline \hline
\end{tabular}
\end{table*}

\begin{table*}[!ht]
\centering
\caption{Item Material KPIs for \texttt{Claude-icl}}
\begin{tabular}{rccccccc}
\textbf{Material} & \textbf{Total} & \textbf{Defects} & \textbf{AP\textsubscript{any}} [\%] & \textbf{AP\textsubscript{major}} [\%] & \textbf{AUROC} & \textbf{R@50\%P} [\%] & \textbf{R@1\%FPR} [\%] \\ \hline \hline
\texttt{book other} & 36 & 10 & 42.22 & \textbf{100.00} & 58.46 & 20.00 & \textbf{20.00} \\
\texttt{book paper} & 1664 & 671 & 43.09 & 20.03 & 53.08 & 11.03 & 2.24 \\
\texttt{book plastic tight wrap} & 127 & 38 & 42.42 & 31.41 & 61.75 & 26.32 & 2.63 \\
\texttt{cardboard} & 3212 & 1261 & 46.29 & 32.53 & 58.07 & 30.29 & 1.43 \\
\texttt{other} & 409 & 90 & 27.92 & 21.71 & 56.08 & 5.56 & 3.33 \\
\texttt{paper} & 131 & 73 & \textbf{69.51} & 65.91 & 66.37 & \textbf{100.00} & - \\
\texttt{plastic bubble wrap} & 187 & 36 & 29.05 & 24.13 & 62.99 & 0.00 & - \\
\texttt{plastic hard} & 1032 & 148 & 24.32 & 22.95 & \textbf{67.30} & 0.00 & 0.68 \\
\texttt{plastic loose bag} & 2419 & 684 & 34.07 & 25.43 & 59.39 & 0.00 & 0.44 \\
\texttt{plastic tight wrap} & 850 & 195 & 26.19 & 19.90 & 55.89 & 0.00 & 0.00 \\
    \hline \hline
\end{tabular}
\end{table*}

\begin{table*}[!ht]
\centering
\caption{Damage Type KPIsfor \texttt{Claude-icl}}
\begin{tabular}{rccccccc}
\textbf{Damage Type} & \textbf{Total} & \textbf{Defects} & \textbf{AP\textsubscript{any}} [\%] & \textbf{AP\textsubscript{major}} [\%] & \textbf{AUROC} & \textbf{R@50\%P} [\%] & \textbf{R@1\%FPR} [\%] \\ \hline \hline
\texttt{penetration} & 7827 & 966 & 19.41 & 16.41 & 62.79 & 0.00 & 0.52 \\
\texttt{deformation} & 8646 & 1785 & 21.85 & 9.70 & 53.40 & 0.00 & 0.17 \\
\texttt{actuation} & 8403 & 1542 & 23.95 & 18.46 & 59.63 & 0.00 & 0.39 \\
\texttt{superficial} & 7585 & 724 & 10.68 & 3.21 & 55.26 & 0.00 & 0.28 \\
\texttt{spillage} & 6943 & 82 & 6.57 & 6.90 & \textbf{77.89} & 0.00 & \textbf{3.66} \\
\texttt{deconstruction} & 7888 & 1027 & \textbf{22.20} & \textbf{19.83} & 65.18 & 0.00 & 0.58 \\
\texttt{missing unit} & 6886 & 25 & 0.50 & 0.50 & 57.94 & 0.00 & 0.00 \\
    \hline \hline
\end{tabular}
\end{table*}

\begin{table*}[!ht]
\centering
\caption{Item Material KPIs for \texttt{Pixtral-zero}}
\begin{tabular}{rccccccc}
\textbf{Material} & \textbf{Total} & \textbf{Defects} & \textbf{AP\textsubscript{any}} [\%] & \textbf{AP\textsubscript{major}} [\%] & \textbf{AUROC} & \textbf{R@50\%P} [\%] & \textbf{R@1\%FPR} [\%] \\ \hline \hline
\texttt{book other} & 36 & 10 & 46.11 & 3.70 & \textbf{67.31} & 50.00 & \textbf{10.00} \\
\texttt{book paper} & 1664 & 671 & 40.97 & 15.13 & 49.75 & 3.28 & 0.30 \\
\texttt{book plastic tight wrap} & 127 & 38 & 31.42 & 17.24 & 53.02 & 0.00 & - \\
\texttt{cardboard} & 3212 & 1261 & 40.57 & 14.07 & 51.28 & 3.81 & 1.03 \\
\texttt{other} & 409 & 90 & 22.89 & 16.38 & 50.36 & 0.00 & 1.11 \\
\texttt{paper} & 131 & 73 & \textbf{60.88} & \textbf{52.79} & 52.83 & \textbf{100.00} & 8.22 \\
\texttt{plastic bubble wrap} & 187 & 36 & 22.84 & 11.32 & 51.38 & 0.00 & 2.78 \\
\texttt{plastic hard} & 1032 & 148 & 15.98 & 14.21 & 50.18 & 0.00 & 0.00 \\
\texttt{plastic loose bag} & 2419 & 684 & 30.49 & 20.64 & 53.87 & 0.44 & 1.02 \\
\texttt{plastic tight wrap} & 850 & 195 & 25.05 & 17.21 & 49.87 & 2.56 & 2.56 \\
    \hline \hline
\end{tabular}
\end{table*}

\begin{table*}[!ht]
\centering
\caption{Damage Type KPIsfor \texttt{Pixtral-zero}}
\begin{tabular}{rccccccc}
\textbf{Damage Type} & \textbf{Total} & \textbf{Defects} & \textbf{AP\textsubscript{any}} [\%] & \textbf{AP\textsubscript{major}} [\%] & \textbf{AUROC} & \textbf{R@50\%P} [\%] & \textbf{R@1\%FPR} [\%] \\ \hline \hline
\texttt{penetration} & 7827 & 966 & 13.77 & 8.51 & 52.43 & 0.00 & 1.55 \\
\texttt{deformation} & 8646 & 1785 & \textbf{20.21} & 5.30 & 49.05 & 0.00 & 0.39 \\
\texttt{actuation} & 8403 & 1542 & 19.46 & 10.80 & 52.12 & 0.00 & 0.78 \\
\texttt{superficial} & 7585 & 724 & 9.49 & 1.39 & 50.03 & 0.00 & 0.14 \\
\texttt{spillage} & 6943 & 82 & 4.01 & 3.38 & \textbf{66.58} & 0.00 & \textbf{3.66} \\
\texttt{deconstruction} & 7888 & 1027 & 14.88 & \textbf{11.80} & 53.09 & 0.00 & 1.27 \\
\texttt{missing unit} & 6886 & 25 & 0.35 & 0.35 & 42.24 & 0.00 & 0.00 \\
    \hline \hline
\end{tabular}
\end{table*}

\begin{table*}[!ht]
\centering
\caption{Item Material KPIs for \texttt{Pixtral-icl}}
\begin{tabular}{rccccccc}
\textbf{Material} & \textbf{Total} & \textbf{Defects} & \textbf{AP\textsubscript{any}} [\%] & \textbf{AP\textsubscript{major}} [\%] & \textbf{AUROC} & \textbf{R@50\%P} [\%] & \textbf{R@1\%FPR} [\%] \\ \hline \hline
\texttt{book other} & 36 & 10 & 44.88 & 8.33 & \textbf{70.77} & 20.00 & - \\
\texttt{book paper} & 1664 & 671 & 39.86 & 12.68 & 49.42 & 0.00 & 0.45 \\
\texttt{book plastic tight wrap} & 127 & 38 & 29.85 & 17.74 & 51.27 & 0.00 & - \\
\texttt{cardboard} & 3212 & 1261 & 39.14 & 10.71 & 49.87 & 0.00 & 0.16 \\
\texttt{other} & 409 & 90 & 19.31 & 13.33 & 43.85 & 0.00 & - \\
\texttt{paper} & 131 & 73 & \textbf{58.52} & \textbf{50.67} & 54.97 & \textbf{100.00} & - \\
\texttt{plastic bubble wrap} & 187 & 36 & 25.43 & 23.28 & 54.76 & 2.78 & \textbf{2.78} \\
\texttt{plastic hard} & 1032 & 148 & 18.13 & 15.42 & 58.25 & 0.00 & 0.68 \\
\texttt{plastic loose bag} & 2419 & 684 & 33.16 & 22.33 & 58.50 & 0.00 & 0.88 \\
\texttt{plastic tight wrap} & 850 & 195 & 23.20 & 15.73 & 50.64 & 0.00 & 0.00 \\
    \hline \hline
\end{tabular}
\end{table*}

\begin{table*}[!ht]
\centering
\caption{Damage Type KPIsfor \texttt{Pixtral-icl}}
\begin{tabular}{rccccccc}
\textbf{Damage Type} & \textbf{Total} & \textbf{Defects} & \textbf{AP\textsubscript{any}} [\%] & \textbf{AP\textsubscript{major}} [\%] & \textbf{AUROC} & \textbf{R@50\%P} [\%] & \textbf{R@1\%FPR} [\%] \\ \hline \hline
\texttt{penetration} & 7827 & 966 & 13.60 & 7.96 & 54.87 & 0.00 & 0.52 \\
\texttt{deformation} & 8646 & 1785 & 19.00 & 4.58 & 45.86 & 0.00 & 0.56 \\
\texttt{actuation} & 8403 & 1542 & \textbf{20.05} & 11.06 & 54.62 & 0.00 & 0.91 \\
\texttt{superficial} & 7585 & 724 & 9.06 & 1.02 & 47.62 & 0.00 & 0.55 \\
\texttt{spillage} & 6943 & 82 & 1.59 & 1.39 & \textbf{59.30} & 0.00 & \textbf{1.22} \\
\texttt{deconstruction} & 7888 & 1027 & 15.56 & \textbf{12.10} & 58.35 & 0.00 & 1.17 \\
\texttt{missing unit} & 6886 & 25 & 0.47 & 0.47 & 56.67 & 0.00 & 0.00 \\
    \hline \hline
\end{tabular}
\end{table*}

\begin{table*}[!ht]
\centering
\caption{Item Material KPIs for \texttt{PatchCore50}}
\begin{tabular}{rccccccc}
\textbf{Material} & \textbf{Total} & \textbf{Defects} & \textbf{AP\textsubscript{any}} [\%] & \textbf{AP\textsubscript{major}} [\%] & \textbf{AUROC} & \textbf{R@50\%P} [\%] & \textbf{R@1\%FPR} [\%] \\ \hline \hline
\texttt{book other} & 36 & 10 & 32.12 & 20.00 & 56.92 & 0.00 & - \\
\texttt{book paper} & 1664 & 671 & \textbf{45.20} & 17.57 & 56.44 & 5.22 & 2.09 \\
\texttt{book plastic tight wrap} & 127 & 38 & 41.24 & \textbf{39.53} & 58.87 & 10.53 & \textbf{5.26} \\
\texttt{cardboard} & 3212 & 1261 & 42.16 & 12.17 & 52.75 & 5.31 & 2.46 \\
\texttt{other} & 409 & 90 & 30.50 & 22.62 & 60.91 & 3.33 & 3.33 \\
\texttt{paper} & 131 & 73 & 53.02 & 43.31 & 46.95 & \textbf{100.00} & - \\
\texttt{plastic bubble wrap} & 187 & 36 & 32.21 & 24.43 & \textbf{71.12} & 0.00 & 2.78 \\
\texttt{plastic hard} & 1032 & 148 & 20.79 & 18.75 & 63.25 & 0.00 & 3.38 \\
\texttt{plastic loose bag} & 2419 & 684 & 31.42 & 21.86 & 54.53 & 0.59 & 1.02 \\
\texttt{plastic tight wrap} & 850 & 195 & 26.97 & 21.36 & 54.77 & 0.00 & 2.56 \\
    \hline \hline
\end{tabular}
\end{table*}

\begin{table*}[!ht]
\centering
\caption{Damage Type KPIsfor \texttt{PatchCore50}}
\begin{tabular}{rccccccc}
\textbf{Damage Type} & \textbf{Total} & \textbf{Defects} & \textbf{AP\textsubscript{any}} [\%] & \textbf{AP\textsubscript{major}} [\%] & \textbf{AUROC} & \textbf{R@50\%P} [\%] & \textbf{R@1\%FPR} [\%] \\ \hline \hline
\texttt{penetration} & 7827 & 966 & 14.45 & 8.55 & 56.60 & 0.00 & 1.66 \\
\texttt{deformation} & 8646 & 1785 & \textbf{24.37} & 6.83 & 53.99 & \textbf{1.18} & \textbf{2.58} \\
\texttt{actuation} & 8403 & 1542 & 20.70 & 12.00 & 55.57 & 0.00 & 1.17 \\
\texttt{superficial} & 7585 & 724 & 12.80 & 1.90 & 52.06 & 1.52 & 4.28 \\
\texttt{spillage} & 6943 & 82 & 1.75 & 1.31 & \textbf{60.44} & 0.00 & 2.44 \\
\texttt{deconstruction} & 7888 & 1027 & 16.05 & \textbf{12.78} & 59.36 & 0.00 & 1.46 \\
\texttt{missing unit} & 6886 & 25 & 0.50 & 0.50 & 59.88 & 0.00 & 0.00 \\
    \hline \hline
\end{tabular}
\end{table*}

\begin{table*}[!ht]
\centering
\caption{Item Material KPIs for \texttt{WinCLIP-few}}
\begin{tabular}{rccccccc}
\textbf{Material} & \textbf{Total} & \textbf{Defects} & \textbf{AP\textsubscript{any}} [\%] & \textbf{AP\textsubscript{major}} [\%] & \textbf{AUROC} & \textbf{R@50\%P} [\%] & \textbf{R@1\%FPR} [\%] \\ \hline \hline
\texttt{book other} & 36 & 10 & 49.95 & 33.33 & 64.62 & 60.00 & - \\
\texttt{book paper} & 1664 & 671 & 42.73 & 15.50 & 53.90 & 0.89 & 1.19 \\
\texttt{book plastic tight wrap} & 127 & 38 & 37.84 & 26.60 & 57.78 & 5.26 & 5.26 \\
\texttt{cardboard} & 3212 & 1261 & 42.12 & 16.58 & 50.97 & 6.74 & 2.30 \\
\texttt{other} & 409 & 90 & 29.01 & 22.66 & 58.62 & 0.00 & 2.22 \\
\texttt{paper} & 131 & 73 & \textbf{73.74} & \textbf{70.17} & 66.01 & \textbf{100.00} & 5.48 \\
\texttt{plastic bubble wrap} & 187 & 36 & 39.74 & 25.79 & \textbf{73.11} & 11.11 & \textbf{8.33} \\
\texttt{plastic hard} & 1032 & 148 & 23.21 & 19.81 & 60.19 & 0.00 & 1.35 \\
\texttt{plastic loose bag} & 2419 & 684 & 31.24 & 20.39 & 54.63 & 1.75 & 1.75 \\
\texttt{plastic tight wrap} & 850 & 195 & 24.32 & 17.81 & 47.60 & 1.54 & 2.05 \\
    \hline \hline
\end{tabular}
\end{table*}

\begin{table*}[!ht]
\centering
\caption{Damage Type KPIsfor \texttt{WinCLIP-few}}
\begin{tabular}{rccccccc}
\textbf{Damage Type} & \textbf{Total} & \textbf{Defects} & \textbf{AP\textsubscript{any}} [\%] & \textbf{AP\textsubscript{major}} [\%] & \textbf{AUROC} & \textbf{R@50\%P} [\%] & \textbf{R@1\%FPR} [\%] \\ \hline \hline
\texttt{penetration} & 7827 & 966 & 15.96 & 11.35 & 55.94 & 0.00 & 2.69 \\
\texttt{deformation} & 8646 & 1785 & 18.47 & 5.51 & 46.01 & 0.00 & 0.50 \\
\texttt{actuation} & 8403 & 1542 & \textbf{22.48} & 13.56 & 56.87 & 0.00 & 2.08 \\
\texttt{superficial} & 7585 & 724 & 8.55 & 1.25 & 45.81 & 0.00 & 0.69 \\
\texttt{spillage} & 6943 & 82 & 2.18 & 1.80 & \textbf{65.47} & 0.00 & 1.22 \\
\texttt{deconstruction} & 7888 & 1027 & 19.49 & \textbf{14.98} & 62.20 & 0.00 & \textbf{3.12} \\
\texttt{missing unit} & 6886 & 25 & 0.80 & 0.80 & 49.99 & 0.00 & 4.00 \\
    \hline \hline
\end{tabular}
\end{table*}

\begin{table*}[!ht]
\centering
\caption{Item Material KPIs for \texttt{ResNet50}}
\begin{tabular}{rccccccc}
\textbf{Material} & \textbf{Total} & \textbf{Defects} & \textbf{AP\textsubscript{any}} [\%] & \textbf{AP\textsubscript{major}} [\%] & \textbf{AUROC} & \textbf{R@50\%P} [\%] & \textbf{R@1\%FPR} [\%] \\ \hline \hline
\texttt{book other} & 36 & 10 & 73.32 & 25.00 & 79.62 & 60.00 & 50.00 \\
\texttt{book paper} & 1664 & 671 & 82.36 & 72.92 & 86.16 & 95.53 & 24.59 \\
\texttt{book plastic tight wrap} & 127 & 38 & 82.54 & 82.69 & 85.99 & 81.58 & 44.74 \\
\texttt{cardboard} & 3212 & 1261 & 81.11 & 73.70 & 84.66 & 92.94 & 27.44 \\
\texttt{other} & 409 & 90 & 60.83 & 60.46 & 80.24 & 66.67 & 16.67 \\
\texttt{paper} & 131 & 73 & 88.14 & 86.22 & 85.00 & \textbf{100.00} & 28.77 \\
\texttt{plastic bubble wrap} & 187 & 36 & \textbf{92.21} & \textbf{86.03} & \textbf{97.15} & 97.22 & \textbf{66.67} \\
\texttt{plastic hard} & 1032 & 148 & 76.58 & 79.37 & 90.78 & 79.05 & 47.97 \\
\texttt{plastic loose bag} & 2419 & 684 & 83.28 & 79.69 & 91.55 & 93.42 & 33.63 \\
\texttt{plastic tight wrap} & 850 & 195 & 76.01 & 78.50 & 87.23 & 81.54 & 37.44 \\
    \hline \hline
\end{tabular}
\end{table*}

\begin{table*}[!ht]
\centering
\caption{Damage Type KPIsfor \texttt{ResNet50}}
\begin{tabular}{rccccccc}
\textbf{Damage Type} & \textbf{Total} & \textbf{Defects} & \textbf{AP\textsubscript{any}} [\%] & \textbf{AP\textsubscript{major}} [\%] & \textbf{AUROC} & \textbf{R@50\%P} [\%] & \textbf{R@1\%FPR} [\%] \\ \hline \hline
\texttt{penetration} & 7827 & 966 & 64.99 & 65.24 & 88.54 & 68.84 & 32.82 \\
\texttt{deformation} & 8646 & 1785 & 72.46 & 60.20 & 89.19 & \textbf{83.31} & 27.90 \\
\texttt{actuation} & 8403 & 1542 & \textbf{75.88} & 73.04 & 90.74 & 83.27 & 35.80 \\
\texttt{superficial} & 7585 & 724 & 52.97 & 36.00 & 87.88 & 54.14 & 24.59 \\
\texttt{spillage} & 6943 & 82 & 18.79 & 16.49 & 82.02 & 6.10 & 29.27 \\
\texttt{deconstruction} & 7888 & 1027 & 73.02 & \textbf{73.43} & \textbf{91.10} & 80.33 & \textbf{39.44} \\
\texttt{missing unit} & 6886 & 25 & 15.40 & 15.40 & 87.07 & 8.00 & 32.00 \\
    \hline \hline
\end{tabular}
\end{table*}

\begin{table*}[!ht]
\centering
\caption{Item Material KPIs for \texttt{ViT-S}}
\begin{tabular}{rccccccc}
\textbf{Material} & \textbf{Total} & \textbf{Defects} & \textbf{AP\textsubscript{any}} [\%] & \textbf{AP\textsubscript{major}} [\%] & \textbf{AUROC} & \textbf{R@50\%P} [\%] & \textbf{R@1\%FPR} [\%] \\ \hline \hline
\texttt{book other} & 36 & 10 & 88.52 & \textbf{100.00} & 94.23 & \textbf{100.00} & 50.00 \\
\texttt{book paper} & 1664 & 671 & 90.68 & 92.63 & 92.35 & 99.40 & 49.03 \\
\texttt{book plastic tight wrap} & 127 & 38 & 92.54 & 94.40 & 94.94 & 94.74 & 65.79 \\
\texttt{cardboard} & 3212 & 1261 & 90.22 & 94.33 & 91.87 & 98.57 & 55.91 \\
\texttt{other} & 409 & 90 & 77.69 & 79.73 & 88.74 & 84.44 & 48.89 \\
\texttt{paper} & 131 & 73 & 96.72 & 97.14 & 95.77 & \textbf{100.00} & 69.86 \\
\texttt{plastic bubble wrap} & 187 & 36 & \textbf{97.22} & 98.44 & \textbf{99.23} & \textbf{100.00} & \textbf{80.56} \\
\texttt{plastic hard} & 1032 & 148 & 92.88 & 94.31 & 97.62 & 95.95 & 79.05 \\
\texttt{plastic loose bag} & 2419 & 684 & 92.51 & 92.41 & 96.29 & 97.52 & 65.64 \\
\texttt{plastic tight wrap} & 850 & 195 & 86.01 & 90.05 & 93.16 & 91.80 & 57.44 \\
    \hline \hline
\end{tabular}
\end{table*}

\begin{table*}[!ht]
\centering
\caption{Damage Type KPIsfor \texttt{ViT-S}}
\begin{tabular}{rccccccc}
\textbf{Damage Type} & \textbf{Total} & \textbf{Defects} & \textbf{AP\textsubscript{any}} [\%] & \textbf{AP\textsubscript{major}} [\%] & \textbf{AUROC} & \textbf{R@50\%P} [\%] & \textbf{R@1\%FPR} [\%] \\ \hline \hline
\texttt{penetration} & 7827 & 966 & 83.17 & 88.80 & 95.11 & 86.65 & 64.49 \\
\texttt{deformation} & 8646 & 1785 & 85.14 & 89.08 & 94.08 & 92.89 & 55.35 \\
\texttt{actuation} & 8403 & 1542 & 89.18 & 92.78 & 96.01 & 93.52 & 68.81 \\
\texttt{superficial} & 7585 & 724 & 70.61 & 75.96 & 93.63 & 73.07 & 46.55 \\
\texttt{spillage} & 6943 & 82 & 44.74 & 45.77 & 89.86 & 48.78 & 52.44 \\
\texttt{deconstruction} & 7888 & 1027 & \textbf{89.53} & \textbf{91.98} & \textbf{96.88} & \textbf{92.50} & \textbf{75.56} \\
\texttt{missing unit} & 6886 & 25 & 59.23 & 59.23 & 89.78 & 60.00 & 68.00 \\
    \hline \hline
\end{tabular}
\end{table*}

\begin{table*}[!ht]
\centering
\caption{Item Material KPIs for \texttt{Pixtral-ft}}
\begin{tabular}{rccccccc}
\textbf{Material} & \textbf{Total} & \textbf{Defects} & \textbf{AP\textsubscript{any}} [\%] & \textbf{AP\textsubscript{major}} [\%] & \textbf{AUROC} & \textbf{R@50\%P} [\%] & \textbf{R@1\%FPR} [\%] \\ \hline \hline
\texttt{book other} & 36 & 10 & 27.78 & 3.70 & 50.00 & 0.00 & - \\
\texttt{book paper} & 1664 & 671 & 40.59 & 12.59 & 50.22 & 0.45 & 0.45 \\
\texttt{book plastic tight wrap} & 127 & 38 & 29.92 & 17.59 & 50.00 & 0.00 & - \\
\texttt{cardboard} & 3212 & 1261 & \textbf{41.24} & 16.88 & \textbf{52.06} & 5.95 & 1.75 \\
\texttt{other} & 409 & 90 & 22.01 & 15.61 & 49.37 & 0.00 & - \\
\texttt{paper} & 131 & 73 & \textbf{55.73} & \textbf{41.41} & 49.14 & \textbf{100.00} & - \\
\texttt{plastic bubble wrap} & 187 & 36 & 19.25 & 11.18 & 50.00 & 0.00 & - \\
\texttt{plastic hard} & 1032 & 148 & 17.50 & 16.05 & 52.32 & 5.41 & \textbf{5.41} \\
\texttt{plastic loose bag} & 2419 & 684 & 30.84 & 22.39 & 51.89 & 3.95 & 3.95 \\
\texttt{plastic tight wrap} & 850 & 195 & 24.13 & 17.07 & 50.77 & 1.54 & 1.54 \\
    \hline \hline
\end{tabular}
\end{table*}

\begin{table*}[!ht]
\centering
\caption{Damage Type KPIsfor \texttt{Pixtral-ft}}
\begin{tabular}{rccccccc}
\textbf{Damage Type} & \textbf{Total} & \textbf{Defects} & \textbf{AP\textsubscript{any}} [\%] & \textbf{AP\textsubscript{major}} [\%] & \textbf{AUROC} & \textbf{R@50\%P} [\%] & \textbf{R@1\%FPR} [\%] \\ \hline \hline
\texttt{penetration} & 7827 & 966 & 14.35 & 9.84 & 51.86 & 2.69 & 4.45 \\
\texttt{deformation} & 8646 & 1785 & \textbf{22.18} & 7.78 & 51.65 & 4.03 & 4.03 \\
\texttt{actuation} & 8403 & 1542 & 20.71 & 13.79 & 51.78 & 4.28 & 4.28 \\
\texttt{superficial} & 7585 & 724 & 10.18 & 1.52 & 51.22 & 0.00 & 3.18 \\
\texttt{spillage} & 6943 & 82 & 1.36 & 1.04 & 51.46 & 0.00 & 3.66 \\
\texttt{deconstruction} & 7888 & 1027 & 16.71 & \textbf{14.52} & \textbf{52.56} & \textbf{5.84} & \textbf{5.84} \\
\texttt{missing unit} & 6886 & 25 & 0.36 & 0.36 & 49.63 & 0.00 & 0.00 \\
    \hline \hline
\end{tabular}
\end{table*}

\begin{table*}[!ht]
\centering
\caption{Item Material KPIs for \texttt{AutoGlounMM}}
\begin{tabular}{rccccccc}
\textbf{Material} & \textbf{Total} & \textbf{Defects} & \textbf{AP\textsubscript{any}} [\%] & \textbf{AP\textsubscript{major}} [\%] & \textbf{AUROC} & \textbf{R@50\%P} [\%] & \textbf{R@1\%FPR} [\%] \\ \hline \hline
\texttt{book other} & 36 & 10 & 91.18 & \textbf{100.00} & 95.77 & \textbf{100.00} & 60.00 \\
\texttt{book paper} & 1664 & 671 & 87.99 & 89.72 & 90.00 & 98.21 & 37.71 \\
\texttt{book plastic tight wrap} & 127 & 38 & 88.07 & 89.84 & 92.82 & 94.74 & 44.74 \\
\texttt{cardboard} & 3212 & 1261 & 87.11 & 89.92 & 89.69 & 97.70 & 41.08 \\
\texttt{other} & 409 & 90 & 73.62 & 73.75 & 87.85 & 84.44 & 31.11 \\
\texttt{paper} & 131 & 73 & 94.08 & 96.07 & 91.73 & \textbf{100.00} & 45.21 \\
\texttt{plastic bubble wrap} & 187 & 36 & 93.55 & 89.45 & \textbf{98.73} & \textbf{100.00} & \textbf{77.78} \\
\texttt{plastic hard} & 1032 & 148 & 88.37 & 88.35 & 95.21 & 89.87 & 74.32 \\
\texttt{plastic loose bag} & 2419 & 684 & 90.27 & 89.32 & 94.59 & 95.91 & 55.85 \\
\texttt{plastic tight wrap} & 850 & 195 & 82.71 & 86.68 & 92.10 & 91.80 & 48.72 \\
    \hline \hline
\end{tabular}
\end{table*}

\begin{table*}[!ht]
\centering
\caption{Damage Type KPIsfor \texttt{AutoGlounMM}}
\begin{tabular}{rccccccc}
\textbf{Damage Type} & \textbf{Total} & \textbf{Defects} & \textbf{AP\textsubscript{any}} [\%] & \textbf{AP\textsubscript{major}} [\%] & \textbf{AUROC} & \textbf{R@50\%P} [\%] & \textbf{R@1\%FPR} [\%] \\ \hline \hline
\texttt{penetration} & 7827 & 966 & 76.85 & 81.41 & 92.75 & 79.81 & 50.83 \\
\texttt{deformation} & 8646 & 1785 & 82.15 & 83.23 & 92.99 & \textbf{90.20} & 43.59 \\
\texttt{actuation} & 8403 & 1542 & \textbf{85.41} & \textbf{87.45} & \textbf{94.55} & 91.38 & 55.97 \\
\texttt{superficial} & 7585 & 724 & 65.02 & 64.72 & 92.38 & 69.89 & 32.46 \\
\texttt{spillage} & 6943 & 82 & 41.72 & 43.75 & 86.44 & 39.02 & 46.34 \\
\texttt{deconstruction} & 7888 & 1027 & 82.94 & 84.86 & 94.05 & 86.47 & \textbf{60.37} \\
\texttt{missing unit} & 6886 & 25 & 37.16 & 37.16 & 84.94 & 36.00 & 48.00 \\
    \hline \hline
\end{tabular}
\end{table*}

\begin{table*}[!ht]
\centering
\caption{Item Material KPIs for \texttt{PatchCore50-ft}}
\begin{tabular}{rccccccc}
\textbf{Material} & \textbf{Total} & \textbf{Defects} & \textbf{AP\textsubscript{any}} [\%] & \textbf{AP\textsubscript{major}} [\%] & \textbf{AUROC} & \textbf{R@50\%P} [\%] & \textbf{R@1\%FPR} [\%] \\ \hline \hline
\texttt{book other} & 36 & 10 & 38.53 & 50.00 & 63.46 & 0.00 & - \\
\texttt{book paper} & 1664 & 671 & \textbf{52.97} & 21.87 & 63.45 & 51.12 & 3.58 \\
\texttt{book plastic tight wrap} & 127 & 38 & 52.01 & \textbf{42.58} & 72.30 & 42.11 & \textbf{7.90} \\
\texttt{cardboard} & 3212 & 1261 & 45.20 & 15.65 & 56.09 & 10.15 & 3.01 \\
\texttt{other} & 409 & 90 & 31.54 & 23.47 & 60.43 & 6.67 & 4.44 \\
\texttt{paper} & 131 & 73 & 59.69 & 51.09 & 50.87 & \textbf{100.00} & 2.74 \\
\texttt{plastic bubble wrap} & 187 & 36 & 36.15 & 32.39 & \textbf{76.29} & 11.11 & 0.00 \\
\texttt{plastic hard} & 1032 & 148 & 23.11 & 21.63 & 62.94 & 6.76 & 6.76 \\
\texttt{plastic loose bag} & 2419 & 684 & 36.65 & 25.60 & 63.65 & 0.00 & 1.61 \\
\texttt{plastic tight wrap} & 850 & 195 & 30.72 & 21.10 & 62.62 & 0.00 & 0.51 \\
    \hline \hline
\end{tabular}
\end{table*}

\begin{table*}[!ht]
\centering
\caption{Damage Type KPIsfor \texttt{PatchCore50-ft}}
\begin{tabular}{rccccccc}
\textbf{Damage Type} & \textbf{Total} & \textbf{Defects} & \textbf{AP\textsubscript{any}} [\%] & \textbf{AP\textsubscript{major}} [\%] & \textbf{AUROC} & \textbf{R@50\%P} [\%] & \textbf{R@1\%FPR} [\%] \\ \hline \hline
\texttt{penetration} & 7827 & 966 & 17.30 & 10.33 & 61.46 & 0.00 & 2.07 \\
\texttt{deformation} & 8646 & 1785 & \textbf{27.28} & 8.27 & 59.16 & \textbf{0.56} & \textbf{2.30} \\
\texttt{actuation} & 8403 & 1542 & 25.02 & 14.93 & 62.63 & 0.00 & 1.88 \\
\texttt{superficial} & 7585 & 724 & 11.61 & 1.58 & 53.25 & 0.00 & 2.21 \\
\texttt{spillage} & 6943 & 82 & 1.83 & 1.45 & 62.05 & 0.00 & 2.44 \\
\texttt{deconstruction} & 7888 & 1027 & 19.16 & \textbf{15.30} & \textbf{66.08} & 0.00 & 1.36 \\
\texttt{missing unit} & 6886 & 25 & 0.55 & 0.55 & 61.83 & 0.00 & 0.00 \\
    \hline \hline
\end{tabular}
\end{table*}

\begin{table*}[!ht]
\centering
\caption{Item Material KPIs for \texttt{AutoGluonMM-gal}}
\begin{tabular}{rccccccc}
\textbf{Material} & \textbf{Total} & \textbf{Defects} & \textbf{AP\textsubscript{any}} [\%] & \textbf{AP\textsubscript{major}} [\%] & \textbf{AUROC} & \textbf{R@50\%P} [\%] & \textbf{R@1\%FPR} [\%] \\ \hline \hline
\texttt{book other} & 36 & 10 & 50.70 & 25.00 & 79.62 & 70.00 & - \\
\texttt{book paper} & 1664 & 671 & 75.12 & 66.10 & 81.07 & 93.29 & 13.86 \\
\texttt{book plastic tight wrap} & 127 & 38 & \textbf{85.85} & \textbf{81.44} & \textbf{92.58} & \textbf{97.37} & \textbf{26.32} \\
\texttt{cardboard} & 3212 & 1261 & 70.14 & 58.16 & 79.10 & 92.07 & 13.64 \\
\texttt{other} & 409 & 90 & 56.90 & 56.83 & 76.70 & 53.33 & 22.22 \\
\texttt{paper} & 131 & 73 & 79.82 & 74.69 & 80.99 & \textbf{100.00} & 5.48 \\
\texttt{plastic bubble wrap} & 187 & 36 & 49.04 & 36.65 & 85.67 & 75.00 & 2.78 \\
\texttt{plastic hard} & 1032 & 148 & 64.01 & 64.40 & 86.34 & 66.22 & 29.73 \\
\texttt{plastic loose bag} & 2419 & 684 & 73.77 & 68.66 & 88.06 & 91.96 & 14.77 \\
\texttt{plastic tight wrap} & 850 & 195 & 65.69 & 63.66 & 84.37 & 71.28 & 16.41 \\
    \hline \hline
\end{tabular}
\end{table*}

\begin{table*}[!ht]
\centering
\caption{Damage Type KPIsfor \texttt{AutoGluonMM-gal}}
\begin{tabular}{rccccccc}
\textbf{Damage Type} & \textbf{Total} & \textbf{Defects} & \textbf{AP\textsubscript{any}} [\%] & \textbf{AP\textsubscript{major}} [\%] & \textbf{AUROC} & \textbf{R@50\%P} [\%] & \textbf{R@1\%FPR} [\%] \\ \hline \hline
\texttt{penetration} & 7827 & 966 & 48.35 & 47.07 & 84.40 & 46.27 & 16.87 \\
\texttt{deformation} & 8646 & 1785 & 58.47 & 43.24 & 85.04 & \textbf{71.99} & 12.21 \\
\texttt{actuation} & 8403 & 1542 & \textbf{62.11} & \textbf{57.57} & \textbf{86.93} & 75.36 & 16.73 \\
\texttt{superficial} & 7585 & 724 & 32.67 & 19.27 & 83.59 & 2.62 & 7.18 \\
\texttt{spillage} & 6943 & 82 & 6.34 & 6.65 & 69.51 & 0.00 & 18.29 \\
\texttt{deconstruction} & 7888 & 1027 & 57.13 & 56.43 & 86.50 & 65.92 & \textbf{19.96} \\
\texttt{missing unit} & 6886 & 25 & 16.27 & 16.27 & 84.46 & 8.00 & 36.00 \\
    \hline \hline
\end{tabular}
\end{table*}

\section{Benchmark: Model Training Details}
\label{sec:model_training_details}

For the supervised methods we used validation set hyperparameter tuning. We employed standard augmentations (color, rotation, flip, zoom) for supervised baselines, and others already inherently apply similar concepts (WinCLIP’s windowed feature extraction). Unless stated otherwise, we used the default values for all methods. Finally, we addressed key
issues, such as AD methods (PatchCore, WinCLIP) being
distracted by the image background, by applying them to
item crops rather than full tray crops.
\vspace{-4mm}
\paragraph{CLIP and POMP}
For both the CLIP and POMP baselines we use the open-source implementation of POMP~\cite{ren2023prompt}. Thereby, we use the ViT-B/16 model, and $224 \times 224$ image input resolution. The evaluated prompts are listed in Section~\ref{sec:clip_prompts}. Example model predictions are shown in Table~\ref{tab:failure_modes_no_references}.
\vspace{-4mm}
\paragraph{WinCLIP}

We use the open-source implementation of WinCLIP~\cite{jeong2023winclip}, based on a pre-trained openCLIP backbone (ViT-B/16 with $240\times240$ input resolution pretrained on LAION400M dataset).  We use the original positive and negative text prompts (22 templates combined with 11 normal/anomaly states), using the generic item type \texttt{object}, e.g.
\vspace{-2mm}
\begin{verbatim}
a photo of an object with damage
a photo of the object with damage
for visual inspection
\end{verbatim}
\vspace{-2mm}
Example model predictions are shown in Table \ref{tab:failure_modes_with_training_with_references}.

\vspace{-4mm}
\paragraph{Claude}
For our experiments with Claude we use the AWS Bedrock API. The model version is Claude Sonnet v3.5 and we use the default sampling parameters: temperature $t=1$, nucleus sampling $top_p = 0.999$ and no $top_k$ sampling.
\vspace{-4mm}
\paragraph{Pixtral}
We use the HuggingFace library to fine-tune the Pixtral-12B model for our purpose. We use a constant learning rate of $3^{-5}$ and an effective batch size of 32. We run distributed training for one epoch on the entire training set across 8 A100 GPUs. The training data is constructed from the available labels as follows:
\vspace{-2mm}
\begin{verbatim}
This item is {condition}.
```json
{
  "is_damaged_ge1": {true/false},
  "is_damaged_ge2": {true/false},
  "damage_intensity_median": {0, 1, 2},
  "condition": "{condition}",
  "severity": {severity}
}
```
\end{verbatim}
\vspace{-3mm}

where \emph{condition} is set to \texttt{DAMAGED} or \texttt{UNDAMAGED}, according to the field
\texttt{is\_damaged\_ge1}, and \emph{severity} is the damage intensity median multiplied by 5 to bring it to the range between 0 and 10. We then use the same prompt as before (see Section~\ref{sec:vlm_prompts}) and extract the value for \emph{severity} as our damage confidence prediction.

\vspace{-4mm}
\paragraph{PatchCore}
We used the PatchCore implementation from the anomalib package \cite{akcay2022anomalib}. For each test image, a memory bank was built from up to 3 reference images (batch size = 1) with a coreset sampling ratio of 1.0 (no subsampling). Patch-level anomaly scores were computed using Euclidean distance to the max(1, |reference images|) nearest neighbors. We used ResNet50 with 1024×1024 inputs, ImageNet normalization, and features extracted from layers 2 and 3. The image-level anomaly threshold was computed using anomalib’s Adaptive F1 method on a validation split containing both normal and anomalous samples. For this step, the memory bank was constructed from 10 normal validation images with a 0.01 coreset ratio. The resulting thresholds (43.76 for PatchCore50, 3.69 for PatchCore50-ft) were fixed during testing via anomalib’s Manual Threshold setting. Example predictions are shown in Table~\ref{tab:failure_modes_with_training_with_references}.

\vspace{-4mm}
\paragraph{ResNet50}
\label{app:resnet50-training_details}
We fine-tune different ResNet backbones pre-training on ImageNet on a single V100 GPU using the \href{https://timm.fast.ai/}{timm} library. After hyperparameter tuning, we arrived at the following settings (final choices in \textbf{bold}):
\begin{itemize}
    \item Backbones: ResNet34, \textbf{ResNet50}, ResNet152
    \item Pre-training weights: ImageNet
    \item Input size: 1024x1024, RGB
    \item Batch size 24 (ResNet152), \textbf{48} (\textbf{ResNet50}), 64 (ResNet34); Epochs: 20
    \item Optimizer: SGD with initial learning rate 0.005, momentum 0.8, ReduceLROnPlateau learning rate scheduler, reduction factor (0.15), patience epochs (1), minimum learning rate (1e-5)
    \item Dropout (dense layer for classification only, 0.5)
    \item Data augmentation: flip (horizontal/vertical), rotation ($0 \ldots 180$), shift 10\%, zoom 10\%
\end{itemize}
Example model predictions are shown in Table~\ref{tab:failure_modes_no_references}.

\vspace{-4mm}
\paragraph{ViT-S}
We fine-tuned a Vision Transformer with different pretraining weights on 8 V100 GPUs.
After hyperparameter tuning, we arrived at the following settings:
\begin{itemize}
    \item Backbones: ViT-small (22.1M parameters)\cite{dosovitskiy2021imageworth16x16words}.
    \item Pre-training weights: \textbf{DINOv2}~\cite{oquab2024dinov}
    \item Input size: 1024x1024, RGB, Patch size: 14x14, RGB
    \item Batch size 8, Epochs: 30
    \item Optimizer: Adam with initial learning rate $5 \times 10^{-6}$ (not scaled by Batch size), momentum 0.9, weight decay (0.05), Cosine scheduler
    \item Data augmentation: flip (horizontal/vertical) and AutoAgument set to \texttt{rand-m9-mstd0.5-inc1}
\end{itemize}
Example model predictions are shown in Table~\ref{tab:failure_modes_no_references}.
\vspace{-4mm}
\paragraph{AutoGluon}
To train a classifier with reference images we employ AutoGluonMM, a freely available AutoML framework (\url{https://auto.gluon.ai}). Training is performed on a single V100 GPU using the default parameter settings. 20\% of the training set are used for validation. The resulting multi-modal fusion MLP model uses a ViT backbone and has 97.4M parameters. Example model predictions are shown in Table \ref{tab:failure_modes_autogluon}.

\begin{figure*}
    \centering
    \begin{subfigure}[b]{0.45\textwidth}
        \centering
        \includegraphics[width=\textwidth]{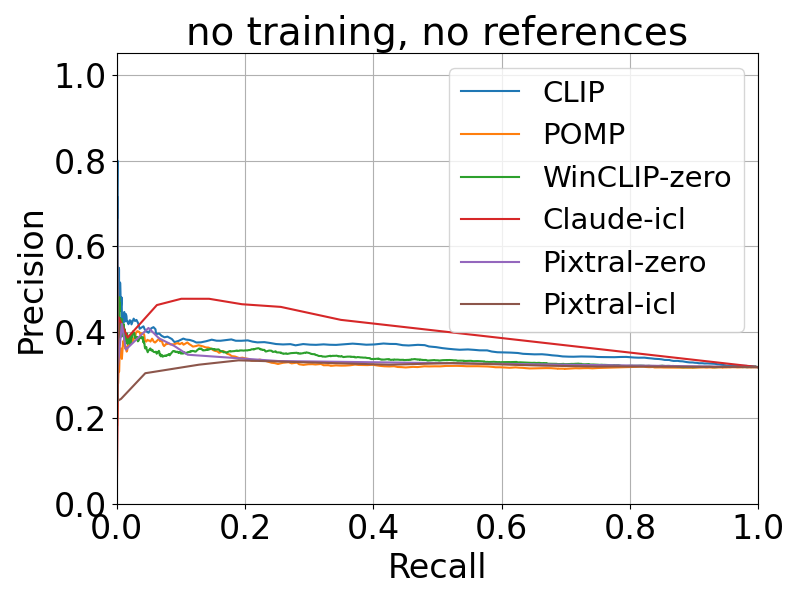}
    \end{subfigure}
    \hfill
    \begin{subfigure}[b]{0.45\textwidth}
        \centering
        \includegraphics[width=\textwidth]{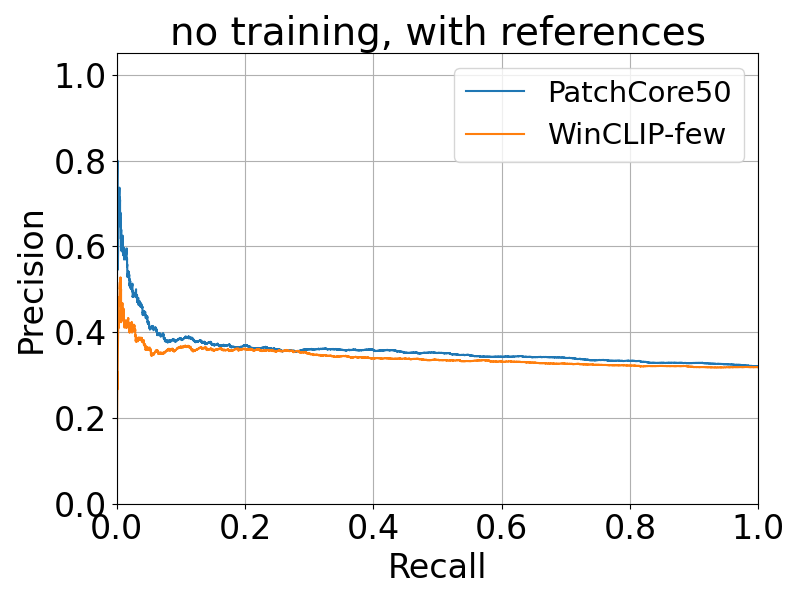}
    \end{subfigure}

    \vspace{0.5cm}

    \begin{subfigure}[b]{0.45\textwidth}
        \centering
        \includegraphics[width=\textwidth]{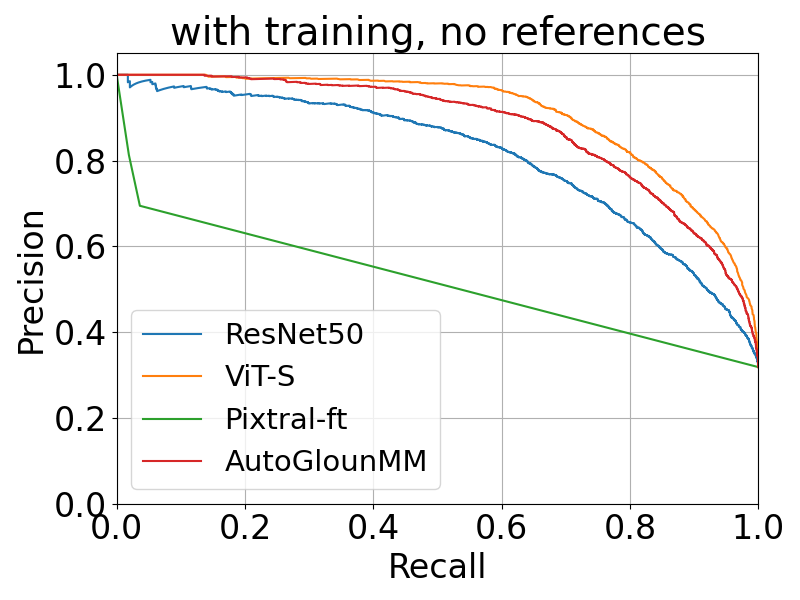}
    \end{subfigure}
    \hfill
    \begin{subfigure}[b]{0.45\textwidth}
        \centering
        \includegraphics[width=\textwidth]{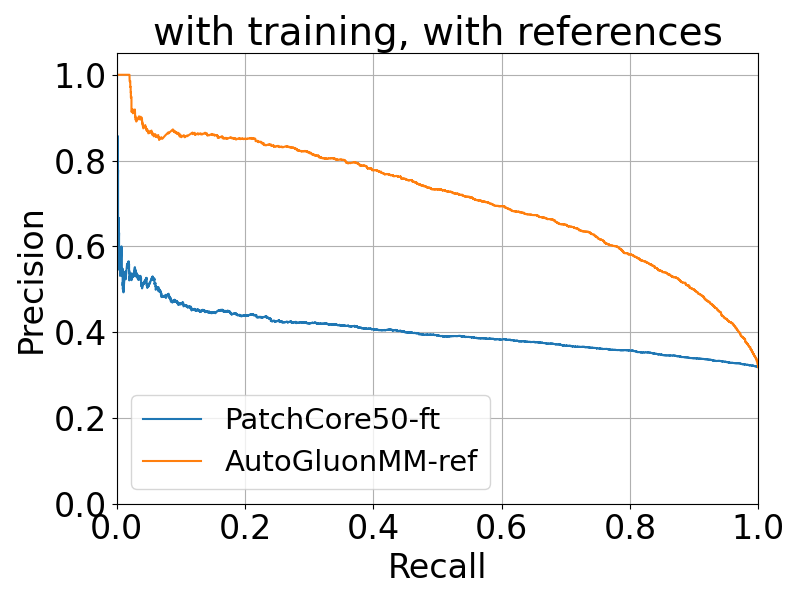}
    \end{subfigure}

    \caption{Precision-Recall curves for all four experiments. These curves illustrate the trade-off between precision and recall across different operating thresholds for each experimental configuration.}
    \label{fig:pr_curves}
\end{figure*}

\begin{figure*}[p]
    \centering
    \begin{subfigure}[b]{0.45\textwidth}
        \centering
        \includegraphics[width=\textwidth]{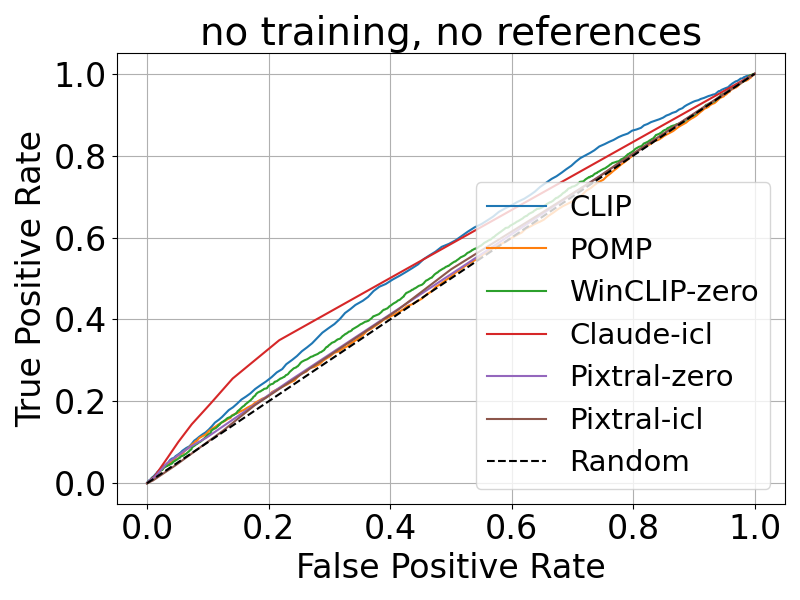}
    \end{subfigure}
    \hfill
    \begin{subfigure}[b]{0.45\textwidth}
        \centering
        \includegraphics[width=\textwidth]{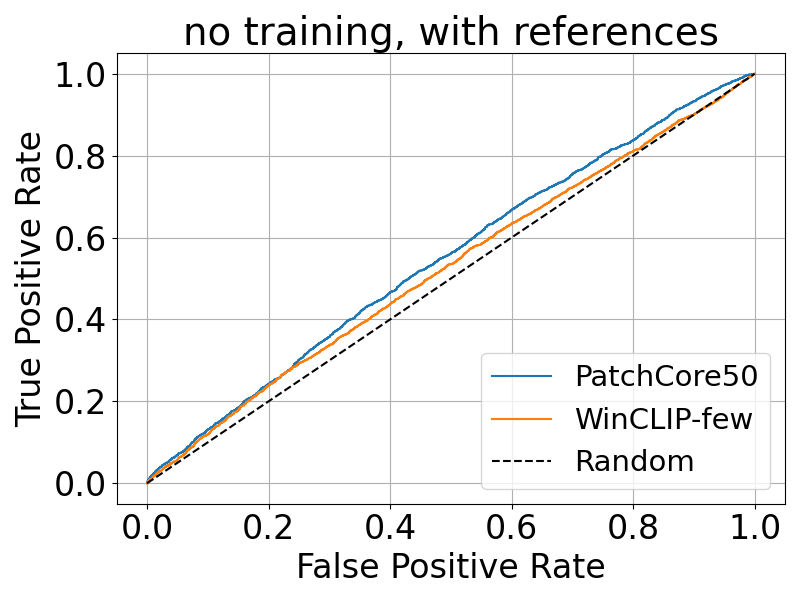}
    \end{subfigure}

    \vspace{0.5cm}

    \begin{subfigure}[b]{0.45\textwidth}
        \centering
        \includegraphics[width=\textwidth]{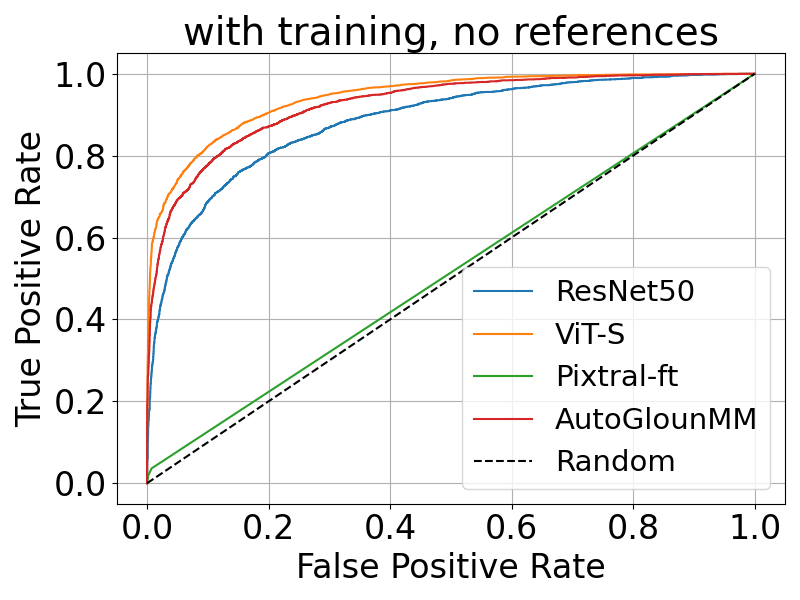}
    \end{subfigure}
    \hfill
    \begin{subfigure}[b]{0.45\textwidth}
        \centering
        \includegraphics[width=\textwidth]{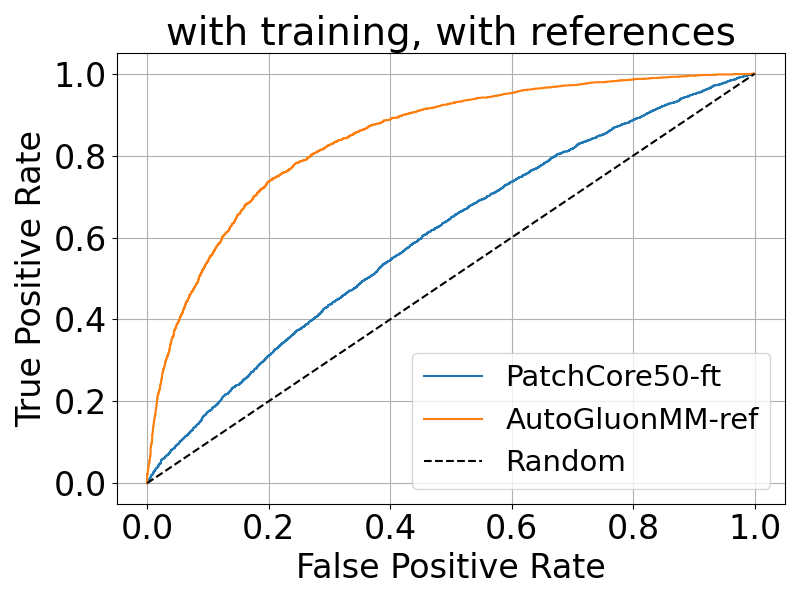}
    \end{subfigure}

    \caption{Receiver Operating Characteristic (ROC) curves for all four experiments. These curves show the performance of the classification models by plotting the true positive rate against the false positive rate at various threshold settings.}
    \label{fig:roc_curves}
\end{figure*}

\end{document}